%% file: acl_latex.tex
\pdfoutput=1

\documentclass[11pt]{article}

\usepackage[final]{acl}

\usepackage{times}
\usepackage{latexsym}

\usepackage[T1]{fontenc}

\usepackage[utf8]{inputenc}

\usepackage{microtype}

\usepackage{inconsolata}

\usepackage{graphicx}

\usepackage{multirow}

\usepackage{booktabs}
\usepackage{makecell}
\usepackage{xcolor}
\usepackage{vcell}
\usepackage{colortbl}
\usepackage{caption}
\usepackage{amssymb}
\usepackage{pifont}
\usepackage{adjustbox}
\usepackage{amsmath}
\usepackage{stfloats}
\usepackage{hyperref}
\usepackage[toc]{multitoc}
\usepackage[skins]{tcolorbox}
\tcbuselibrary{breakable} 
\newtcolorbox[auto counter, number within=section, list type=subsubsection, list inside=toc]{sectionbox}[2][]{
colback=white!98!gray, colframe=black, 
colbacktitle=white!90!gray, coltitle=black, 
fonttitle=\bfseries,
title={#2}, 
list entry={Comment \thetcbcounter\quad}
}

\newcommand{\xmark}{\ding{55}}%

\definecolor{wkred}{RGB}{255, 200, 200}
\definecolor{wkblue}{RGB}{210, 230, 250}
\definecolor{wkgold}{RGB}{255, 223, 129}
\definecolor{wksilver}{RGB}{192, 192, 192}

\definecolor{codegreen}{rgb}{0,0.6,0}
\definecolor{codegray}{rgb}{0.5,0.5,0.5}
\definecolor{codepurple}{rgb}{0.58,0,0.82}
\definecolor{backcolour}{rgb}{0.95,0.95,0.92}
\definecolor{wkgreen}{RGB}{220,244,229}
\definecolor{wkpurple}{RGB}{210,210,253}
\definecolor{wkyellow}{RGB}{255,241,177}

\definecolor{upred}{HTML}{DC143C}
\definecolor{downgreen}{HTML}{32CD32}
\newcommand{\up}[1]{\textcolor{upred}{{+#1}}}
\newcommand{\down}[1]{\textcolor{downgreen}{{-#1}}}

\newcommand{\icdatasetname}{\texttt{{ImgCode-8.6M}}}
\newcommand{\mathdatasetname}{\texttt{{MM-MathInstruct-3M}}}
\newcommand{\icmodelname}{{{FigCodifier}}}

\newcommand{\best}{\cellcolor{wkred}}
\newcommand{\second}{\cellcolor{wkblue}}
\newcommand{\third}{\cellcolor{wkgreen}}

\definecolor{templategray}{RGB}{240,240,240}
\definecolor{swk_green}{RGB}{0,128,0}

\definecolor{shadecolor}{RGB}{237,237,237}

\newlength{\wklength}

\title{MathCoder-VL: Bridging Vision and Code for Enhanced \\ Multimodal Mathematical Reasoning}

\author{%
    Ke Wang$^{1}$ 
    \And 
    Junting Pan$^{1,2}$\thanks{Project lead} 
    \And 
    Linda Wei$^{1}$ 
    \And 
    Aojun Zhou$^{1}$ 
    \And Weikang Shi$^{1}$ 
    \And 
    Zimu Lu$^{1}$ 
    \AND 
    Han Xiao$^{1}$ 
    \And
    Yunqiao Yang$^{1}$ 
    \And 
    Houxing Ren$^{1}$ 
    \And Mingjie Zhan$^{1}$\thanks{Corresponding author} 
    \And 
    Hongsheng Li$^{1,2}$\footnotemark[2]
    \AND
    \vspace{-8mm}\\
\textsuperscript{1}Multimedia Laboratory (MMLab), The Chinese University of Hong Kong,
\textsuperscript{2}CPII under InnoHK\\
 \small{
   \href{mailto:email@domain}{wangk@link.cuhk.edu.hk} $\quad$
   \href{mailto:email@domain}{hsli@ee.cuhk.edu.hk}
 }
}

\begin{document}
\maketitle

\input{00-Abstract}

\input{01-Introduction}
\input{04-RelatedWork}

\input{02-Methods}

\input{03-Experiments}
\input{05-Conclusion}
\input{06-Limitations}

\bibliography{custom}
\appendix
\clearpage
\input{Appendix}

\end{document}

%% file: 00-Abstract.tex
\begin{abstract}
Natural language image-caption datasets, widely used for training Large Multimodal Models, mainly focus on natural scenarios and overlook the intricate details of mathematical figures that are critical for problem-solving, hindering the advancement of current LMMs in multimodal mathematical reasoning.
To this end, we propose leveraging code as supervision for cross-modal alignment, since code inherently encodes all information needed to generate corresponding figures, establishing a precise connection between the two modalities. Specifically, we co-develop our image-to-code model and dataset with model-in-the-loop approach, resulting in an image-to-code model, \icmodelname~and \icdatasetname, the largest image-code dataset to date.
Furthermore, we utilize \icmodelname~to synthesize novel mathematical figures and then construct \mathdatasetname, a high-quality multimodal math instruction fine-tuning dataset.
Finally, we present MathCoder-VL, trained with \icdatasetname~for cross-modal alignment and subsequently fine-tuned on \mathdatasetname~for multimodal math problem solving. Our model achieves a new open-source state-of-the-art across all six metrics. Notably, it surpasses GPT-4o and Claude 3.5 Sonnet in the geometry problem-solving subset of MathVista, achieving improvements of 8.9\% and 9.2\%.
The dataset and models will be released at \url{https://github.com/mathllm/MathCoder}.
\end{abstract}

%% file: 01-Introduction.tex
\section{Introduction}
\label{intro}
\begin{figure*}[t]
  \includegraphics[width=0.99\linewidth]{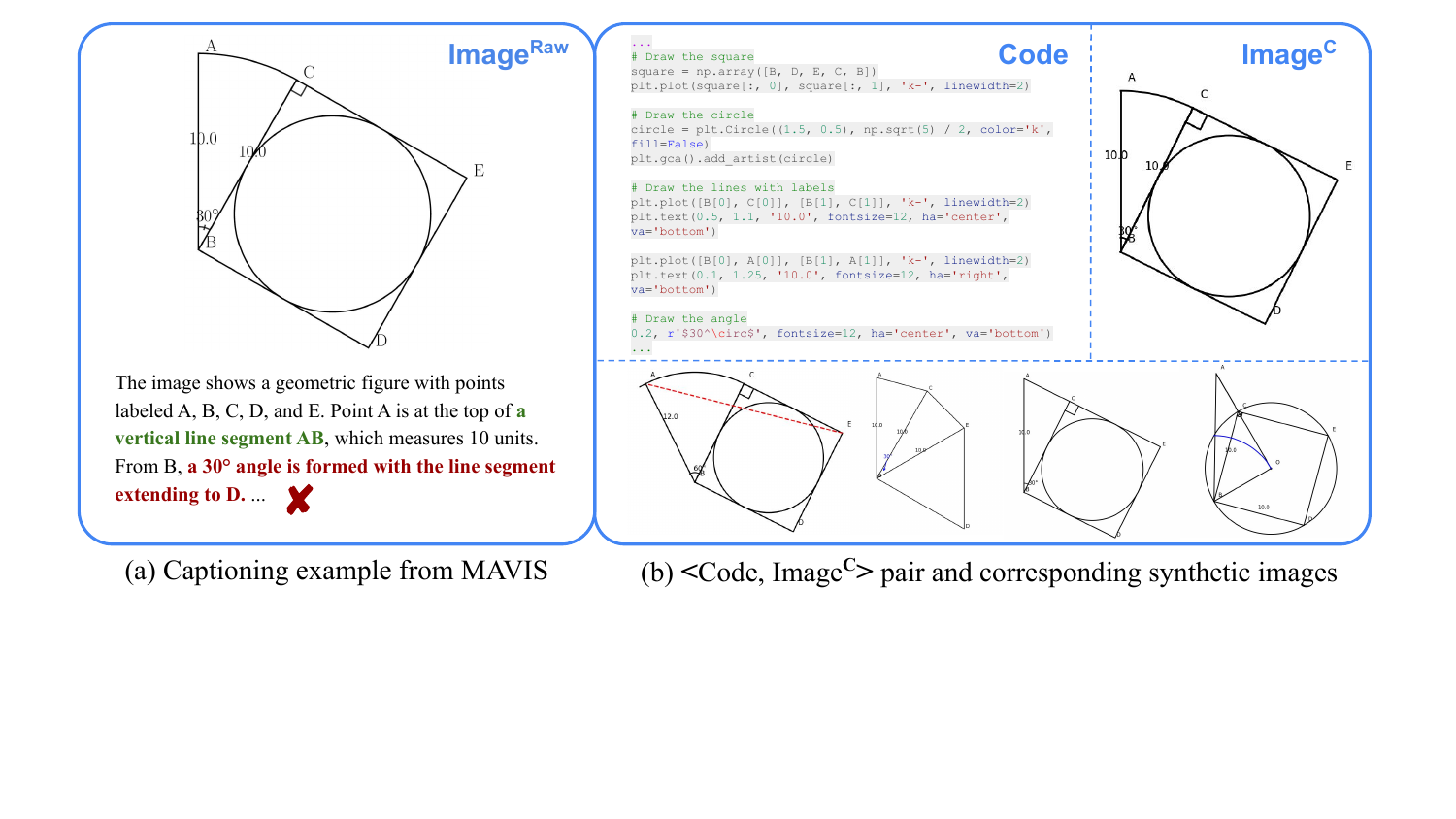}
  \vspace{-27mm}
\caption{(a) Natural language captions often struggle to convey {all details} in a image and {guarantee correctness}. (b) Our approach uses image-translated $\mathbf{Code}$ and code-generated $\mathbf{Image^C}$ to create $\mathbf{\langle Image^C, Code \rangle}$ pairs. Since the $\mathbf{Image^{C}}$ is rendered from the $\mathbf{Code}$, the cross-modal alignment is {always accurate and contains all the details}. Below are four examples of new figures synthesized based on $\mathbf{Image^{Raw}}$.}
  \label{fig:intro}
\end{figure*}

Recently, Large Language Models (LLMs) have outperformed humans in complex reasoning at the Olympiad competition level~\citep{openai2024openaio1card,deepseekai2025deepseekr1}. However, the reasoning abilities of Large Multimodal Models (LMMs) still fall short of their potential, often struggling with even simple tasks, such as simple geometry problems~\citep{wang2024mathvision}. Overcoming these limitations is essential for advancing toward Artificial General Intelligence (AGI).

In our efforts to enhance the mathematical capabilities of LMMs, we identify two key challenges that distinguish them from LLMs:
(i) Aligning math-related visual and textual details accurately to enable effective problem-solving.
(ii) Scaling the generation of diverse new math figures for multi-modal math problem synthesis.

Despite significant advancements, LMMs still struggle with effective modality alignment, especially in the math field, primarily due to the scarcity of high-quality, error-free, math-specific cross-modal data. Traditional image caption datasets~\citep{chen2023sharegpt4vimprovinglargemultimodal,schuhmann2022laion5b} often focus on natural scenarios and lose details important for math problem-solving, and cannot guarantee correctness, as shown in Figure~\ref{fig:intro} (a).

In contrast, code inherently contains all information needed to render corresponding image and establish a strict correspondence between the two modalities. In light of this, we propose image-to-code mid-training to enhance math-related cross-modal alignment. We construct an image-to-code model, \icmodelname, which converts math-related images into detailed code capable of rendering new images, as shown in Figure~\ref{fig:intro} (b). By pairing the generated code with the rendered images, we create high-quality $\mathbf{\langle Image^C, Code \rangle}$ pairs that are inherently \textbf{always accurate and contain all details for cross-modal alignment}. Using this automated data engine, we construct \icdatasetname, significantly enhancing LMMs' cross-modal ability.

Additionally, with a higher temperature, our \icmodelname~can synthesize new images that are more different from the raw images, which enables the \textbf{synthesis of new diverse images} for problem-solving dataset construction. Synthetic data have proven effective for math reasoning~\citep{wang2023mathcoderseamlesscodeintegration,gou2024toratoolintegratedreasoningagent,huang2024key}, and dataset quality and diversity are the most important factors. However, the construction of multi-modal math problem-solving datasets still relies heavily on either question rewriting and generating new solutions~\citep{guo2024mammoth,luo2025ursa}, sourcing existing images~\citep{shi2024mathllava,peng2024multimath}, or manually designed figures~\citep{zhuang2024mathpuma,zhang2025mavis}. The diversity of images lags significantly behind the diversity of text, restricting the overall dataset variety. Unlike these methods, with our \icmodelname, generating new images becomes significantly easier, as shown in Figure~\ref{fig:intro} (b). This allows us to create diverse new math figures at low cost, which has the potential to improve LMMs' mathematical reasoning abilities substantially. Our main contributions are:

1. We co-develop our image-to-code model with model-in-the-loop approach, resulting in a \icmodelname~model and \icdatasetname~dataset, the largest image-code dataset to date.

2. With our \icmodelname, we construct \mathdatasetname. To our knowledge, this is the first high-quality multi-modal problem-solving dataset with {not only new questions but also diverse newly synthesized images}.

3. We present MathCoder-VL, achieving SOTA results across all six metrics among comparable-size LMMs. We will open-source our models, code and datasets.

%% file: 04-RelatedWork.tex
\section{Related Works}
\label{sec:related_works}

\textbf{Multimodal Math Reasoning}
The mathematical reasoning abilities of LMMs have garnered widespread attention~\cite{gao2023gllava,li2024llavaonevision,dong2024insight,hu2024visual,yang2024mathglmvision,han24infimm,guo2024mammoth}. Unlike mathematical reasoning tasks in traditional large language models~\cite{zhou2024solving,luo2023wizardmath,yu2023metamath}, multimodal mathematical reasoning requires LMMs to extract information from the visual domain and perform cross-modal reasoning. Tasks such as geometric problem-solving are particularly challenging~\cite{chen2021geoqa,wang2024mathvision}. 
Several studies have attempted to enhance the input of visual mathematical signals by enhancing visual encoders~\cite{liu2024sphinxx,chen2024internvl2}. However, ensuring accurate correspondence between images and text remains a significant challenge. To address this, we propose using code and code-generated images, which inherently maintain precise and sufficient alignment between modalities.

\textbf{Data Synthesis.}
Methods based on data synthesis are favored by academia and industry due to their demonstrated efficiency~\cite{sprague2024cot,lu2023machine,huang2024key,fu2024vita}. Numerous fine-tuning~\citep{yu2024metamathbootstrapmathematicalquestions, wang2023mathcoderseamlesscodeintegration, lu2024mathgeniegeneratingsyntheticdata} and pretraining~\cite{gunasekar2023textbooksneed, wang2023generativeaimathi, yang2024qwen25mathtechnicalreportmathematical} studies have explored training on synthetic data generated using language models or predefined templates. MathGLM~\citep{yang2023gptsolvemathematicalproblems} and InternLM-Math~\citep{ying2024internlmmath} use templates to generate synthetic numerical operation data, while Phi~\citep{gunasekar2023textbooksneed} produces textbook-quality data with models. EntiGraph~\citep{yang2024syntheticcontinuedpretraining} generates diverse text by drawing connections between sampled entities.
However, efforts on the synthesis of multimodal mathematical reasoning data are primarily focused on the diversity and complexity of problem or solution text. Math-LLaVA~\cite{shi2024mathllava} proposes the MathV360K dataset by classifying images based on complexity and enhancing questions accordingly. R-CoT~\cite{deng2024rcot}, GeoGPT4V~\cite{cai2024geogpt4v}, MammoTH-VL~\citep{guo2024mammoth}, and Multimath~\cite{peng2024multimath} collect and enhance problems or solutions. MAVIS~\citep{zhang2025mavis} generates new geometry and function images with code but lacks diversity, as the codes are design by humans and only contain three types. Our work proposes a novel method that can synthesize diverse new images automatically for crafting problems.

%% file: 02-Methods.tex
\begin{figure*}[t]
  \includegraphics[width=0.99\linewidth]{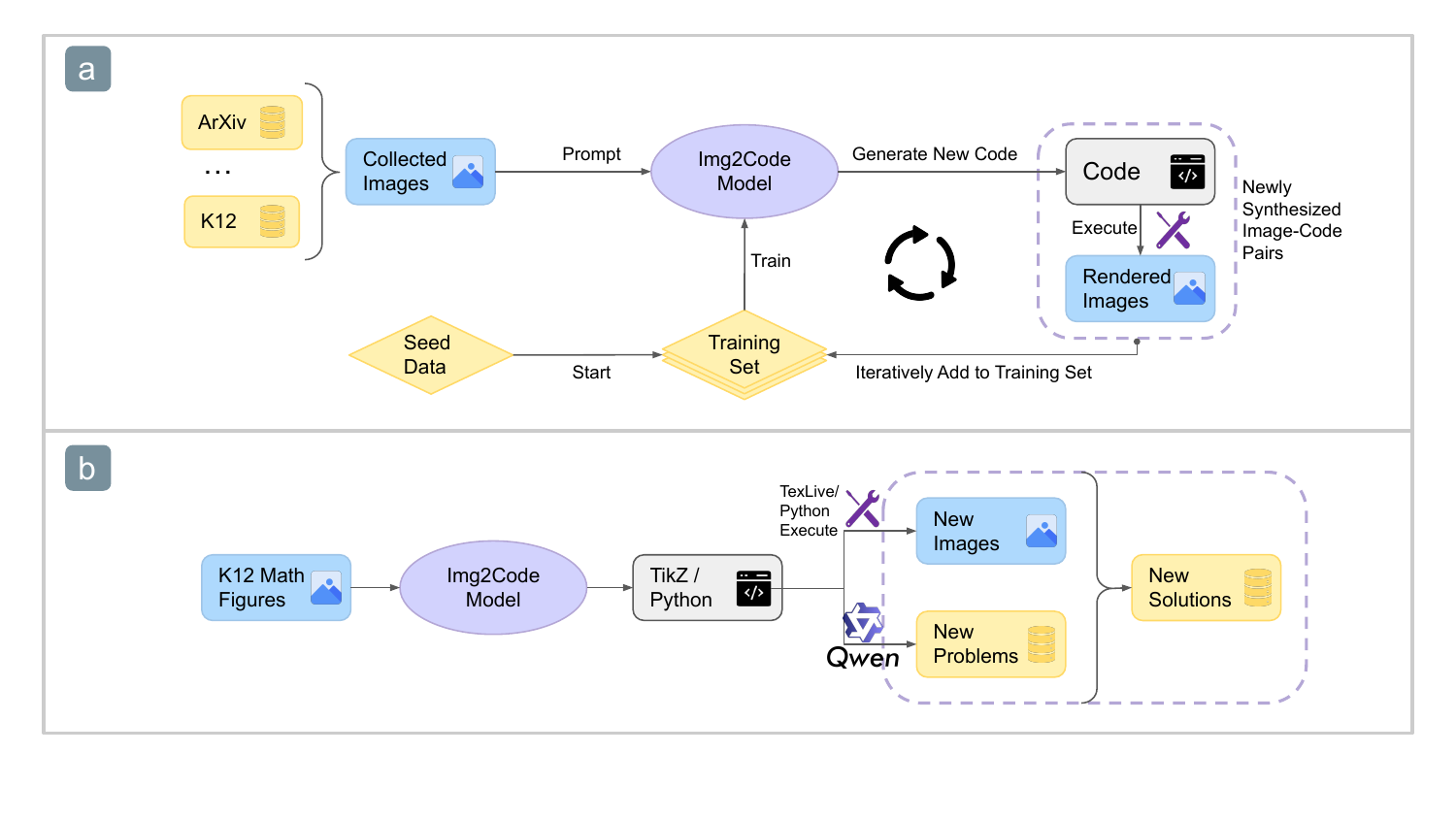}
  \vspace{-8mm}
  \caption {(a) The iterative training pipeline of our image-to-code model. We use DaTikZ-119K as seed data to train our first image-to-code model. We start by collecting 3 million math-related images and ultimately synthesize 8.6 million image-code pairs. Our final image-to-code model, \icmodelname, is based on InternVL2-8B~\citep{chen2024internvl}, with all model parameters being fully learnable.  
(b) The pipeline for generating new math problems with diverse new images. Using the final model from (a), we convert raw images into code and leverage Qwen models to generate new questions and step-by-step solutions based on the newly synthesized images.} 
  \label{fig:data_pipeline}
\end{figure*}

\section{MathCoder-VL}
We developed MathCoder-VL through a two-stage process: image-to-code mid-training using the \icdatasetname~dataset, followed by math instruction fine-tuning on \mathdatasetname. This section details the construction of the two datasets.

\subsection{Image-to-Code Model and Data}
To synthesize image-code pairs and new images, we need models that can generate code to render high-quality mathematical figures. However, even commercial models like Claude 3.5 and GPT-4 struggle to perform image-to-code conversion effectively~\citep{belouadi2024detikzify}. Additionally, the largest TikZ dataset to date, DaTikZ~\citep{belouadi2024automatikz}, contains only 119k TikZ graphics. To address these limitations, we build \icdatasetname~and develop our \icmodelname.

\subsubsection{Collect Math-related Images}
\label{sec:math_related_images}
We start by collecting 3 million math-related images, of which 164K are paired with corresponding TikZ code. The data composition is as follows.

\textbf{DaTikZ Training Set.}
DaTikZ is designed to facilitate the development of machine learning models capable of generating or manipulating vector graphics in \LaTeX. We use the 119K image-TikZ code pairs from DaTikZ as our seed data.

\textbf{K12 Problem-Solving Dataset.}
To diversify our dataset, we included math problems from K12 books, exercises, and exams with permission from the data providers. We gathered 4.6 million math problems, of which 996K include at least one image. This dataset contains 1.57 million images from a wide range of math problems across all K12 grades, spanning 19 subjects, including Statistics, Probability, Algebra, Geometry, Functions, Permutations, Combinations, and more. See Sec.~\ref{sec:k12} for details on the curation process.

\textbf{Mathematical Textbooks.}
Textbooks provide structured presentations of math concepts and are a valuable resource. We collected 8K PDFs of math-related textbooks from online sources, focusing on titles with keywords like algebra, geometry, and probability. These PDFs were converted into markdown format, and the images were extracted, resulting in 202K diverse math-related images.

\textbf{ArXiv.}
We utilized bulk data from arXiv between September 2023 and October 2024, yielding 45K images with corresponding TikZ code and 681K images without code, many of which are statistical visualizations.

\textbf{Open-Source Datasets.}
MathV360K~\cite{shi2024mathllava} consists of 360K question-answer pairs and 40K images from 24 previous datasets. MultiMath~\citep{peng2024multimath} contains 300K newly collected math problems with 280K images, mostly consisting of geometry diagrams.

\subsubsection{Iteratively Build Image-to-Code Model}
\label{sec:imd2code_model}

We train our first image-to-code model using 119K image-TikZ pairs sourced from DaTikZ, leveraging InternVL-Chat-V1-2-40B~\cite{chen2024internvl}. As the dataset scales beyond one million samples, we adopt InternVL2-8B~\citep{chen2024internvl2} as the base model after comprehensively weighing the image-to-code performance and cost. The complete training pipeline is illustrated in Figure~\ref{fig:data_pipeline} (a).

\textbf{Synthesis of Image-Code Pairs.}
To scale the size of Image-Code pairs, we used the image-to-code model to translate the 3M collected images into corresponding code. We then run the generated code to render new images, and only the successfully generated $\mathbf{\langle Image^C, Code \rangle}$ pairs were included in our dataset. This iterative process allowed us to continually generate fresh $\mathbf{\langle Image^C, Code \rangle}$ pairs and refine the model with each new version. Ultimately, we get \icmodelname~and the \icdatasetname.

\textbf{TikZ to Python Conversion.}
In addition to TikZ code, we also leverage GPT-4o mini to translate TikZ code into Python code, which is then executed to generate new images. This step significantly expands our dataset, further enhancing the model's capabilities. By diversifying the types of code used, the model can generate a broader range of images, as different code structures produce distinct visual outputs. Through this process, we curate 3.1 million image-Python pairs.

\textbf{Data Cleaning and Deduplication.}\label{sec:imd2code_model:data_clean}
We implement a rigorous cleaning and deduplication process to ensure data quality:
1. {Code Validation}: We only retain code that generates a valid image. Over the course of the iterative process, the code success rate improves, rising from 46.5\% for TikZ to 81.2\% for TikZ and 84.5\% for Python on the DaTikZ test set.
2. {Deduplication}: We apply carefully designed rules to eliminate duplicate or highly similar code, removing 4.4\% of the dataset.
3. {Quality Filtering}: Through keyword matching, we filter out low-quality data, such as randomly generated or irrelevant images, which accounts for 3.7\% of the data.
4. {Code Length}: We remove code that is excessively long, which can introduce unnecessary complexity.
5. {Image Quality}: Images that are almost entirely white—identified through standard deviation and mean pixel value analysis—are removed, accounting for approximately 0.5\% of the data.
Details of this process can be found in Appendix~\ref{app:data_clean}.
After cleaning, we retain 4.3M image-TikZ pairs and 4.3M image-Python pairs.

\begin{figure}[t]
  \includegraphics[width=\columnwidth]{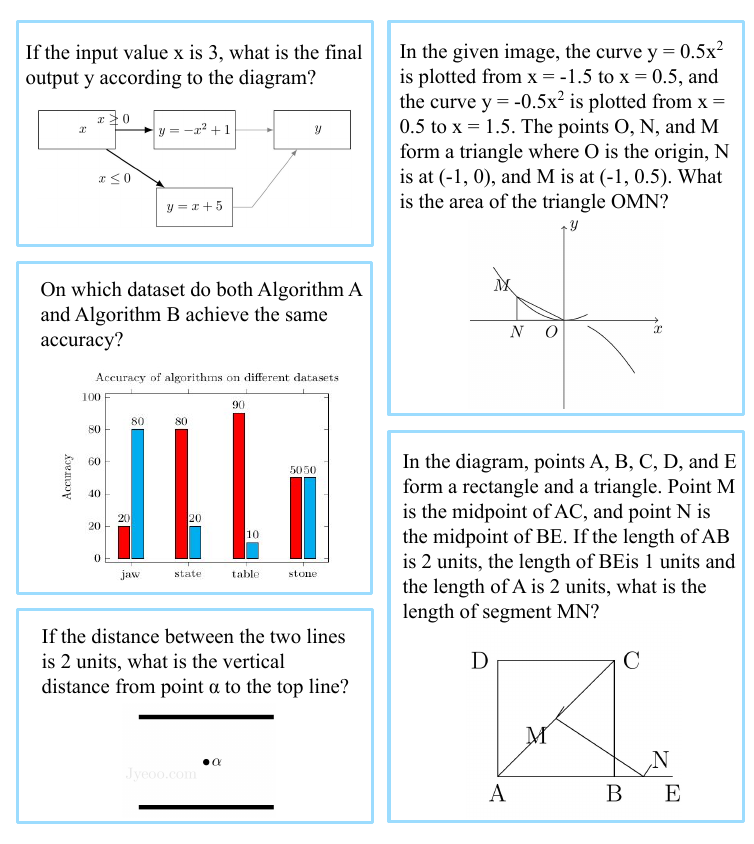}
  \vspace{-8mm}
  \caption{Sample questions paired with newly synthesized images, as generated in Figure~\ref{fig:data_pipeline} (b).}
  \label{fig:training_samples}
\end{figure}
\vspace{-2mm}
\begin{figure}[t]
  \includegraphics[width=\columnwidth]{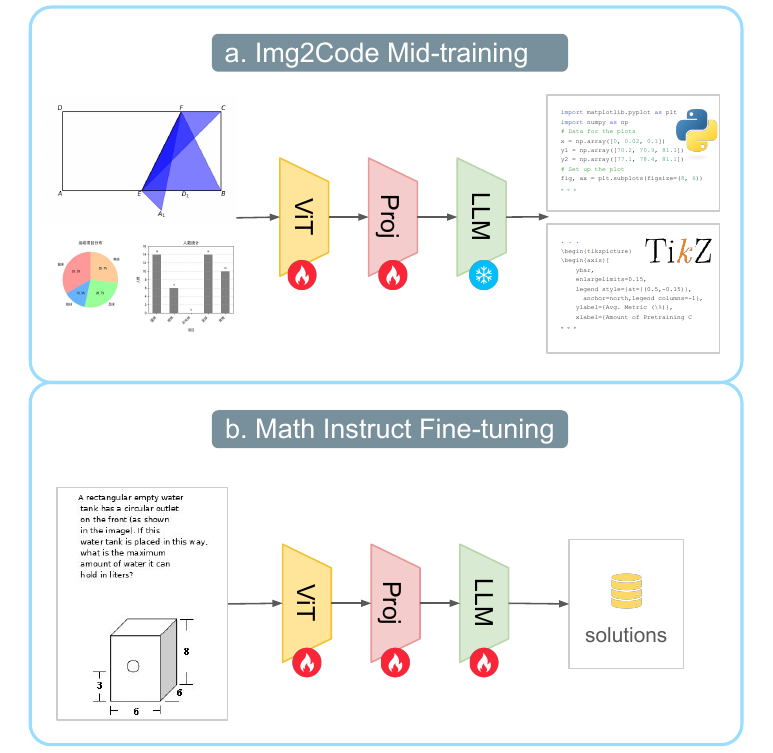}
  \vspace{-6mm}
  \caption{Two training stages of MathCoder-VL.}
  \label{fig:training_pipeline}
\end{figure}
\vspace{-2mm}

\subsection{Math Instruction Fine-tuning Data}
\label{sec:math_problem_solving}
In this section, we introduce the construction of our \mathdatasetname~as shown in Figure~\ref{fig:data_pipeline} (b).

\subsubsection{Construction of K12-2M Dataset}
\label{sec:k12}
We collected 4.6 million math problems with simple solutions, where the equations are in image format. 
First, we distinguish math figures from equations based on their size, as equations tend to be much smaller. Next, we convert the equations into \LaTeX~text using MinerU~\citep{wang2024mineru}. This process results in 2 million samples containing at least one actual image. 
To enhance data quality, we then use GPT-4o mini to translate the original simple solutions into detailed, step-by-step CoT solutions, ultimately resulting in K12-2M.

\subsubsection{Synthetic Math Data with New Images}
\label{sce:new_k12}

To generate new multi-modal math problems, we follow a structured approach:

\textbf{Newly Synthesized Images.}
We leverage the 1.57 million raw images from K12-2M, using our \icmodelname~with a temperature of 0.7 to generate new math figures. With a higher temperature, the model can produce images that diverge more from the raw dataset. More examples of the newly synthesized images are shown in Appendix~\ref{app:new_images}.

\textbf{Questions Based on New Images.}
From the 1.1 million newly generated image-code pairs, we use Qwen2.5-72B-Instruct~\citep{qwen2_5} to craft math reasoning questions appropriate for a K12 audience. These questions are based on the visual elements (such as patterns, shapes, and numbers) present in each image. The questions are designed to be concise, self-contained, and to engage the reasoning skills of the reader. At this stage, the model is not required to provide answers to the questions. Details can be found in Appendix~\ref{app:new_problems}.

\textbf{Synthesize Solutions.}
For generating solutions, we employ both Qwen2.5-Math-72B-Instruct~\citep{yang2024qwen25math} and Qwen2.5-72B-Instruct~\citep{qwen2_5}. Each model independently attempts to solve the question, taking both the question and image code as inputs. We retain a solution only if both models produce consistent answers, assuming that there is typically one correct answer and multiple possible incorrect ones. The solution pass rate is 51\%. Following the data cleaning procedure outlined in Section~\ref{sec:imd2code_model:data_clean}, we remove duplicates and overly long samples. The final output consists of 1 million new samples, some of which are illustrated in Figure~\ref{fig:training_samples}.

%% file: 03-Experiments.tex
\input{tables/main_model_performance}

\input{tables/mathvision_geometry_performance}
\input{tables/ablation}
\input{tables/ablation_midtraining}

\section{Experiments}
\label{sec:experiments}
In this section, we introduce our two-stage training approach: image-to-code mid-training with \icdatasetname, followed by math instruction fine-tuning with \mathdatasetname.

\subsection{Training Stages}
As illustrated in Figure~\ref{fig:training_pipeline}, the training process for a single MathCoder-VL model consists of two stages aimed at improving the model’s math-related visual perception and multimodal reasoning capabilities.

\textbf{Image-to-Code Mid-training.}  
In this stage, we use \icdatasetname~to improve cross-modal alignment between mathematical diagrams and language embedding spaces. Both the vision encoder and MLP projector are trainable during this phase. The primary objective is to enhance the vision encoder’s ability to extract mathematic visual features. Since the correspondence between code and image is highly accurate and contains all the detailed information, this stage allows the model to capture intricate patterns, especially those related to mathematics. These math-related patterns, including geometric shapes, process flows, and other mathematical representations, are underrepresented in large web-scale datasets like LAION-5B~\citep{schuhmann2022laion5b}. Importantly, we freeze the LLM backbone during this stage to preserve its general language abilities, as we do not require it to generate code for downstream tasks.

\textbf{Math Instruction Fine-tuning.}  
In this stage, as shown in Figure~\ref{fig:training_pipeline}, the entire model is fine-tuned on our high-quality multimodal math problem-solving dataset, \mathdatasetname. This dataset includes 3 million samples, with 1 million generated by our image-to-code model-based data engine. To the best of our knowledge, this is the first data engine capable of generating multimodal math problem-solving data that includes not only new textual content but also new diverse math figures.

\subsection{Experimental Setup}
We use InternVL-Chat-2B-V1-5~\citep{gao2024internvlmini} and InternVL2-8B~\citep{chen2024internvl2} as the base models for our experiments.

\textbf{Implementation Details.}
We train the model for one epoch across two stages. In the first stage, we use a batch size of 1024 and a learning rate of 2e-5. In the second stage, we use a batch size of 512 and a learning rate of 4e-5. To efficiently train the computationally intensive models, we utilize DeepSpeed at ZeRO-1 stage~\citep{rajbhandari2020zero} and flash attention~\citep{dao2022flashattention}. The 2B and 8B models are trained on 32 and 64 NVIDIA A800 80GB GPUs, respectively. To ensure reproducibility, we fix the random seed and employ greedy decoding during testing.

\textbf{Benchmarks.}
We assess our models across a diverse set of widely recognized mathematical benchmarks. The MATH-Vision~\citep{wang2024mathvision} dataset includes 3,040 visually contextualized math problems sourced from real-world competitions. MathVista~\citep{lu2023mathvista} is a well-known dataset designed for evaluating reasoning in visual contexts. MathVerse~\citep{zhang2025mathverse} emphasizes core mathematical skills such as plane geometry, solid geometry, and functions. GAOKAO-MM~\citep{zong2024gaokaomm} is based on the Chinese College Entrance Examination. Many tasks in MathVista require more emphasis on natural image recognition rather than math reasoning abilities~\citep{wang2024mathvision}, so we only report results on the Geometry Problem Solving (GPS) subset. Collectively, these datasets cover a wide spectrum of mathematical challenges, ranging from elementary word problems to advanced college-level exercises in both English and Chinese, providing a comprehensive evaluation of model capabilities.

\textbf{Baselines.}
We compare our approach against a range of base models with strong mathematical capabilities and similar sizes. Our selected baselines include both closed-source and open-source LMMs. Both general LMMs and math-focused LMMs are incorporated. For general LMMs, we include powerful models like GPT-4o~\citep{openai2024gpt4o}, Qwen2-VL~\citep{wang2024qwen2vl} and IXC-2.5-Reward~\citep{zang2025internlmxcomposer25reward}. For math-focused LMMs, we choose recent models such as MathGLM-Vision~\citep{yang2024mathglmvision}, Math-PUMA~\citep{zhuang2024mathpuma}, Multimath~\citep{peng2024multimath}, and MAVIS~\citep{zhang2025mavis}.

\subsection{Main Results}
We evaluate MathCoder-VL across several benchmarks, analyzing its performance from the perspectives of mathematical subjects and input modalities.

\textbf{Overall Performances.}
As shown in Table~\ref{tab:main_model_performance}, MathCoder-VL demonstrates strong performance across multiple mathematical benchmarks, particularly in comparison to other open-source models. MathCoder-VL-8B achieves the highest accuracy among open-source LMMs of similar sizes, with 26.1\% on MATH-Vision, 46.5\% on MathVerse, and an impressive 73.6\% on the MathVista (GPS). These results show a notable improvement over its base model, InternVL2-8B, by 6.1\%, 10.6\%, and 11.6\% on the respective benchmarks. The smaller model also demonstrates strong capabilities, with MathCoder-VL-2B outperforming MathGLM-Vision-9B by 2.5\% and Multimath-7B by 5.4\% on MATH-Vision. MathCoder-VL-8B significantly outperforms InternVL2-76B, with a gap of 2.5\% on MATH-Vision, 3.7\% on MathVerse, 5.8\% on MathVista (GPS), and 10.0\% on GAOKAO-MM Math. The model's performance in Chinese is also noteworthy, with MathCoder-VL-8B reaching 51.2\% on GAOKAO-MM, outperforming all other open-source LMMs.

Compared to closed-source models, MathCoder-VL-8B remains competitive, outperforming several proprietary models. It surpasses GPT-4V on all four benchmarks and exceeds GPT-4-turbo by 3.0\% on MathVerse. It also outperforms the newest Claude3.5-Sonnet (64.4\% vs 73.6\%) and GPT-4o (64.7\% vs 73.6\%) on MathVista (GPS). However, it still falls short of top-tier closed-source LMMs in some areas. For example, it lags behind GPT-4o by 3.0\% on MATH-Vision.

\textbf{Performance on multi-step problems.}
MathCoder-VL-8B exhibits robust performance on multi-step problems, outperforming GPT-4o on both two-step (58.6\% vs 58.1\%) and three-step problems (52.1\% vs 43.6\%) on We-Math~\citep{qiao2024wemath}. Our \mathdatasetname, which provides step-by-step solutions for every problem, enhances the model's Chain-of-Thought~\citep{wei2022chain} reasoning ability. Notably, MathCoder-VL-8B surpasses InternVL2-76B by a significant margin, achieving a 20.7\% improvement on two-step problems and a 3.0\% improvement on three-step problems, while only slightly edging it out by 0.2\% on one-step problems. This demonstrates that, as a math-specific language model, MathCoder-VL excels over general open-source models, particularly on complex problems.

\textbf{Outstanding Ability in Geometry.}
When evaluating MathCoder-VL's capabilities in geometry, its performance on the MathVista (GPS) stands out. Additionally, we present the detailed accuracy of the model on the plane geometry subsets from MATH-Vision, as shown in Table~\ref{tab:mathvision_geometry_performance}. MathCoder-VL excels across all three plane geometry subsets in MATH-V, achieving an impressive average score of 37.6\%, which surpasses GPT-4o by 11.9\%. Notably, the model scored exceptionally well in each of the three subsets—angle, area, and length—with scores of 48.6\%, 32.2\%, and 32.1\%, respectively. This superior performance can be attributed to MathCoder-VL's enhanced understanding of geometry figures, enabling it to effectively process and interpret geometric shapes and measurements.

\subsection{Ablation Study}
In this session, we analyze the impact of various components of the training pipeline.

\textbf{Ablation on Impact of Image-to-Code Mid-training.}
From Table~\ref{tab:ablation}, we can observe the impact of image-to-code mid-training on the model's reasoning ability. Comparing the results without mid-training to those with mid-training, performance improvements are noted in MATH-Vision (+1.7\%), MathVerse (+5.8\%), MathVista (GPS) (+18.7\%), and GAOKAO-MM (Math) (+3.8\%), highlighting its contribution to enhanced multi-modal mathematical reasoning. The most significant gain is observed in MathVista (GPS), suggesting that image-to-code mid-training strengthens spatial and graphical problem-solving capabilities and improves understanding of geometry figures.

\textbf{Ablation on Impact of Input Modality.}
Table~\ref{tab:ablation_midtraining} illustrates the impact of image-to-code mid-training on MathVerse across different modality dominance levels: Text-Dominant (TD), Text-Lite (TL), Vision-Dominant (VD), and Vision-Only (VO). Across all categories, mid-training with image-to-code leads to improved performance, with an overall gain of 5.8\% and 4.6\%. Notably, the largest improvement is seen in the VO setting, where performance increases by 11.0\% and 7.4\%, indicating that image-to-code mid-training significantly enhances the model's ability to process purely visual inputs, while the smallest improvements are observed in TD (+3.8\%) and TL (+3.0\%). This suggests that image-to-code mid-training effectively enhances multi-modal reasoning, particularly in scenarios where vision plays a more dominant role.

\textbf{Ablation on Impact of Newly Synthesized Images.}
As shown in Table~\ref{tab:ablation}, the MathCoder-VL-2B model generally benefits from the math instruction fine-tuning dataset based on newly synthesized images. Performance improvements are observed across multiple benchmarks: MathVerse (+2.4\%), MathVista-Testmini (+5.0\%), MathVista-GPS (+2.0\%), and GAOKAO-MM (Math) (+3.7\%), with only a slight decrease on MATH-Vision of 0.3\%. Notably, MathVista shows a significant increase of 5.0\%, suggesting that the new synthetic math problems contribute to a broader diversity of instructions. This enhanced diversity likely improves the model's generalization capabilities, particularly as many tasks in MathVista differ substantially from traditional math problem-solving.
 

%% file: tables/main_model_performance.tex
\begin{table*}[ht]
\centering
\resizebox{\textwidth}{!}{%
\small
\begin{tabular}{l|c|c|c|c|c|c|c|c}
\toprule
\multirow{2}{*}{Model}& \multirow{2}{*}{\#Params}           &     MATH-Vision   &   MathVerse          & MathVista        & GAOKAO-MM    &     \multicolumn{3}{c}{We-Math}             \\
                 &                                          &        (Test)     &   (Testmini)         &  (GPS)           & (Math)       &    (S1)     &    (S2)     &      (S3)       \\
\toprule
Random Chance    &                                        - &          7.2      &   12.4               &  21.6            &     -        &     -       &     -       &      -          \\
Human            &                                        - &         68.8      &   64.9               &  48.4            &     -        &     -       &     -       &      -          \\
\midrule

\multicolumn{9}{c}{Closed-source LMMs}\\
\midrule
Qwen-VL-Plus~\citep{QwenVL}       &                       - &          10.8      &   21.3               &  35.5           &  33.8        &             &             &                 \\
Qwen-VL-Max~\citep{QwenVL} &                              - &          15.6      &   35.9               &  46.1           &     -        &    40.8     &    30.3     &    20.6         \\
GPT-4V~\citep{openai2023gpt4v}    &                       - &          22.8      &   39.4               &  50.5           &  45.0        &    65.5     &    49.2     &    38.2         \\
GPT-4-turbo~\citep{openai2024gpt4technicalreport}&        - &          30.3      &   43.5               &  58.3           &  \best{50.0} &     -       &     -       &      -          \\
GPT-4o~\citep{openai2024gpt4o}&                           - &          30.4      &   50.8               &  \best{64.7}    &     -        & \best{72.8} & \best{58.1} &  \best{43.6}     \\
Claude3-Opus~\citep{claude3}&                             - &          27.1      &   31.8               &  52.9           &     -        &     -       &     -       &      -          \\
Claude3.5-Sonnet~\citep{claude3}&                         - &    \best{37.9}     &   49.0               &  64.4           &     -        &     -       &     -       &      -          \\
Gemini-1.5-Pro~\citep{geminiteam2024gemini1_5}&           - &          19.2      &   \best{51.1}        &  58.9           &     -        &    56.1     &    51.4     &    33.9         \\
\midrule  

\multicolumn{9}{c}{Open-source LMMs}\\
\midrule
LLaVA-1.5-13B~\citep{liu2024llava1_5}&                  13B &          11.0      &   12.7               &   22.7          &  16.3        &    35.4     &    30.0     &    32.7         \\
SPHINX-V2-13B~\citep{lin2023sphinx}&                    13B &             -      &   16.1               &   16.4          &     -        &     -       &     -       &     -           \\
IXC-2-VL~\citep{dong2024internlmxc2vl}&    7B &          14.5      &   25.9               &   63.0          &     -        &    47.0     &    33.1     &    33.3         \\
Deepseek-VL~\citep{lu2024deepseek}&                      8B &             -      &   19.3               &   28.4          &  20.0        &    32.6     &    26.7     &    25.5         \\
Qwen2-VL~\citep{wang2024qwen2vl}&                        8B &          19.2      &   33.6               &   40.9          &  25.0        &    59.1     &    43.6     &    26.7         \\
InternVL-Chat-2B-V1-5~\citep{gao2024internvlmini}&  2B &          15.3      &   23.1               &   37.5          &  17.5        &    34.3     &    26.1     &    20.0         \\
InternVL2-8B~\citep{chen2024internvl2}&                  8B &          20.0      &   35.9               &   62.0          &  32.5        &    59.4     &    43.6     &    35.2         \\
InternVL2-26B~\citep{chen2024internvl2}&                26B &          23.1      &   40.0               &   54.3          &  33.4        &    51.0     &    39.2     &    46.1         \\
InternVL2-76B~\citep{chen2024internvl2}&                76B &  \third{23.6}      &   42.8               &   \third{67.8}  & \third{41.2} &\third{65.2} & \third{49.4}& \third{49.1}    \\
IXC-2.5-Reward~\citep{zang2025internlmxcomposer25reward}&7B&  19.0  &   18.8               &   63.5          &     -        &    44.4     &    35.3     &    27.9         \\
\midrule
\multicolumn{9}{c}{Open-source Math LMMs}\\
\midrule
G-LLaVA-7B~\citep{gao2023gllava}&                        7B &             -      &   16.6               &   48.7          &     -        &    32.4     &    30.6     &    32.7         \\
Math-LLaVA-13B~\citep{shi2024mathllava}&                13B &          15.7      &   22.9               &   57.7          &     -        &    38.7     &    34.2     &    34.6         \\
InfiMM-Math~\citep{han24infimm}&                         7B &             -      &   34.5               &   -             &     -        &     -       &     -       &     -           \\
MathGLM-Vision-9B~\citep{yang2024mathglmvision}&         9B &          19.2      &  \third{44.2}        &   64.4          &     -        &     -       &     -       &     -           \\
Math-PUMA-Qwen2~\citep{zhuang2024mathpuma}&           8B &          14.0      &   33.6               &   48.1          &     -        &    53.3     &    39.4     &    36.4         \\
Math-PUMA-DS~\citep{zhuang2024mathpuma}&           7B &             -      &   31.8               &   39.9          &     -        &    45.6     &    38.1     &    33.9         \\
Multimath-7B~\citep{peng2024multimath}          &        7B &          16.3      &   27.7               &   66.8          &     -        &     -       &     -       &     -           \\
MAVIS-7B~\citep{zhang2025mavis}&                         7B &          19.2      &   35.2               &   64.1          &     -        &    57.2     &    37.9     &    34.6         \\
\midrule
MathCoder-VL-2B   &                                      2B &          21.7      &   35.4               &   66.4          & 37.5         &    52.0     &    42.2     &    38.8         \\
\textit{$\Delta$ Over Base Model}&                          &     \up{6.4}       &   \up{12.3}          &   \up{28.9}     & \up{20.0}    &  \up{17.7}  &  \up{16.1}  &  \up{18.8}      \\
\midrule
MathCoder-VL-8B   &                                      8B &   \second{26.1}    &   \second{46.5}      &  \second{73.6}  & \second{51.2}&\second{65.4}&\second{58.6}&\second{52.1}    \\
\textit{$\Delta$ Over Base Model}&                          &     \up{6.1}       &    \up{10.6}          &  \up{11.6}      & \up{18.7}    &  \up{6.0}   &  \up{15.0}  &  \up{16.9}      \\
\bottomrule
\end{tabular}%
}
\caption{Comparison of model performances across various math benchmarks. MATH-Vision~\citep{wang2024mathvision}, MathVerse~\citep{zhang2025mathverse}, MathVista~\citep{lu2023mathvista}, and We-Math~\citep{qiao2024wemath} are in English, while GAOKAO-MM~\citep{zong2024gaokaomm} is in Chinese. The best results of closed-source LMMs are highlighted in \colorbox{wkred}{red}. The best and second-best results of open-source LMMs are highlighted in \colorbox{wkblue}{blue} and \colorbox{wkgreen}{green} respectively. (GPS: geometry problem solving, S1: one-step problems, S2: two-step problems, S3: three-step problems)}
\label{tab:main_model_performance}
\end{table*}

%% file: tables/mathvision_geometry_performance.tex
\begin{table}[ht]
  \centering
  \small
  \begin{tabular}{l|ccc|c}
    \toprule
    \multirow{2}{*}{Model} & \multicolumn{3}{c|}{MATH-V Geometry} & \multirow{2}{*}{Average}       \\
                           &   angle     &  area       &  length     &                                     \\
    \toprule
                             \multicolumn{5}{c}{Closed-source LMMs}                               \\
    \midrule
    GPT-4o                 &   17.3      &\second{29.8}&\second{30.1}&\second{25.7}                              \\
    GPT-4V                 &\second{22.0}&   22.2      &   20.9      &   21.7                              \\
    Gemini-1.5-Pro         &   14.5      &   14.4      &   16.5      &   15.1                              \\
    \midrule
                             \multicolumn{5}{c}{Open-source LMMs}                                 \\
    \midrule
    Qwen2-VL-8B            &   19.1      &   22.4      &   22.5      &   21.3                              \\
    InternVL2-8B           &   20.8      &   22.4      &   20.5      &   21.2                              \\
    InternVL2.5-8B         &   22.0      &   19.4      &   15.4      &   18.9                              \\
    \midrule
                             \multicolumn{5}{c}{Open-source Math LMMs}                            \\
    \midrule
    Math-LLaVA-13B         &   20.2      &   18.4      &   17.6      &   18.7                              \\
    Multimath-7B           &   20.1      &   16.4      &   21.3      &   19.3                              \\
    Math-PUMA-8B           &   11.7      &   15.8      &   12.2      &   13.2                              \\
    \midrule
    MathCoder-VL-8B         &\best{48.6}&\best{32.2}&\best{32.1}&\best{37.6}                              \\
    \bottomrule
  \end{tabular}
  \caption{Comparison of model performances on the three plane geometry subsets of MATH-Vision~\citep{wang2024mathvision}. The best and second-best results are highlighted in \colorbox{wkred}{red} and \colorbox{wkblue}{blue} respectively.}
  \label{tab:mathvision_geometry_performance}
\end{table}

%% file: tables/ablation.tex
\begin{table*}[ht]
\centering
\resizebox{\textwidth}{!}{%
\small
\begin{tabular}{l|c|c|c|c|c|c|c}
\toprule
\multirow{2}{*}{Model}&        Image-to-Code       &       Math  Instruction     &     MATH-Vision   &   MathVerse    & MathVista     & MathVista    & GAOKAO-MM        \\
                  &           Mid-training      &         Fine-tuning      &        (Test)     &   (Testmini)   & (Testmini)    &  (GPS)       & (Math)           \\
\toprule 
InternVL-Chat-2B-V1-5&       \xmark        &         \xmark           &         15.3      &   23.1         &     41.1      &   37.5       &   17.5           \\
\midrule
                  &   \multirow{2}{*}{\xmark}   &  \multirow{2}{*}{K12-2M} &         20.3      &   27.2         &     37.0      &   45.7       &   30.0           \\
                  &    &                                                   &        \up{5.0}   &   \up{4.1}     &     \down{4.1}  &   \up{8.2}   &   \up{12.5}    \\
\midrule
                  & \multirow{2}{*}{\checkmark} &  \multirow{2}{*}{K12-2M} &         22.0      &   33.0         &     39.4      &   64.4       &   33.8           \\
                  &    &                                                   &        \up{1.7}   &   \up{5.8}     &     \up{2.4}  &   \up{18.7}   &   \up{3.8}      \\
\midrule
\multirow{2}{*}{MathCoder-VL-2B}
                  & \multirow{2}{*}{\checkmark} &                {K12-2M } &   21.7    &   35.4         &     44.4      &   66.4       &   37.5   \\
                  &    &         + New-1M                                  &     \down{0.3}    &   \up{2.4}     &     \up{5.0}  &   \up{2.0}   &   \up{3.7}     \\
\bottomrule
\end{tabular}%
}
\caption{Ablation study of image-to-code mid-training and math instruction fine-tuning dataset on MathCoder-VL-2B. K12-2M + New-1M dataset is our \mathdatasetname.}
\label{tab:ablation}
\end{table*}

%% file: tables/ablation_midtraining.tex
\begin{table}[ht]
  \centering
\small
\begin{tabular}{c|c|cc|cc|c}
\toprule
 {Mid-} & {Fine-}&   \multicolumn{5}{c}{MathVerse} \\
 training & tuning &  TD      &  TL      &  VD      &  VO      &  All         \\
\toprule
      \xmark     &    \xmark    & 27.5     & 25.8     & 20.1     & 18.1     & 23.1         \\
\midrule
      \xmark     &     2M       & 36.7     & 30.7     & 25.3     & 15.9     & 27.2         \\
      \checkmark &     2M       & 40.9     & 34.5     & 31.1     & 26.9     & 33.0         \\
                 &              & \up{4.2} & \up{3.8} & \up{5.8} & \up{11.0}& \up{5.8}     \\
\midrule
      \xmark     &     2M+1M    & 40.7     & 32.4     & 30.1     & 19.8     & 30.8         \\
      \checkmark &     2M+1M    & 43.7     & 36.9     & 34.1     & 27.2     & 35.4         \\
                 &              & \up{3.0} & \up{4.5} & \up{4.0} & \up{7.4} & \up{4.6}     \\
\bottomrule
\end{tabular}%
  \caption{Effects of image-to-code mid-training on model performances with varying degrees of input content in multi-modality on MathVerse~\citep{zhang2025mathverse}.}
  \label{tab:ablation_midtraining}
\end{table}


%% file: 05-Conclusion.tex
\section{Conclusion}
\label{sec:conclusion}

In this paper, we propose a model-based multi-modal data engine. Using this data engine, we construct two datasets: \icdatasetname~for accurate cross-modal alignment and \mathdatasetname, a math problem-solving dataset featuring diverse newly synthesized images. Leveraging these datasets, we develop MathCoder-VL-2B and 8B models trained with image-to-code mid-training and math instruction fine-tuning. MathCoder-VL achieves a new state-of-the-art among open-source models for multi-modal mathematical reasoning.

%% file: 06-Limitations.tex
\section{Limitations}
\label{sec:limitations}

One limitation of our work is that \mathdatasetname~focuses primarily on mathematics and does not intentionally include other STEM subjects, such as physics and chemistry. Additionally, our dataset consists entirely of English text and does not incorporate math-related content in other languages, such as Chinese. Due to computational resource constraints, we only trained 2B and 8B models. Future work could address these limitations by expanding the dataset to include other subjects and languages and by training larger language models. Furthermore, this paper primarily focuses on image-to-code mid-training and math instruction fine-tuning, so we did not apply reinforcement learning methods, such as GRPO, in the post-training phase, which could further improve performance on mathematical reasoning tasks. In the future, we plan to explore these methods with MathCoder-VL.

%% file: Appendix.tex
\label{sec:appendix}

\section{Details of K12 Data Process}
\label{app:data_clean}
\begin{figure*}[htbp]
  \includegraphics[width=0.99\linewidth]{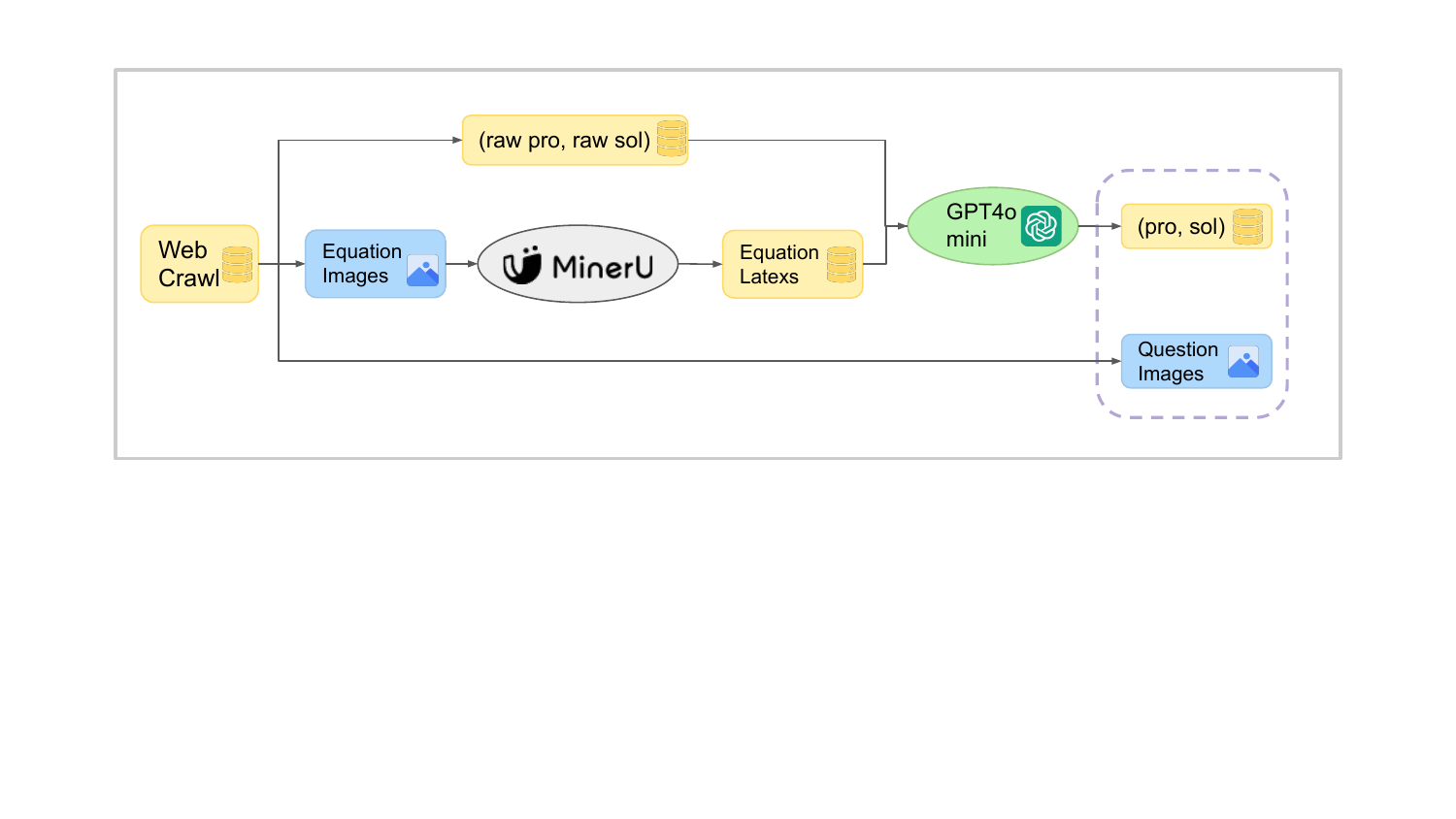}
  \vspace{-36mm}
  \caption{The pipeline for processing the K12 math problem-solving dataset.} 
  \label{fig:k12_process_pipeline}
\end{figure*}

In this section, we provide additional details about processing the newly collected K12 math problem-solving dataset. The overall pipeline for data processing is illustrated in Figure~\ref{fig:k12_process_pipeline}.

\subsection{Data Cleaning}
\begin{figure}[htbp]
  \includegraphics[width=0.99\linewidth]{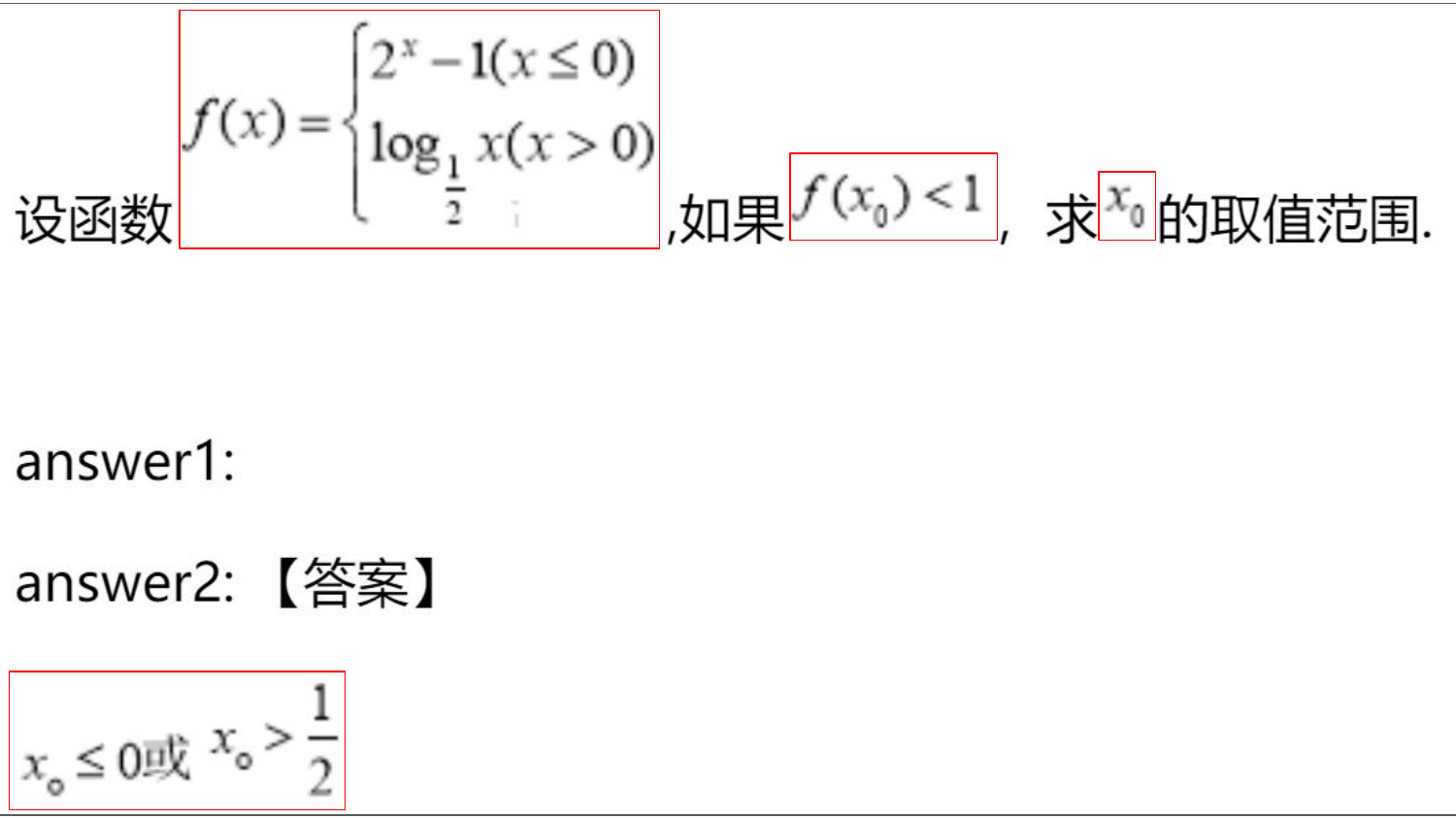}
  \caption{Example of a raw math problem that only contains equation images. Such problems are filtered out after converting the images into \LaTeX~equations using MinerU.} 
  \label{fig:problem_with_ony_equation_images}
\end{figure}

The primary objective of the data cleaning process is to curate a dataset that consists exclusively of multi-modal math problems. These problems should include both textual descriptions and mathematical expressions represented in \LaTeX~code and math figures. In the raw dataset, a significant number of equations were provided solely as images, as shown in Figure~\ref{fig:problem_with_ony_equation_images}. To address this, we employed the MinerU tool to convert these equation images into LaTex-formatted equations, ensuring a consistent and standardized representation of mathematical content. Furthermore, problems that contained only equation images are excluded from the dataset. This cleaning process ensures that the final dataset is rich, diverse, and appropriately structured for addressing K12 math problems that require multi-modal reasoning.

\subsection{Data Augmentation}

\begin{figure*}[htbp]
  \includegraphics[width=0.99\linewidth]{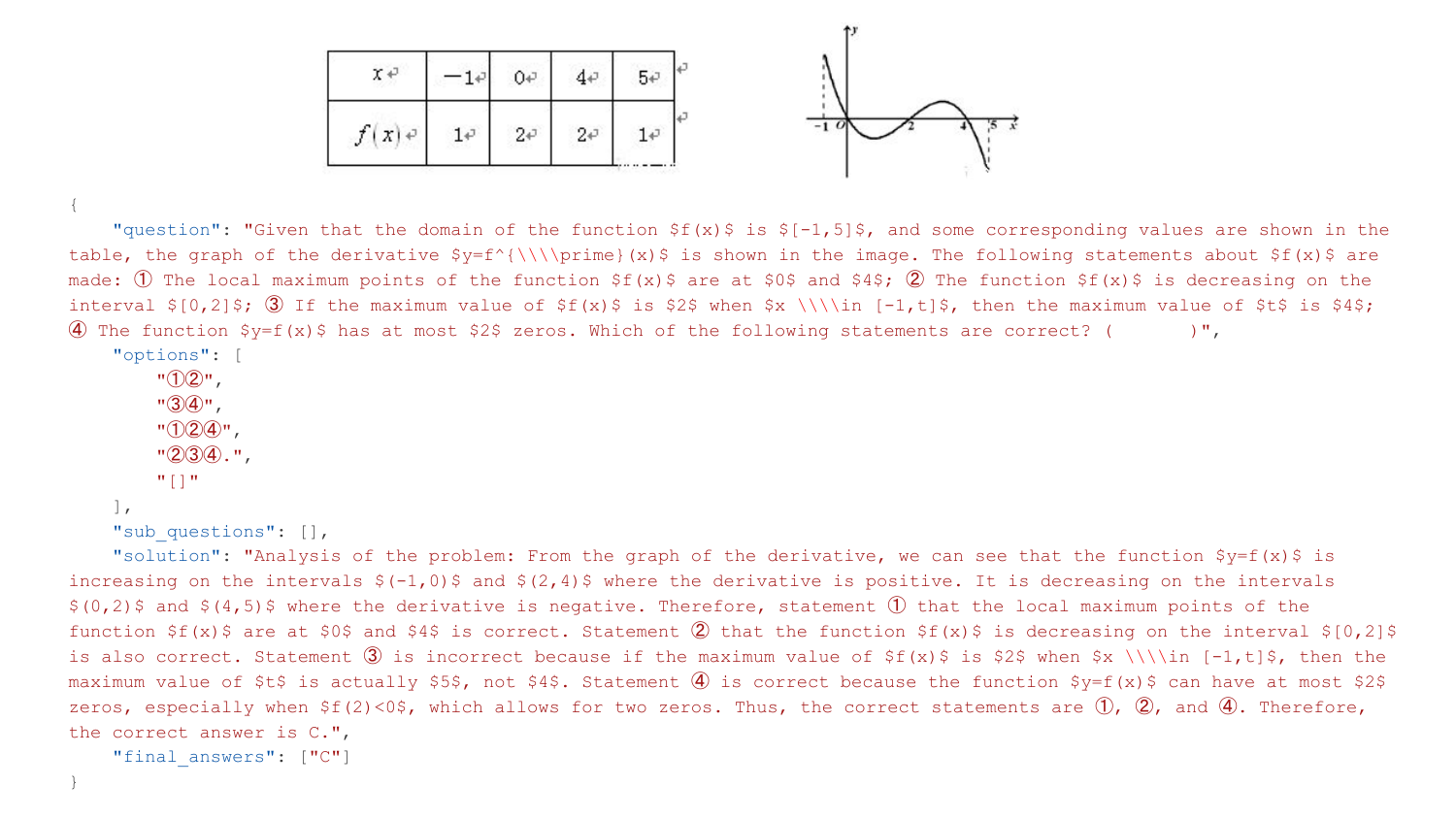}
  \caption{Example output generated by GPT4o-mini, showcasing the translation, step-by-step solution, and short answer extraction for a given math problem.} 
  \label{fig:k12_gpt4omini_out_example}
\end{figure*}

Figure~\ref{fig:prompt_translation} presents a structured system prompt designed for processing K-12 mathematical problems. It outlines a comprehensive workflow for translating, solving, and formatting math problems from a JSON object. The prompt includes explicit instructions for translation into English, step-by-step solution generation, and concise answer presentation, ensuring clarity and correctness in the output. One example of the GPT4o-mini's output is shown in Figure~\ref{fig:k12_gpt4omini_out_example}.

\input{appendix/k12_translate_prompts}

\section{Details of Image-to-Code}
\label{app:img2code}

\subsection{Code Ability}
TikZ is a powerful and flexible package for creating vector graphics in \LaTeX. It is based on the PGF (Portable Graphics Format) system and is known for its high-quality output and extensive customization options. TikZ allows users to create a wide range of graphics, from simple shapes and diagrams to complex illustrations and plots. Its strength lies in its ability to seamlessly integrate with \LaTeX~documents, ensuring that the fonts, styles, and layout of the graphics match the document's overall design. TikZ is particularly useful for creating precise, technical illustrations, flowcharts, and scientific figures. The syntax of TikZ is based on a series of commands that define paths, nodes, and styles, making it highly programmable and suitable for generating graphics algorithmically. Some examples of images generated by TikZ are shown in Figure~\ref{fig:tikz_examples}.
\input{appendix/python_tikz_intro}

On the other hand, Matplotlib is a popular plotting library in Python that provides a wide range of tools for creating static, animated, and interactive visualizations. It is widely used in scientific computing, data analysis, and machine learning for generating publication-quality figures. Matplotlib supports various types of plots, including line plots, scatter plots, bar charts, histograms, and more. One of its key strengths is its flexibility and ease of use, allowing users to quickly generate visualizations with a few lines of code. Matplotlib also offers extensive customization options, enabling users to adjust every aspect of a plot, from line styles and colors to axis labels and legends. Additionally, it integrates well with other Python libraries such as NumPy and Pandas, making it a versatile tool for data visualization in the Python ecosystem. Some examples of images generated by Python are shown in Figure~\ref{fig:python_examples}.

When comparing the syntax of Python's Matplotlib and \LaTeX's TikZ for creating plots and graphics, the differences are quite pronounced as shown in Figure~\ref{fig:python_tikz_code}. Matplotlib, being a Python library, follows a procedural programming style, where functions are called to add elements to a plot. In contrast, TikZ, which is part of the \LaTeX~ecosystem, uses a declarative style, where user describe the elements of the graphic in a more structured, often nested, manner. While Matplotlib's syntax is more straightforward and easier to learn for those familiar with Python, TikZ offers greater control over the visual details of the plot, making it a preferred choice for complex, publication-quality graphics.

\subsection{Prompt Templates}
To facilitate the generation of code from images, we designed two structured prompt templates that guide the process of converting visual elements into executable code as shown in Figure\ref{fig:i2c_prompt_templates}.

\input{appendix/img2code_prompts}

\subsection{TikZ to Python}
To enhance the capabilities of our image-to-code model, we use GPT4o-mini to translate TikZ code into Python code. Figure~\ref{fig:tikz2python_prompt} illustrates the detailed prompt used for this translation. The prompt in Figure~\ref{fig:tikz2python_prompt} is designed to guide the conversion of \LaTeX~TikZ code into Python code using popular plotting libraries like Matplotlib. It ensures that the resulting Python code is executable, accurately reproduces the visual details of the TikZ diagram, and avoids overlaps between elements such as points, labels, and text for better readability. The prompt also emphasizes the correct formatting of \LaTeX~mathematical expressions to maintain visual clarity and precision in the generated plots. This structured approach helps bridge the gap between \LaTeX-based graphics and Python-based visualization.

\input{appendix/tikz2python_prompt}

\input{appendix/tikz2python_examples}

In Figure~\ref{fig:tikz2python_examples}, we compare images generated from the original TikZ code with those generated from the translated Python code. The results demonstrate that the images produced by the Python code are highly similar to the original images, showcasing the effectiveness of our translation approach.

\subsection{Data Cleaning}

\begin{figure*}[htbp]
  \includegraphics[width=0.99\linewidth]{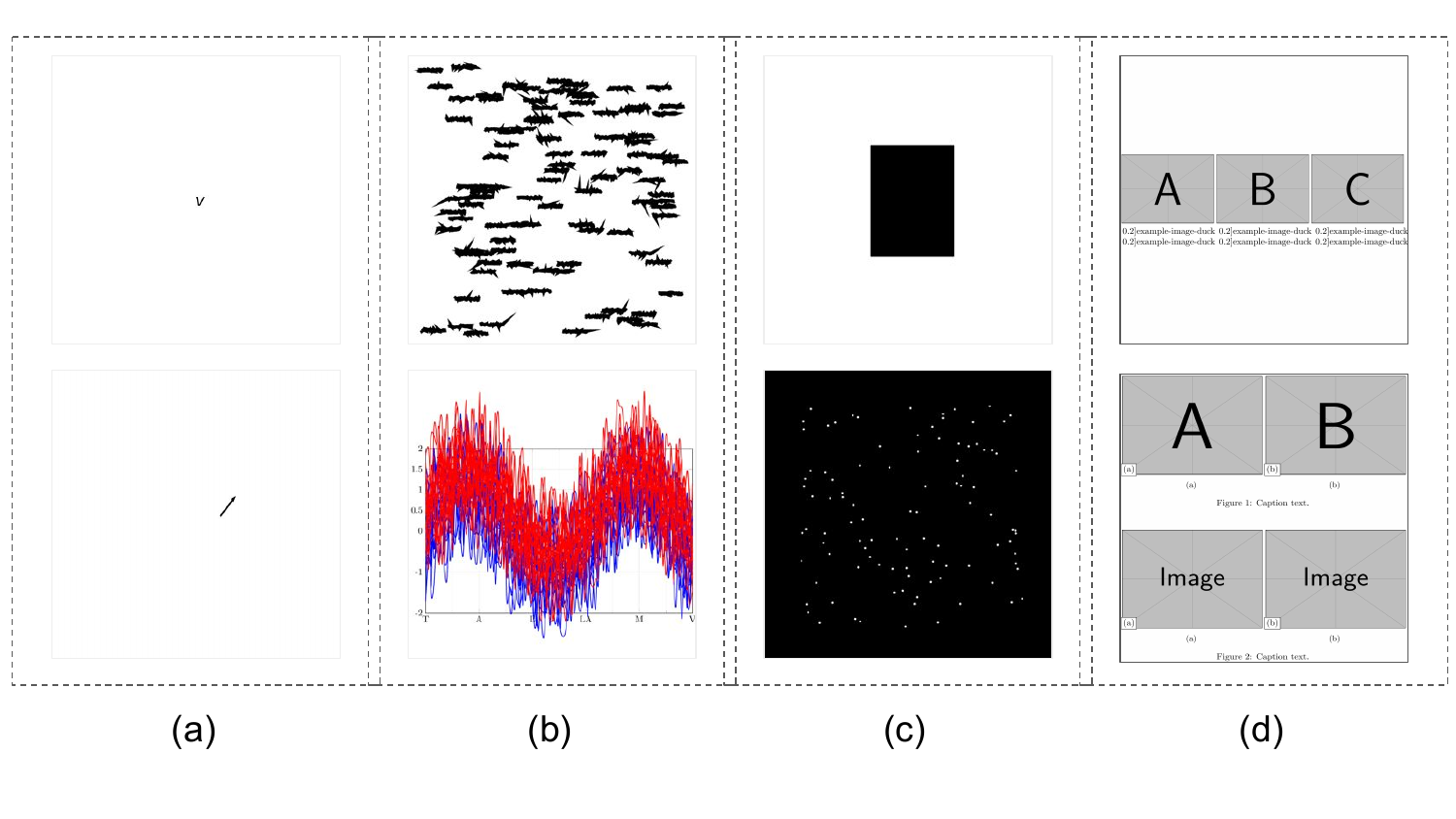}
  \vspace{-5mm}
  \caption{Examples of low-quality image-code pairs removed from the dataset. (a) Almost blank images with very low pixel variation. (b) Images containing random lines or shapes. (c) Images with black squares caused by incorrect preprocessing. (d) Images generated using external files accessed through the code.} 
  \label{fig:data_clean_examples}
\end{figure*}

We remove low-quality image-code pairs from our dataset. Figure~\ref{fig:data_clean_examples} illustrates four types of low-quality samples: 
(a) Almost blank images: We remove images with a standard deviation (std) of pixel values less than five.  
(b) Images with random lines or shapes: These are filtered out by analyzing and filtering the corresponding code.  
(c) Images with black squares: This issue arises when images with blank backgrounds are converted incorrectly during preprocessing, resulting in completely black images. We addressed this by removing the affected data and optimizing the conversion logic.  
(d) Images with externally loaded content: We identify and remove such data by detecting commands in the code that access local files.

\subsection{Performance of img2code model}

The img2code model aims to bridge the gap between visual data and code generation by translating images into accurate and meaningful code representations. This section evaluates the model's progression through iterative training and highlights its ability to synthesize new, diverse images. By comparing the performance of the initial and final versions of the model and exploring its capabilities with high-temperature synthesis, we demonstrate its advancements in accuracy and creative output.

\textbf{Comparison Between Initial and Final Models.}
Our img2code model was trained iteratively, culminating in a final version trained on 8.6 million image-code pairs. The performance improvements from the initial to the final model are demonstrated in Figures~\ref{fig:img2code_examples_1}, \ref{fig:img2code_examples_2}, \ref{fig:img2code_examples_3}, and \ref{fig:img2code_examples_4}. These figures highlight the significant advancements in accuracy and the quality of the generated code and corresponding images as the model evolved through successive training cycles.

\textbf{Synthesize New Images with High Temperature.}
\label{app:new_images}
Using the final iteration of the Img2Code-8B model, we synthesized new images from 1.57 million raw images in the foundational dataset. By setting a temperature of 0.7, the model was able to generate more diverse and creative outputs, deviating meaningfully from the original dataset. The results of this high-temperature synthesis are illustrated in Figures~\ref{new_images_from_k12_examples_1of3}, \ref{new_images_from_k12_examples_2of3}, \ref{new_images_from_k12_examples_3of3}, \ref{new_images_from_arxiv_examples_1of1}, \ref{new_images_from_mathv360k_examples_1of2}, and \ref{new_images_from_mathv360k_examples_2of2}. These figures demonstrate the model's ability to produce innovative and varied image outputs suitable for diverse applications.

\textbf{Synthesize New Problems Based on New images.}
\label{app:new_problems}
The Problem Synthesis Prompt shown in Figure~\ref{fig:synthesize_problem_prompt} is designed to encourage creative and meaningful engagement with visual data by crafting math reasoning questions that are both accessible and challenging for a K-12 audience. This process involves analyzing patterns, shapes, and numerical relationships present in an image, then constructing a single, concise question that stimulates analytical thinking. The prompt ensures that the generated question is self-contained, solvable using the visible information in the image, and includes any essential details that may not be immediately apparent. By adhering to these guidelines, educators and content creators can develop visually engaging problems that promote critical reasoning and mathematical exploration, fostering a deeper connection between visual interpretation and problem-solving skills.

\input{appendix/synthesize_problem_prompt}

\clearpage

\input{appendix/img2code_examples}

\clearpage

\input{appendix/new_images}

%% file: appendix/k12_translate_prompts.tex
\begin{figure*}[htbp]
\begin{tcolorbox}
[colback=wkpurple!50!white,colframe=wkpurple!95!black,title=\textcolor{black}{K12 Process Prompt}]
\begin{small}
\textbf{System Prompt:}
    
\vspace{2mm}

You are an expert in mathematical problem-solving, LaTeX formatting, and structured data extraction. Please present all results in English and well-formatted LaTeX, converting HTML to LaTeX as needed. You will be provided with a JSON object containing the following fields: ["question", "option\_a", "option\_b", "option\_c", "option\_d", "option\_e", "answer1", "answer2", "parse"].

\textcolor{wkpurple}{\rule{\linewidth}{0.4pt}}

\textbf{User Prompt:}
    
\vspace{2mm}

Please process the provided JSON object by following these steps:
    
\vspace{2mm}

1. **Translation:**

\hspace*{4mm}- Translate the math problem and any accompanying options into English.

\hspace*{4mm}- If the problem includes multiple-choice options, format them as a bulleted list.

\hspace*{4mm}- If no options are available, return an empty option list (`[]`).

\hspace*{4mm}- For problems with multiple sub-questions, separate each sub-question as an individual item in another list.
    
\vspace{4mm}

2. **Step-by-Step Solution:**

\hspace*{4mm}- Provide a detailed, step-by-step solution to the problem, referencing "answer1", "answer2", and "parse".

\hspace*{4mm}- Adhere to the solution process provided by "answer1", "answer2", and "parse", as they are correct.
    
\vspace{4mm}

3. **Short Answer:**

\hspace*{4mm}- Specify the answer(s) in a list format, where each item is a single word or phrase.

\hspace*{4mm}- Answer(s) should adhere to that provided by "answer1", "answer2", and "parse".

\hspace*{4mm}- For multiple-choice questions, return one of A, B, C, D, or E.

\hspace*{4mm}- For proof-based questions, return "proven".

\hspace*{4mm}- For problems with sub-questions, provide the answer for each sub-question in the same order as the sub-question list.
    
\vspace{4mm}

**Input JSON:**

\begin{verbatim}
```json\end{verbatim}
[Raw Json Data]
\begin{verbatim}```
\end{verbatim}

\end{small}
\end{tcolorbox}
\caption{Prompt for processing, solving, and formatting K-12 math problems from structured JSON input.}
\label{fig:prompt_translation}
\end{figure*}

%% file: appendix/python_tikz_intro.tex
\begin{figure*}[ht]
  \includegraphics[width=0.99\linewidth]{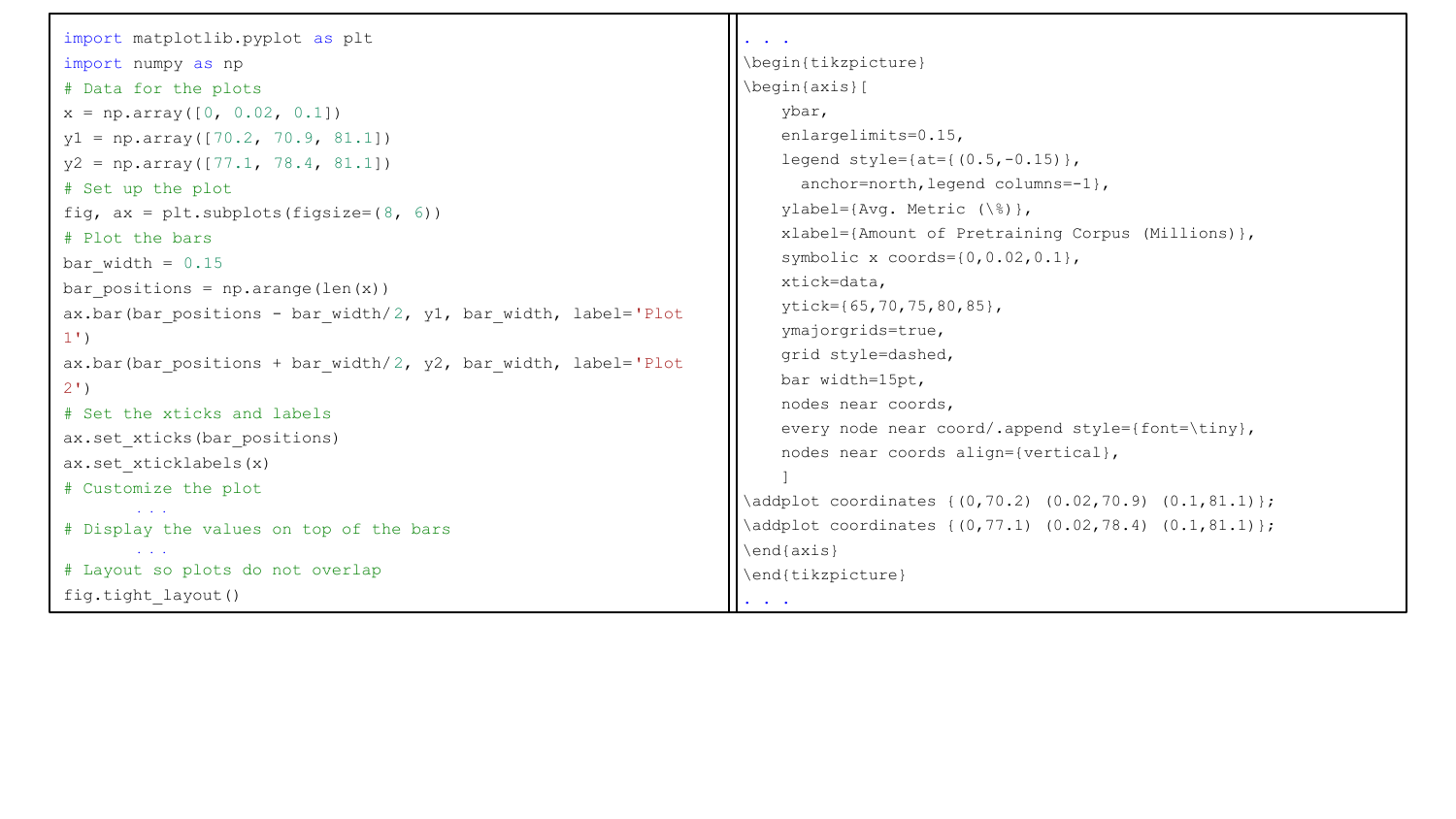}
  \vspace{-21mm}
  \caption{Comparison between Python code and TiKZ code.} 
  \label{fig:python_tikz_code}
\end{figure*}

\begin{figure*}[ht]
  \includegraphics[width=0.99\linewidth]{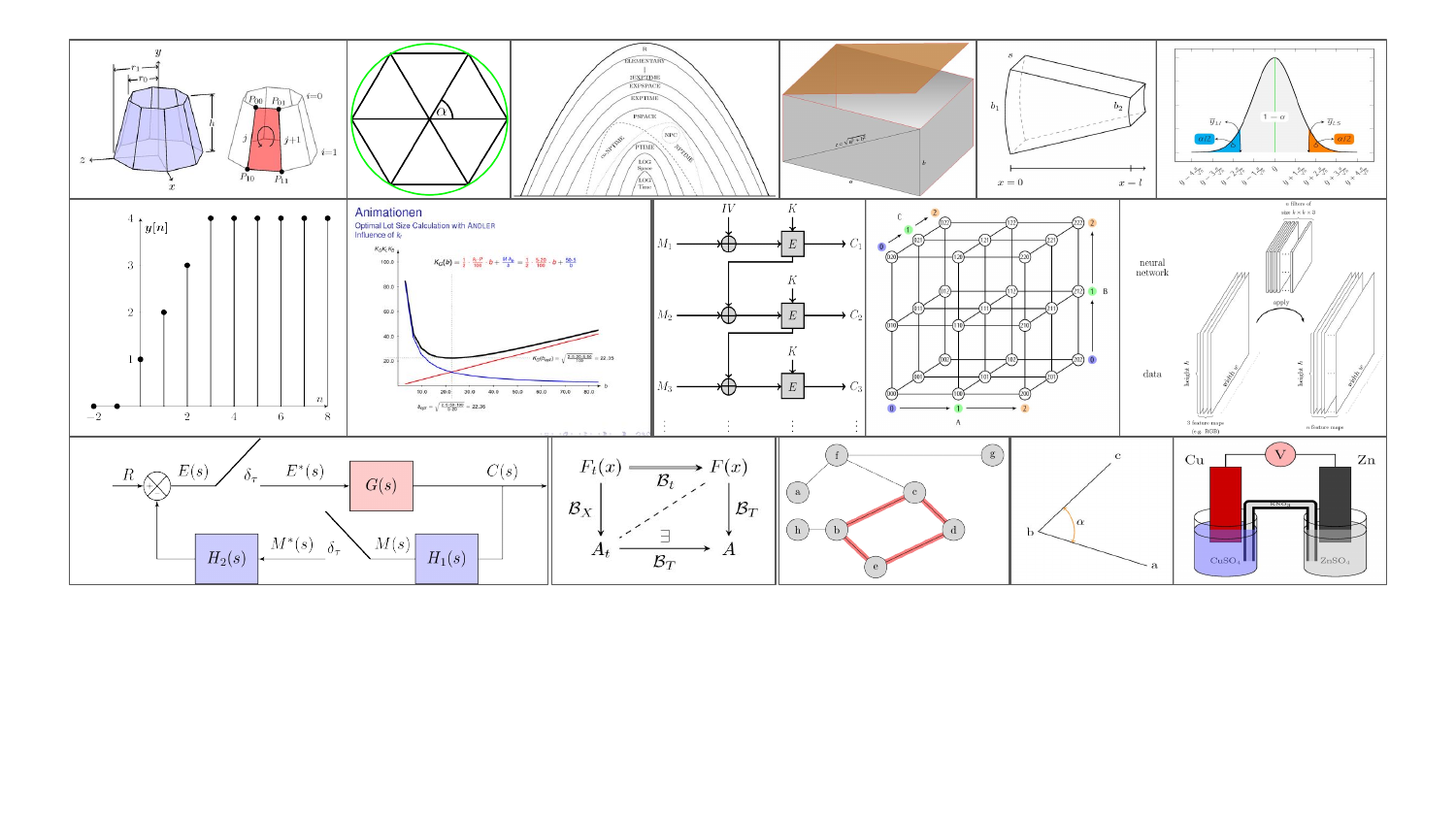}
  \vspace{-24mm}
  \caption {Some example of images generated by TikZ.} 
  \label{fig:tikz_examples}
\end{figure*}

\begin{figure*}[ht]
  \includegraphics[width=0.99\linewidth]{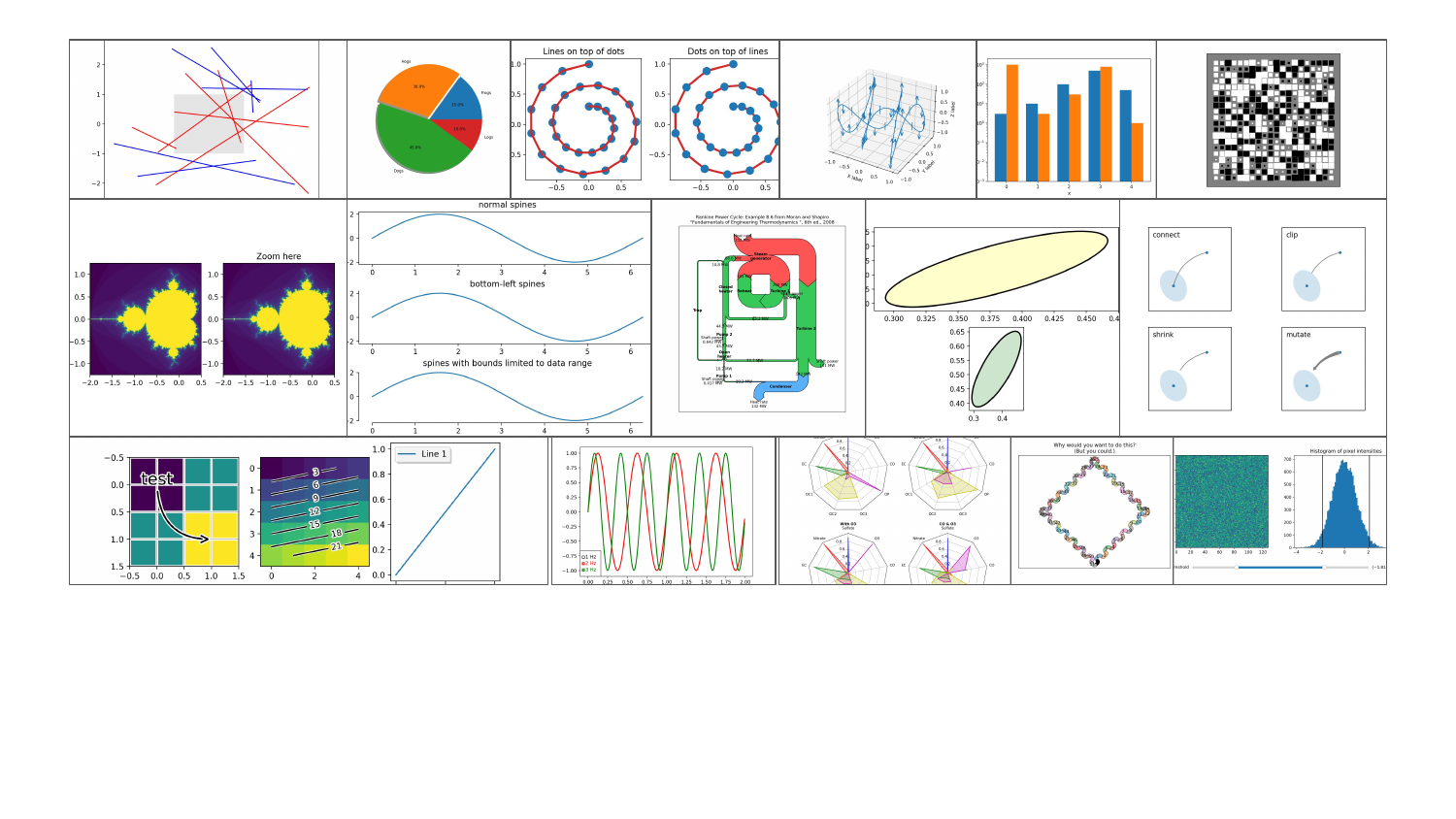}
  \vspace{-24mm}
  \caption{Some example of images generated by Python.} 
  \label{fig:python_examples}
\end{figure*}

%% file: appendix/img2code_prompts.tex
\begin{figure}[htbp]
\begin{tcolorbox}[colback=wkyellow!50!white,colframe=wkyellow!80!orange,title=\textcolor{black}{(a) Image-to-TikZ Prompt:}]
\begin{small}
Please generate the corresponding TikZ code that accurately represents the visual elements in the image. TikZ is a powerful tool for creating vector graphics within LaTeX documents. Your generated code should be precise, well-structured, and should recreate the image as faithfully as possible.\\
<image>

\textcolor{wkyellow}{\rule{\linewidth}{2pt}}

The image can be generated using the following TikZ code:\\

\begin{verbatim}
```tikz\end{verbatim}
[code]
\begin{verbatim}```
\end{verbatim}
\end{small}
\end{tcolorbox}

\begin{tcolorbox}[colback=wkblue!50!white,colframe=wkblue!80!blue,title=\textcolor{black}{(b) Image-to-Python Prompt: }]
\begin{small}
Please provide the Python code needed to reproduce this image.\\
<image>

\textcolor{wkblue}{\rule{\linewidth}{2pt}}

The image can be generated using the following Python code:\\

\begin{verbatim}
```python\end{verbatim}
[code]
\begin{verbatim}```
\end{verbatim}

\end{small}
\end{tcolorbox}

\caption{Prompt templates of our Image-to-Code Dataset.}
\label{fig:i2c_prompt_templates}
\end{figure}

%% file: appendix/tikz2python_prompt.tex
\begin{figure*}[t]
\begin{tcolorbox}
[colback=wkpurple!50!white,colframe=wkpurple!95!black,title=\textcolor{black}{TikZ-to-Python Prompt}]
\begin{small}
\textbf{System Prompt:}
    
\vspace{2mm}

You are an expert in both LaTeX (specifically TiKZ) and Python (specifically Matplotlib).

\textcolor{wkpurple}{\rule{\linewidth}{0.4pt}}

\textbf{User Prompt:}
    
\vspace{2mm}

Translate the provided TiKZ code into Python code using appropriate plotting libraries, such as Matplotlib. Pay close attention to the following requirements:
    
\vspace{4mm}

1. **Avoid Overlapping**: Ensure that points, labels and text elements have different positions to avoid any overlap, enhancing readability.
    
\vspace{4mm}

2. **LaTeX Formatting**: Accurately interpret and format any LaTeX equations or mathematical expressions to ensure they render correctly in the image.
    
\vspace{4mm}

3. **Executable Code**: Ensure that the Python code is complete and can be executed directly without errors.
    
\vspace{4mm}

Here\'s the TiKZ code:
    
\vspace{4mm}

\begin{verbatim}
```latex\end{verbatim}
[TiKZ Code]
\begin{verbatim}```
\end{verbatim}
    
\vspace{4mm}

Make sure to wrap your resulting Python code in the following format:
    
\vspace{4mm}

\begin{verbatim}
```python\end{verbatim}
[your python code here]
\begin{verbatim}```
\end{verbatim}

\end{small}
\end{tcolorbox}
\caption{Prompt for translating LaTeX TiKZ code into Python Matplotlib code with a focus on accuracy, readability, and executability.}
\label{fig:tikz2python_prompt}
\end{figure*}

%% file: appendix/tikz2python_examples.tex
\begin{table*}[htb]
  \scriptsize
  \pgfmathsetlength{\wklength}{\textwidth/6 -2\tabcolsep}
  \begin{tabular}{c|c||c|c||c|c}
    \toprule
    TiKZ & Python & TiKZ & Python & TiKZ & Python \\
    \midrule
    \includegraphics[height=2.1cm,width=\wklength,keepaspectratio]{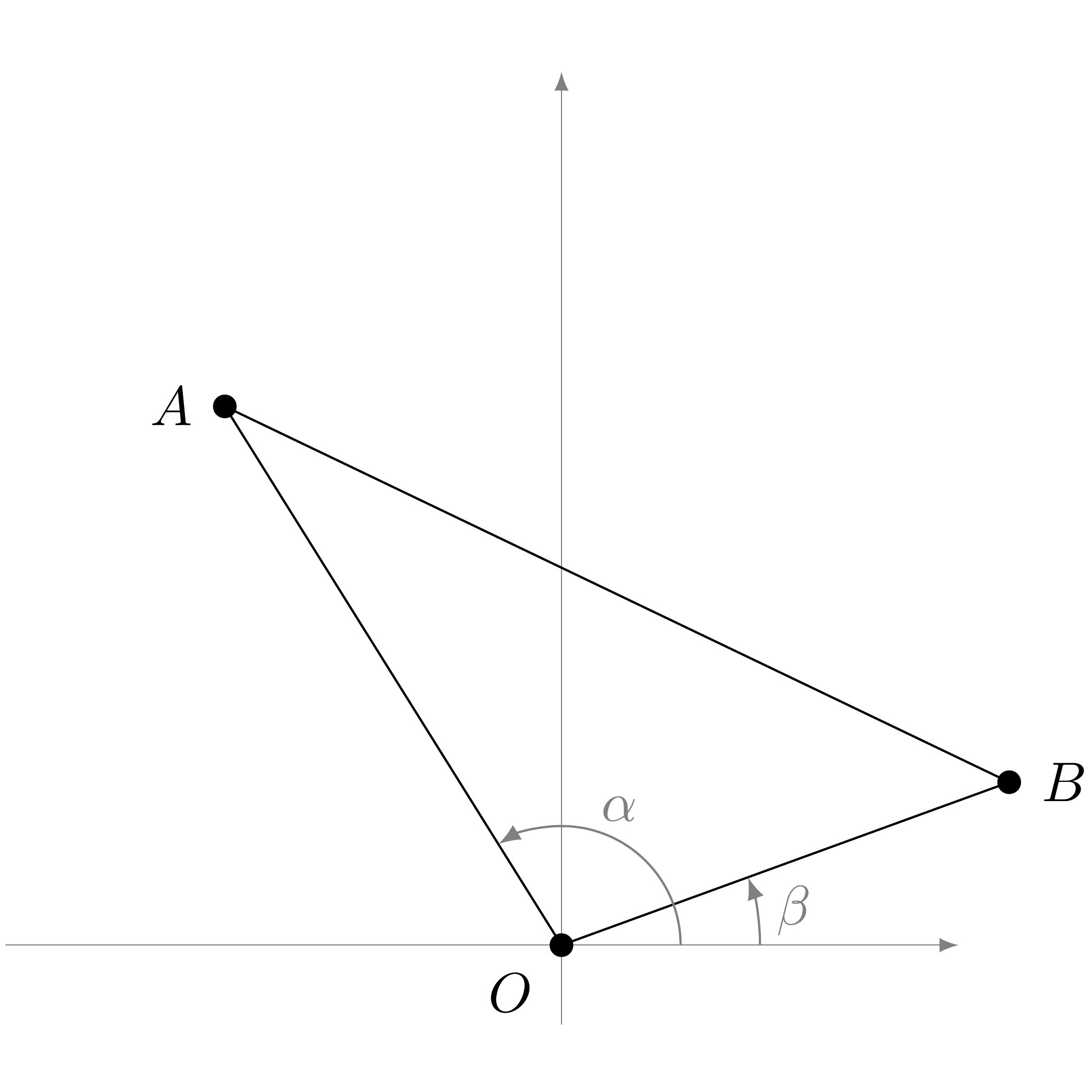} &
    \includegraphics[height=2.1cm,width=\wklength,keepaspectratio]{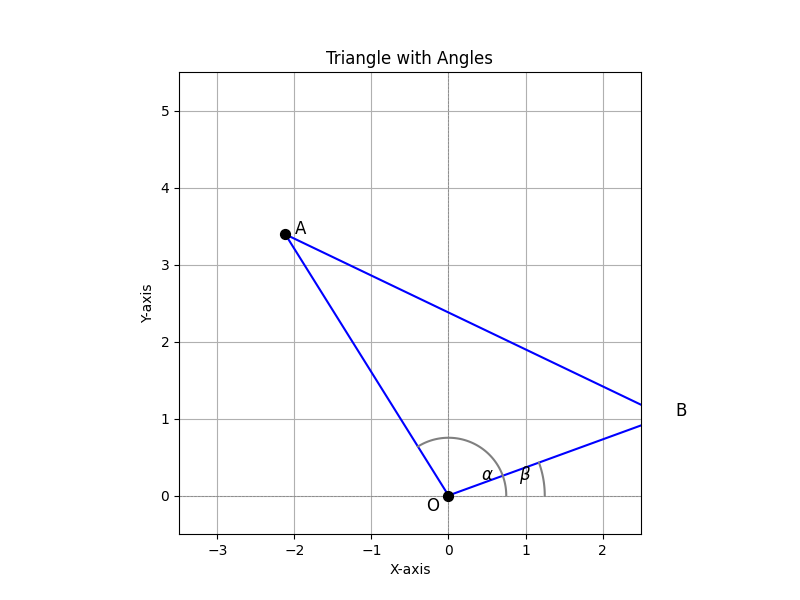} &
    \includegraphics[height=2.1cm,width=\wklength,keepaspectratio]{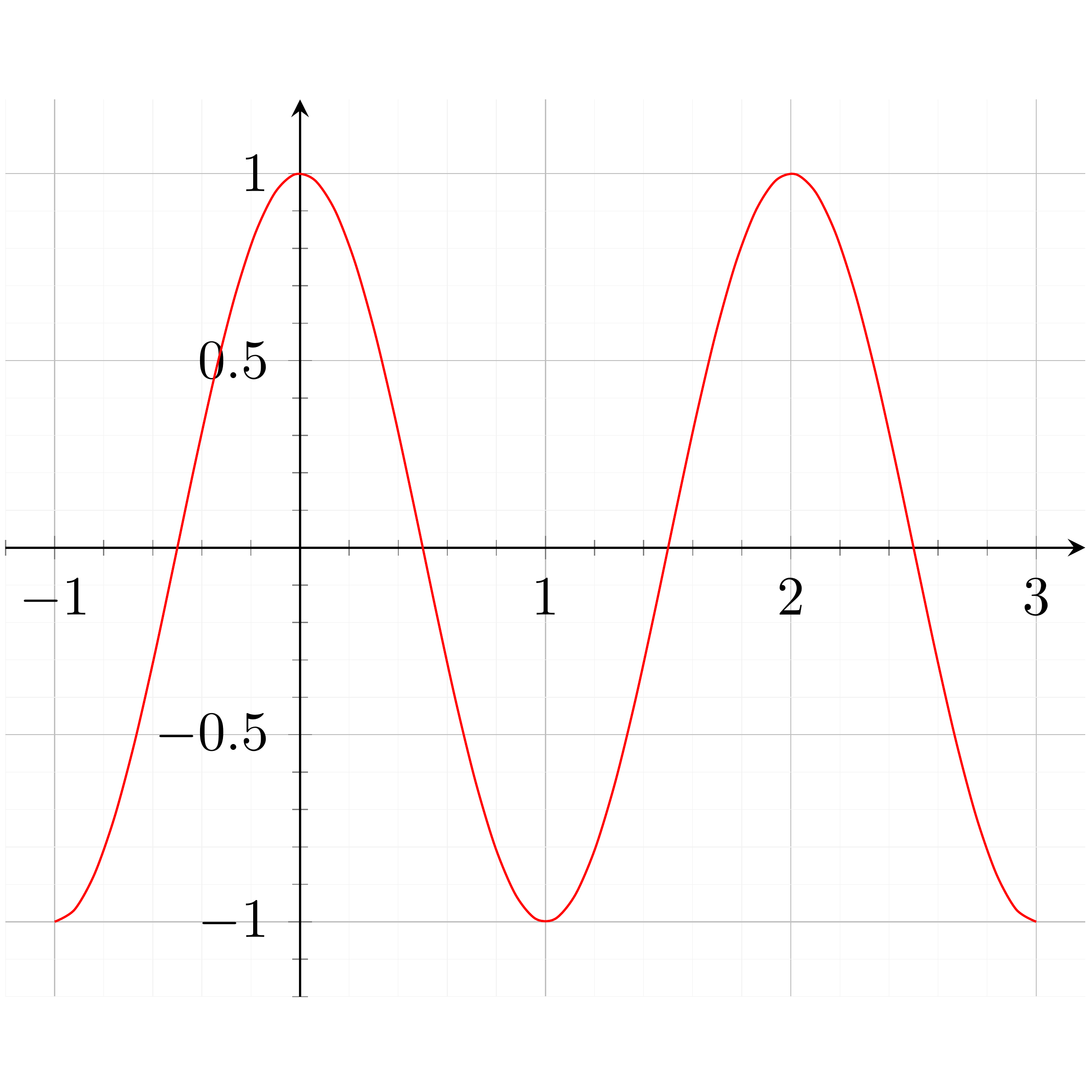} &
    \includegraphics[height=2.1cm,width=\wklength,keepaspectratio]{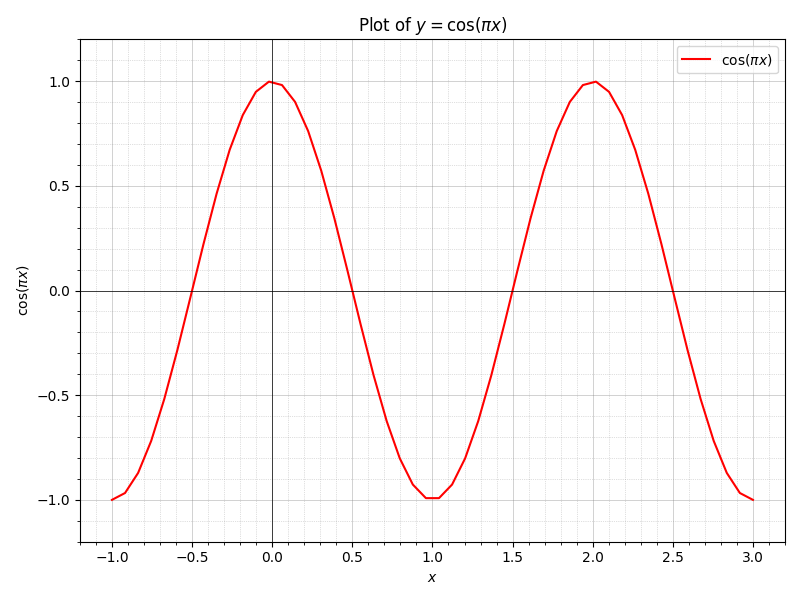} &
    \includegraphics[height=2.1cm,width=\wklength,keepaspectratio]{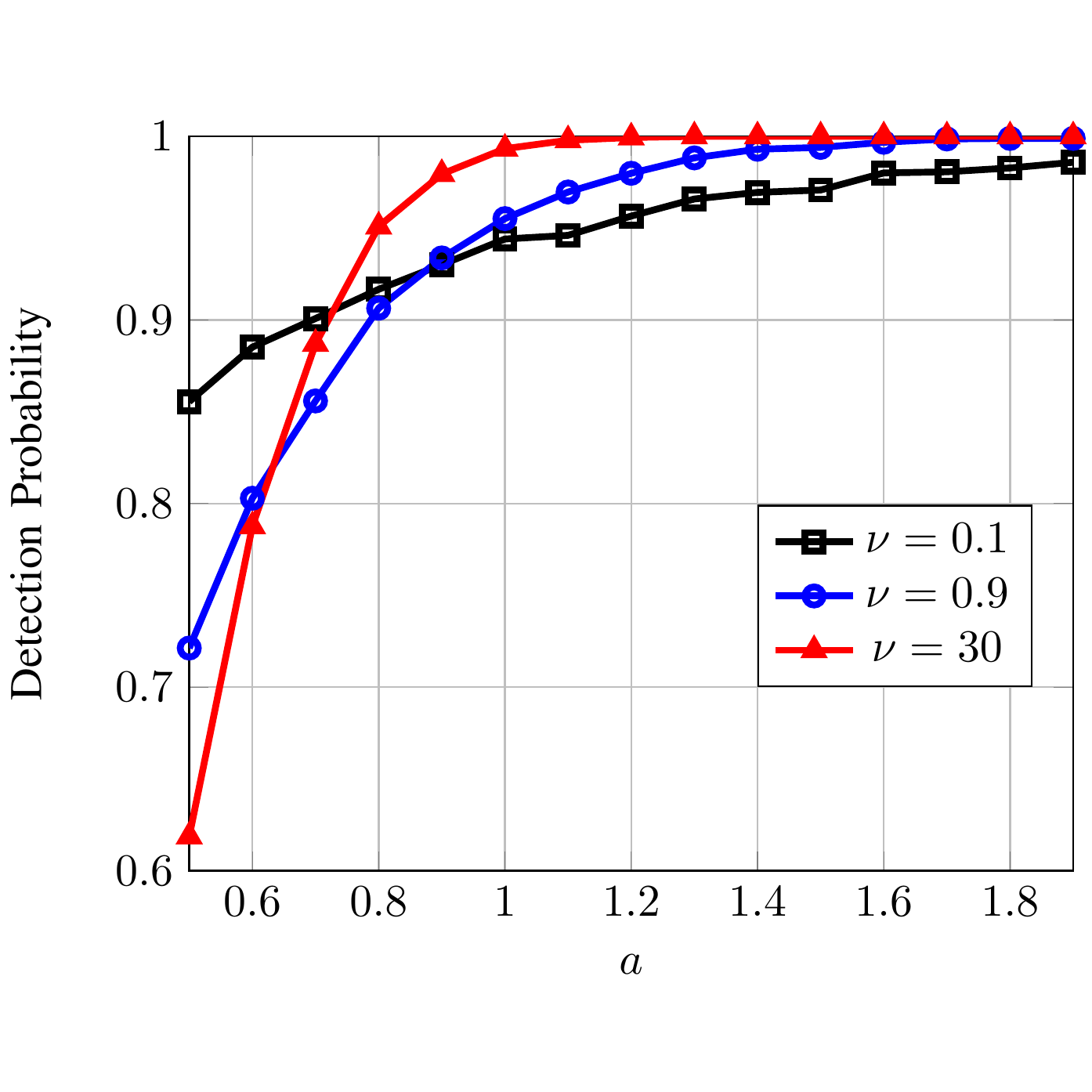} &
    \includegraphics[height=2.1cm,width=\wklength,keepaspectratio]{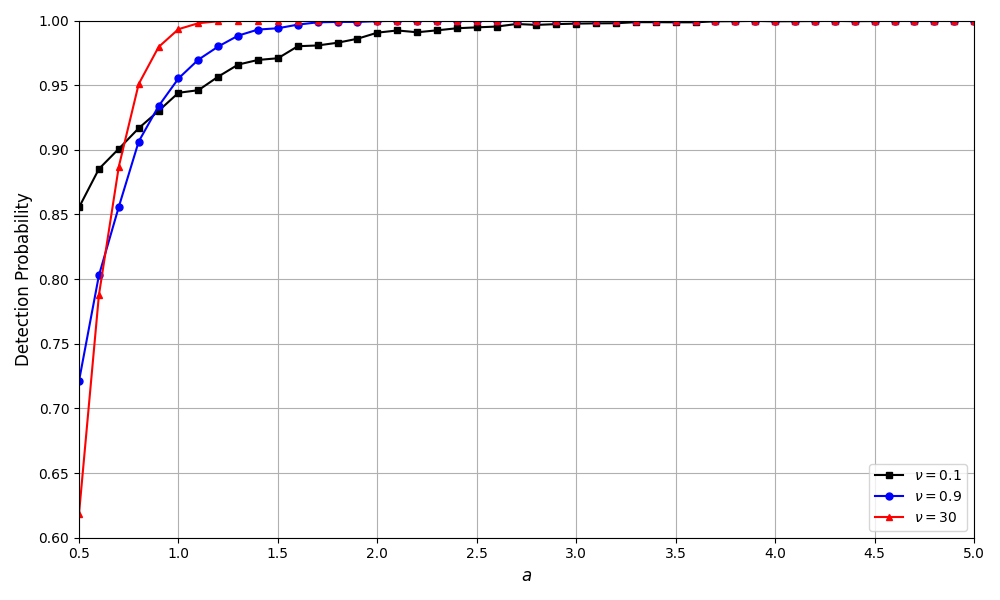} \\
    \midrule
    \includegraphics[height=2.1cm,width=\wklength,keepaspectratio]{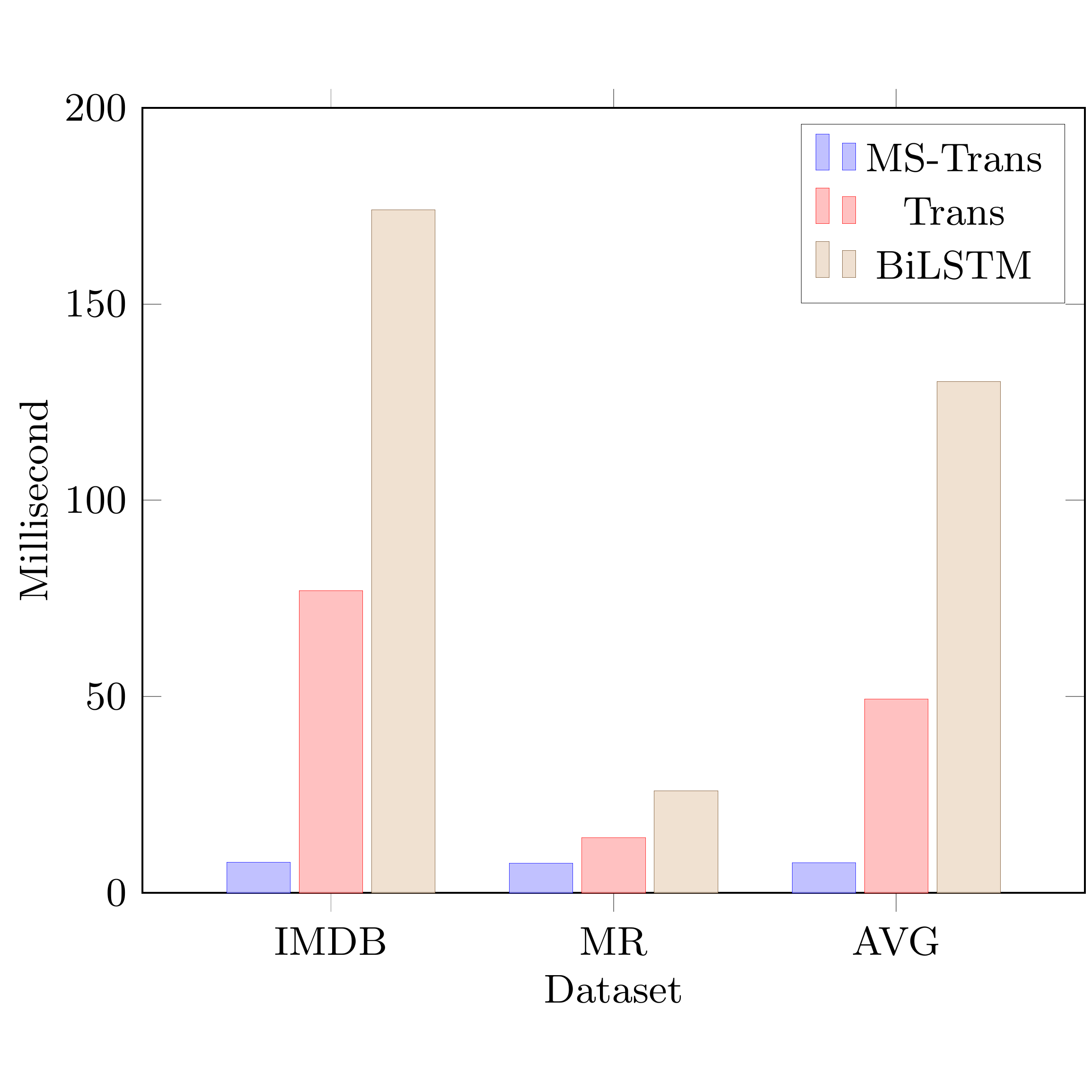} &
    \includegraphics[height=2.1cm,width=\wklength,keepaspectratio]{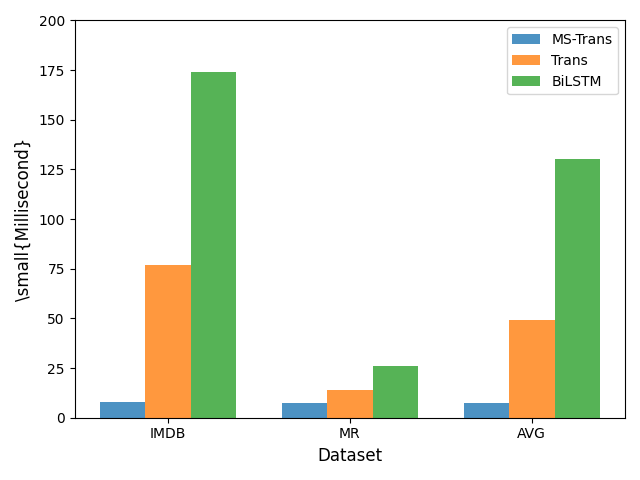} &
    \includegraphics[height=2.1cm,width=\wklength,keepaspectratio]{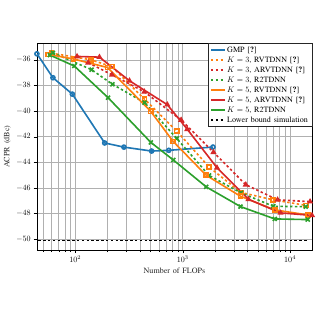} &
    \includegraphics[height=2.1cm,width=\wklength,keepaspectratio]{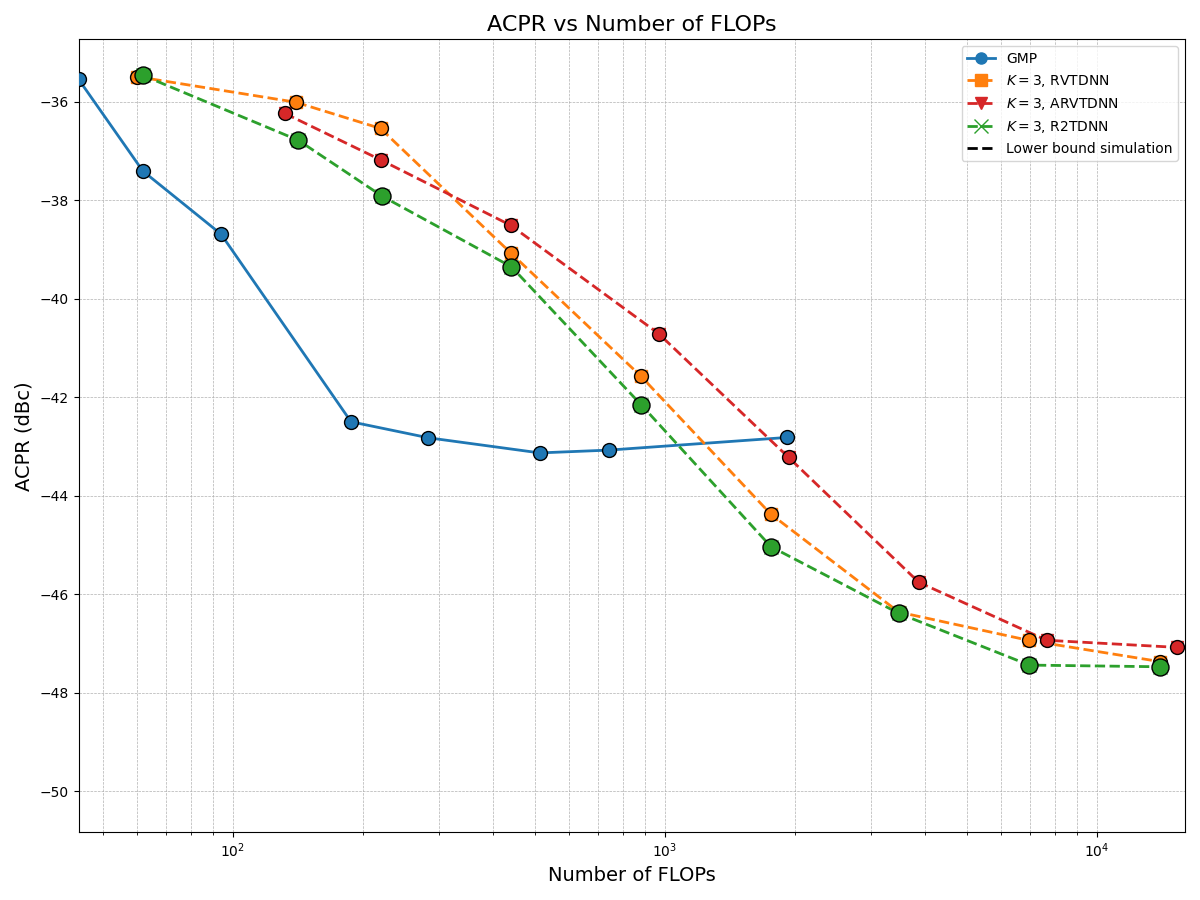} &
    \includegraphics[height=2.1cm,width=\wklength,keepaspectratio]{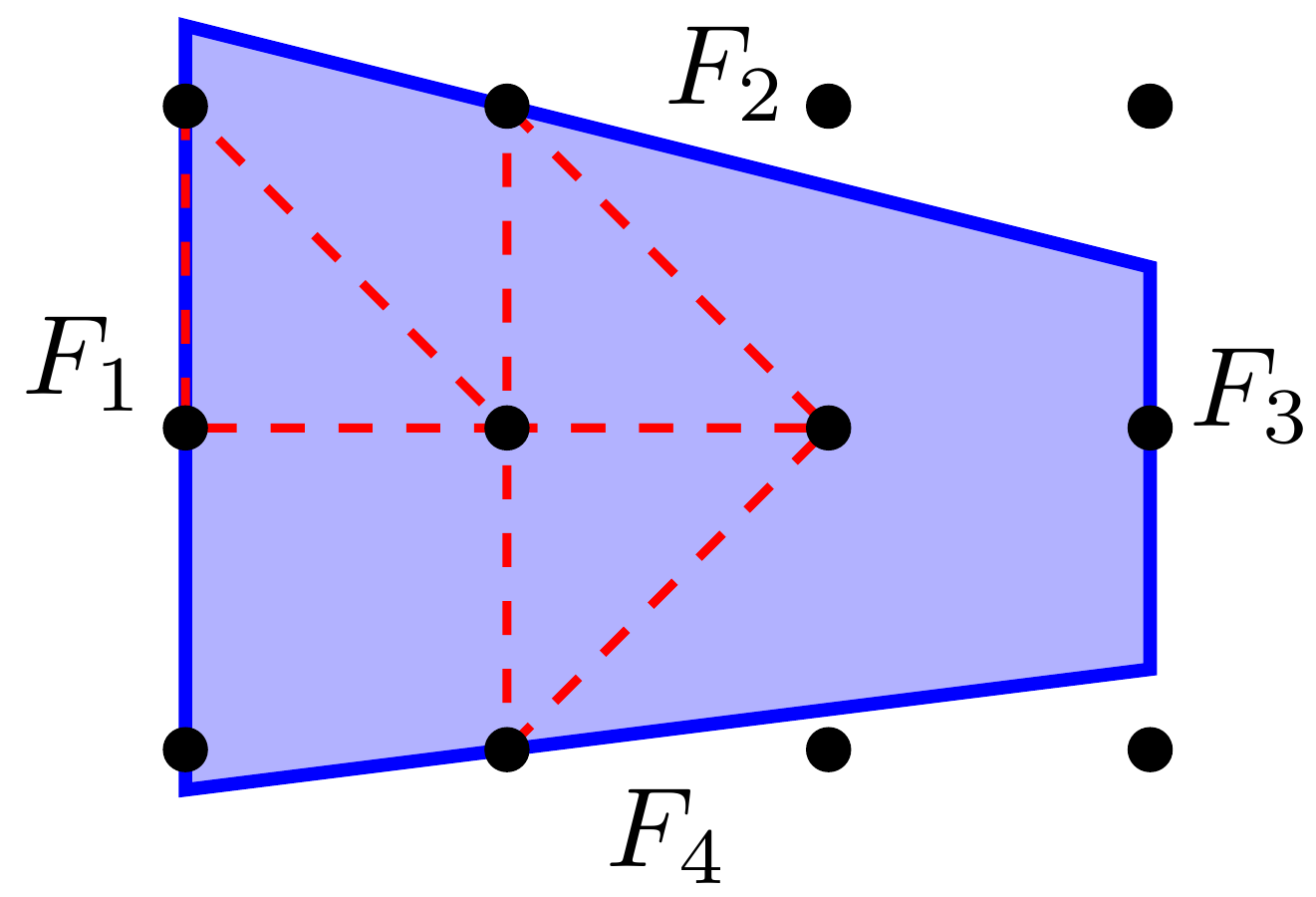} &
    \includegraphics[height=2.1cm,width=\wklength,keepaspectratio]{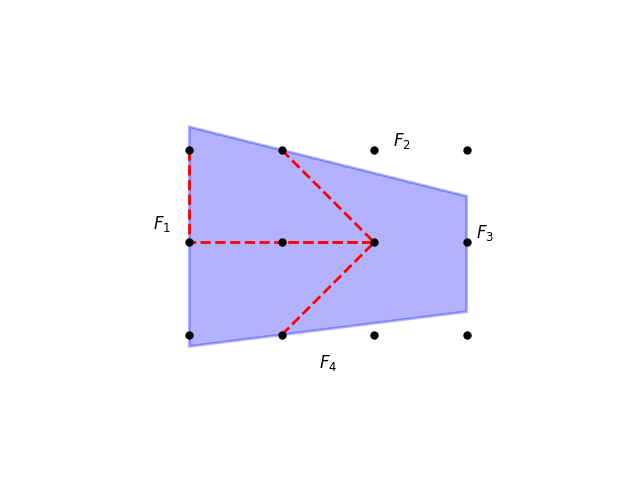} \\
    \midrule
    \includegraphics[height=2.1cm,width=\wklength,keepaspectratio]{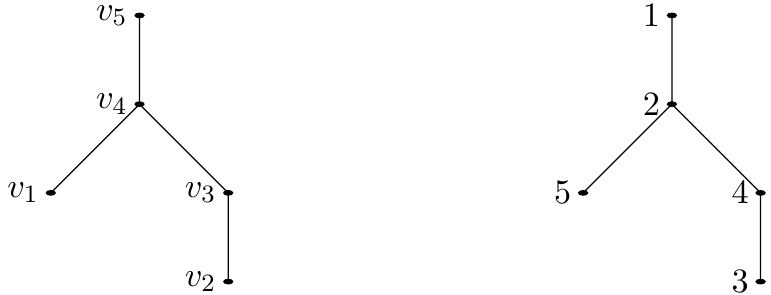} &
    \includegraphics[height=2.1cm,width=\wklength,keepaspectratio]{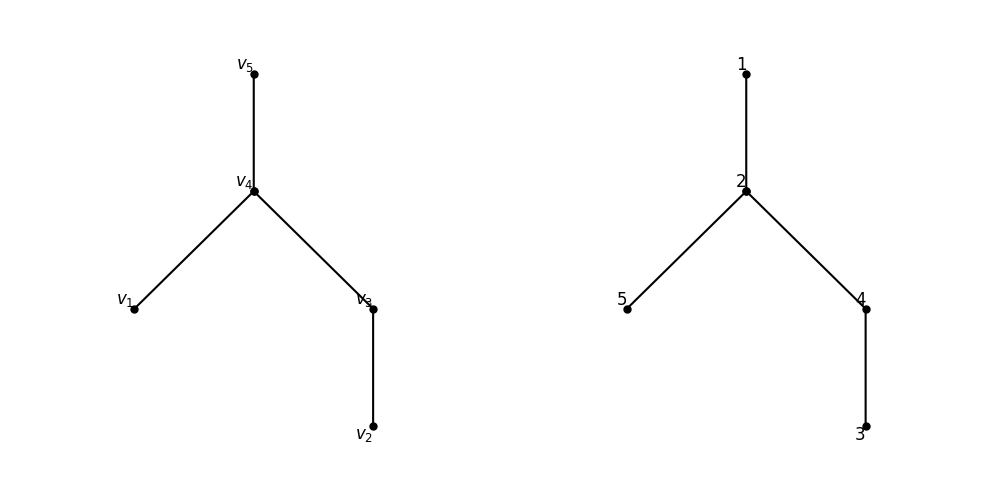} &
    \includegraphics[height=2.1cm,width=\wklength,keepaspectratio]{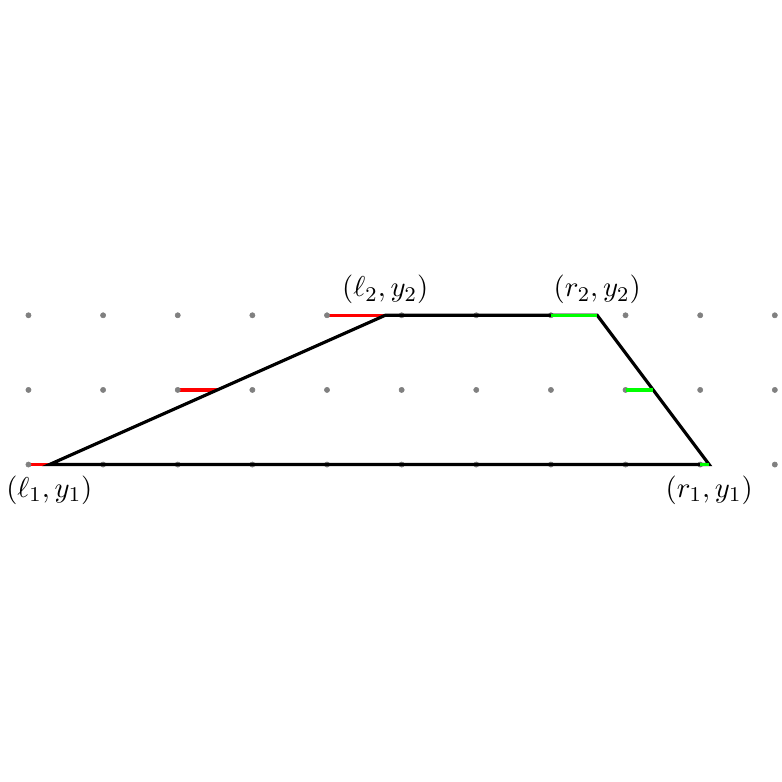} &
    \includegraphics[height=2.1cm,width=\wklength,keepaspectratio]{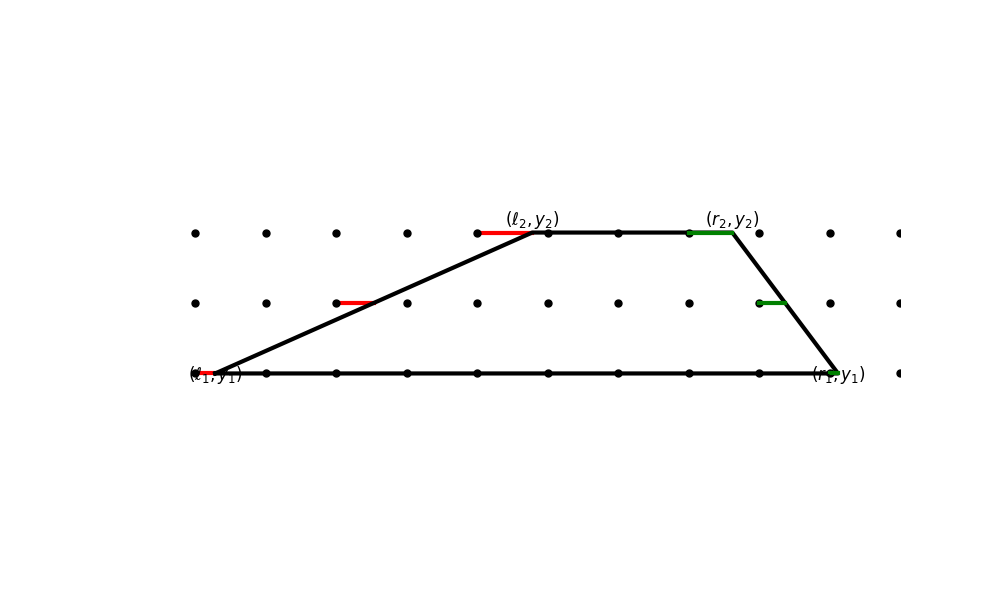} &
    \includegraphics[height=2.1cm,width=\wklength,keepaspectratio]{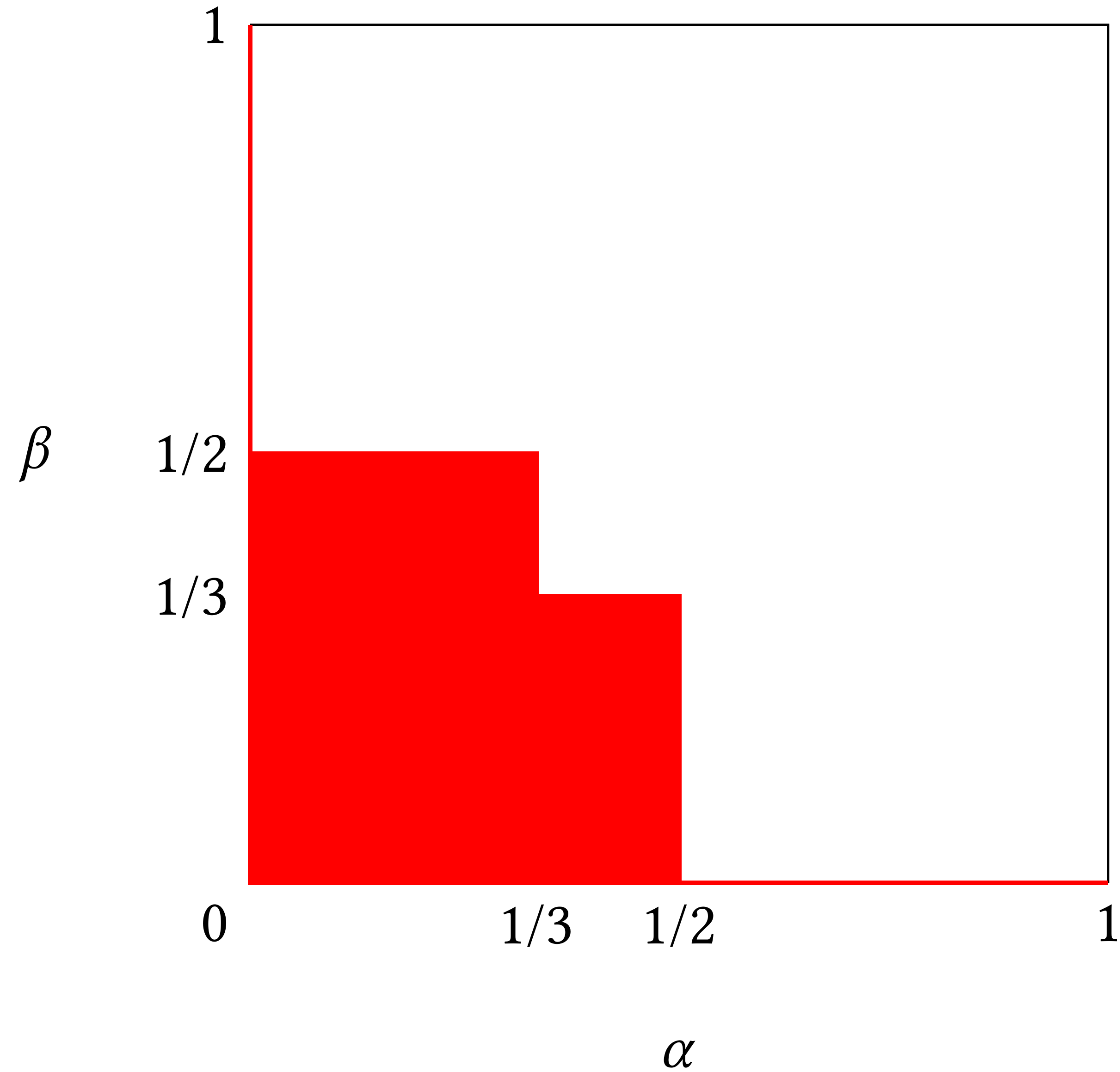} &
    \includegraphics[height=2.1cm,width=\wklength,keepaspectratio]{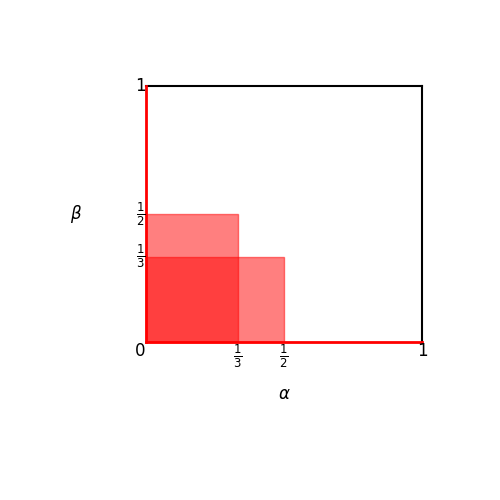} \\
    \midrule
    \includegraphics[height=2.1cm,width=\wklength,keepaspectratio]{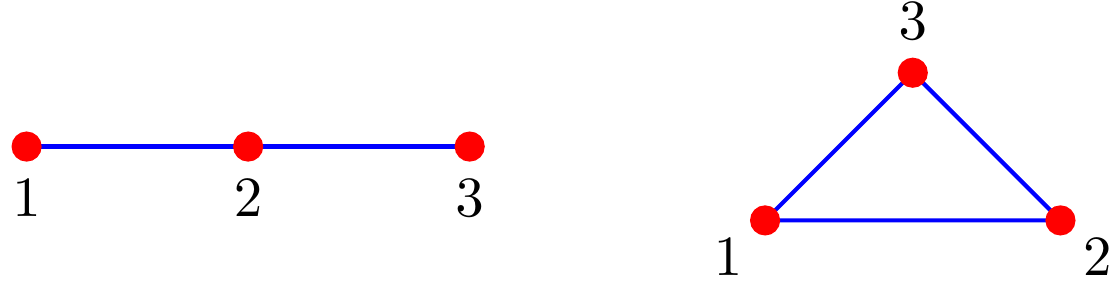} &
    \includegraphics[height=2.1cm,width=\wklength,keepaspectratio]{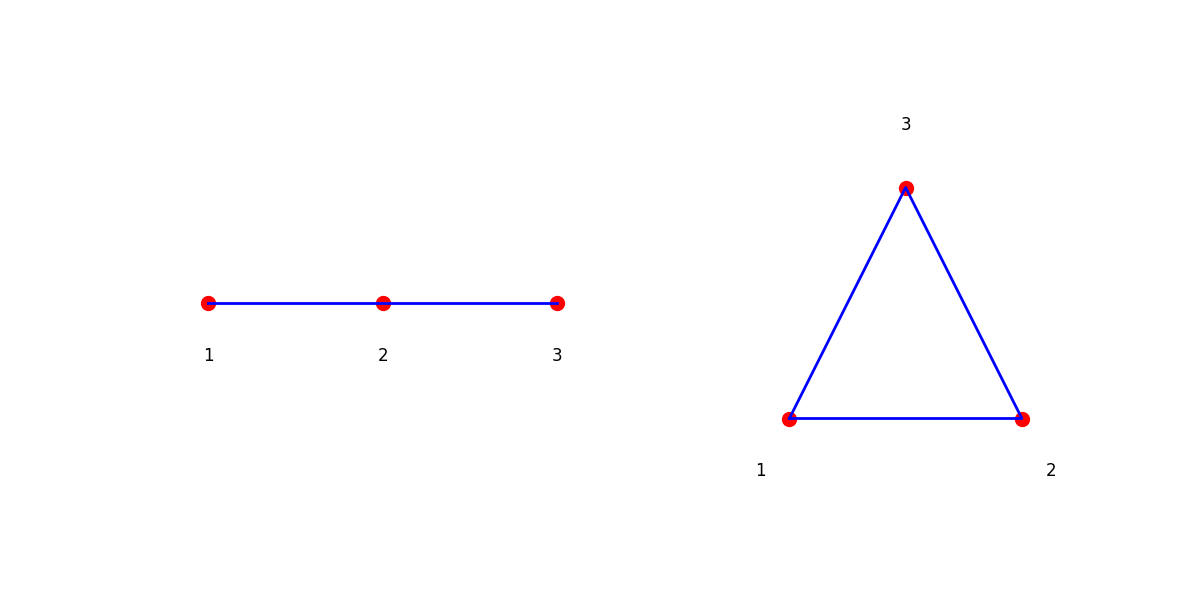} &
    \includegraphics[height=2.1cm,width=\wklength,keepaspectratio]{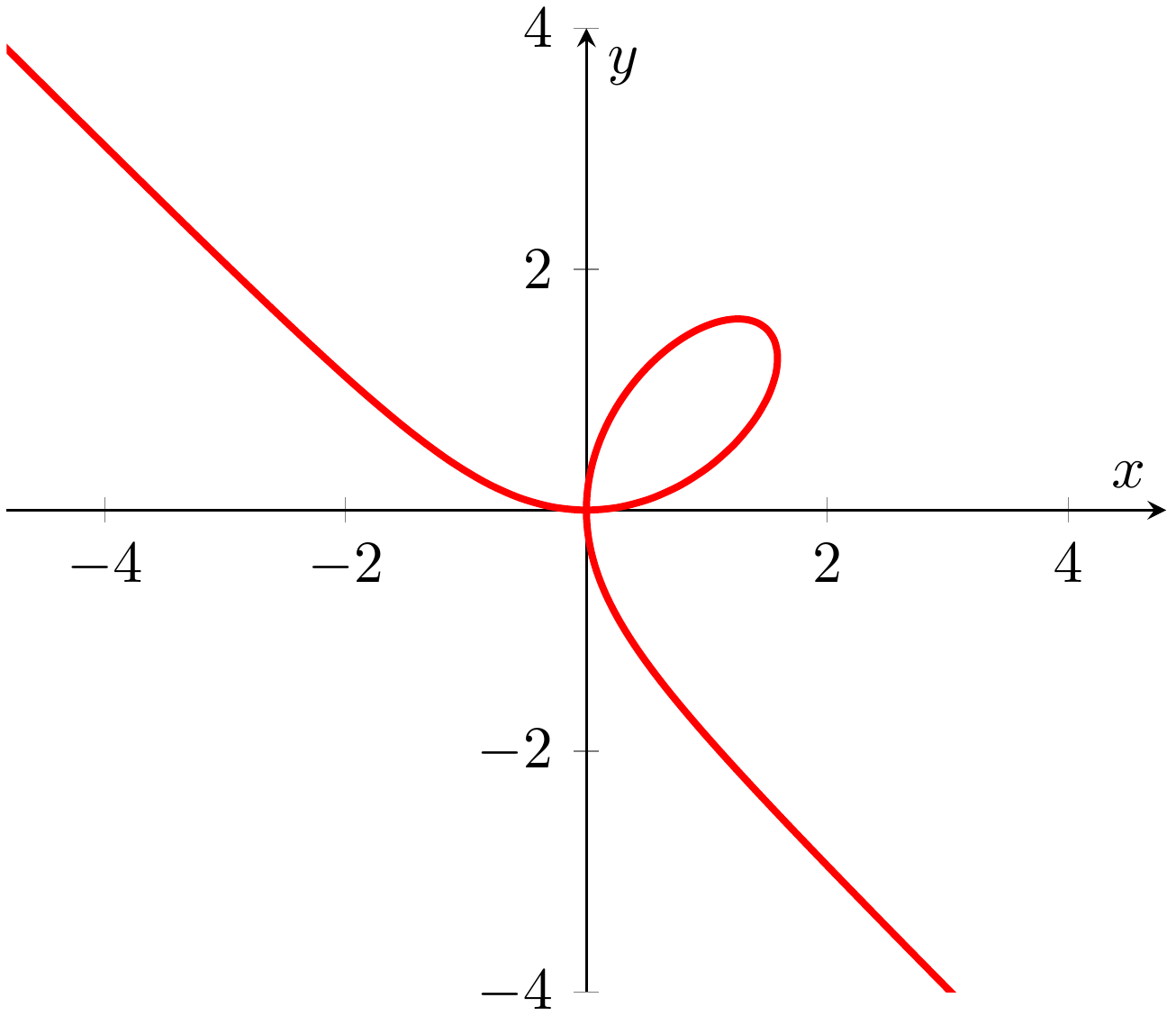} &
    \includegraphics[height=2.1cm,width=\wklength,keepaspectratio]{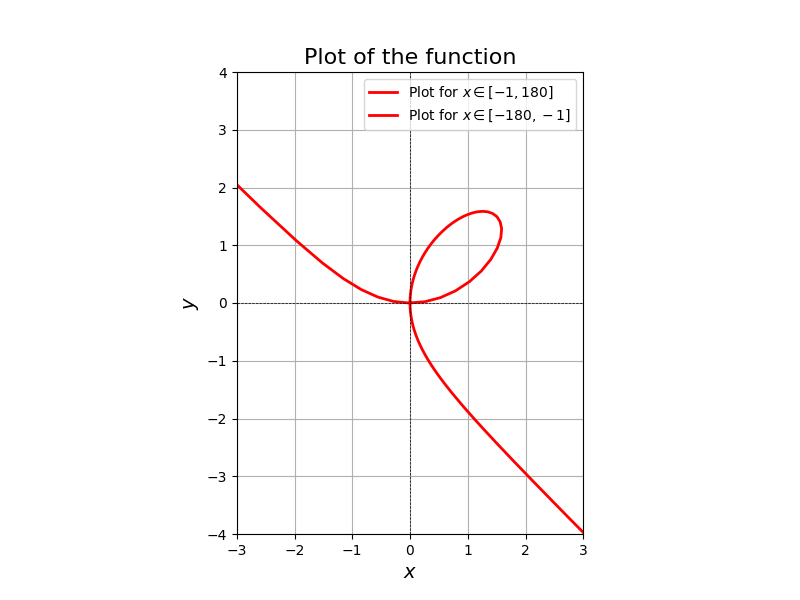} &
    \includegraphics[height=2.1cm,width=\wklength,keepaspectratio]{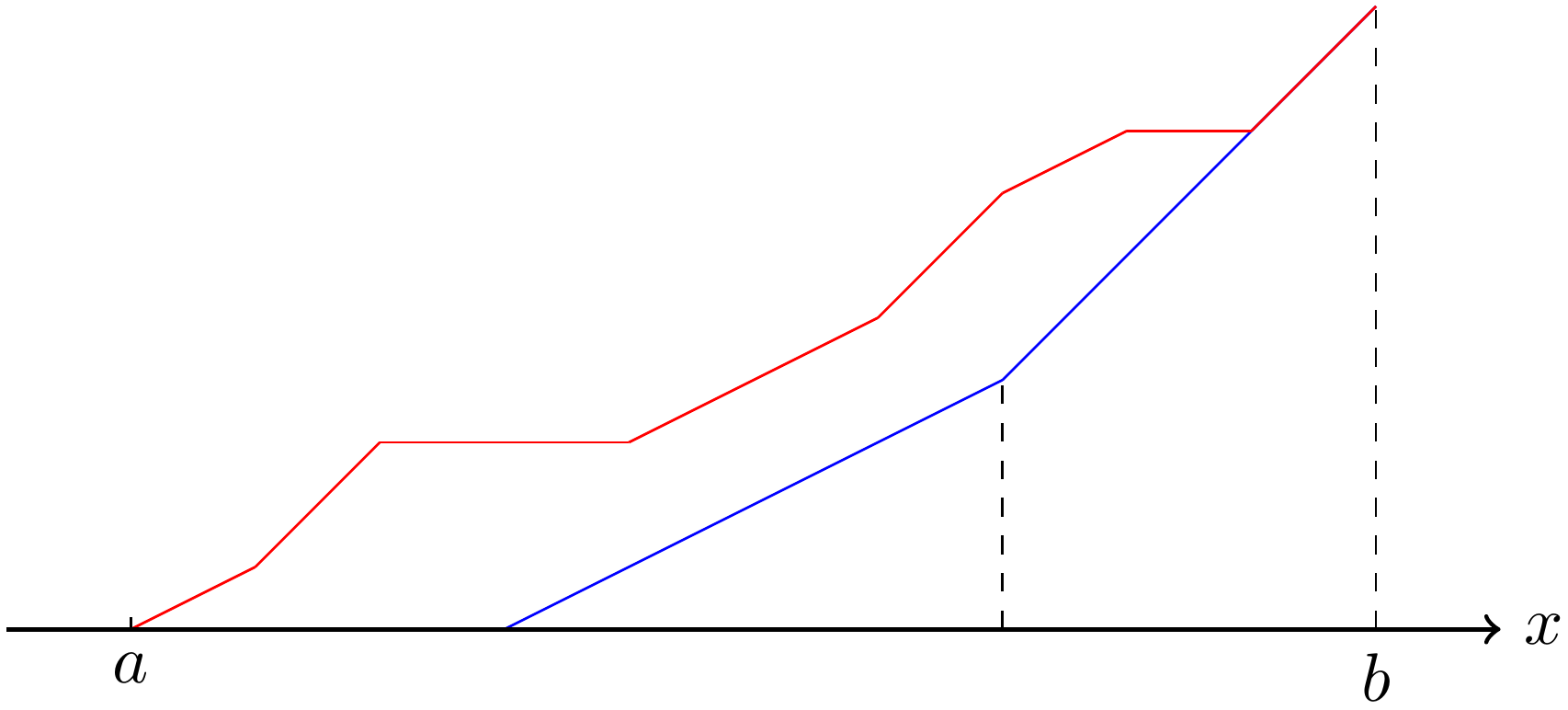} &
    \includegraphics[height=2.1cm,width=\wklength,keepaspectratio]{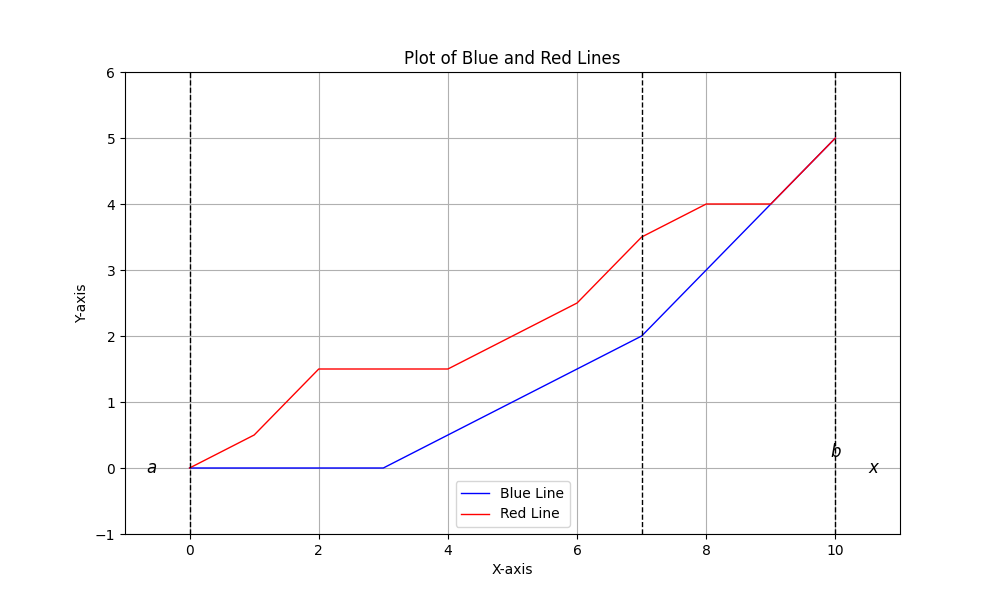} \\
    \midrule
    \includegraphics[height=2.1cm,width=\wklength,keepaspectratio]{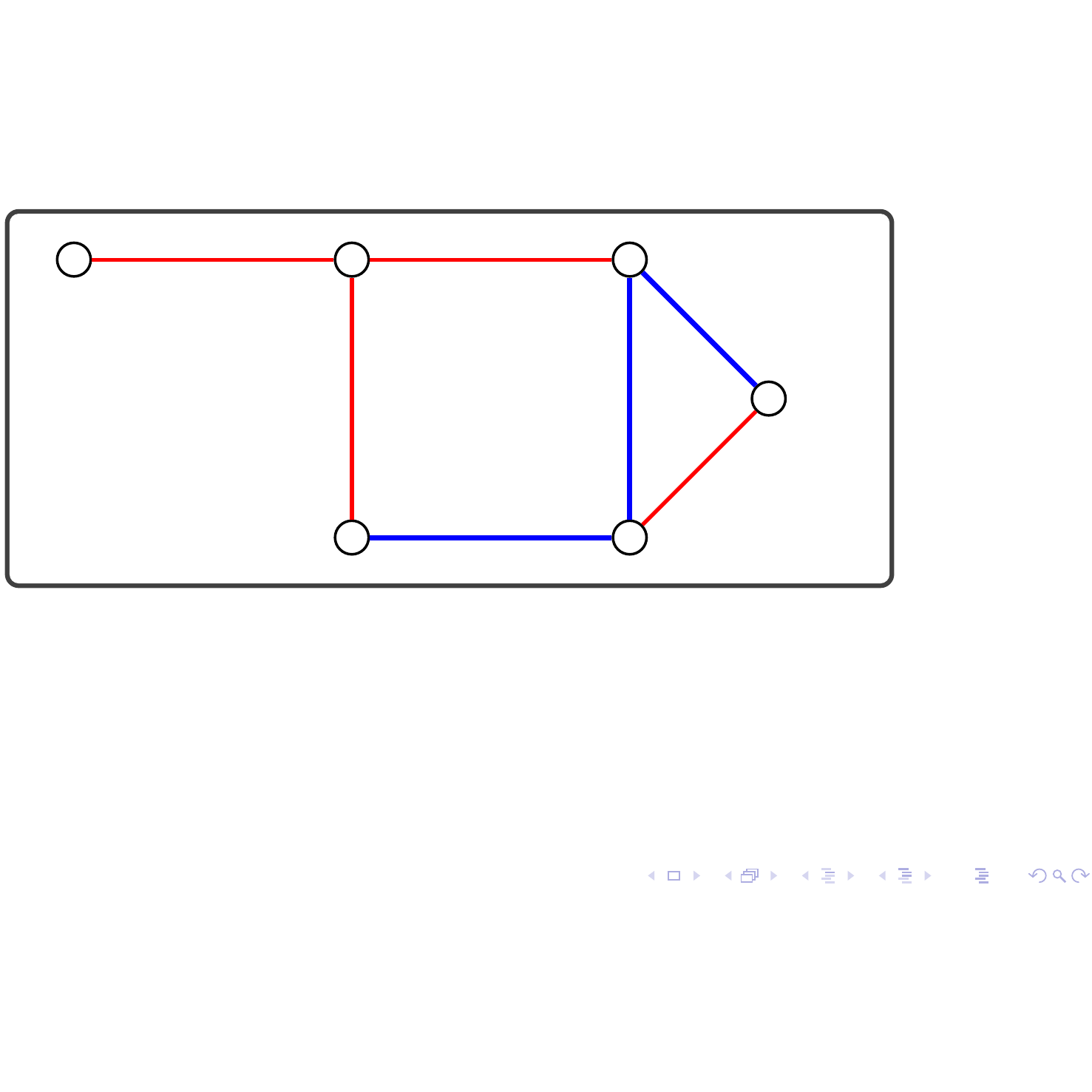} &
    \includegraphics[height=2.1cm,width=\wklength,keepaspectratio]{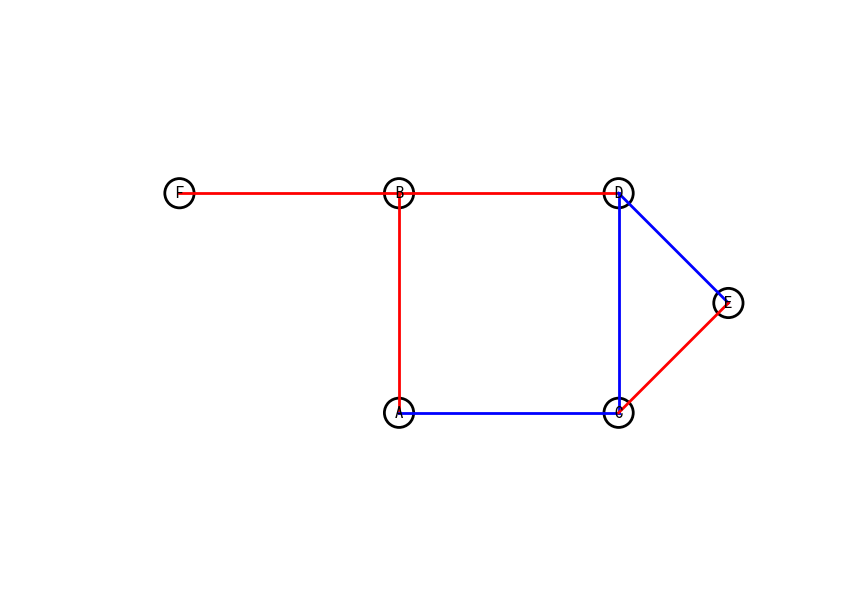} &
    \includegraphics[height=2.1cm,width=\wklength,keepaspectratio]{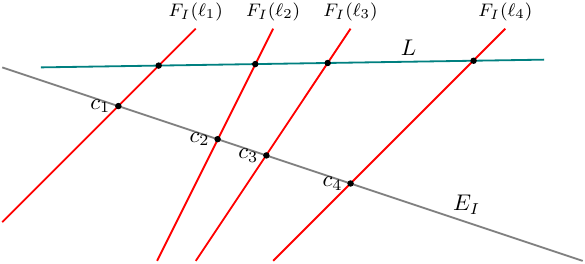} &
    \includegraphics[height=2.1cm,width=\wklength,keepaspectratio]{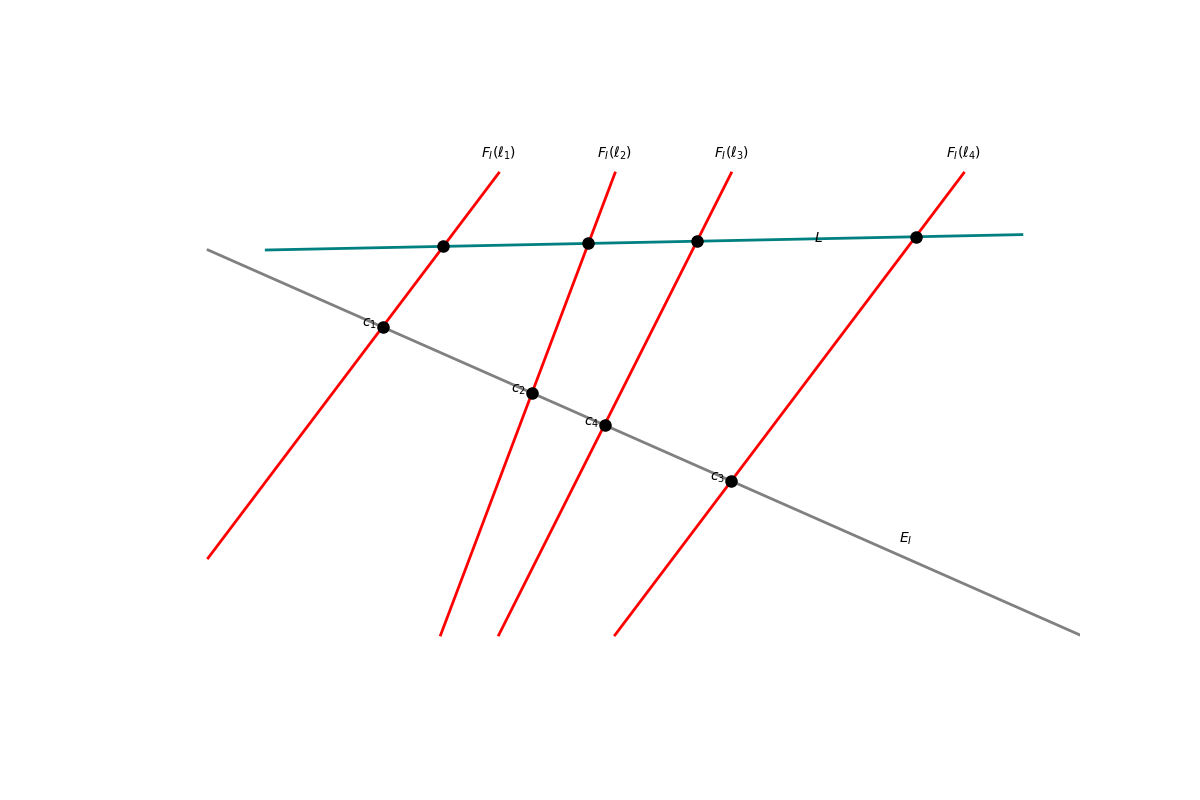} &
    \includegraphics[height=2.1cm,width=\wklength,keepaspectratio]{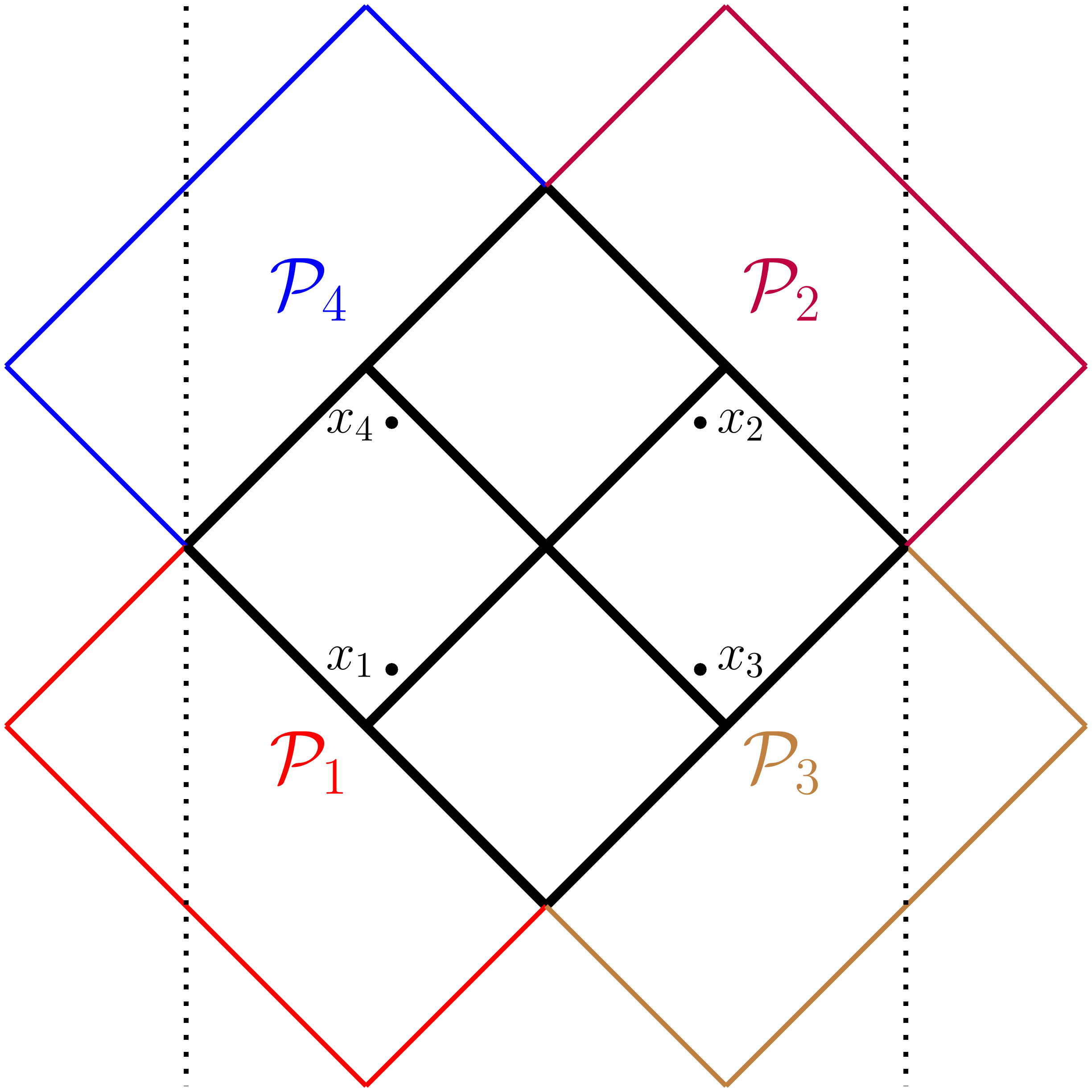} &
    \includegraphics[height=2.1cm,width=\wklength,keepaspectratio]{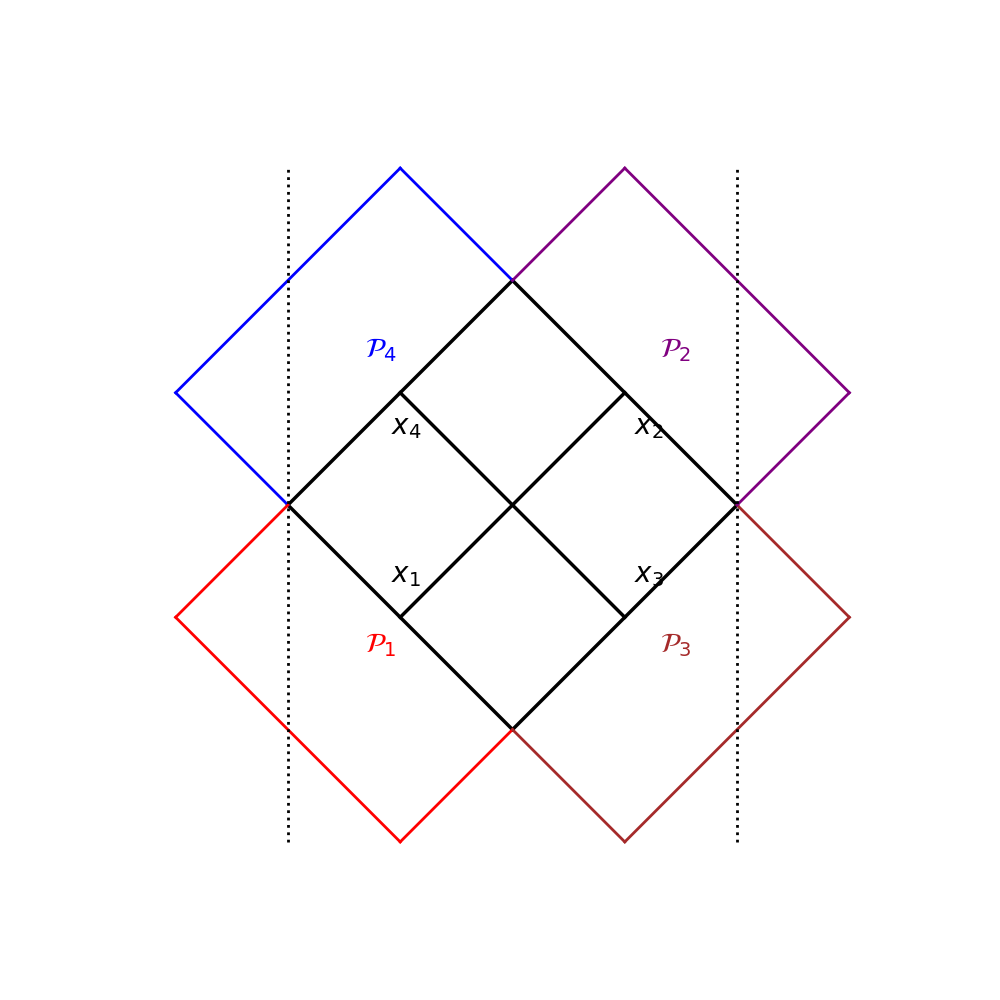} \\
    \midrule
    \includegraphics[height=2.1cm,width=\wklength,keepaspectratio]{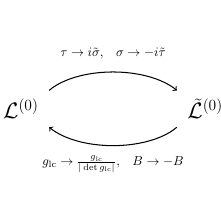} &
    \includegraphics[height=2.1cm,width=\wklength,keepaspectratio]{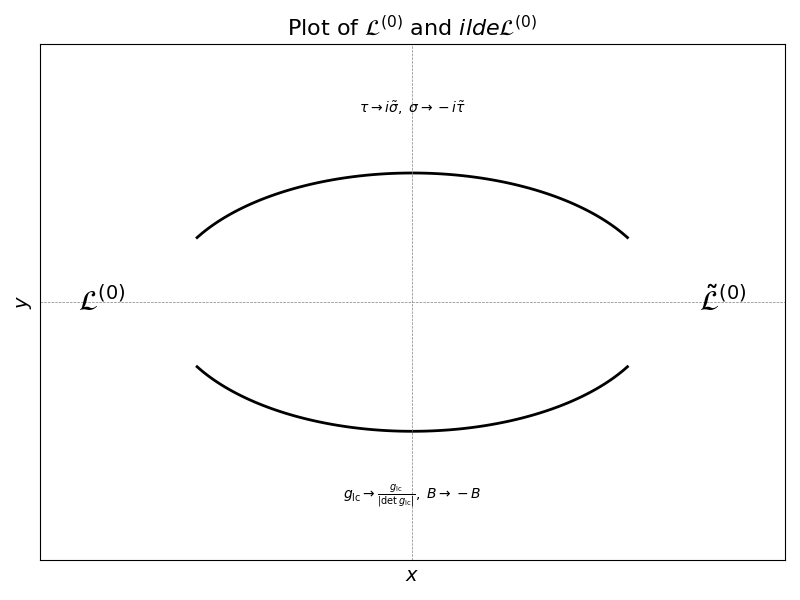} &
    \includegraphics[height=2.1cm,width=\wklength,keepaspectratio]{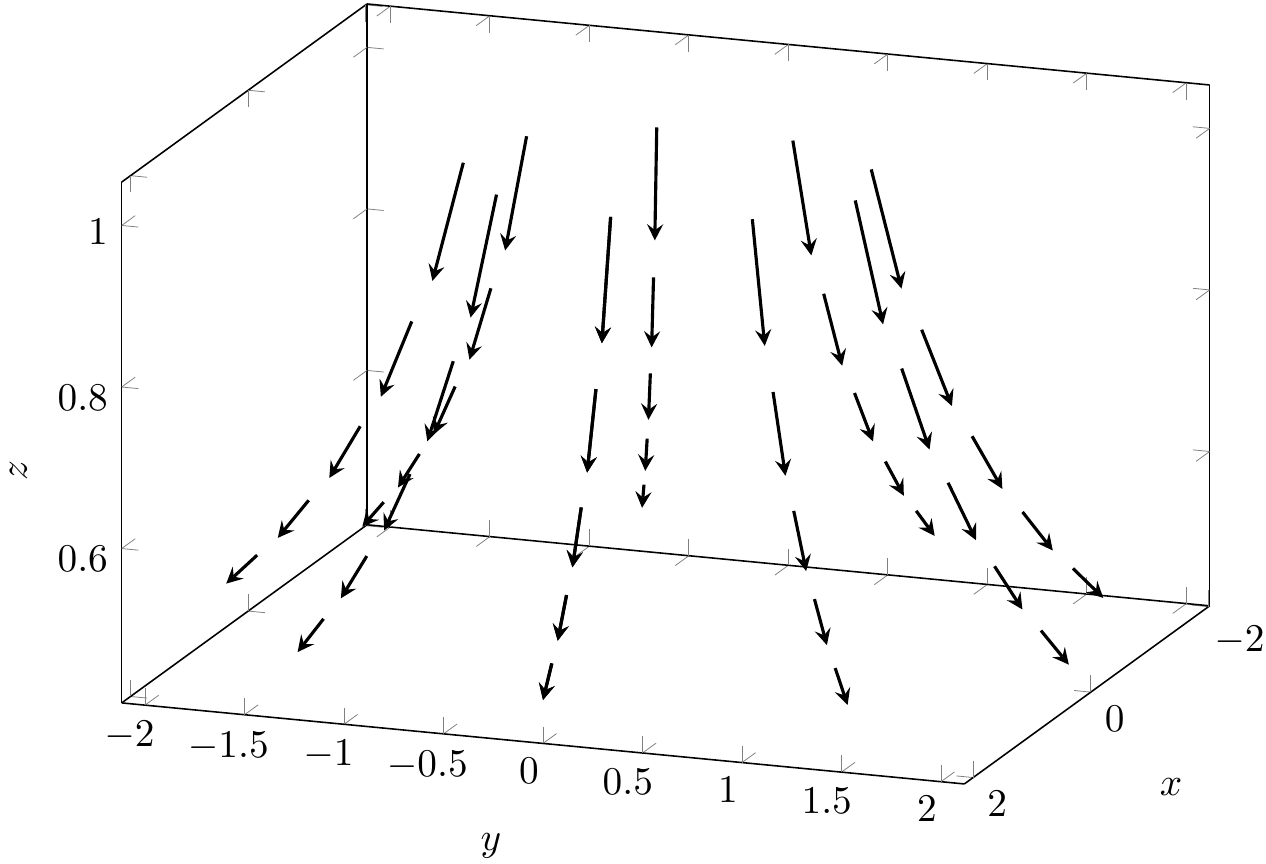} &
    \includegraphics[height=2.1cm,width=\wklength,keepaspectratio]{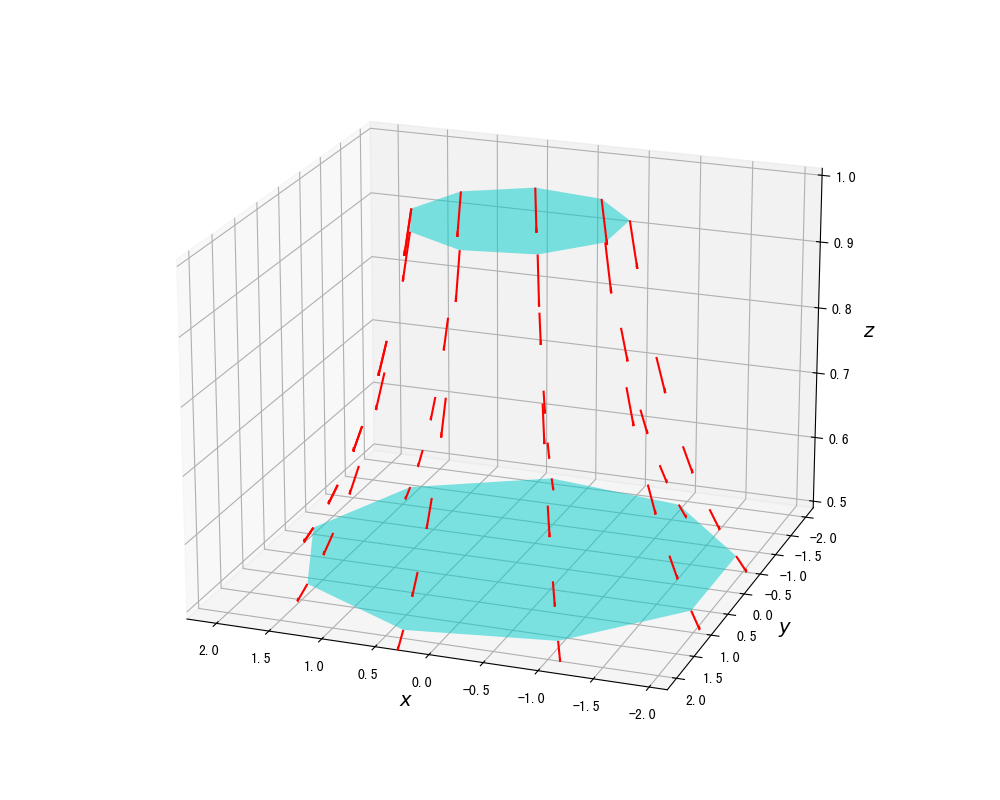} &
    \includegraphics[height=2.1cm,width=\wklength,keepaspectratio]{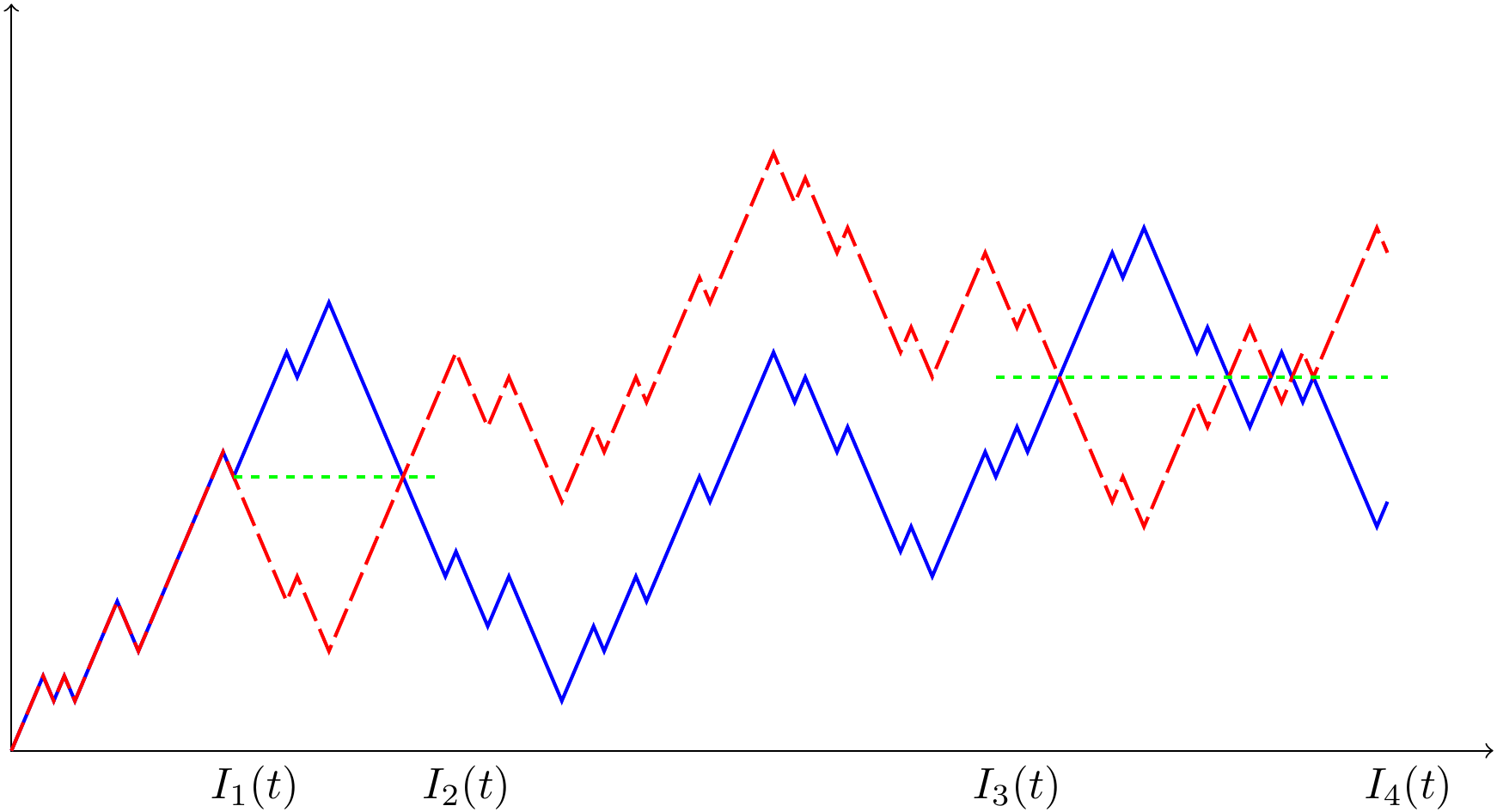} &
    \includegraphics[height=2.1cm,width=\wklength,keepaspectratio]{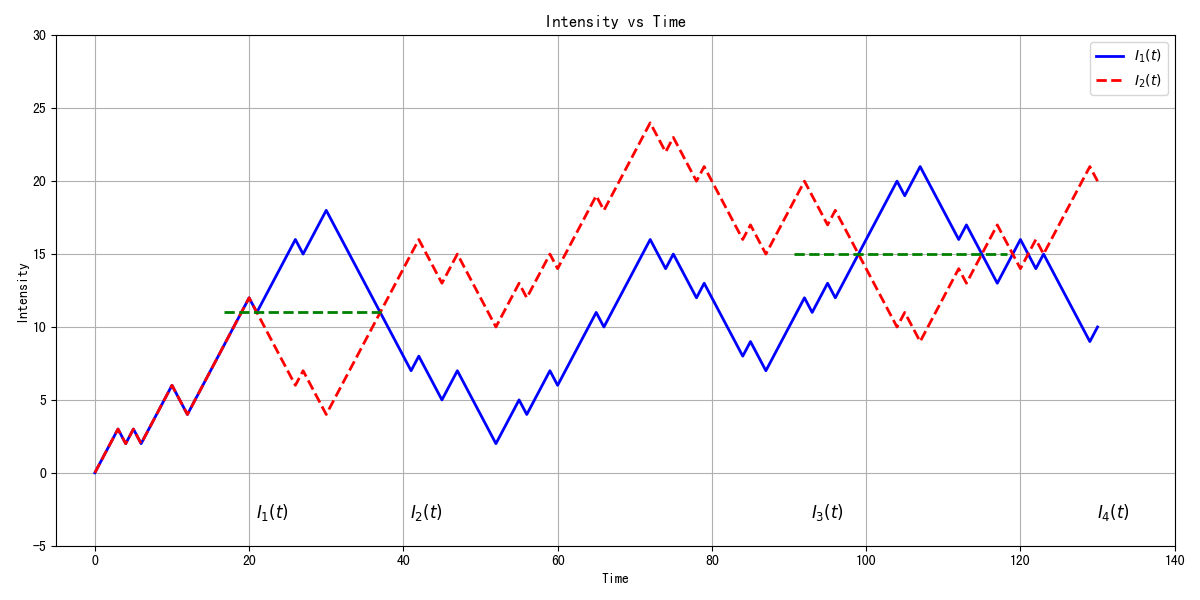} \\
    \midrule
    \includegraphics[height=2.1cm,width=\wklength,keepaspectratio]{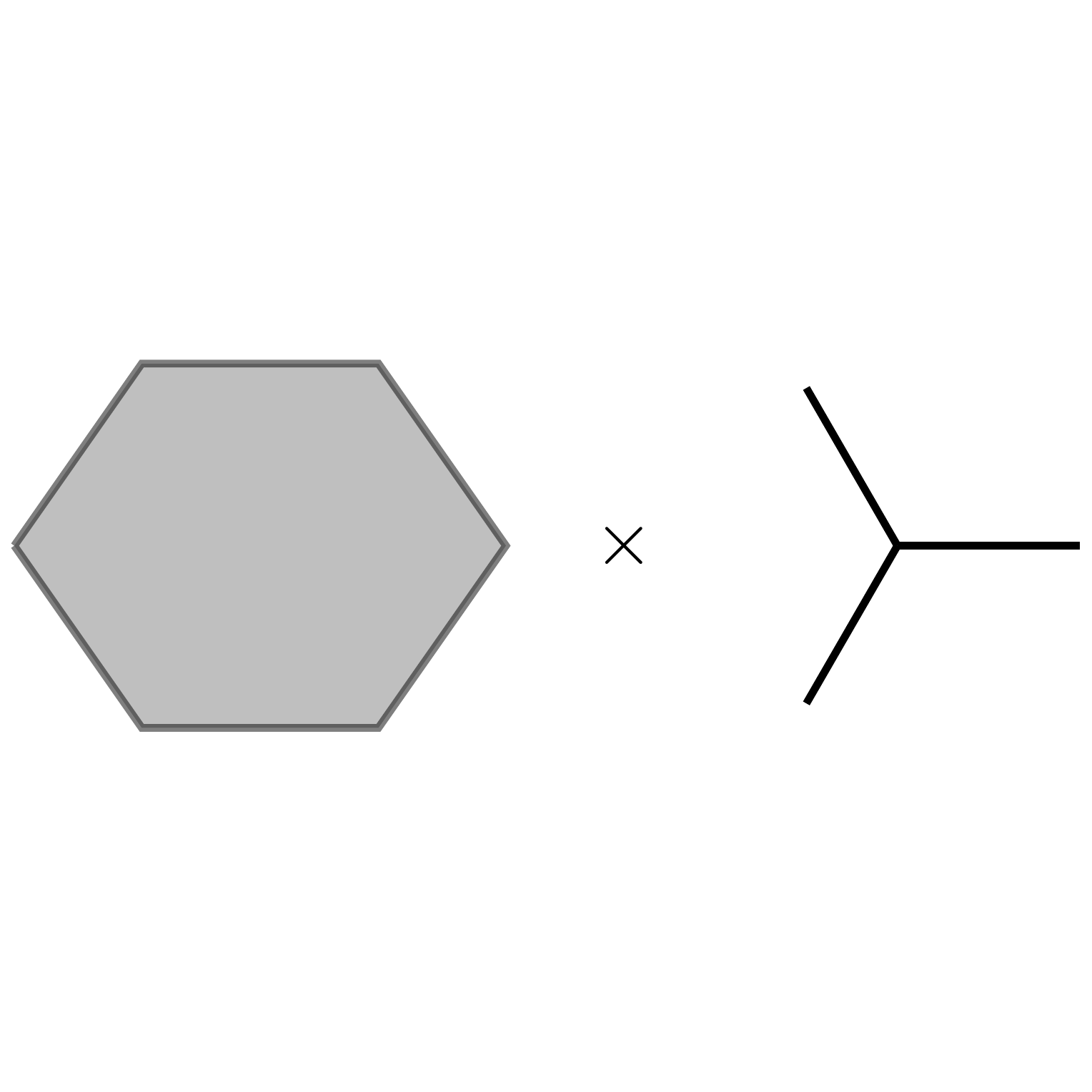} &
    \includegraphics[height=2.1cm,width=\wklength,keepaspectratio]{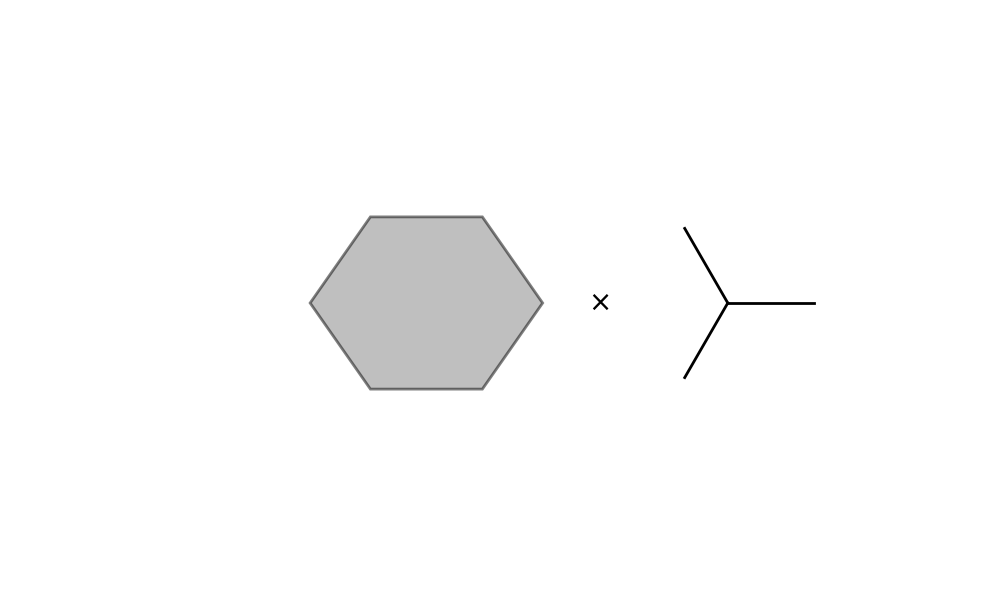} &
    \includegraphics[height=2.1cm,width=\wklength,keepaspectratio]{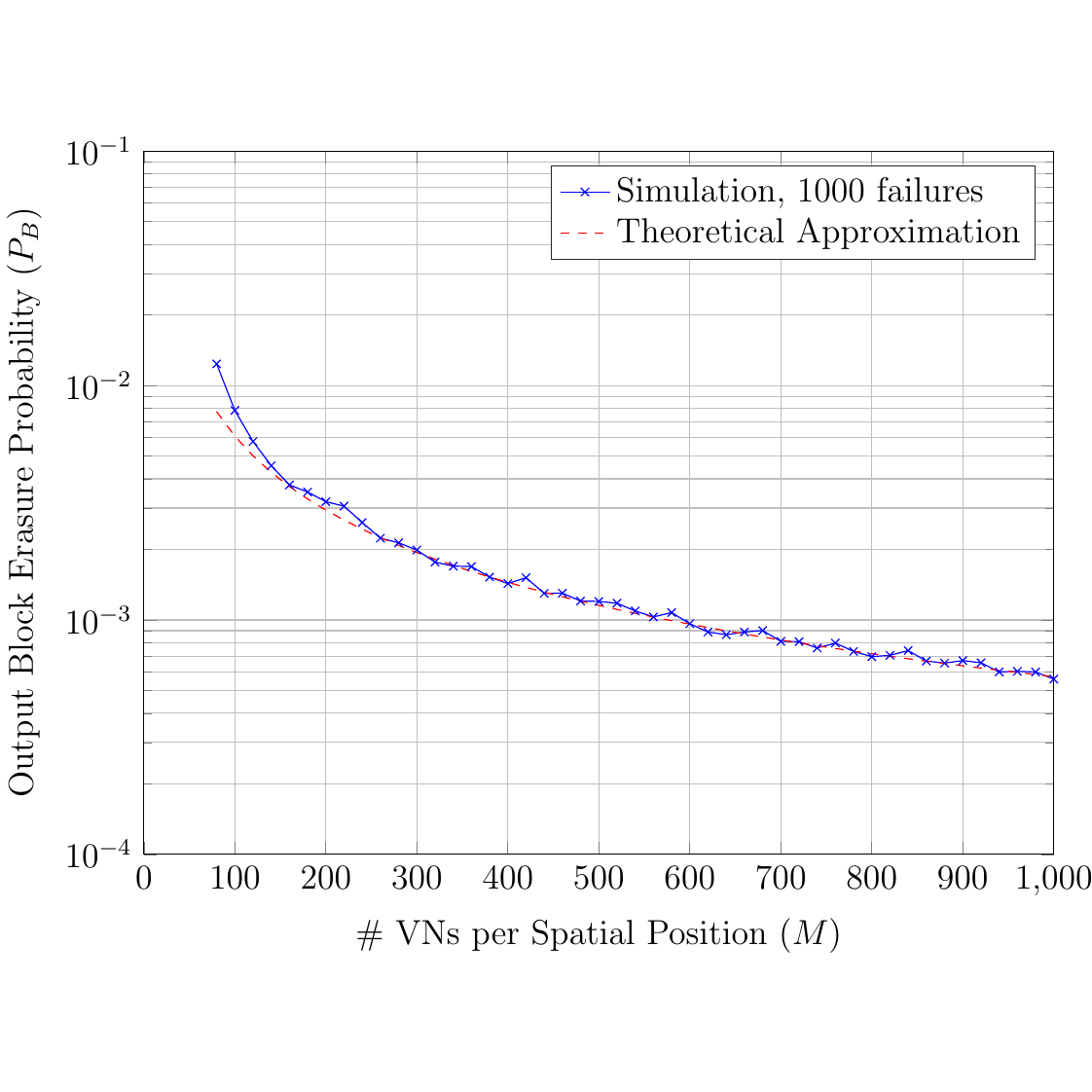} &
    \includegraphics[height=2.1cm,width=\wklength,keepaspectratio]{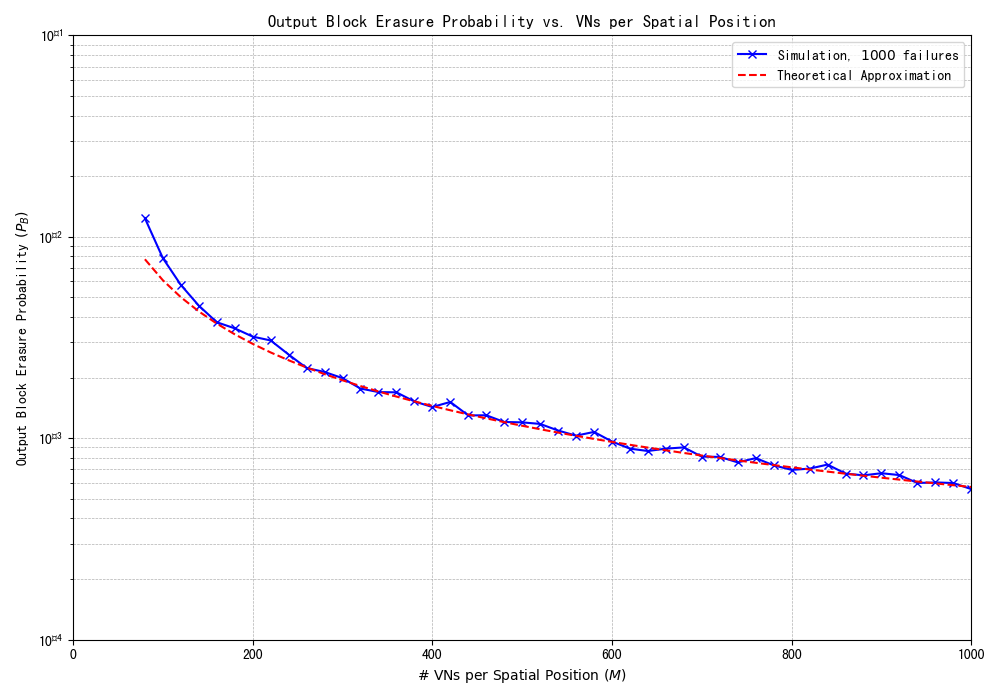} &
    \includegraphics[height=2.1cm,width=\wklength,keepaspectratio]{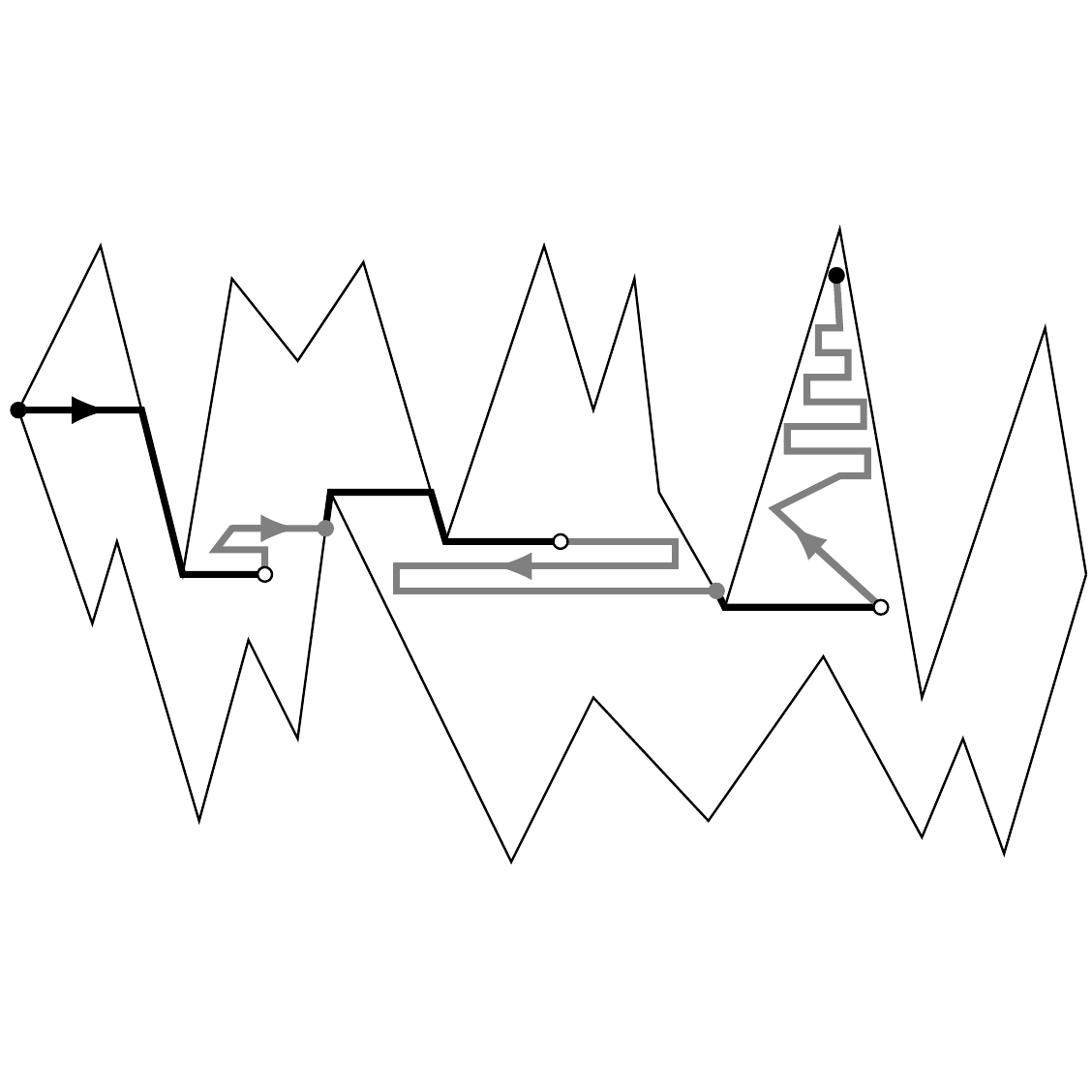} &
    \includegraphics[height=2.1cm,width=\wklength,keepaspectratio]{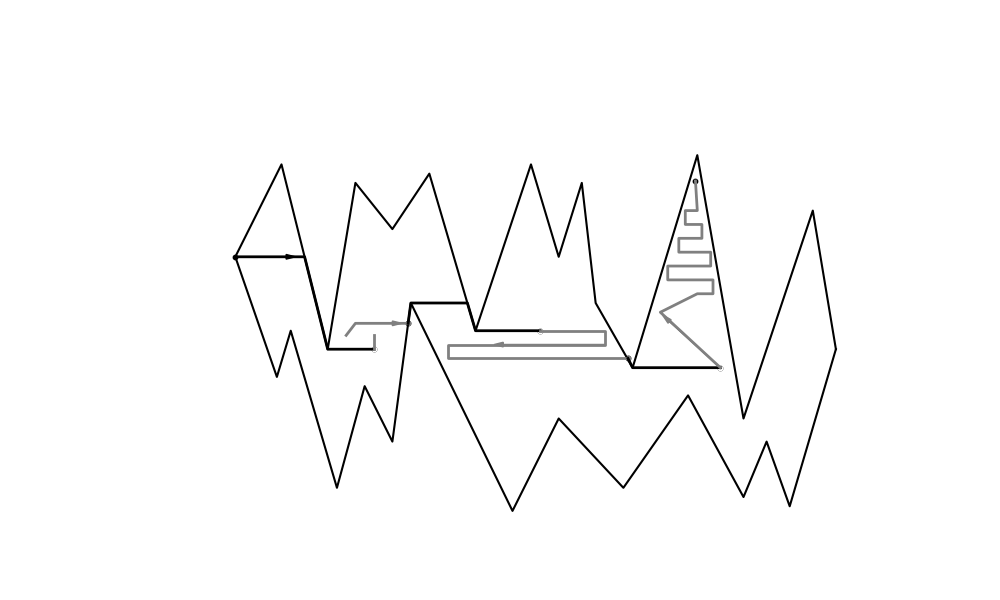} \\
    \midrule
    \includegraphics[height=2.1cm,width=\wklength,keepaspectratio]{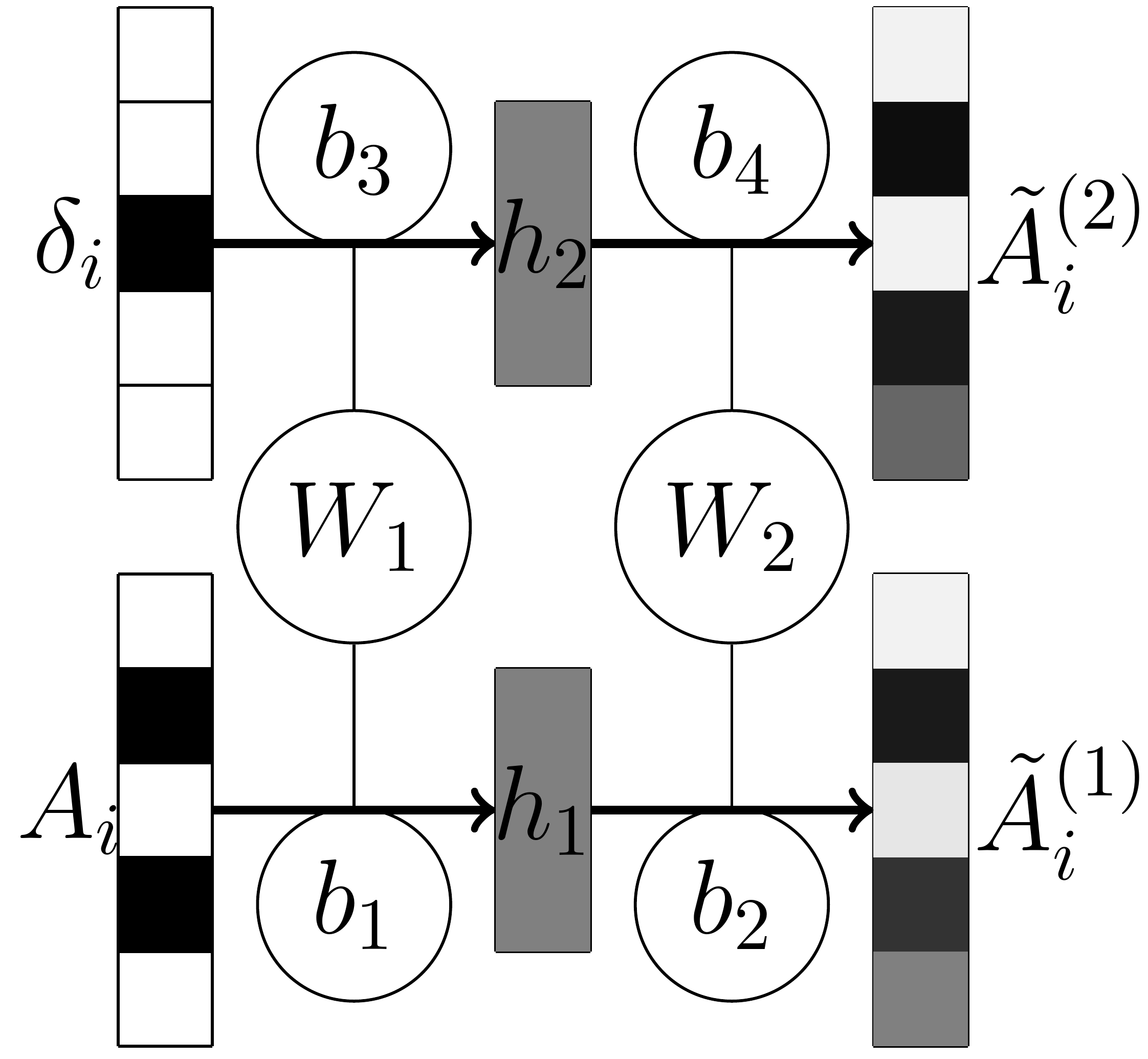} &
    \includegraphics[height=2.1cm,width=\wklength,keepaspectratio]{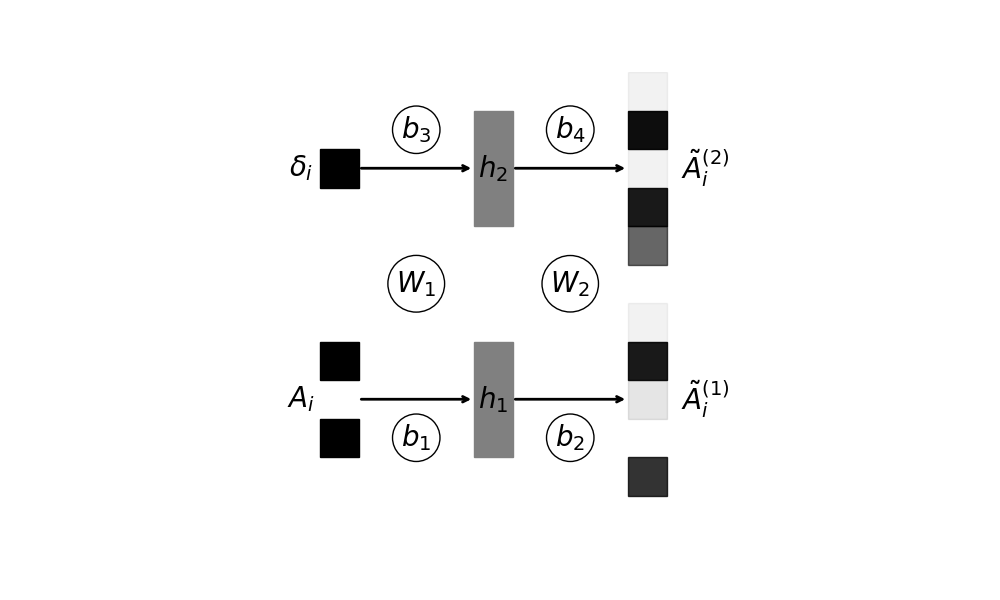} &
    \includegraphics[height=2.1cm,width=\wklength,keepaspectratio]{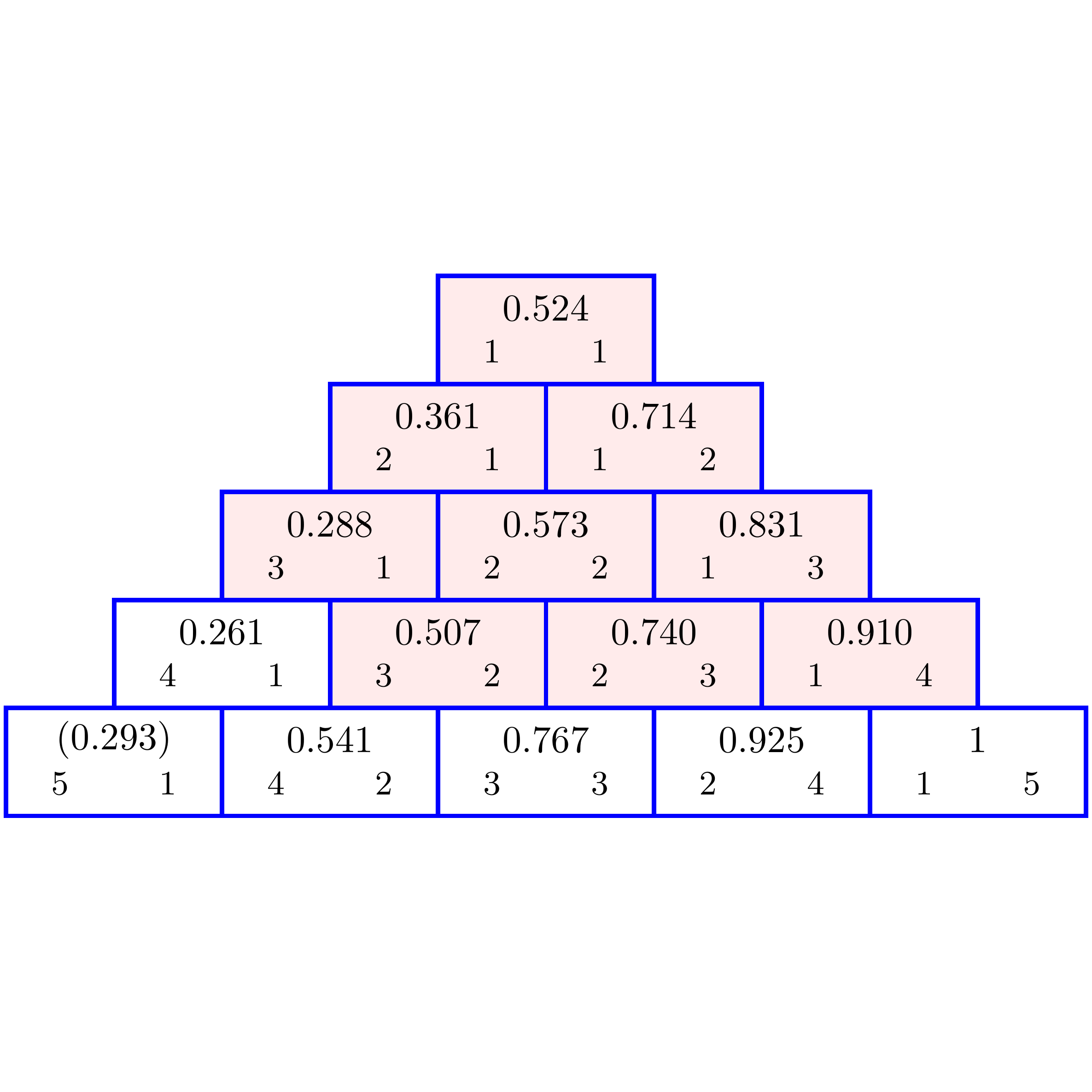} &
    \includegraphics[height=2.1cm,width=\wklength,keepaspectratio]{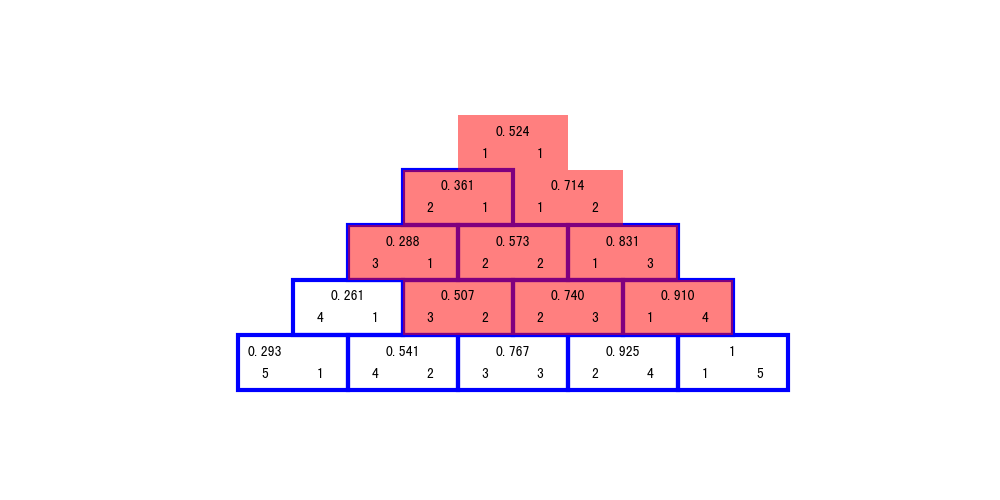} &
    \includegraphics[height=2.1cm,width=\wklength,keepaspectratio]{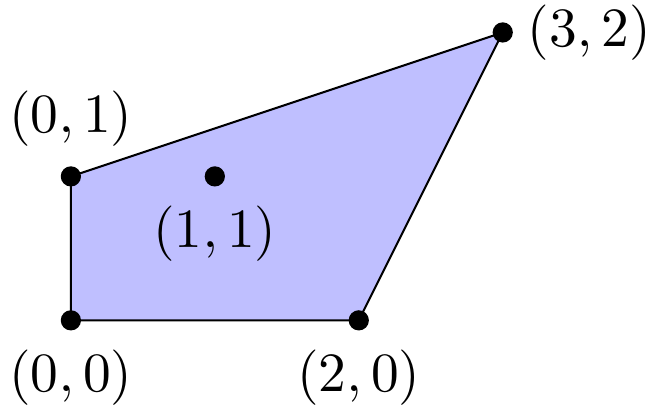} &
    \includegraphics[height=2.1cm,width=\wklength,keepaspectratio]{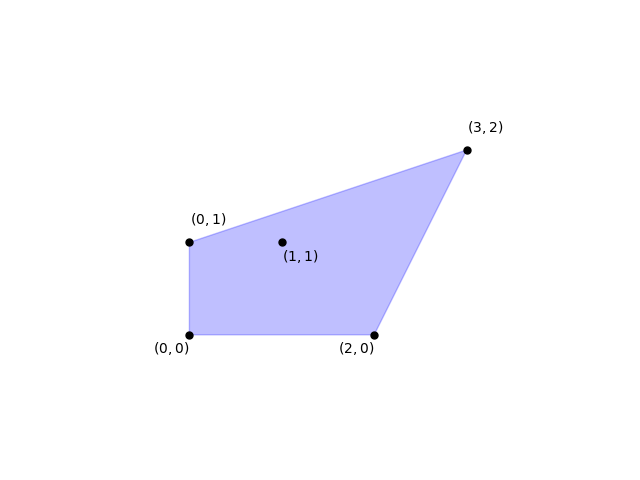} \\
    \midrule
    \includegraphics[height=2.1cm,width=\wklength,keepaspectratio]{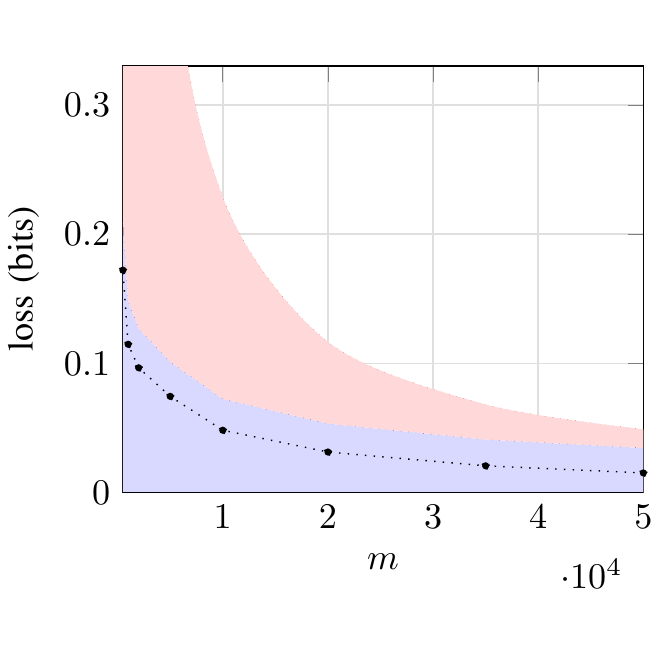} &
    \includegraphics[height=2.1cm,width=\wklength,keepaspectratio]{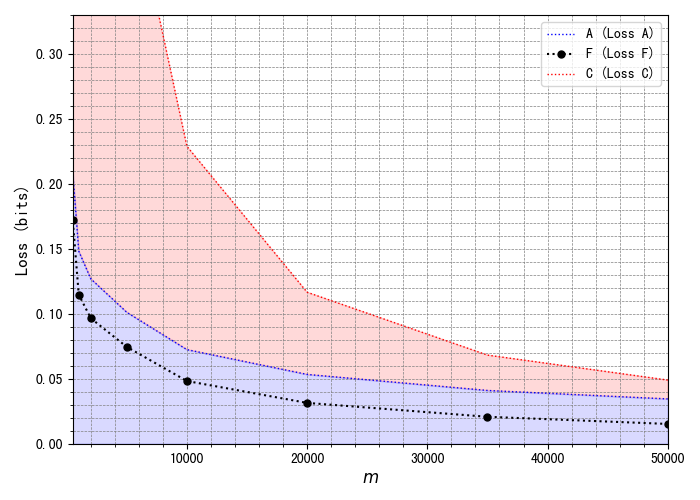} &
    \includegraphics[height=2.1cm,width=\wklength,keepaspectratio]{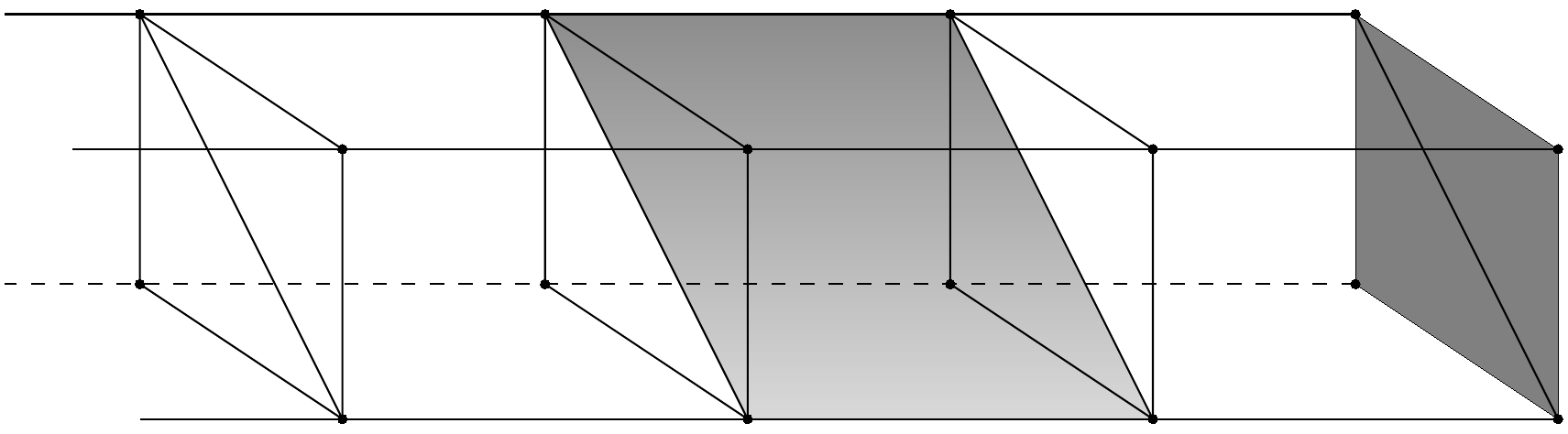} &
    \includegraphics[height=2.1cm,width=\wklength,keepaspectratio]{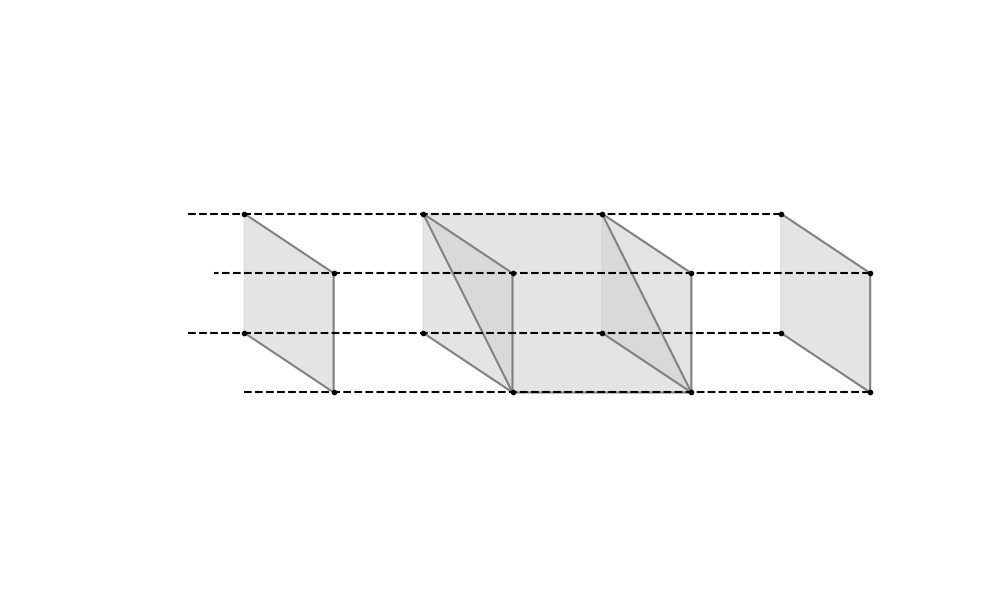} &
    \includegraphics[height=2.1cm,width=\wklength,keepaspectratio]{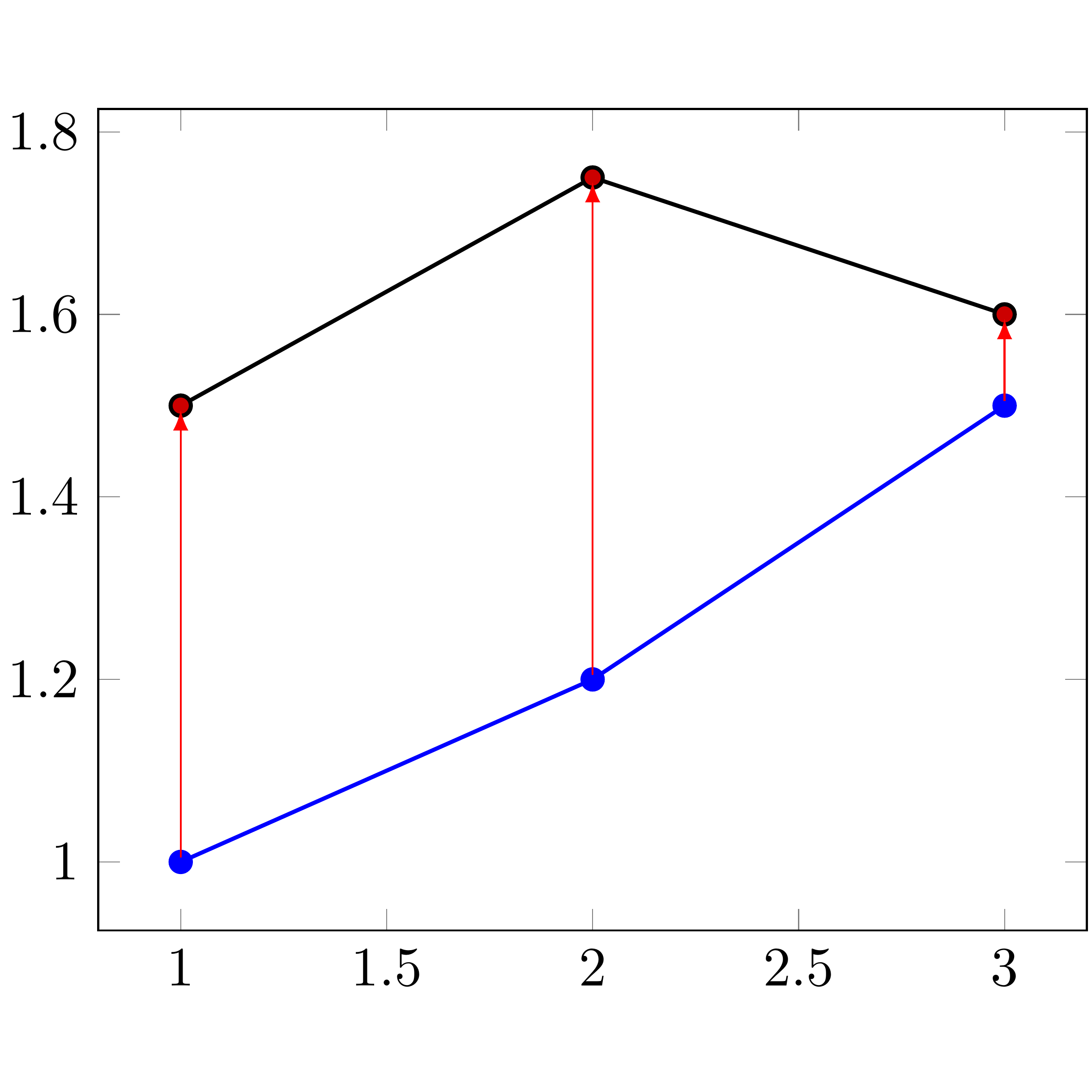} &
    \includegraphics[height=2.1cm,width=\wklength,keepaspectratio]{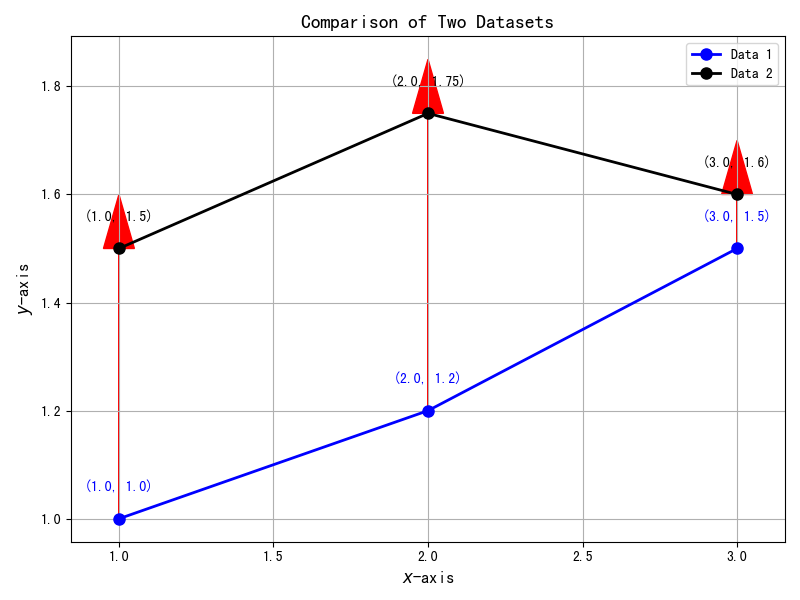} \\
    \bottomrule
  \end{tabular}
  \caption{Comparison of images generated from the original TiKZ code and the translated Python code.}%
  \label{fig:tikz2python_examples}
\end{table*}

%% file: appendix/synthesize_problem_prompt.tex
\begin{figure}[htbp]
\begin{tcolorbox}
[colback=wkpurple!50!white,colframe=wkpurple!95!black,title=\textcolor{black}{Problem Synthesis Prompt}]
\begin{small}

Please create a **math reasoning question** for a K-12 audience based on the image generated by the following {} code. The question must adhere to these criteria:

\vspace{4mm}

1. **Image Engaging**: The question must utilize visible patterns, shapes, numbers, or other elements present in the image to engage reasoning skills.

\vspace{2mm}

2. **Single Question**: Write a single, standalone question. The question should be concise and self-contained, without any subparts. You do not need to provide an answer to the question.

\vspace{2mm}

3. **Self-Sufficiency**: The recipient will only see the image, not the code. Include any essential details from the code (e.g., coordinates, hidden axes, specific data points, or labels) that are necessary for solving the question but may not be visible in the image.

\vspace{2mm}

4. **Solvability**: Ensure the question can be solved using only the visible information in the image and the question text.

\vspace{4mm}

Below is the {} code that generates the image:

\begin{verbatim}
```python/tikz\end{verbatim}
[Image Code]
\begin{verbatim}```
\end{verbatim}
\vspace{4mm}
\#\#\# Question:

\end{small}
\end{tcolorbox}
\caption{Prompt for synthesizing math reasoning problems based on synthesized images.}
\label{fig:synthesize_problem_prompt}
\end{figure}

%% file: appendix/img2code_examples.tex
\begin{table*}[htb]
  \scriptsize
  \pgfmathsetlength{\wklength}{\textwidth/6 -2\tabcolsep}
\begin{tabular}{c|c|c||c|c|c}
    \toprule
    Origin Image & Initial Model & Final Model & Origin Image & Initial Model & Final Model \\
    \midrule
    \includegraphics[height=2.1cm,width=\wklength,keepaspectratio]{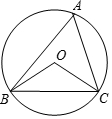} &
    \includegraphics[height=2.1cm,width=\wklength,keepaspectratio]{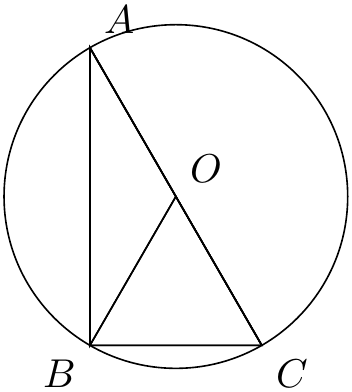} &
    \includegraphics[height=2.1cm,width=\wklength,keepaspectratio]{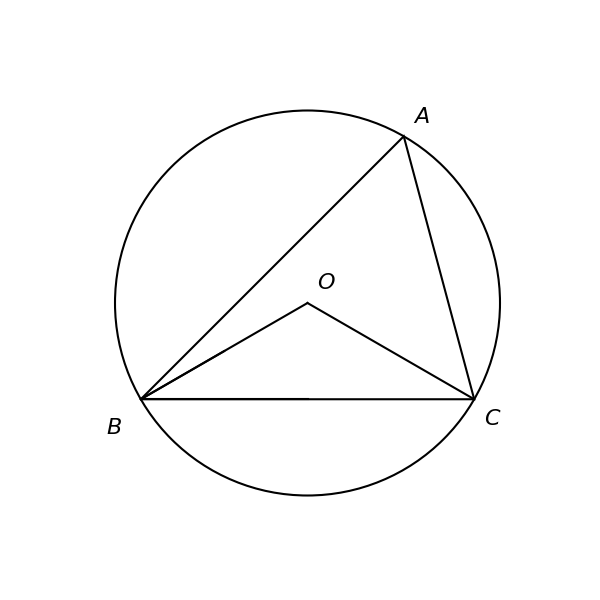} &
    \includegraphics[height=2.1cm,width=\wklength,keepaspectratio]{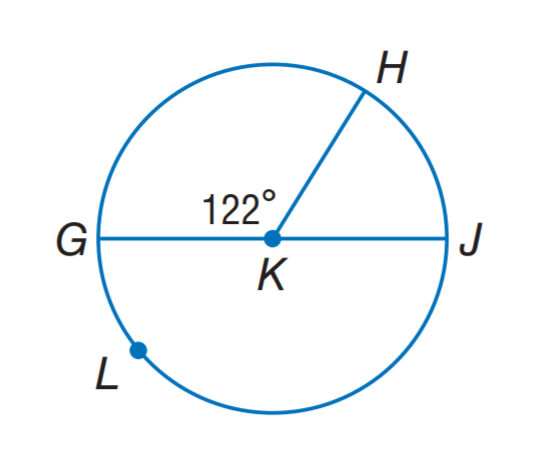} &
    \includegraphics[height=2.1cm,width=\wklength,keepaspectratio]{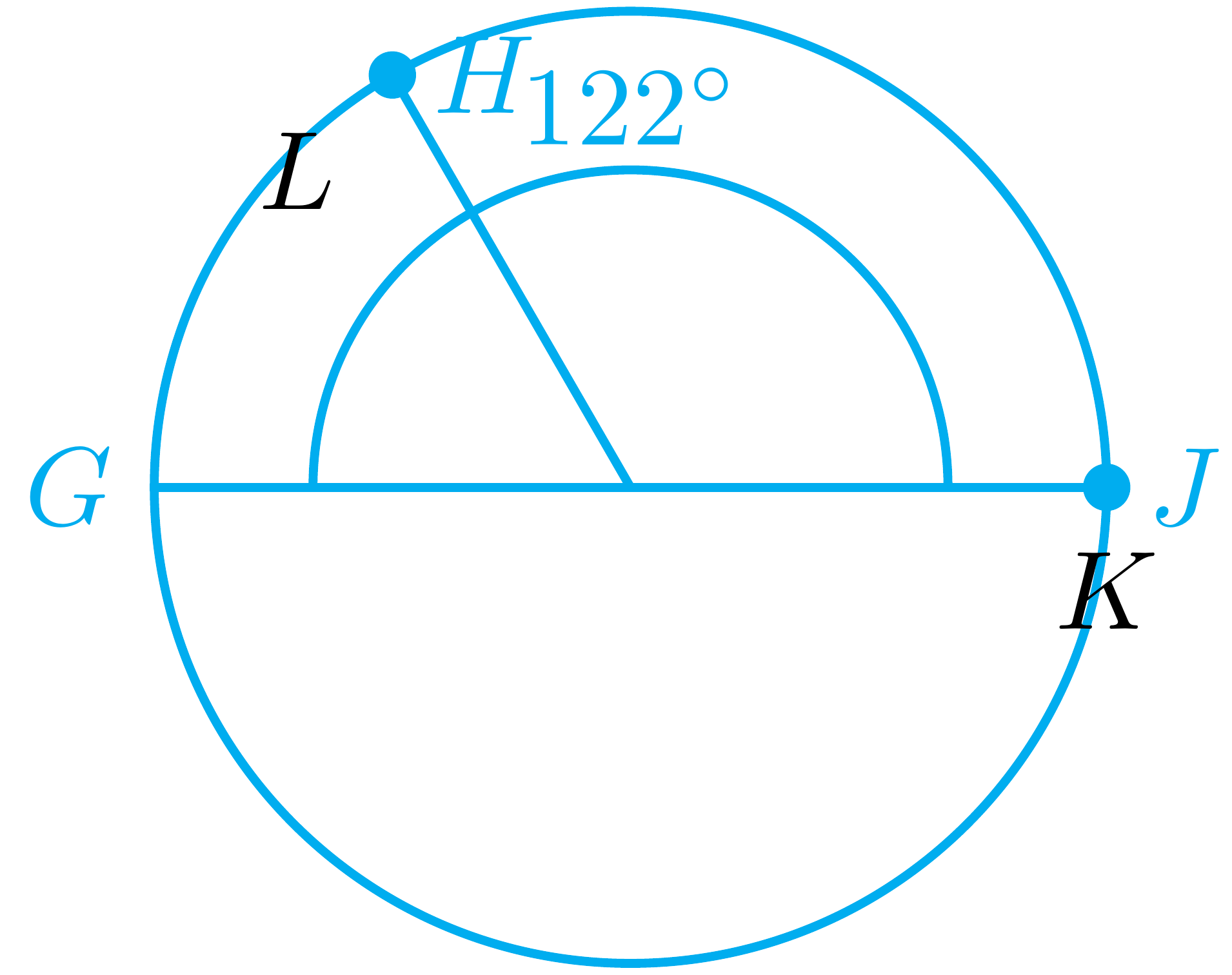} &
    \includegraphics[height=2.1cm,width=\wklength,keepaspectratio]{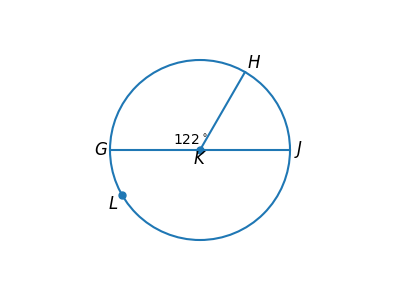} \\
    \midrule
    \includegraphics[height=2.1cm,width=\wklength,keepaspectratio]{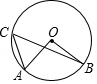} &
    \includegraphics[height=2.1cm,width=\wklength,keepaspectratio]{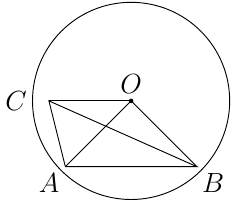} &
    \includegraphics[height=2.1cm,width=\wklength,keepaspectratio]{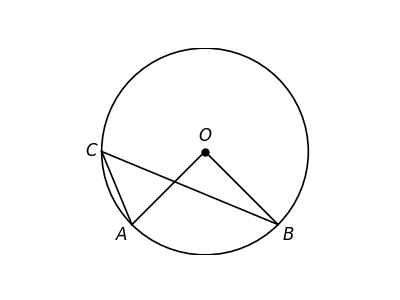} &
    \includegraphics[height=2.1cm,width=\wklength,keepaspectratio]{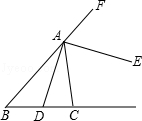} &
    \includegraphics[height=2.1cm,width=\wklength,keepaspectratio]{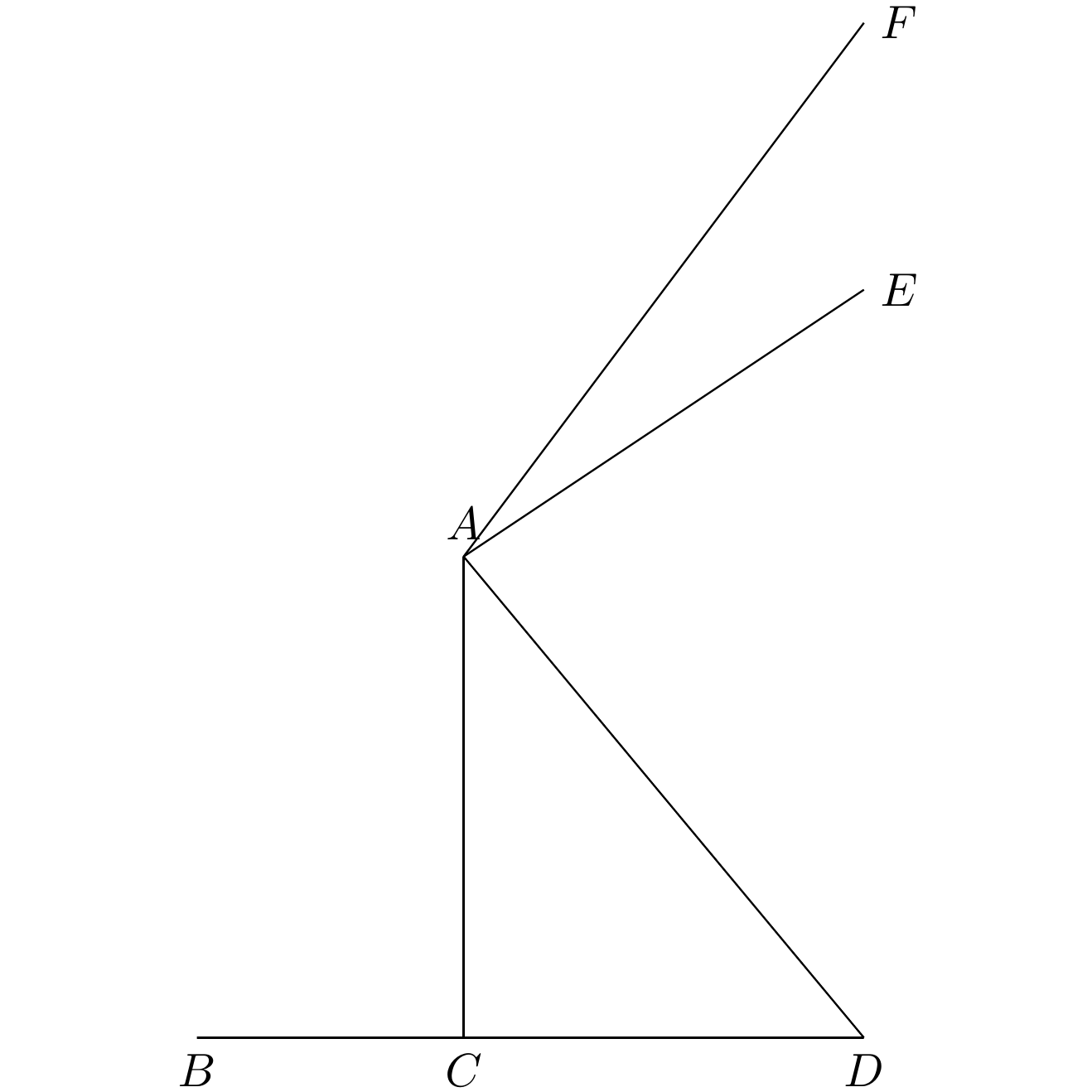} &
    \includegraphics[height=2.1cm,width=\wklength,keepaspectratio]{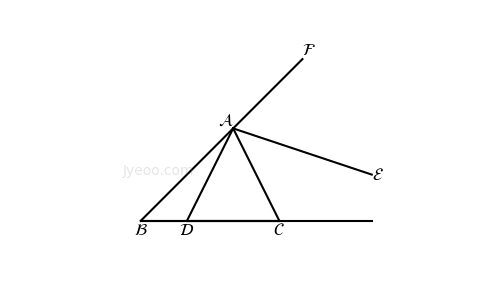} \\
\bottomrule
  \end{tabular}
  \caption{Comparison of image-to-code performance between the initial and final models.}%
  \label{fig:img2code_examples_1}
\end{table*}

\begin{table*}[htb]
  \scriptsize
  \pgfmathsetlength{\wklength}{\textwidth/6 -2\tabcolsep}
  \begin{tabular}{c|c|c||c|c|c}
    \toprule
    Origin Image & Initial Model & Final Model & Origin Image & Initial Model & Final Model \\
    \midrule
    \includegraphics[height=2.1cm,width=\wklength,keepaspectratio]{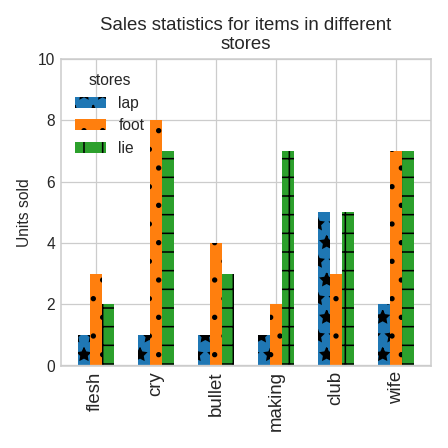} &
    \includegraphics[height=2.1cm,width=\wklength,keepaspectratio]{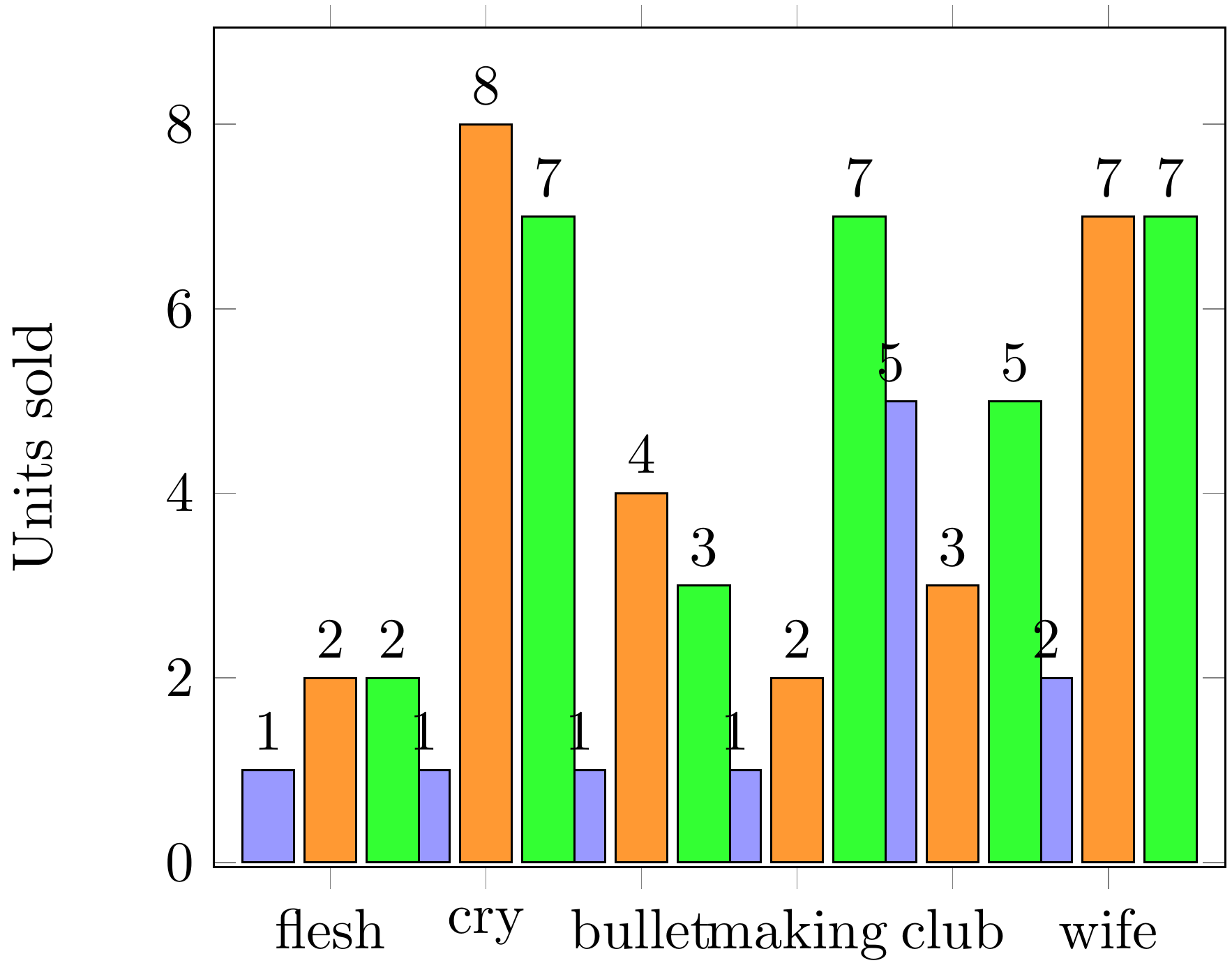} &
    \includegraphics[height=2.1cm,width=\wklength,keepaspectratio]{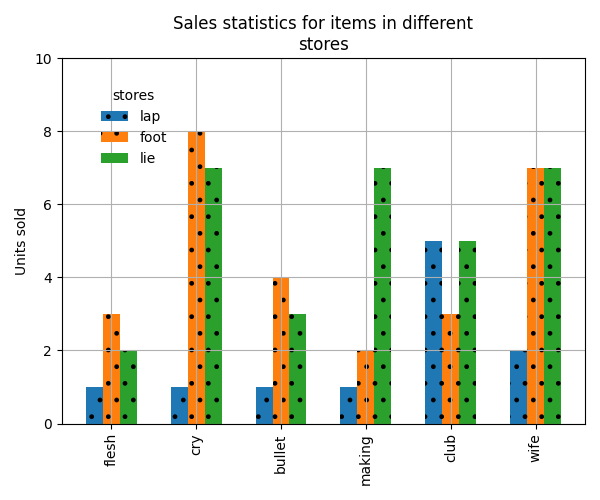} &
    \includegraphics[height=2.1cm,width=\wklength,keepaspectratio]{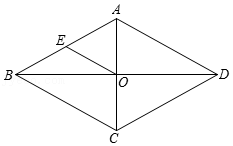} &
    \includegraphics[height=2.1cm,width=\wklength,keepaspectratio]{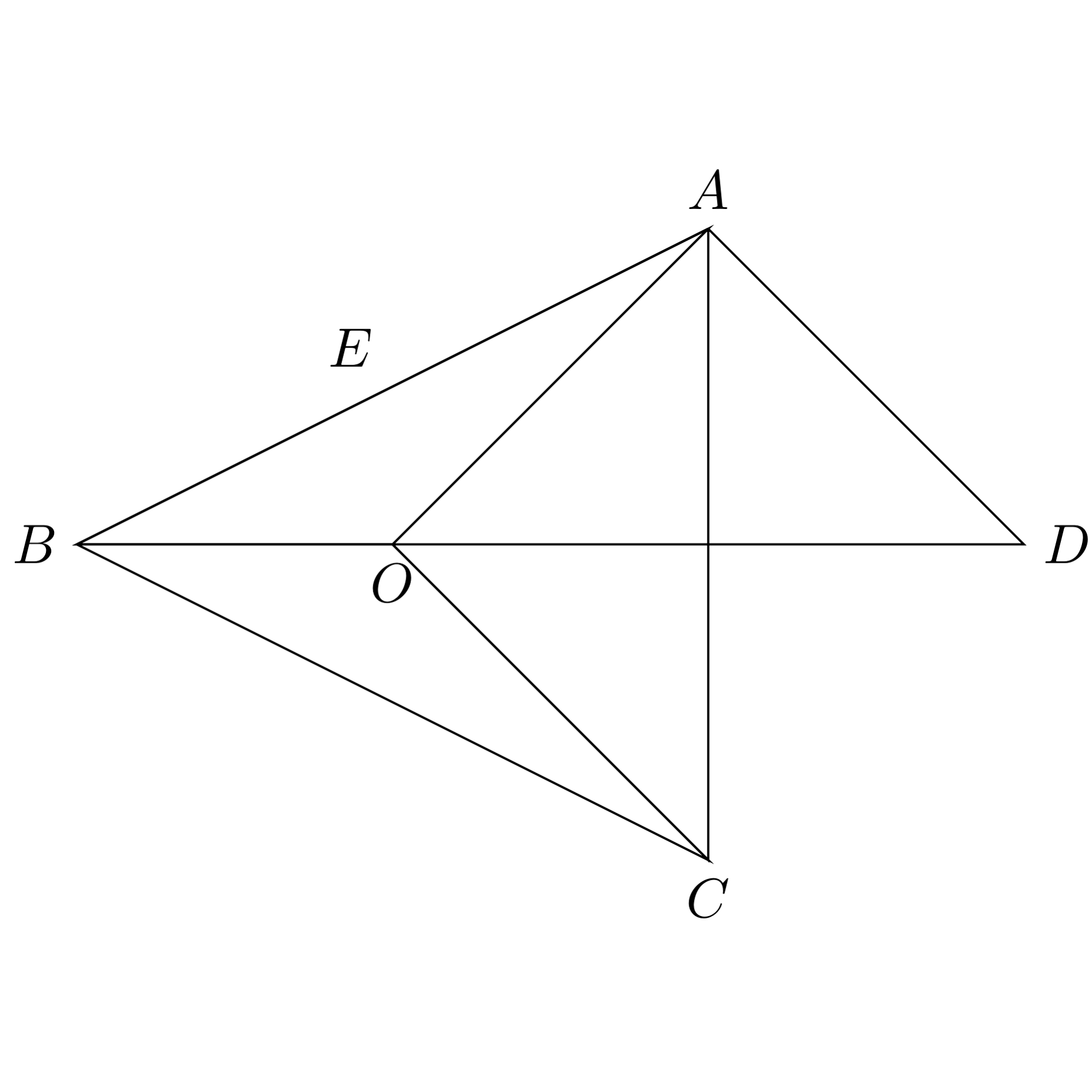} &
    \includegraphics[height=2.1cm,width=\wklength,keepaspectratio]{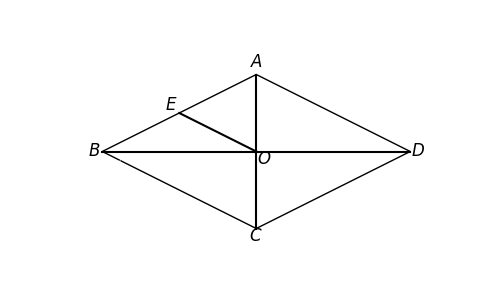} \\
    \midrule
    \includegraphics[height=2.1cm,width=\wklength,keepaspectratio]{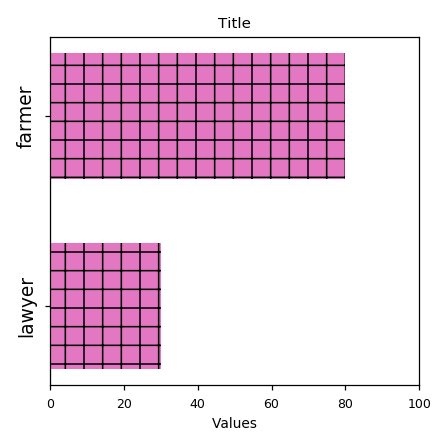} &
    \includegraphics[height=2.1cm,width=\wklength,keepaspectratio]{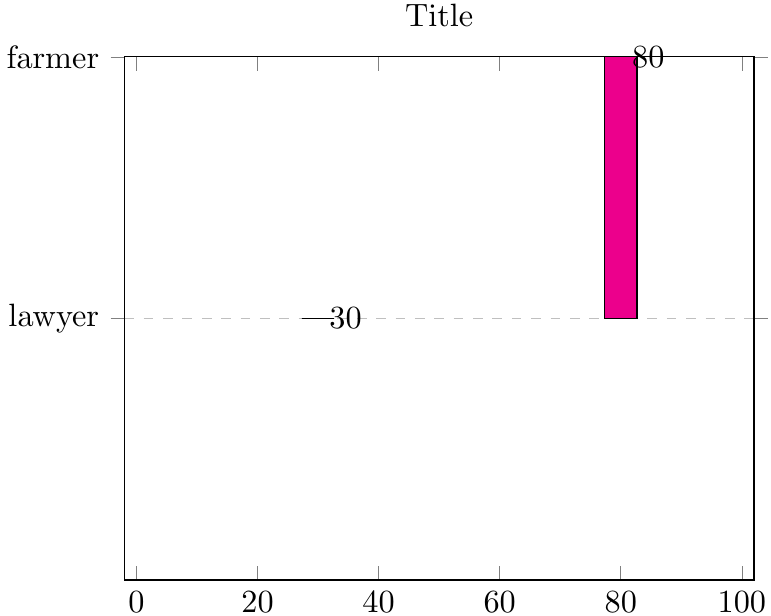} &
    \includegraphics[height=2.1cm,width=\wklength,keepaspectratio]{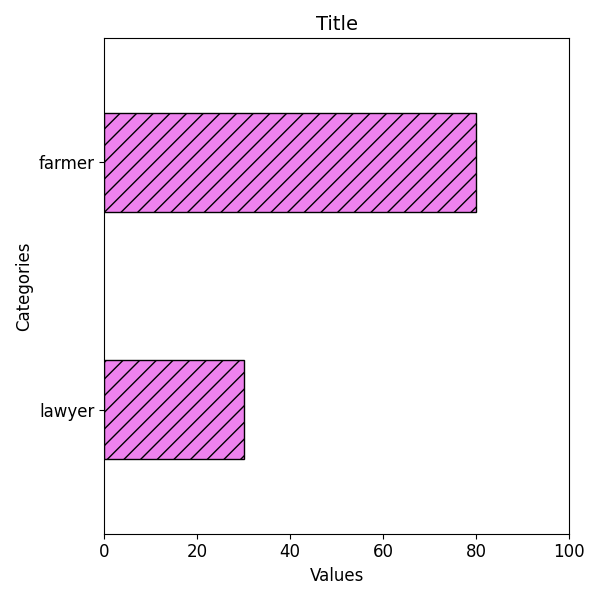} &
    \includegraphics[height=2.1cm,width=\wklength,keepaspectratio]{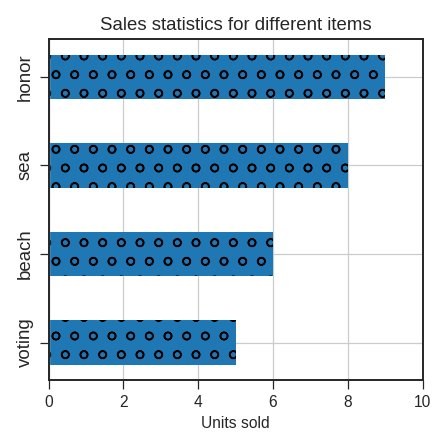} &
    \includegraphics[height=2.1cm,width=\wklength,keepaspectratio]{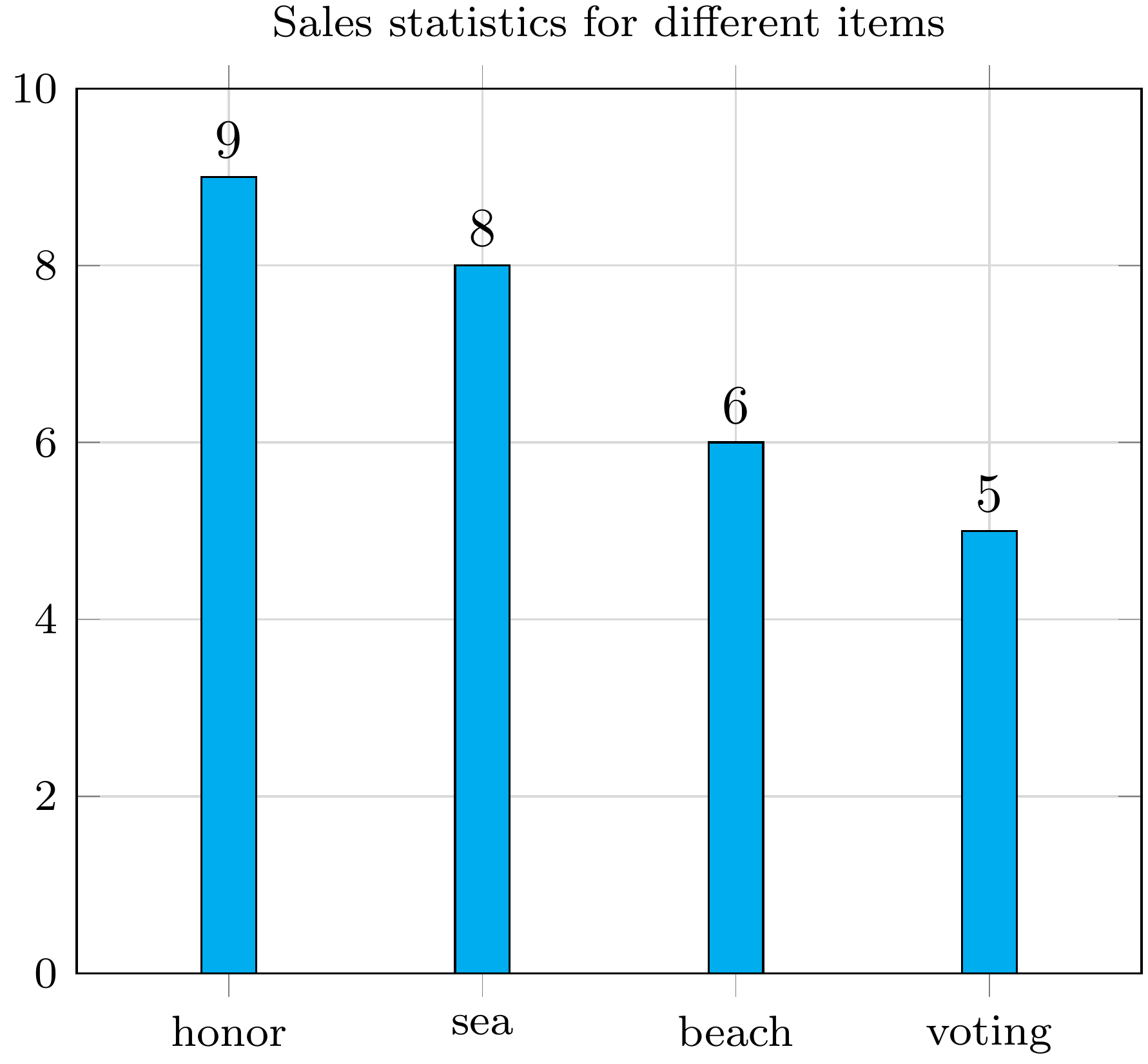} &
    \includegraphics[height=2.1cm,width=\wklength,keepaspectratio]{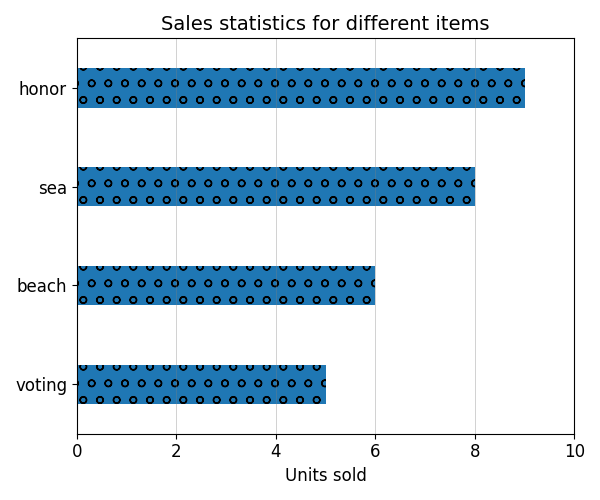} \\
\bottomrule
  \end{tabular}
  \caption{Comparison of image-to-code performance between the initial and final models.}%
  \label{fig:img2code_examples_2}
\end{table*}

\begin{table*}[htb]
  \scriptsize
  \pgfmathsetlength{\wklength}{\textwidth/6 -2\tabcolsep}
  \begin{tabular}{c|c|c||c|c|c}
    \toprule
    Origin Image & Initial Model & Final Model & Origin Image & Initial Model & Final Model \\
    \midrule
    \includegraphics[height=2.1cm,width=\wklength,keepaspectratio]{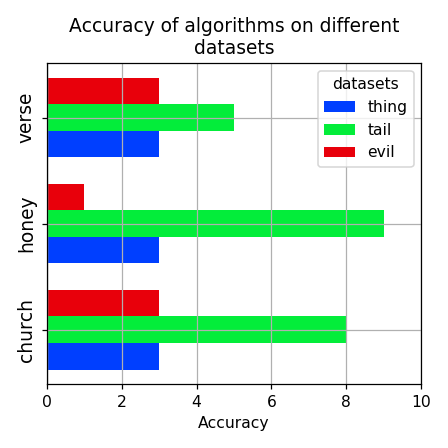} &
    \includegraphics[height=2.1cm,width=\wklength,keepaspectratio]{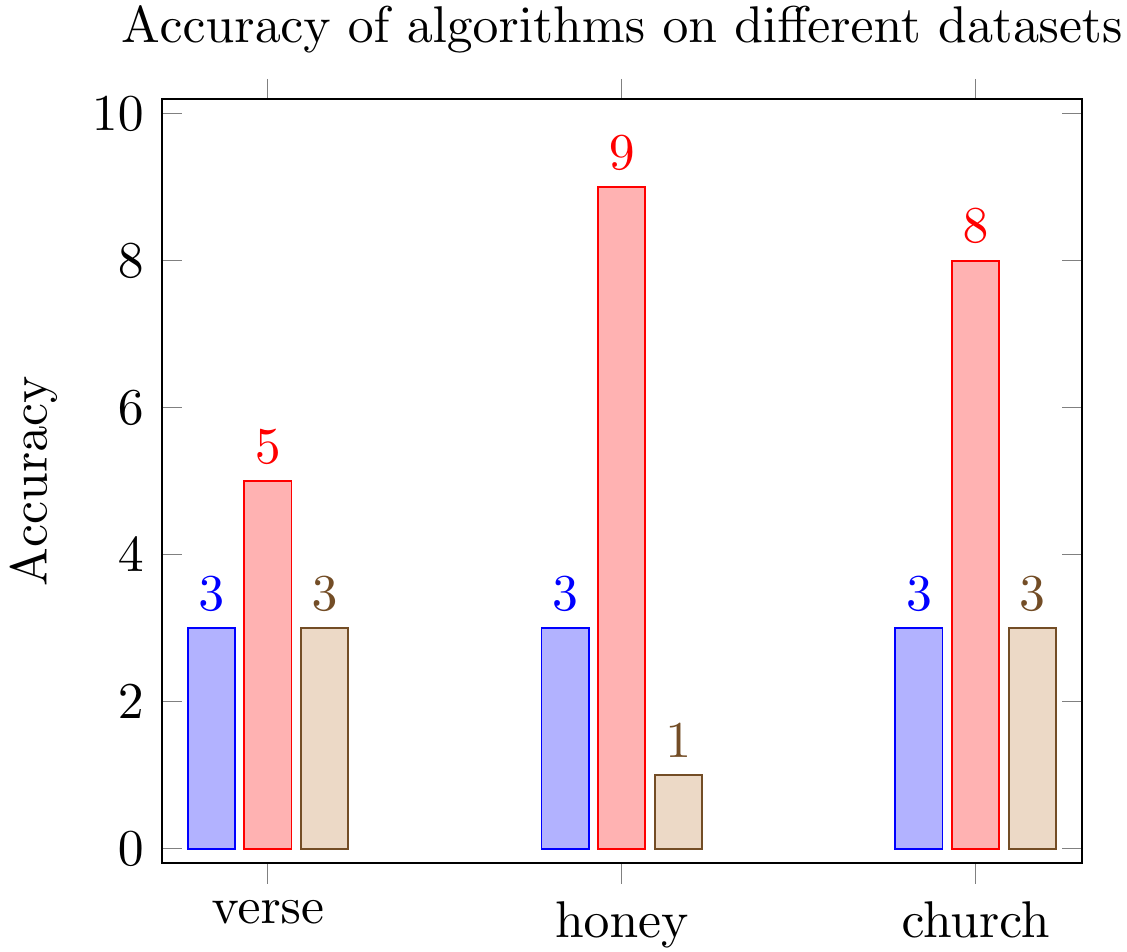} &
    \includegraphics[height=2.1cm,width=\wklength,keepaspectratio]{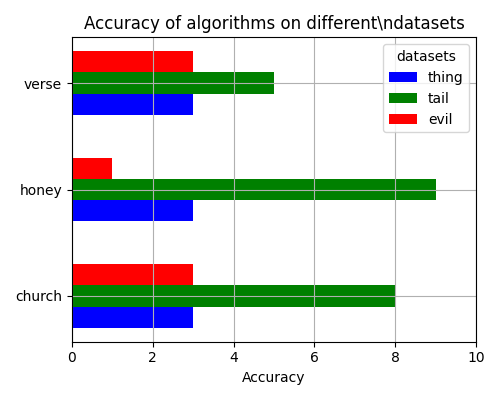} &
    \includegraphics[height=2.1cm,width=\wklength,keepaspectratio]{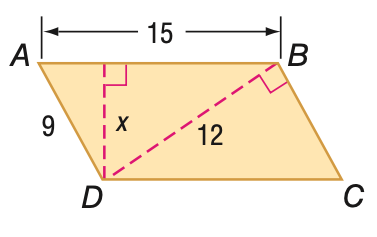} &
    \includegraphics[height=2.1cm,width=\wklength,keepaspectratio]{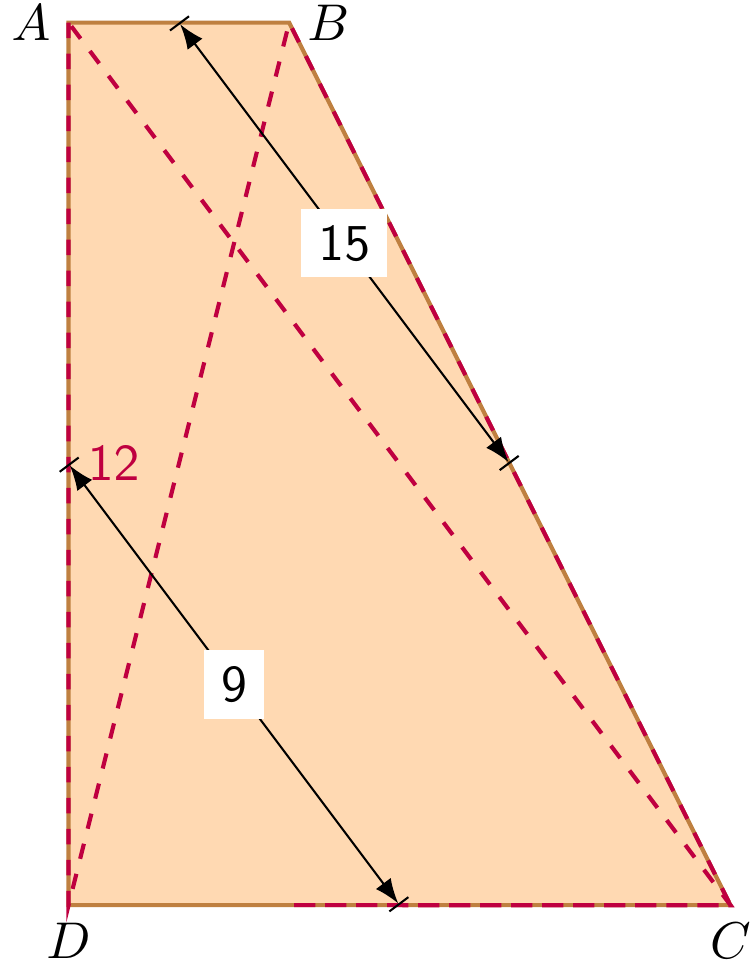} &
    \includegraphics[height=2.1cm,width=\wklength,keepaspectratio]{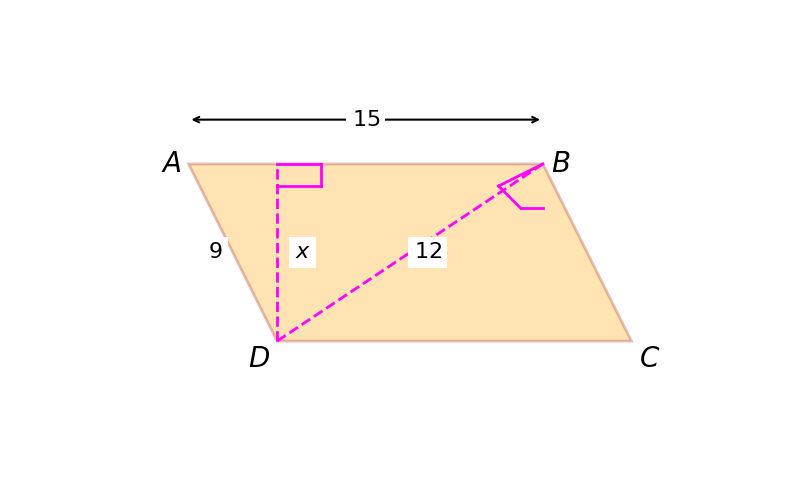} \\
    \midrule
    \includegraphics[height=2.1cm,width=\wklength,keepaspectratio]{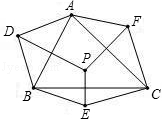} &
    \includegraphics[height=2.1cm,width=\wklength,keepaspectratio]{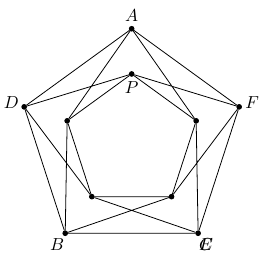} &
    \includegraphics[height=2.1cm,width=\wklength,keepaspectratio]{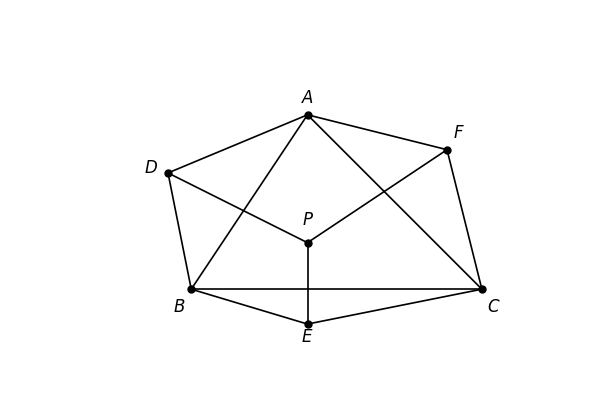} &
    \includegraphics[height=2.1cm,width=\wklength,keepaspectratio]{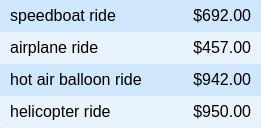} &
    \includegraphics[height=2.1cm,width=\wklength,keepaspectratio]{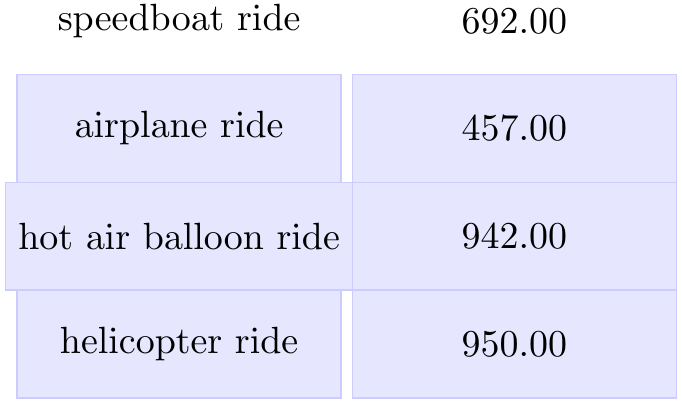} &
    \includegraphics[height=2.1cm,width=\wklength,keepaspectratio]{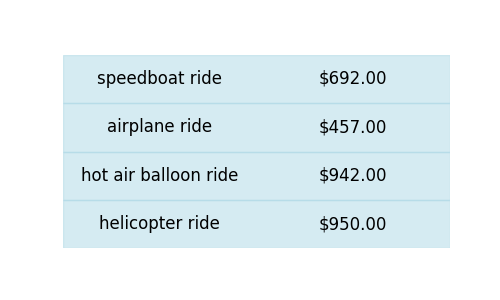} \\
\bottomrule
  \end{tabular}
  \caption{Comparison of image-to-code performance between the initial and final models.}%
  \label{fig:img2code_examples_3}
\end{table*}

\begin{table*}[htb]
  \scriptsize
  \pgfmathsetlength{\wklength}{\textwidth/6 -2\tabcolsep}
  \begin{tabular}{c|c|c||c|c|c}
    \toprule
    Origin Image & Initial Model & Final Model & Origin Image & Initial Model & Final Model \\
    \midrule
    \includegraphics[height=2.1cm,width=\wklength,keepaspectratio]{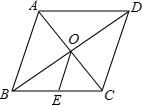} &
    \includegraphics[height=2.1cm,width=\wklength,keepaspectratio]{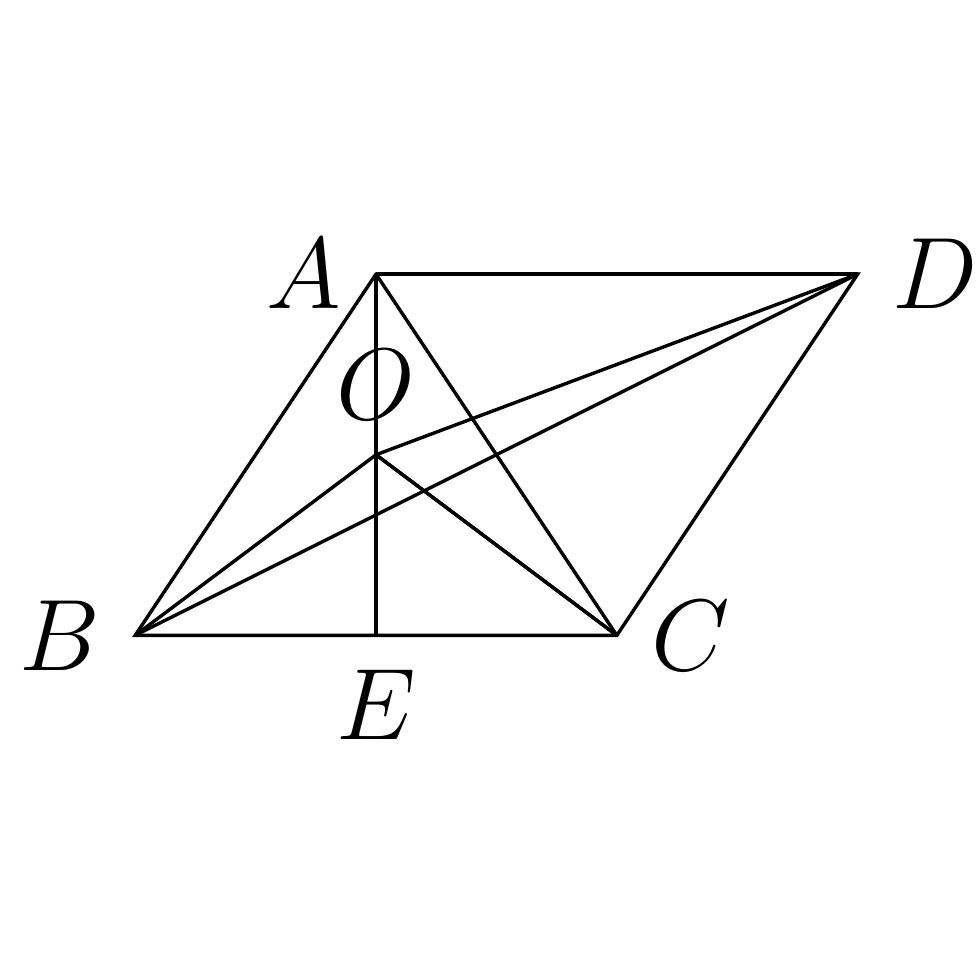} &
    \includegraphics[height=2.1cm,width=\wklength,keepaspectratio]{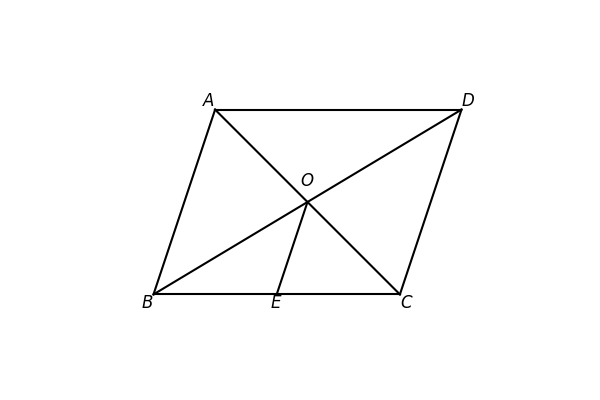} &
    \includegraphics[height=2.1cm,width=\wklength,keepaspectratio]{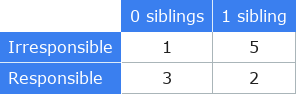} &
    \includegraphics[height=2.1cm,width=\wklength,keepaspectratio]{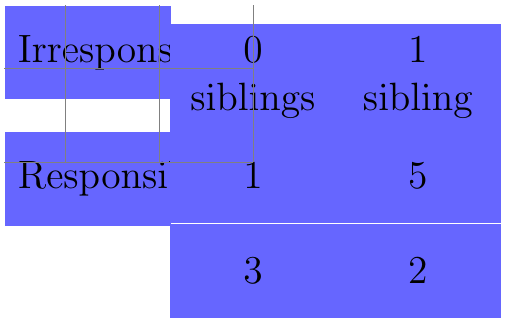} &
    \includegraphics[height=2.1cm,width=\wklength,keepaspectratio]{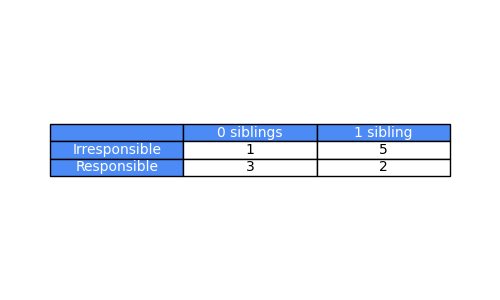} \\
    \midrule
    \includegraphics[height=2.1cm,width=\wklength,keepaspectratio]{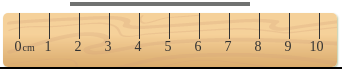} &
    \includegraphics[height=2.1cm,width=\wklength,keepaspectratio]{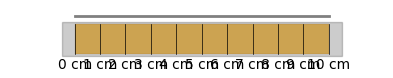} &
    \includegraphics[height=2.1cm,width=\wklength,keepaspectratio]{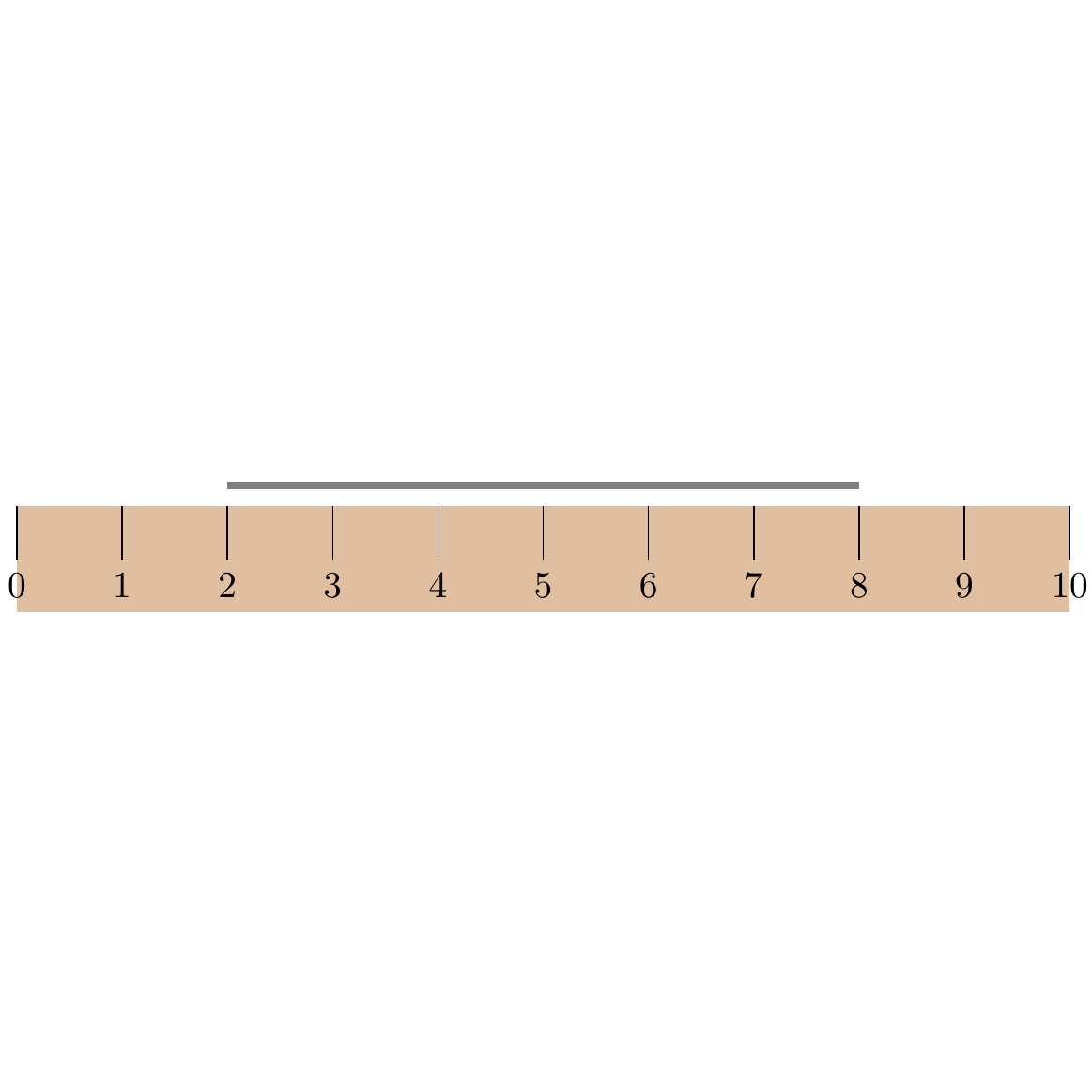} &
    \includegraphics[height=2.1cm,width=\wklength,keepaspectratio]{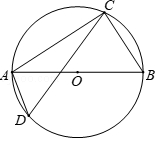} &
    \includegraphics[height=2.1cm,width=\wklength,keepaspectratio]{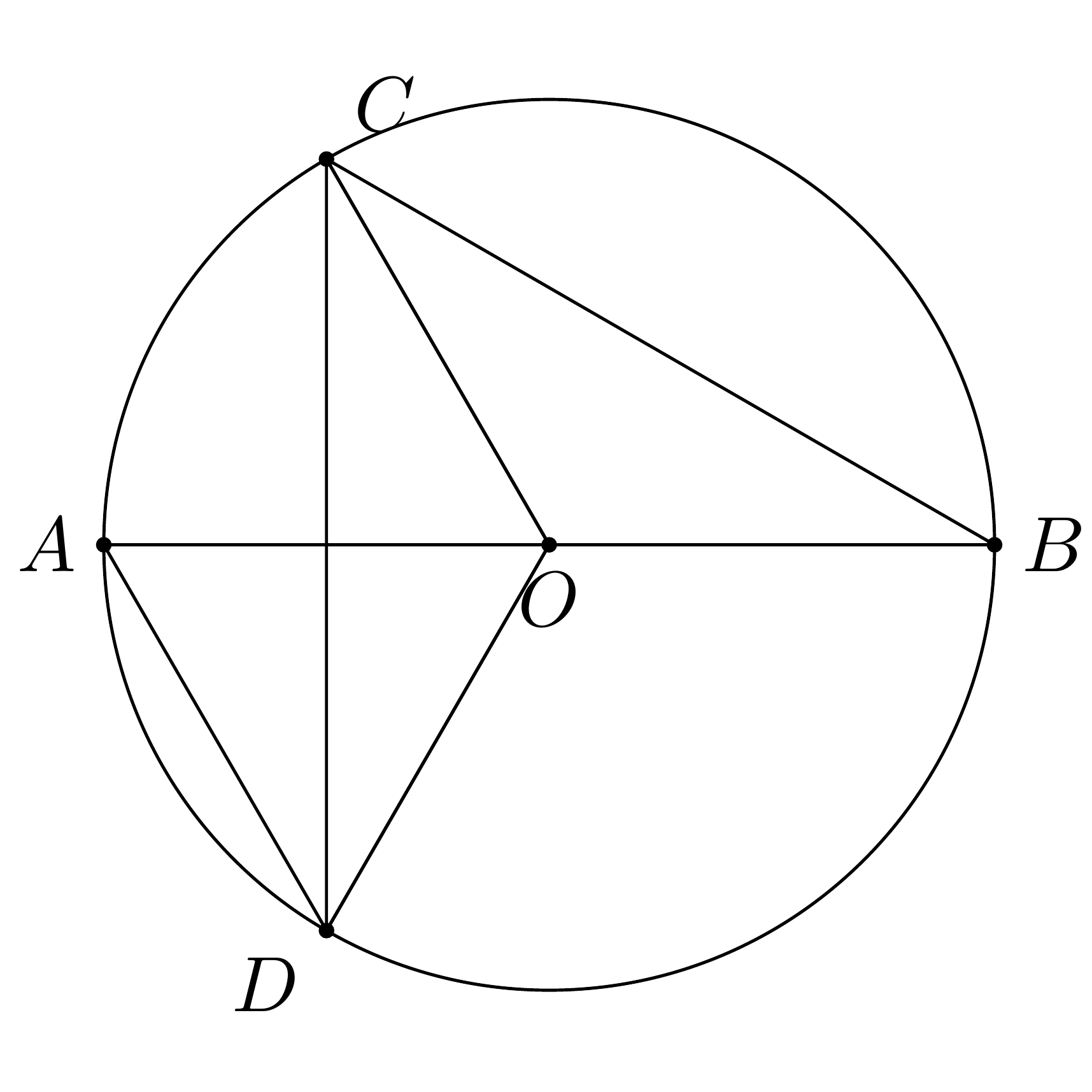} &
    \includegraphics[height=2.1cm,width=\wklength,keepaspectratio]{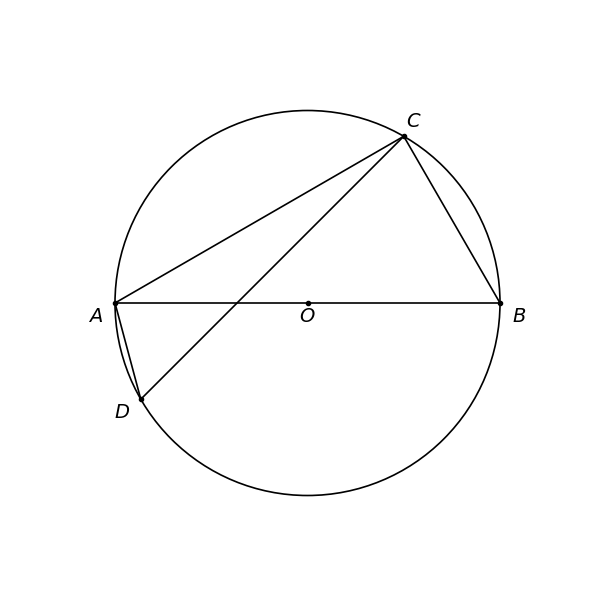} \\
\bottomrule
  \end{tabular}
  \caption{Comparison of image-to-code performance between the initial and final models.}%
  \label{fig:img2code_examples_4}
\end{table*}

%% file: appendix/new_images.tex
\begin{table*}[htb]
  \scriptsize
  \pgfmathsetlength{\wklength}{\textwidth/6 -2\tabcolsep}
  \begin{tabular}{ccc|ccc}
    \toprule
    \multicolumn{3}{c|}{New Images from One Seed Image} &
    \multicolumn{3}{c}{New Images from One Seed Image} \\
    \midrule
    \includegraphics[height=2.1cm,width=\wklength,keepaspectratio]{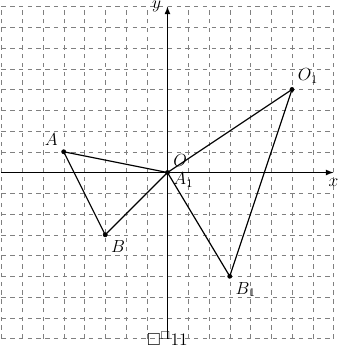} &
    \includegraphics[height=2.1cm,width=\wklength,keepaspectratio]{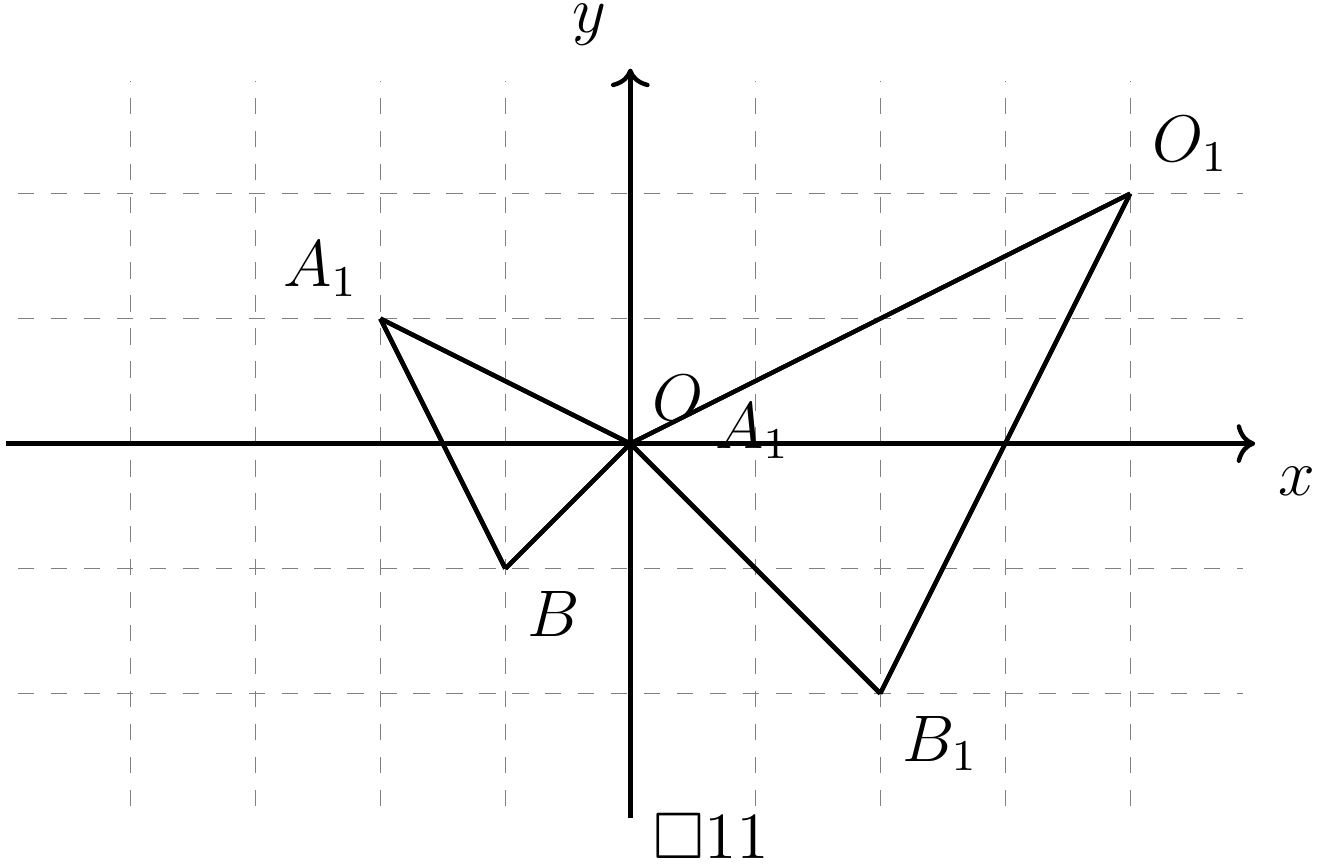} &
    \includegraphics[height=2.1cm,width=\wklength,keepaspectratio]{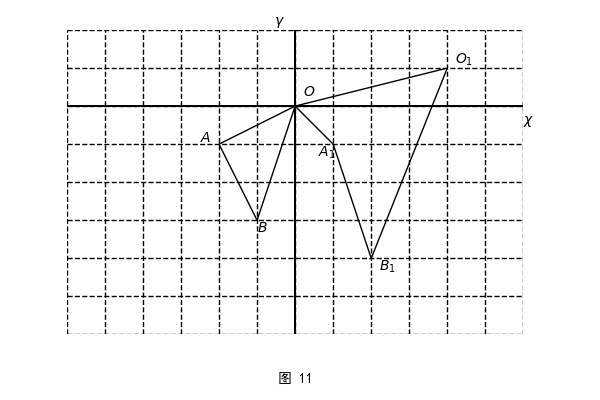} &
    \includegraphics[height=2.1cm,width=\wklength,keepaspectratio]{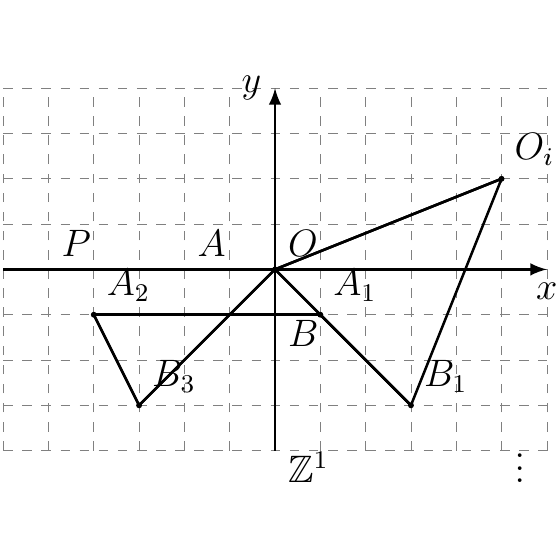} &
    \includegraphics[height=2.1cm,width=\wklength,keepaspectratio]{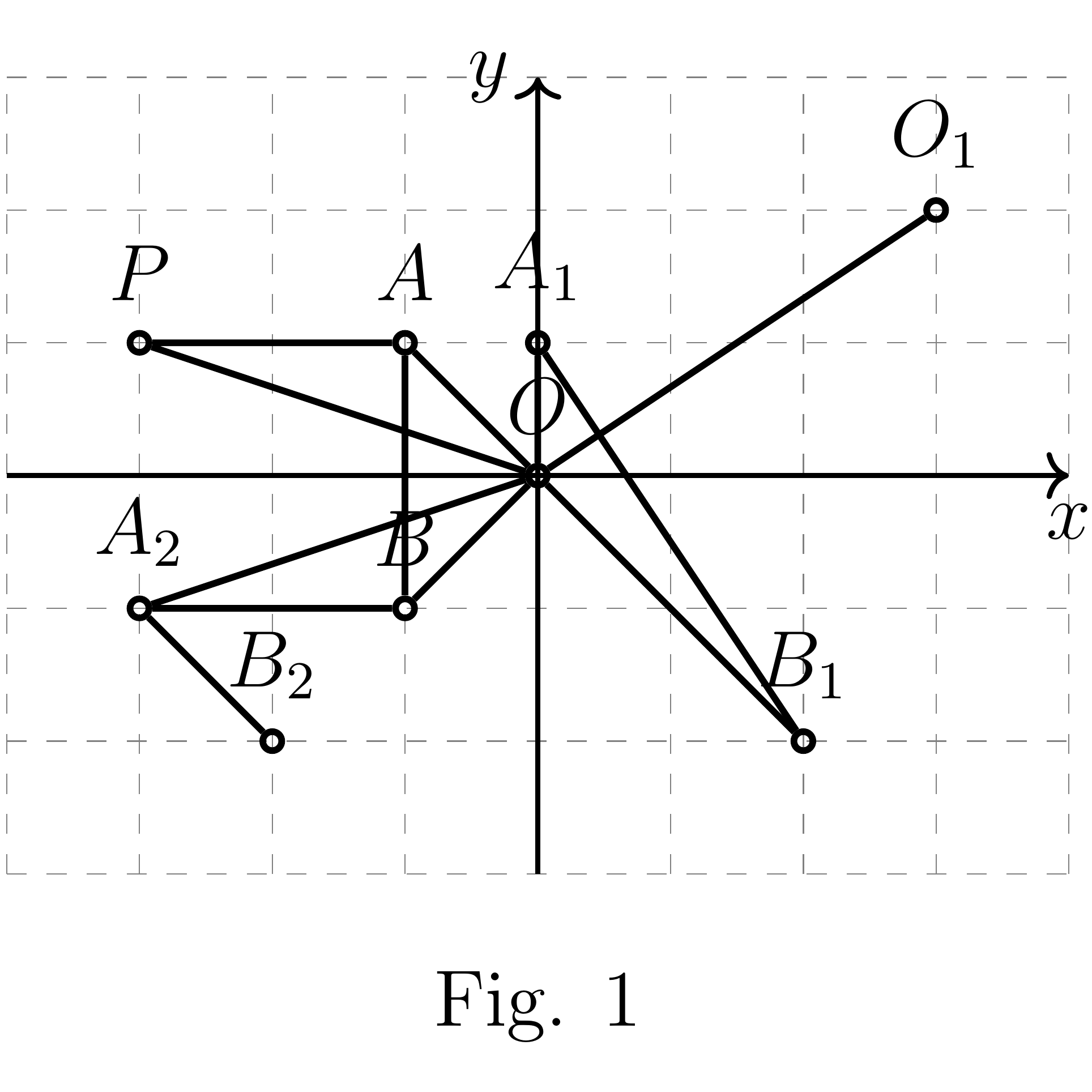} &
    \includegraphics[height=2.1cm,width=\wklength,keepaspectratio]{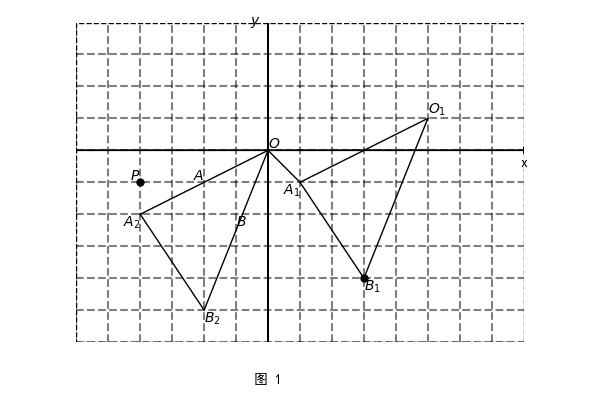} \\
    \midrule
    \includegraphics[height=2.1cm,width=\wklength,keepaspectratio]{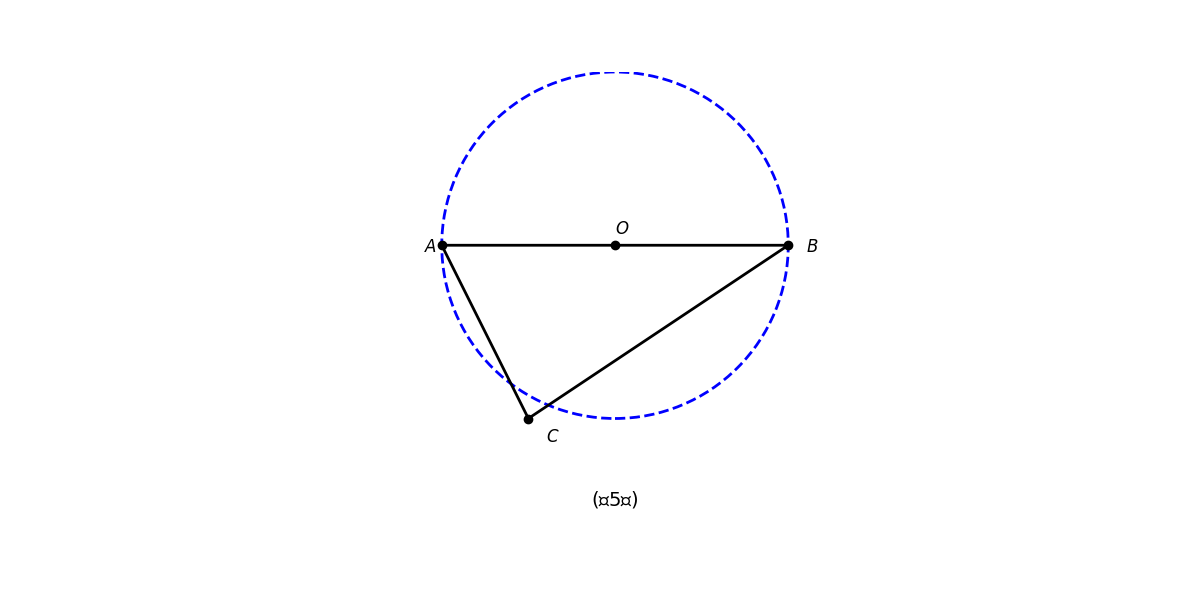} &
    \includegraphics[height=2.1cm,width=\wklength,keepaspectratio]{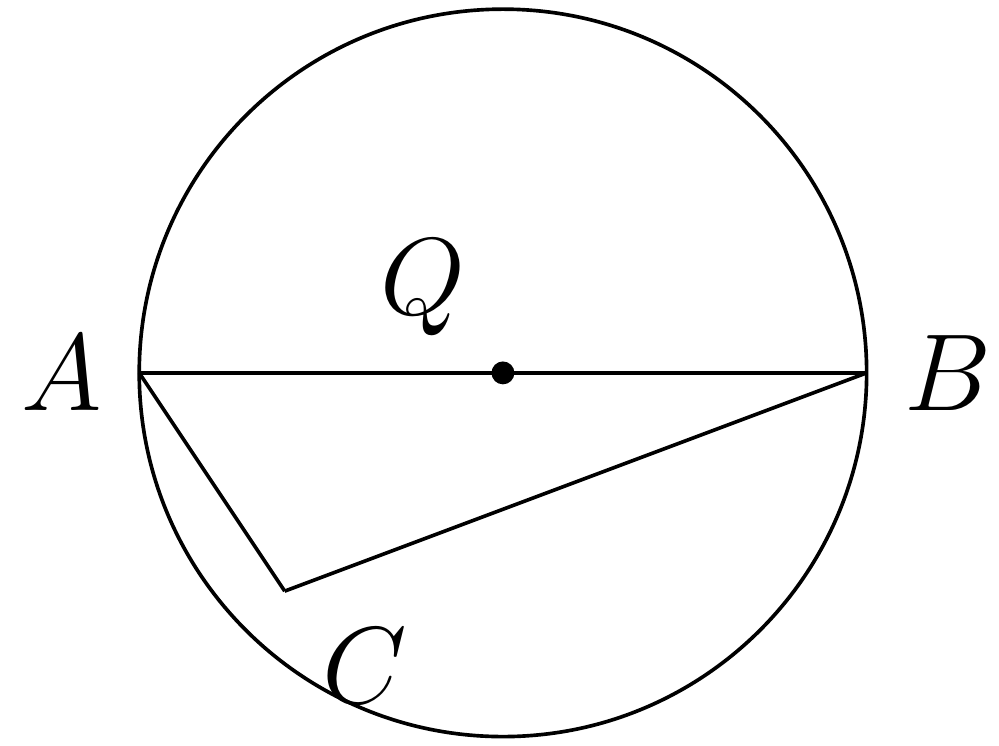} &
    \includegraphics[height=2.1cm,width=\wklength,keepaspectratio]{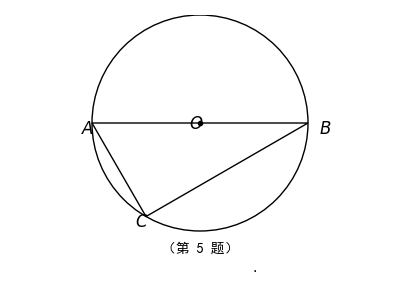} &
    \includegraphics[height=2.1cm,width=\wklength,keepaspectratio]{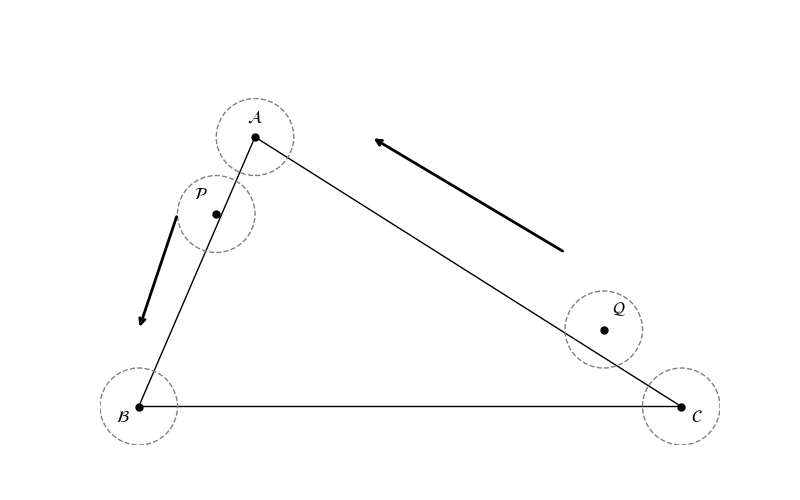} &
    \includegraphics[height=2.1cm,width=\wklength,keepaspectratio]{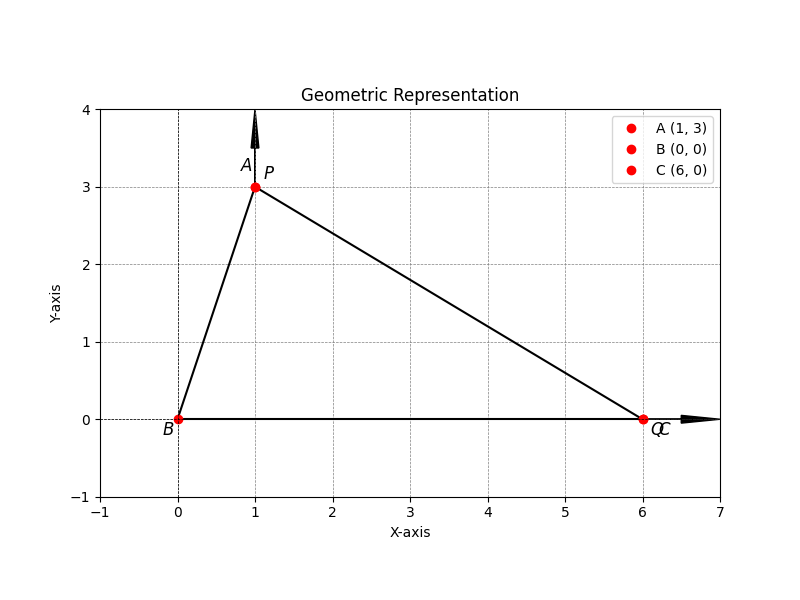} &
    \includegraphics[height=2.1cm,width=\wklength,keepaspectratio]{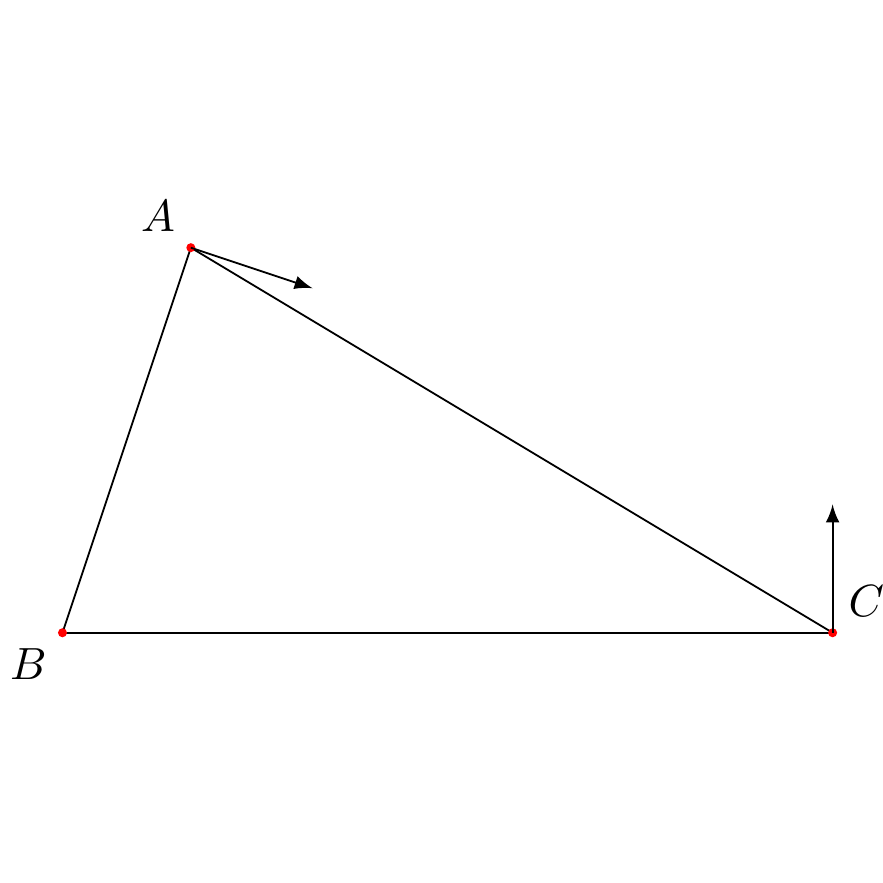} \\
    \midrule
    \includegraphics[height=2.1cm,width=\wklength,keepaspectratio]{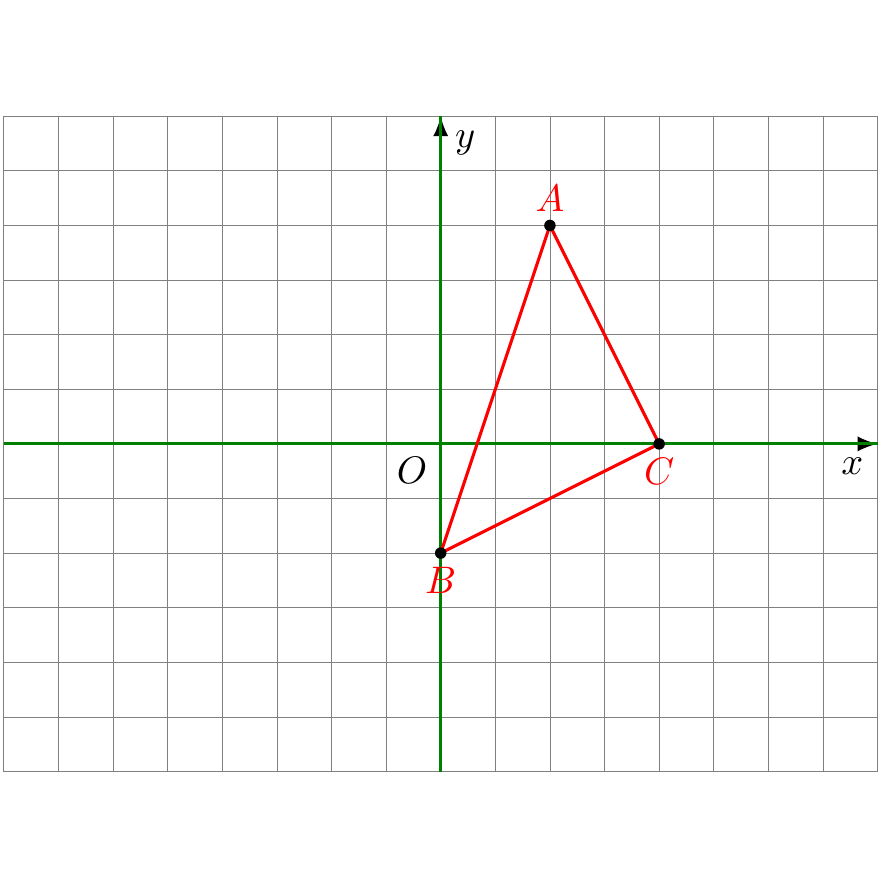} &
    \includegraphics[height=2.1cm,width=\wklength,keepaspectratio]{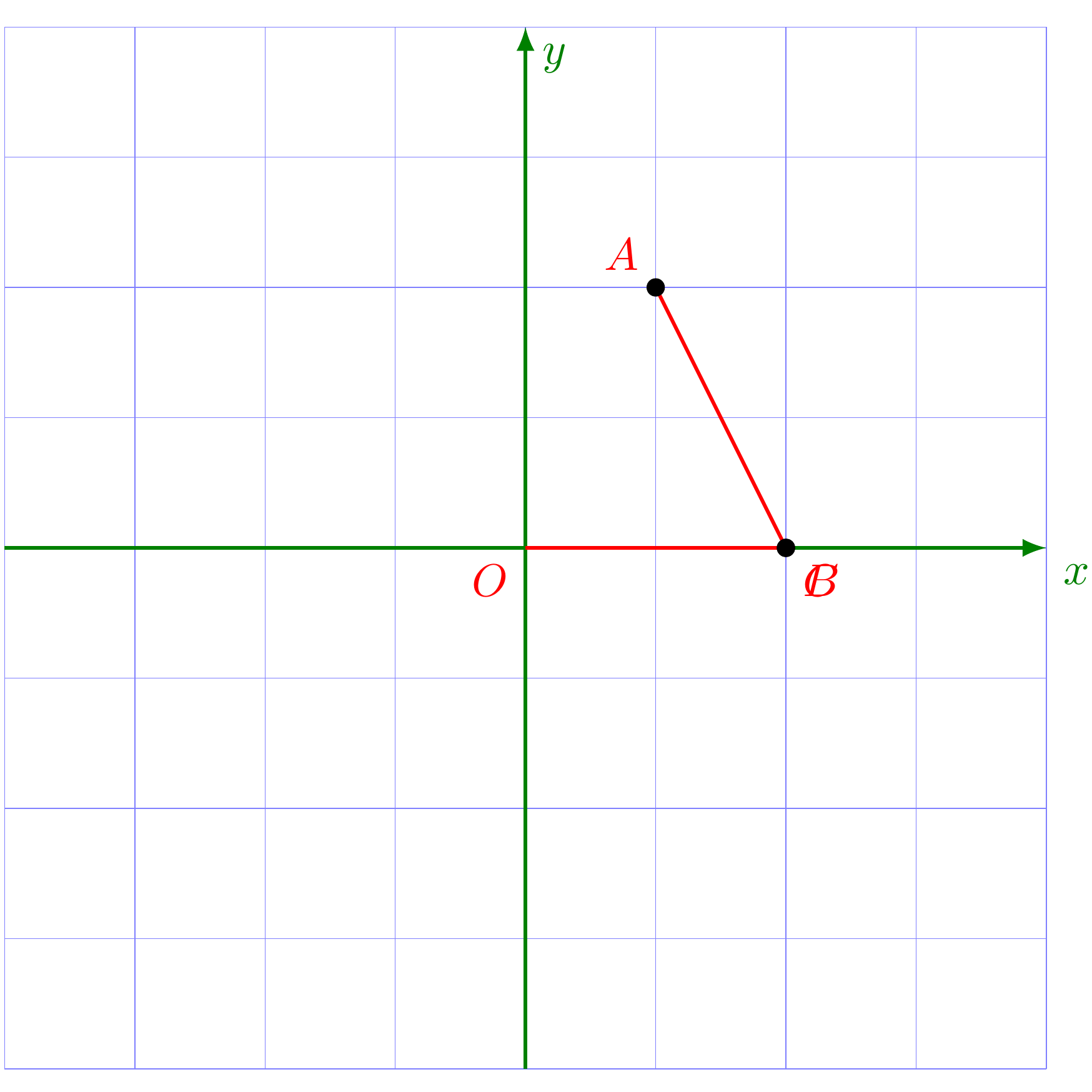} &
    \includegraphics[height=2.1cm,width=\wklength,keepaspectratio]{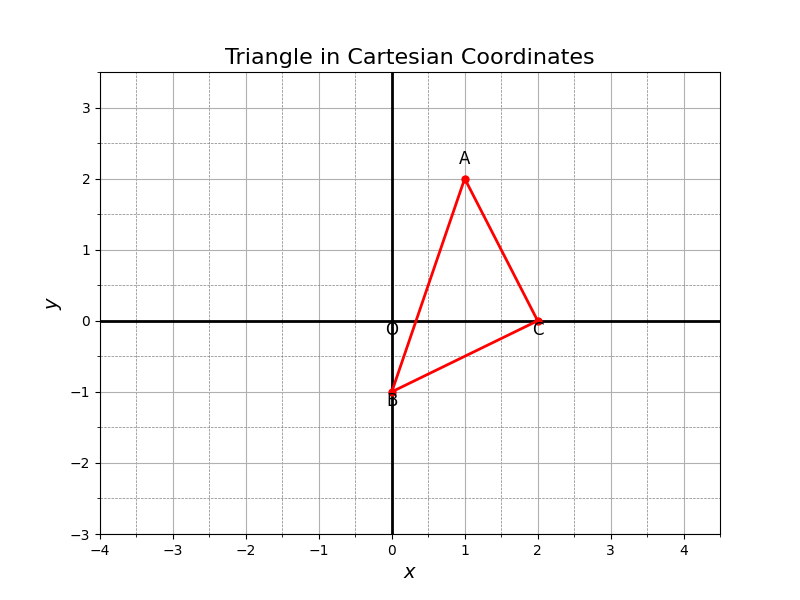} &
    \includegraphics[height=2.1cm,width=\wklength,keepaspectratio]{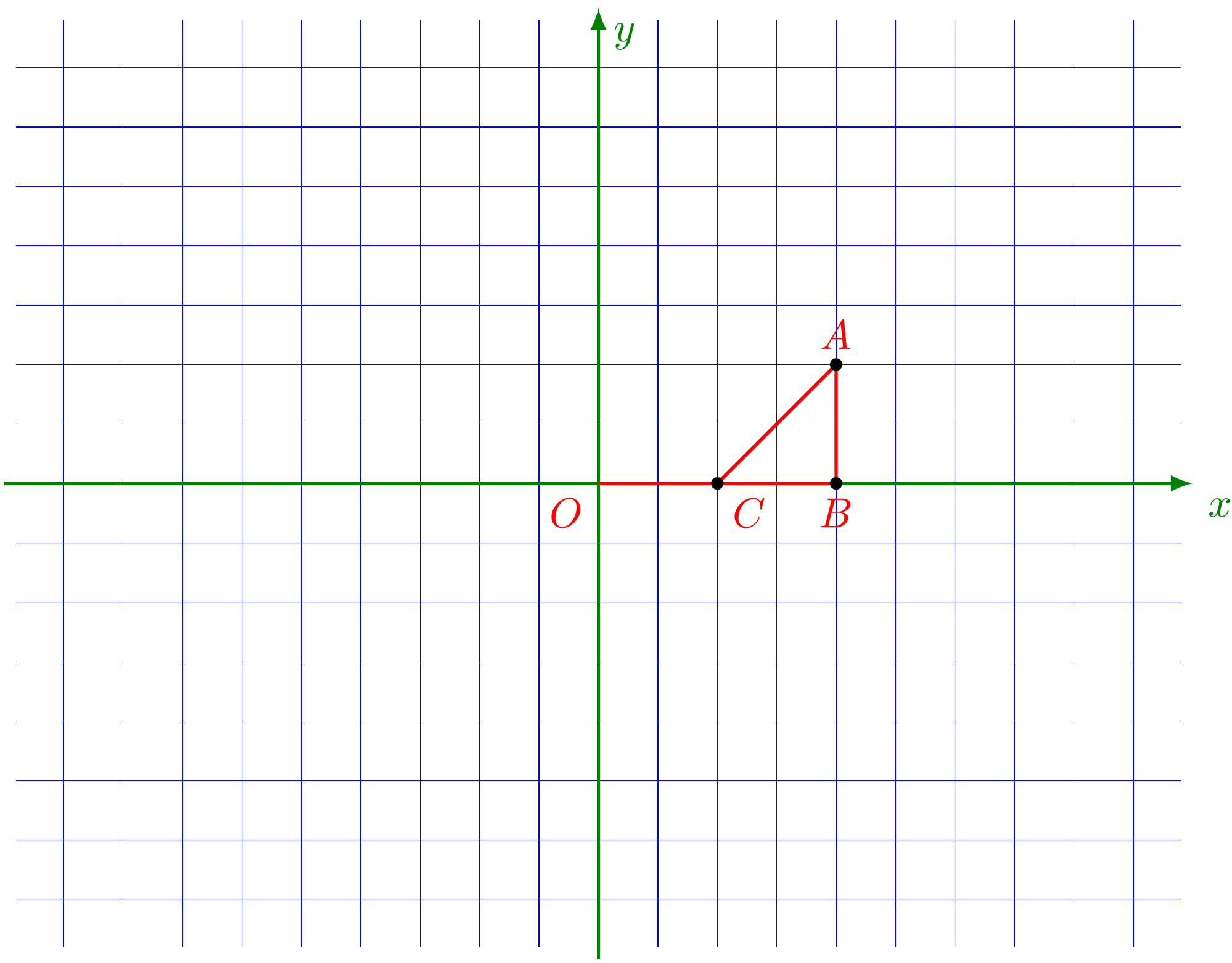} &
    \includegraphics[height=2.1cm,width=\wklength,keepaspectratio]{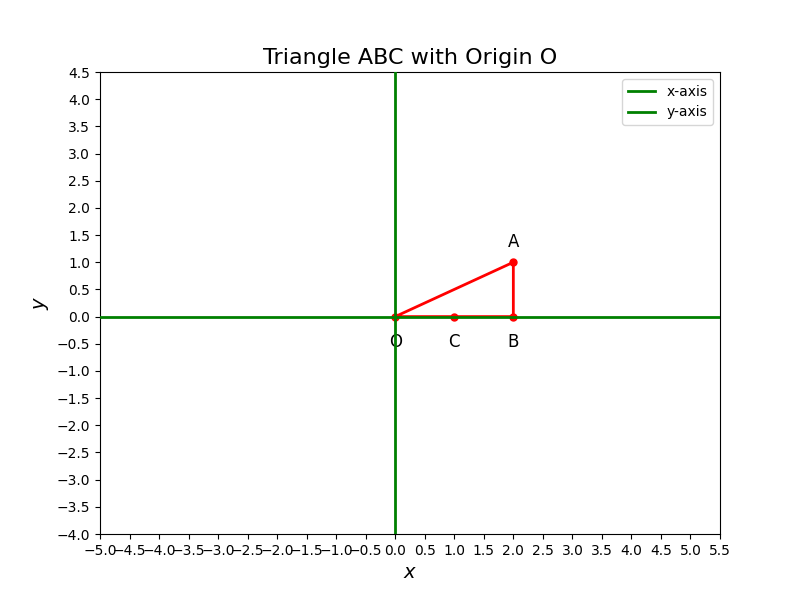} &
    \includegraphics[height=2.1cm,width=\wklength,keepaspectratio]{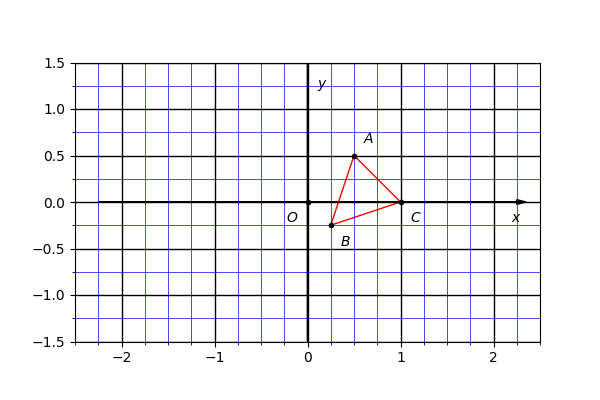} \\
\bottomrule
  \end{tabular}
  \caption{New images synthesized with seed images form K12-2M.}%
  \label{new_images_from_k12_examples_1of3}
\end{table*}


\begin{table*}[htb]
  \scriptsize
  \pgfmathsetlength{\wklength}{\textwidth/6 -2\tabcolsep}
  \begin{tabular}{ccc|ccc}
    \toprule
    \multicolumn{3}{c|}{New Images from One Seed Image} &
    \multicolumn{3}{c}{New Images from One Seed Image} \\
    \midrule
    \includegraphics[height=2.1cm,width=\wklength,keepaspectratio]{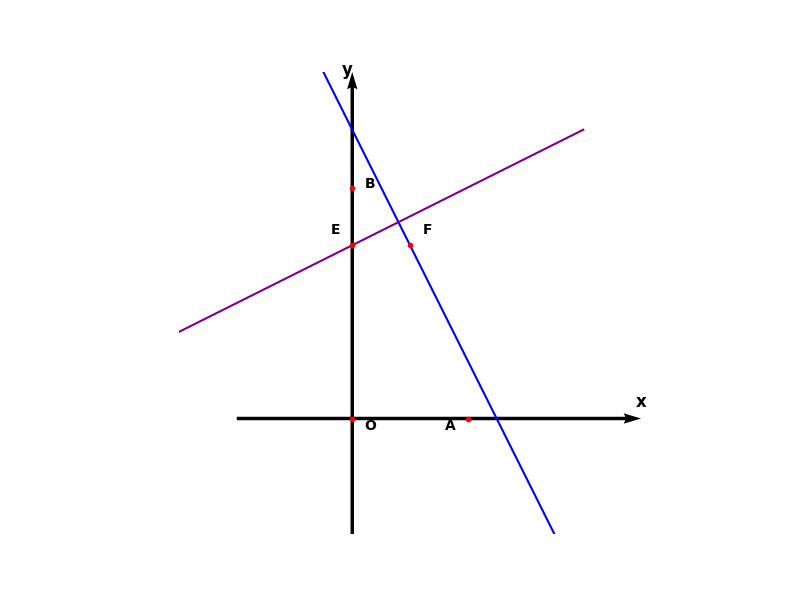} &
    \includegraphics[height=2.1cm,width=\wklength,keepaspectratio]{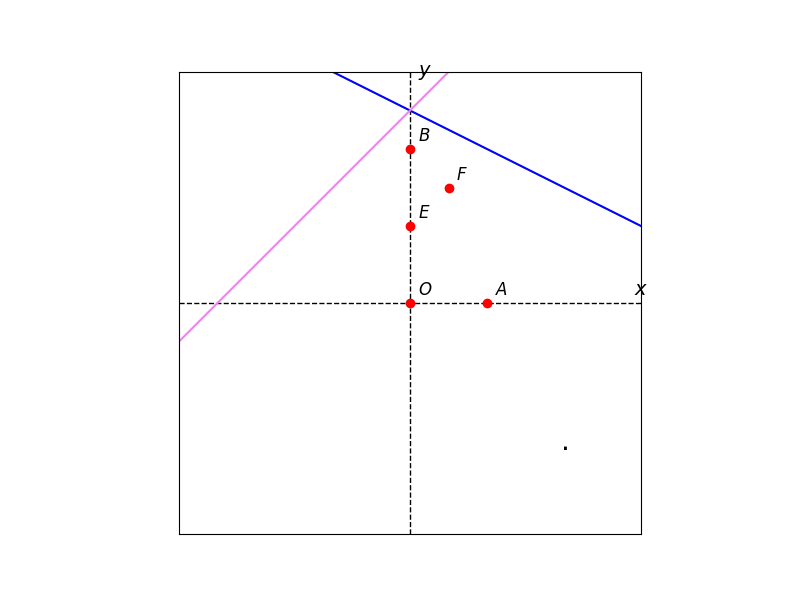} &
    \includegraphics[height=2.1cm,width=\wklength,keepaspectratio]{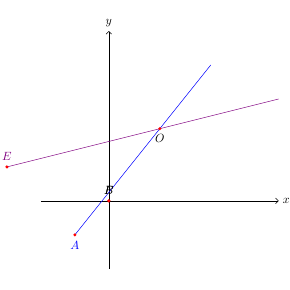} &
    \includegraphics[height=2.1cm,width=\wklength,keepaspectratio]{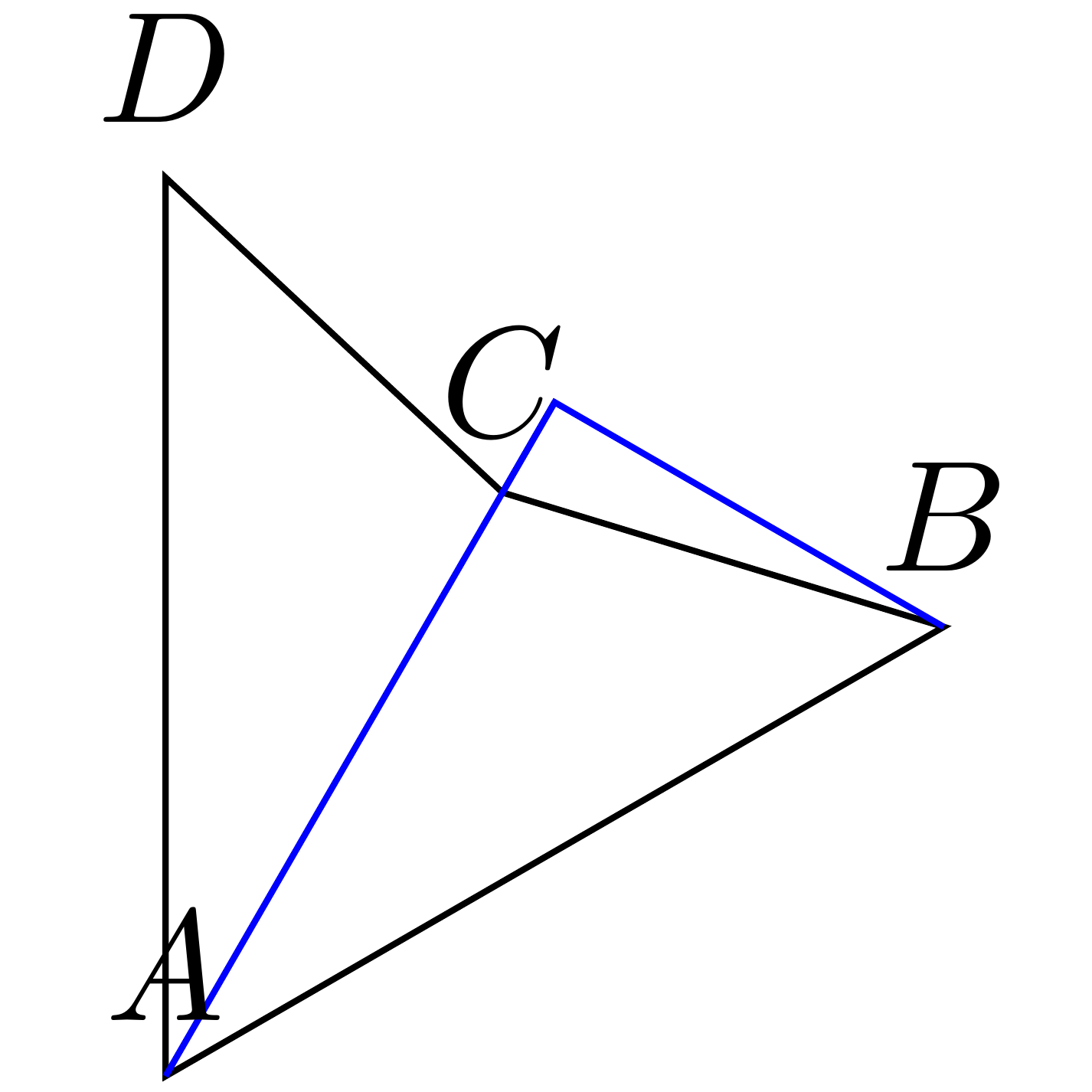} &
    \includegraphics[height=2.1cm,width=\wklength,keepaspectratio]{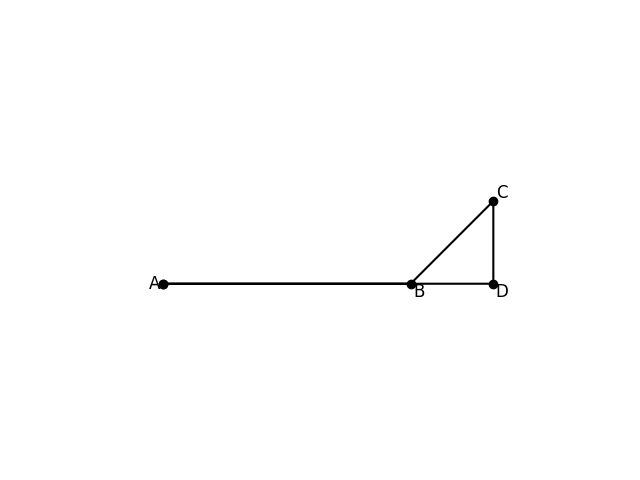} &
    \includegraphics[height=2.1cm,width=\wklength,keepaspectratio]{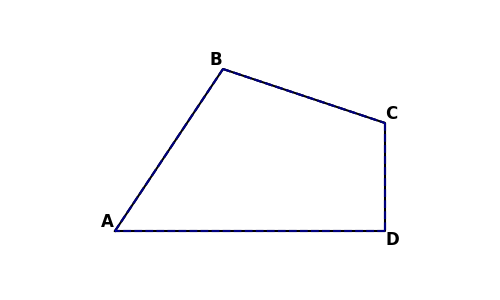} \\
    \midrule
    \includegraphics[height=2.1cm,width=\wklength,keepaspectratio]{examples/k12/00033/01.png} &
    \includegraphics[height=2.1cm,width=\wklength,keepaspectratio]{examples/k12/00033/02.png} &
    \includegraphics[height=2.1cm,width=\wklength,keepaspectratio]{examples/k12/00033/04.png} &
    \includegraphics[height=2.1cm,width=\wklength,keepaspectratio]{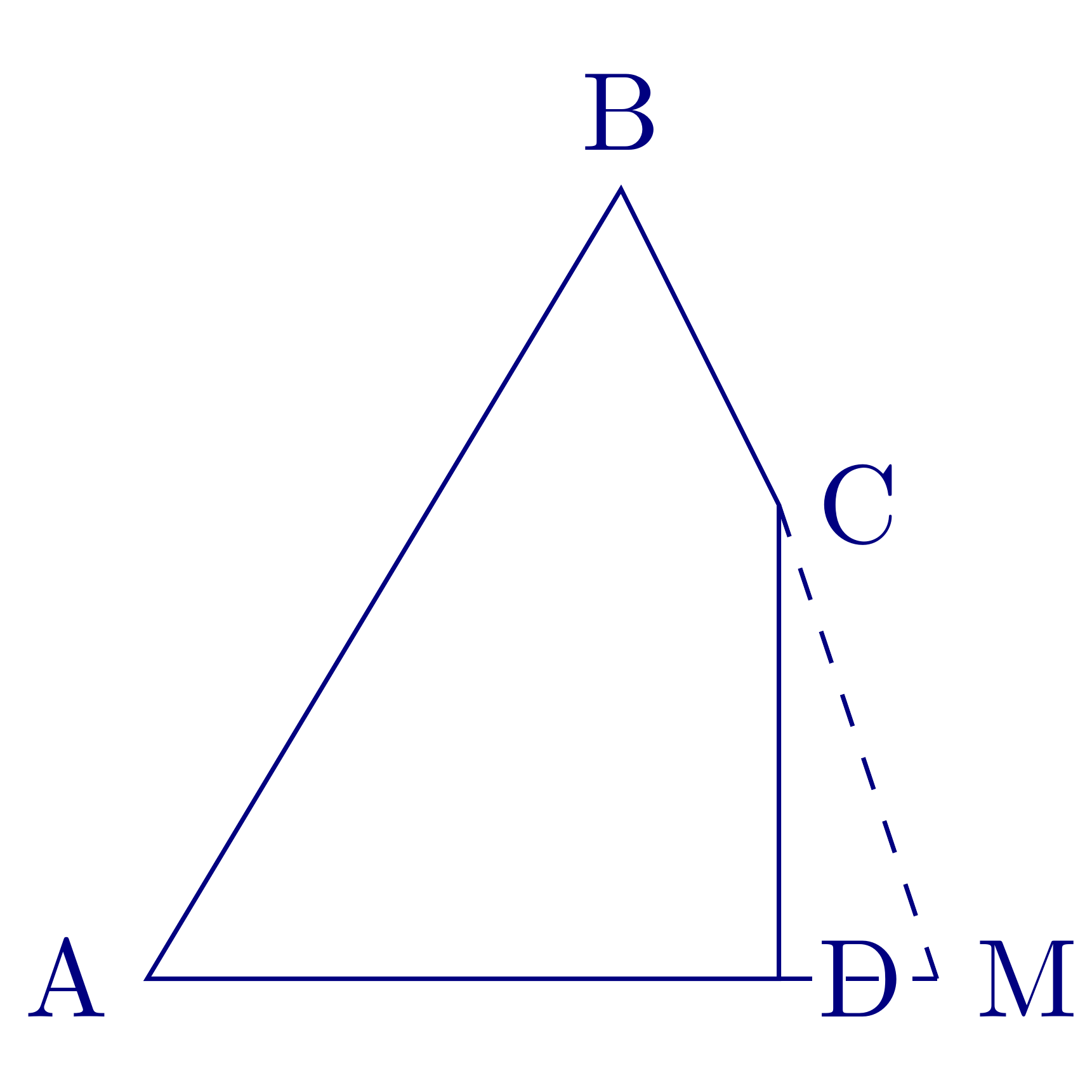} &
    \includegraphics[height=2.1cm,width=\wklength,keepaspectratio]{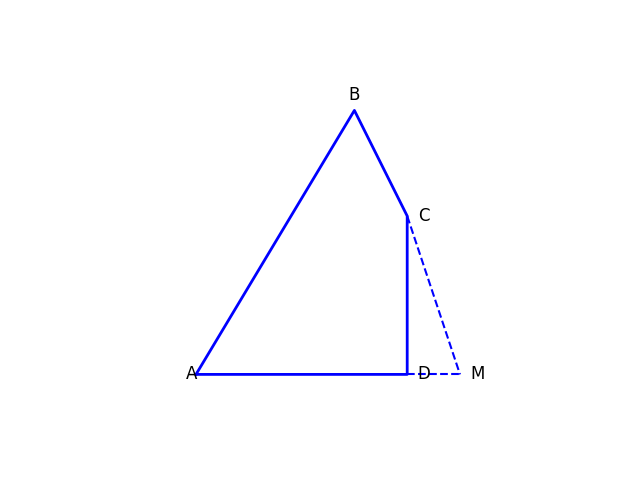} &
    \includegraphics[height=2.1cm,width=\wklength,keepaspectratio]{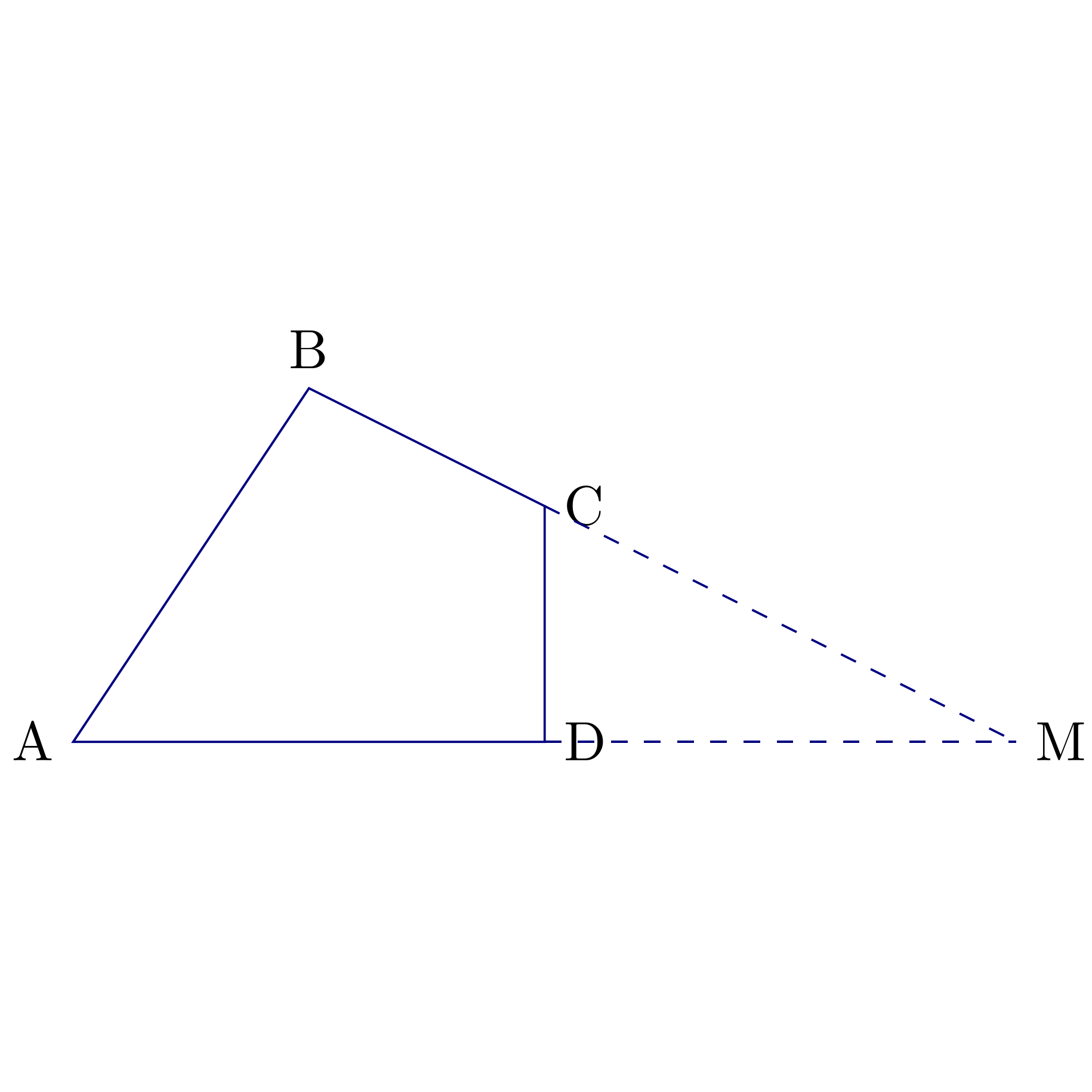} \\
    \midrule
    \includegraphics[height=2.1cm,width=\wklength,keepaspectratio]{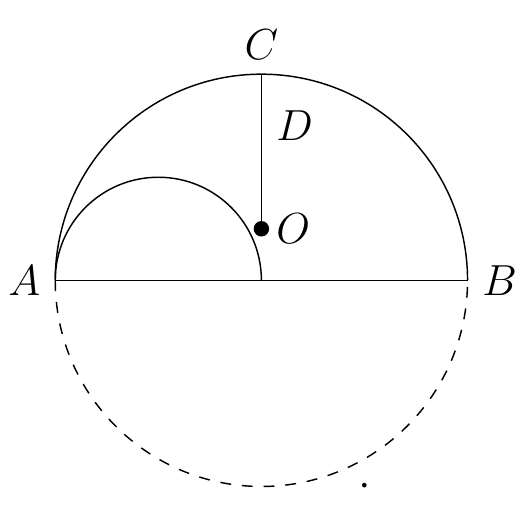} &
    \includegraphics[height=2.1cm,width=\wklength,keepaspectratio]{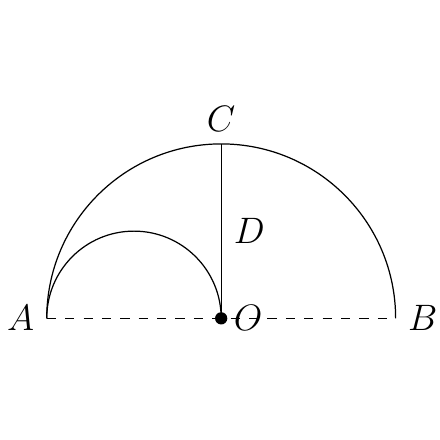} &
    \includegraphics[height=2.1cm,width=\wklength,keepaspectratio]{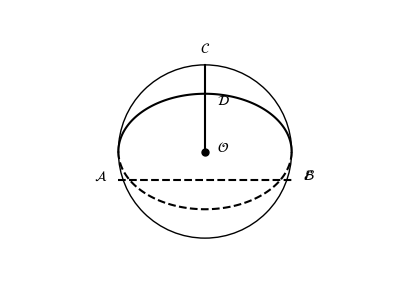} &
    \includegraphics[height=2.1cm,width=\wklength,keepaspectratio]{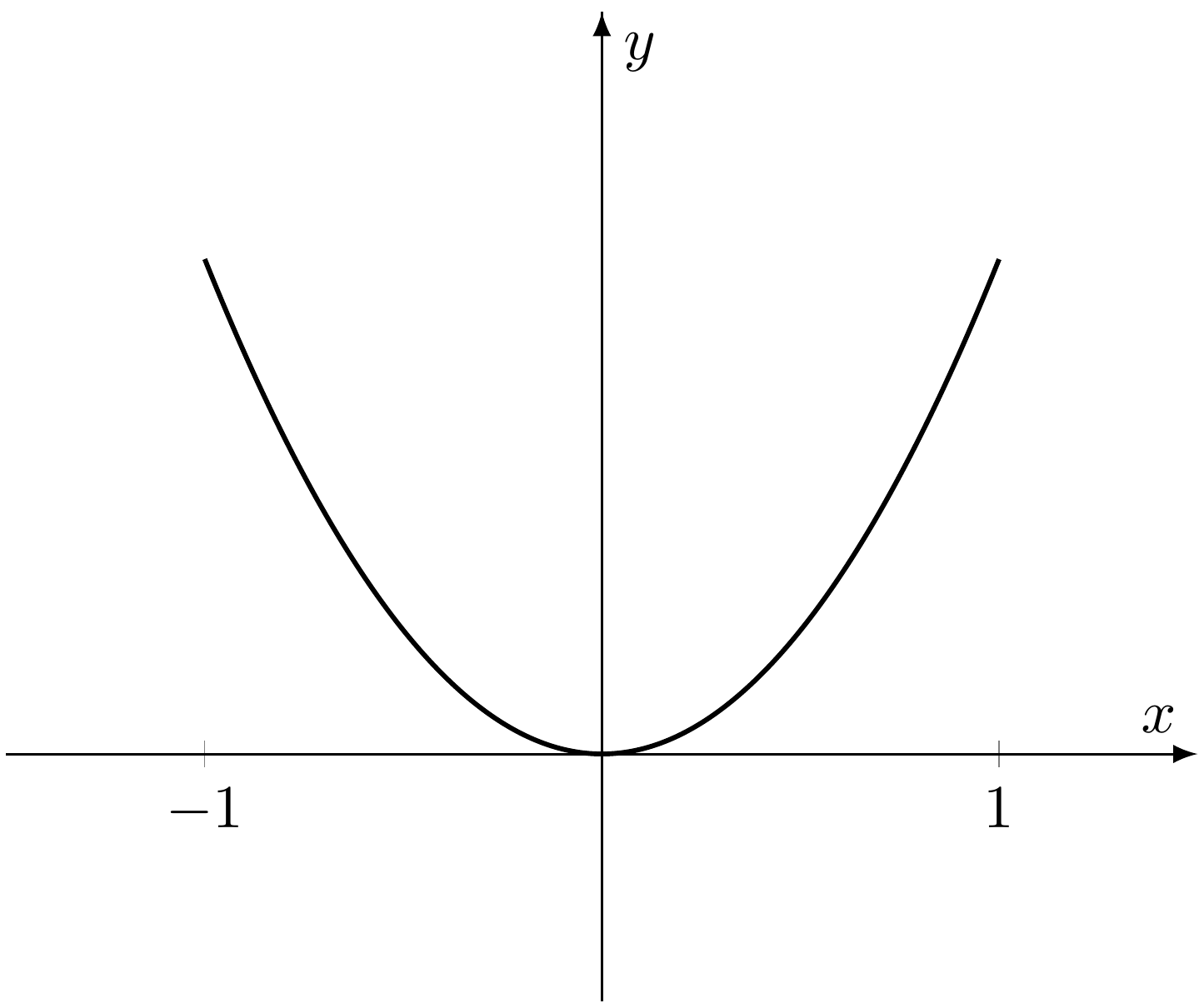} &
    \includegraphics[height=2.1cm,width=\wklength,keepaspectratio]{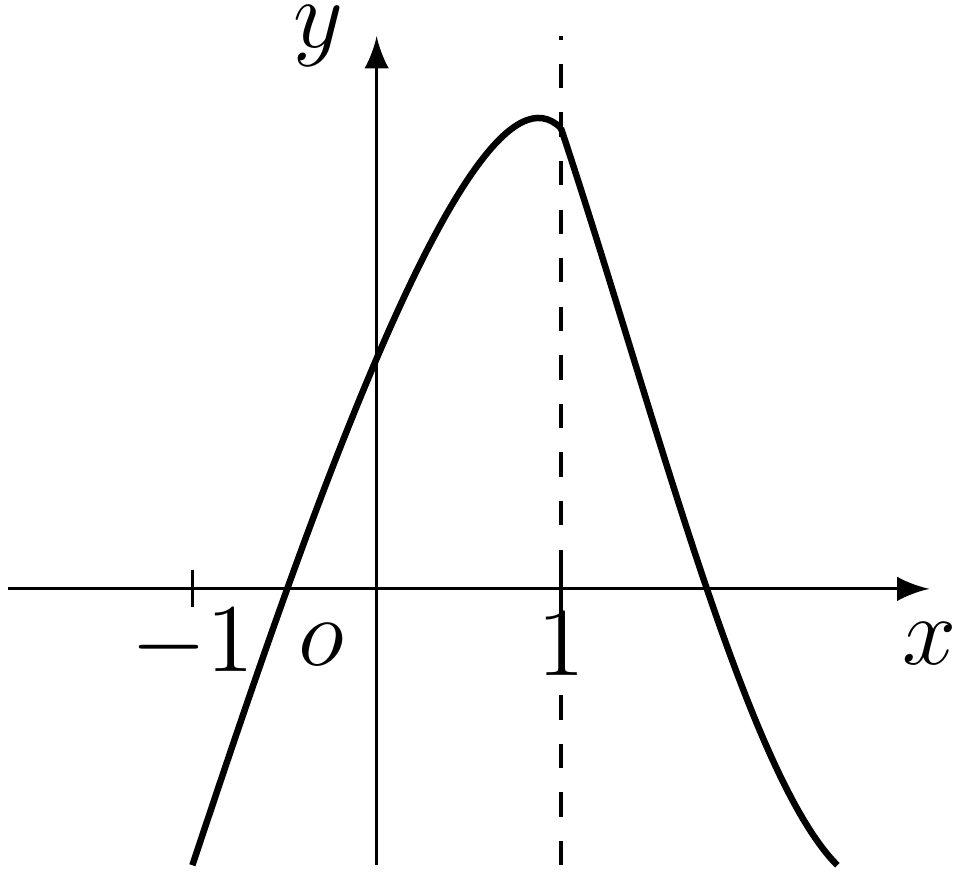} &
    \includegraphics[height=2.1cm,width=\wklength,keepaspectratio]{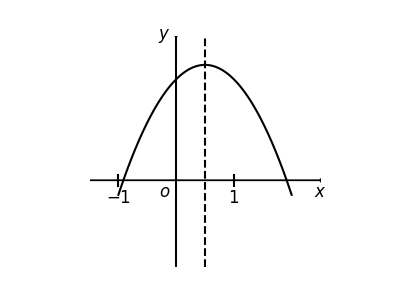} \\
\bottomrule
  \end{tabular}
  \caption{New images synthesized with seed images form K12-2M}%
  \label{new_images_from_k12_examples_2of3}
\end{table*}

\begin{table*}[htb]
  \scriptsize
  \pgfmathsetlength{\wklength}{\textwidth/6 -2\tabcolsep}
  \begin{tabular}{ccc|ccc}
    \toprule
    \multicolumn{3}{c|}{New Images from One Seed Image} &
    \multicolumn{3}{c}{New Images from One Seed Image} \\
    \midrule
    \includegraphics[height=2.1cm,width=\wklength,keepaspectratio]{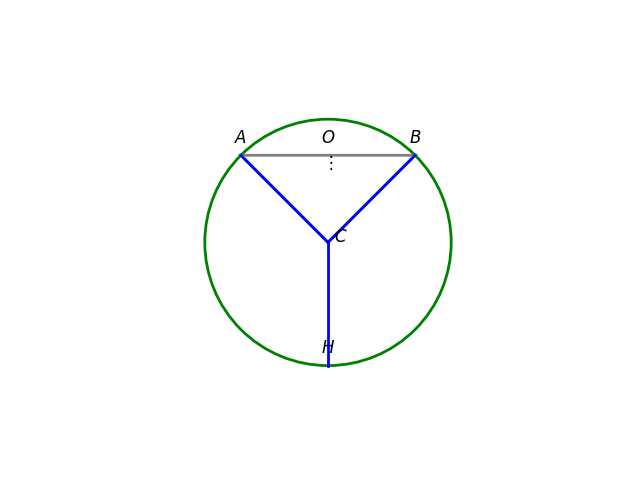} &
    \includegraphics[height=2.1cm,width=\wklength,keepaspectratio]{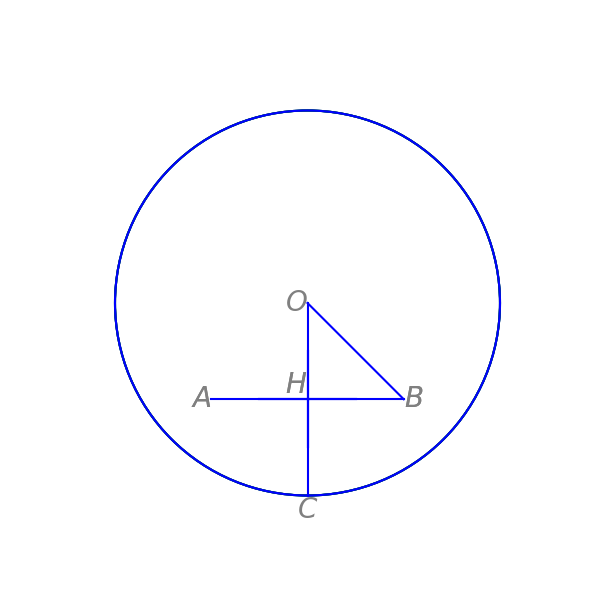} &
    \includegraphics[height=2.1cm,width=\wklength,keepaspectratio]{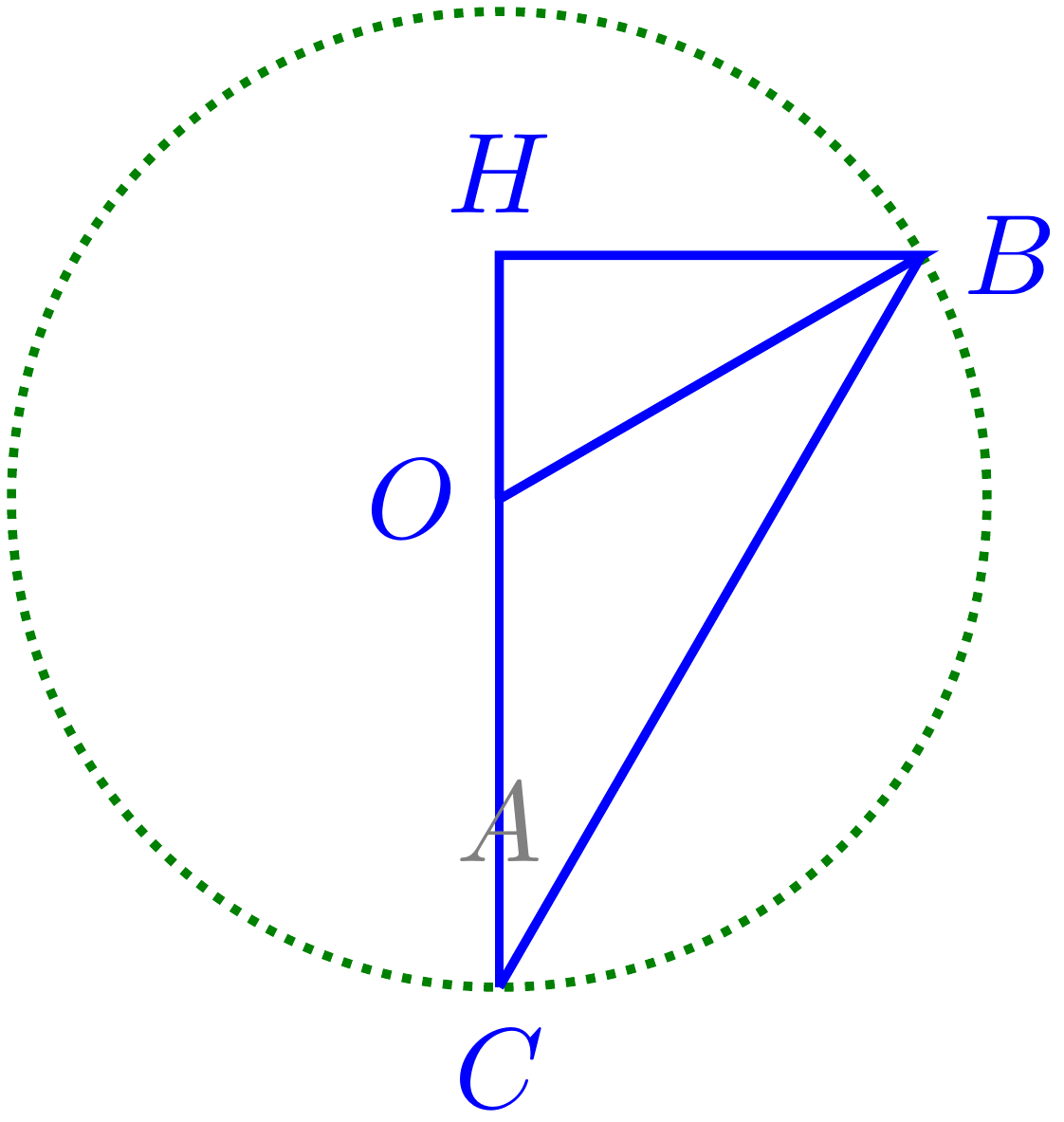} &
    \includegraphics[height=2.1cm,width=\wklength,keepaspectratio]{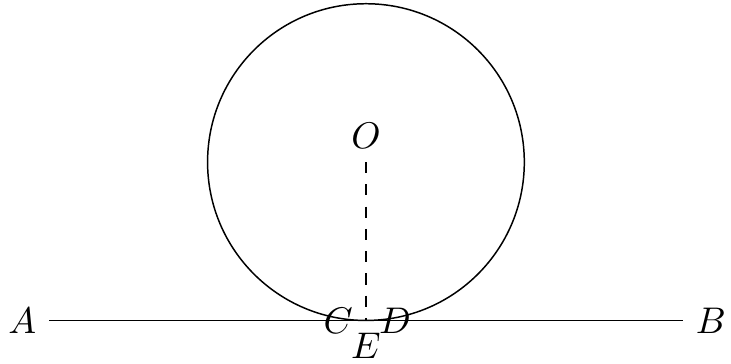} &
    \includegraphics[height=2.1cm,width=\wklength,keepaspectratio]{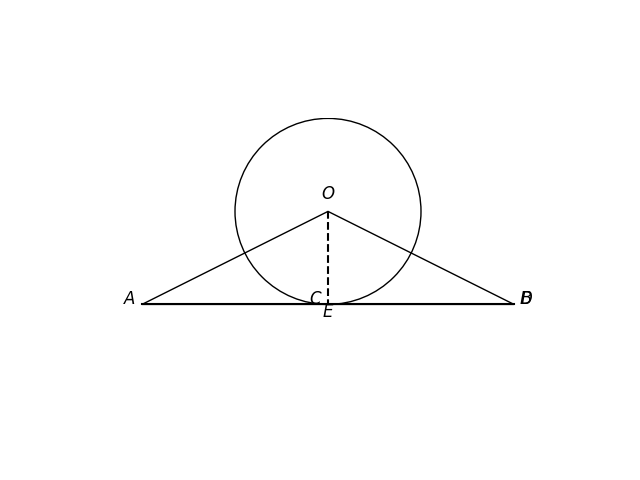} &
    \includegraphics[height=2.1cm,width=\wklength,keepaspectratio]{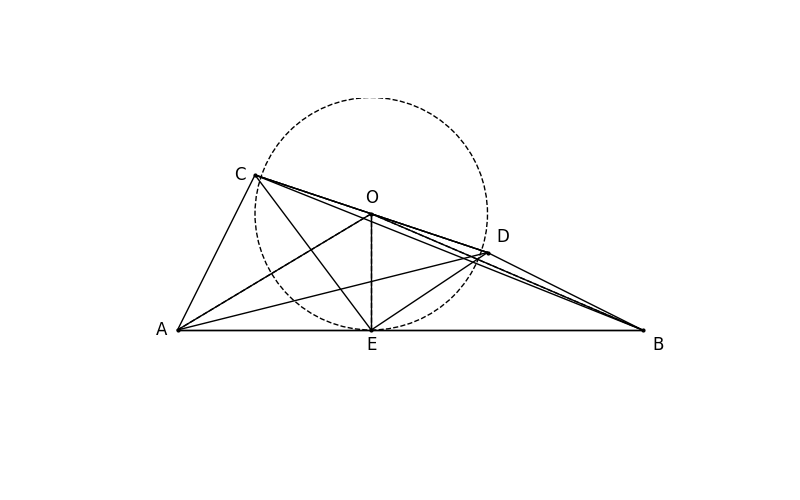} \\
    \midrule
    \includegraphics[height=2.1cm,width=\wklength,keepaspectratio]{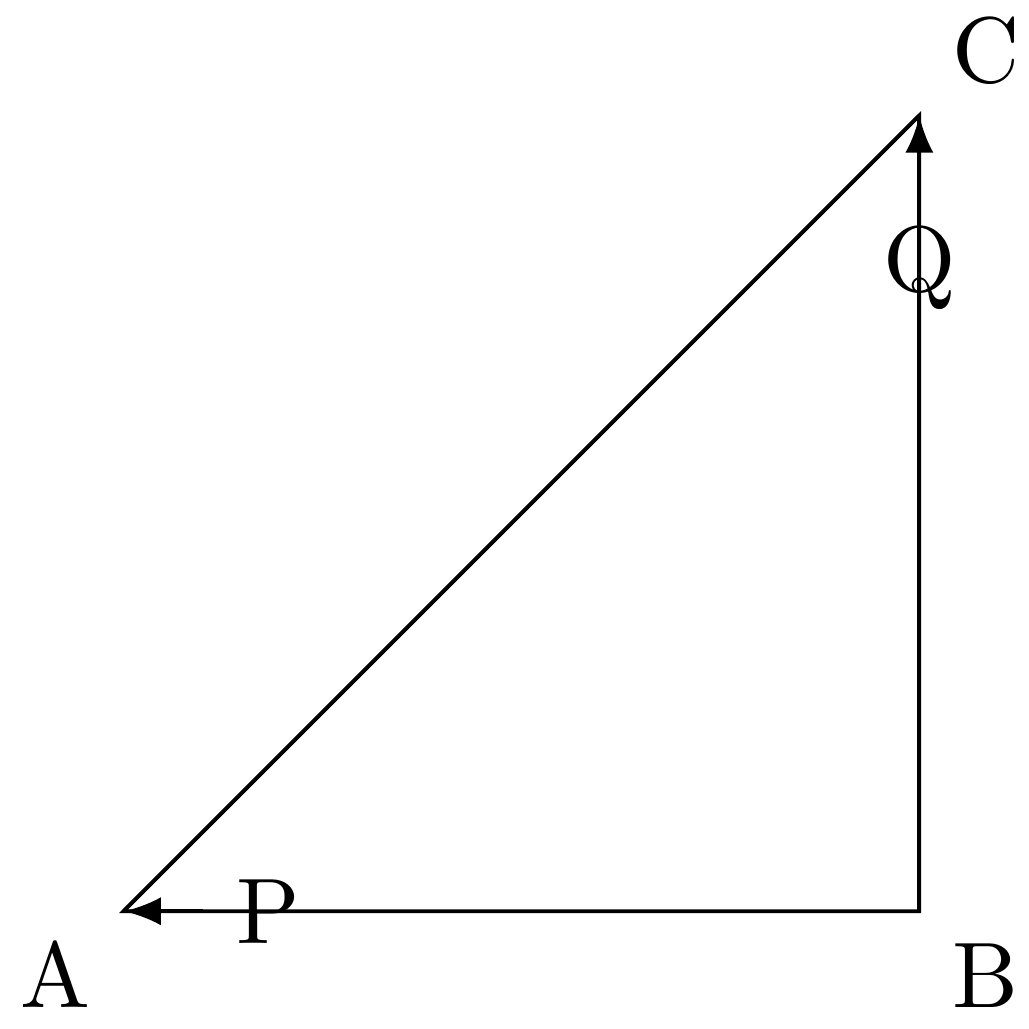} &
    \includegraphics[height=2.1cm,width=\wklength,keepaspectratio]{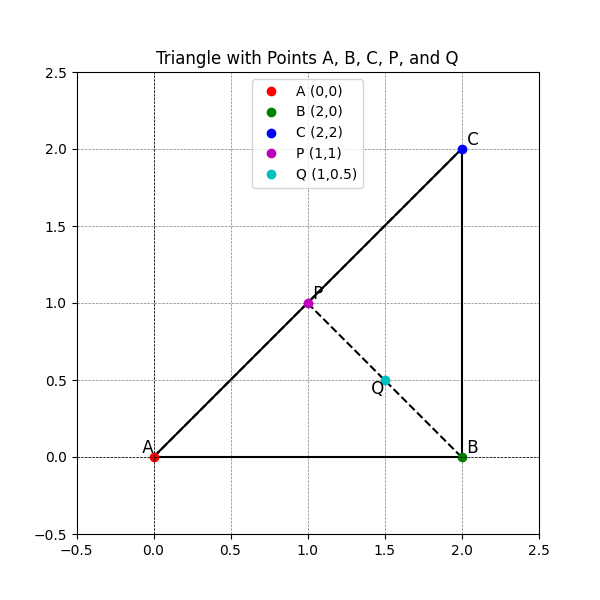} &
    \includegraphics[height=2.1cm,width=\wklength,keepaspectratio]{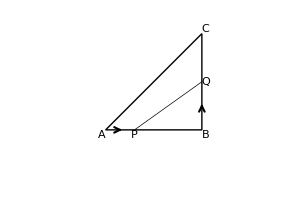} &
    \includegraphics[height=2.1cm,width=\wklength,keepaspectratio]{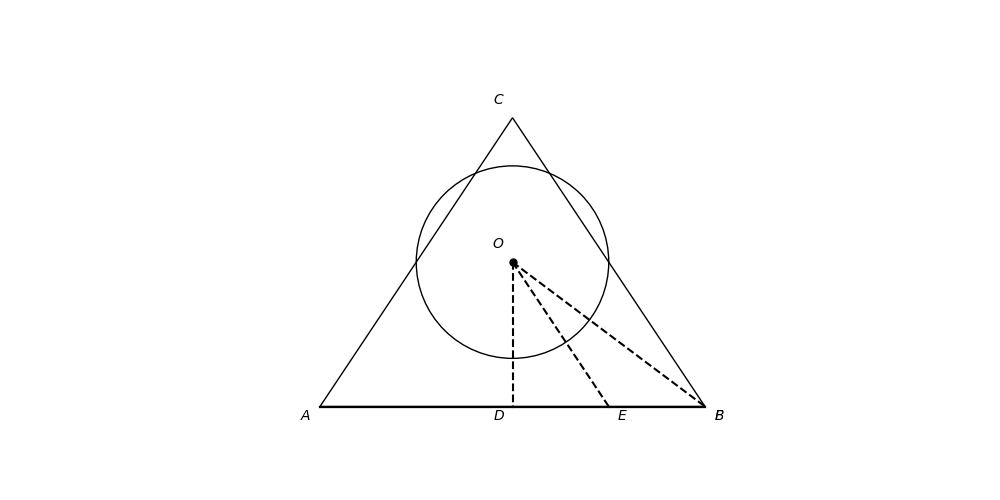} &
    \includegraphics[height=2.1cm,width=\wklength,keepaspectratio]{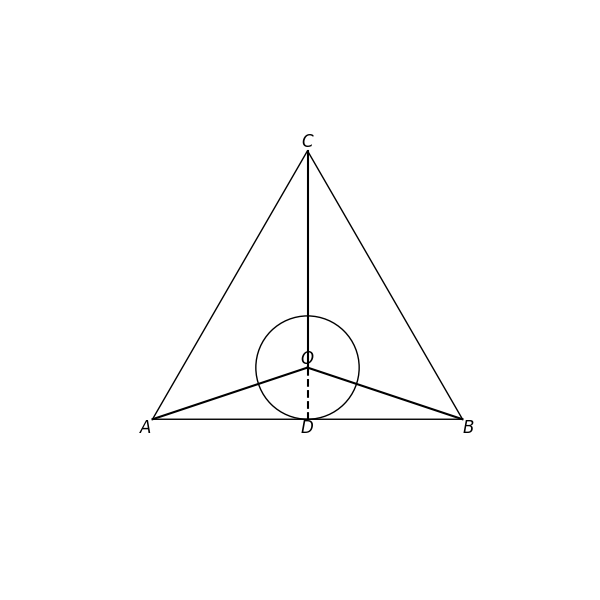} &
    \includegraphics[height=2.1cm,width=\wklength,keepaspectratio]{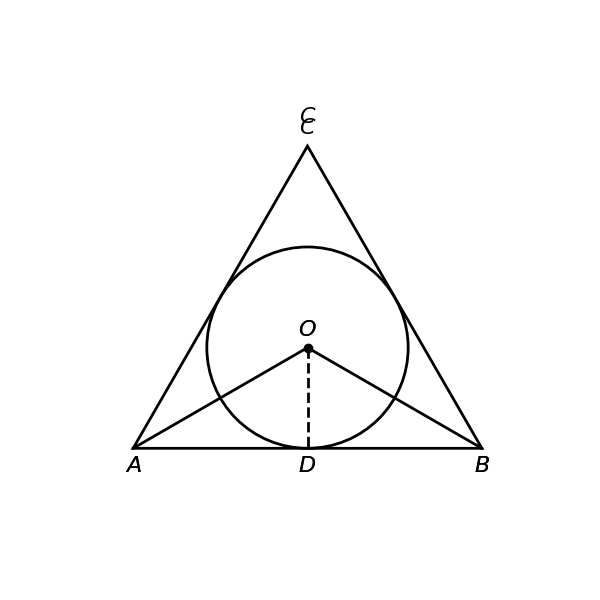} \\
    \midrule
    \includegraphics[height=2.1cm,width=\wklength,keepaspectratio]{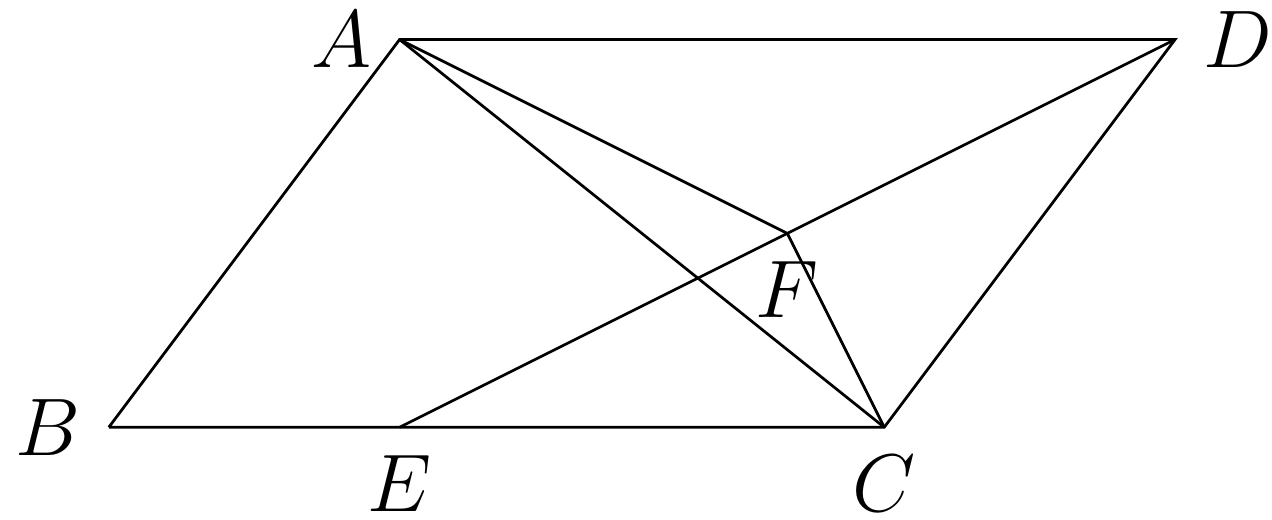} &
    \includegraphics[height=2.1cm,width=\wklength,keepaspectratio]{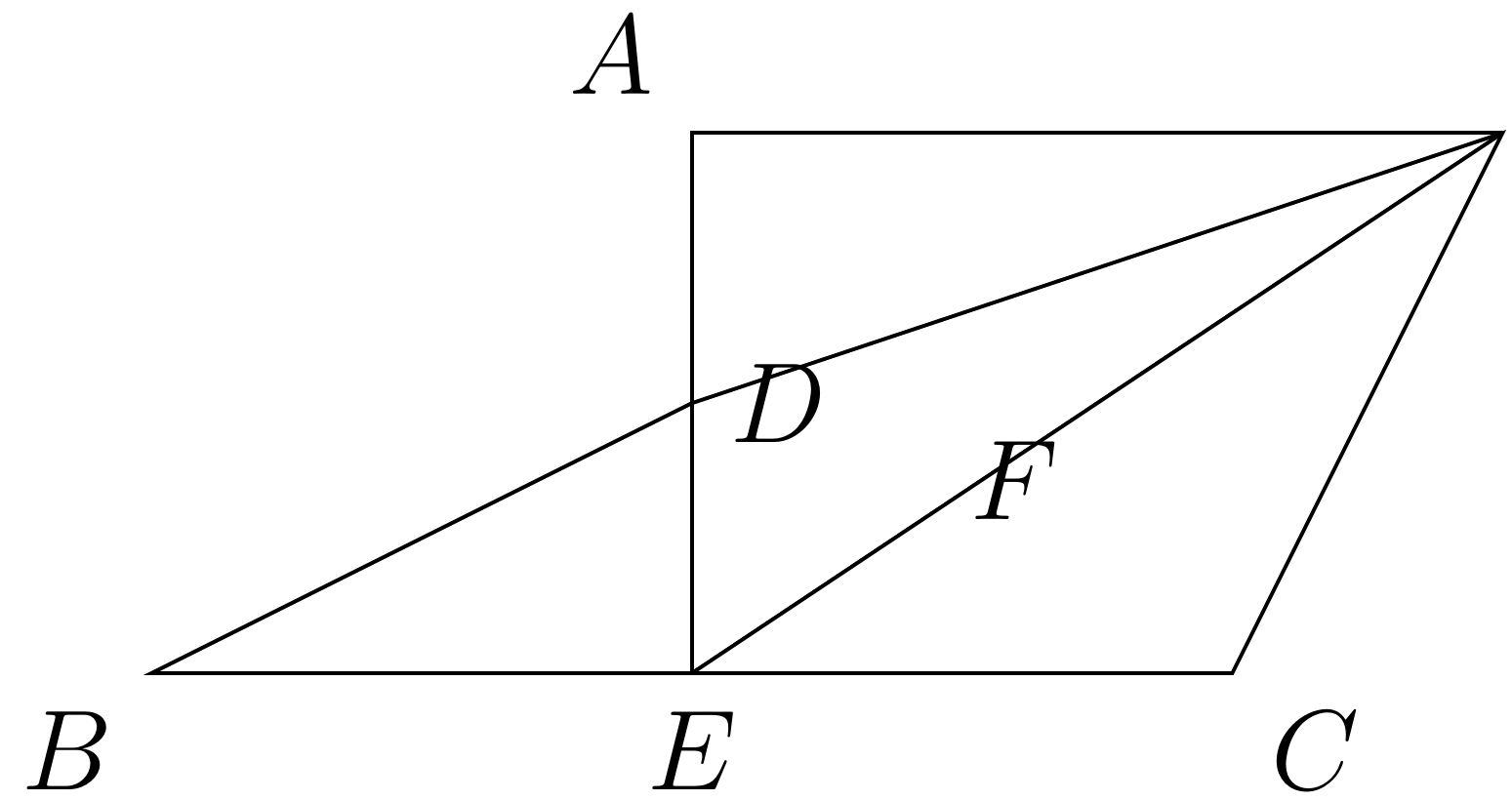} &
    \includegraphics[height=2.1cm,width=\wklength,keepaspectratio]{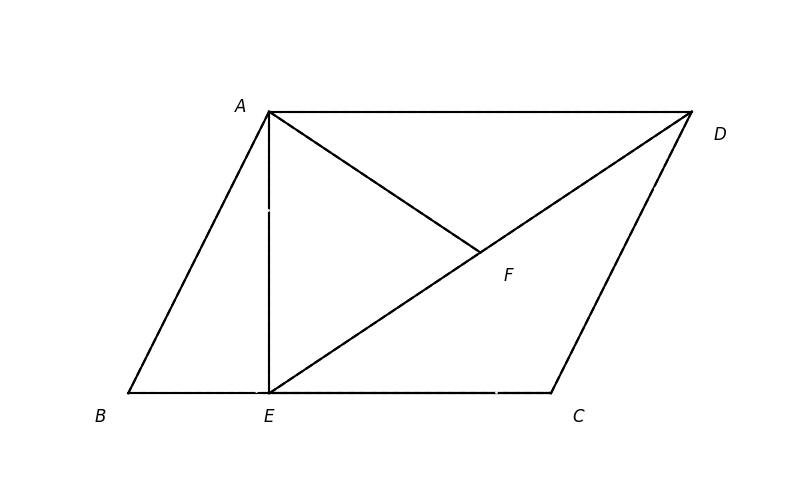} &
    \includegraphics[height=2.1cm,width=\wklength,keepaspectratio]{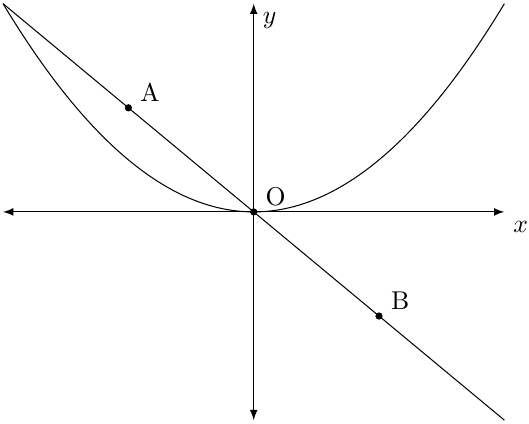} &
    \includegraphics[height=2.1cm,width=\wklength,keepaspectratio]{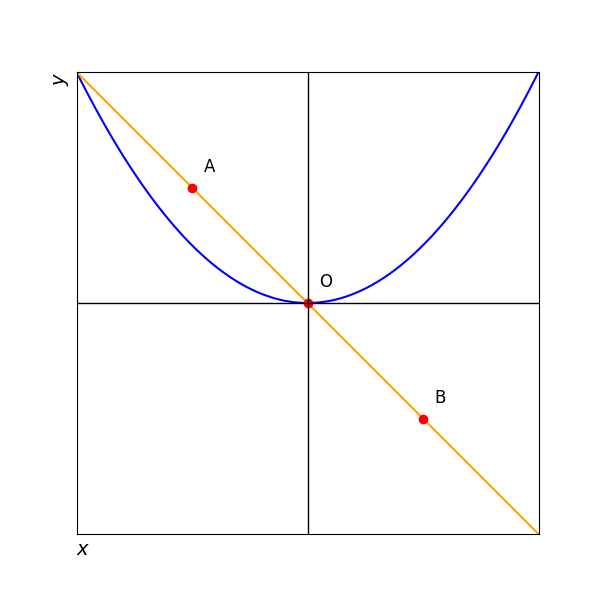} &
    \includegraphics[height=2.1cm,width=\wklength,keepaspectratio]{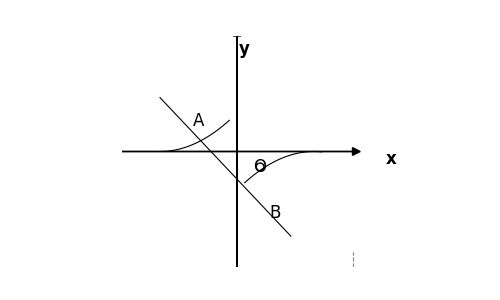} \\
\bottomrule
  \end{tabular}
  \caption{New images synthesized with seed images form K12-2M}%
  \label{new_images_from_k12_examples_3of3}
\end{table*}

\begin{table*}[htb]
  \scriptsize
  \pgfmathsetlength{\wklength}{\textwidth/6 -2\tabcolsep}
  \begin{tabular}{ccc|ccc}
    \toprule
    \multicolumn{3}{c|}{New Images from One Seed Image} &
    \multicolumn{3}{c}{New Images from One Seed Image} \\
    \midrule
    \includegraphics[height=2.1cm,width=\wklength,keepaspectratio]{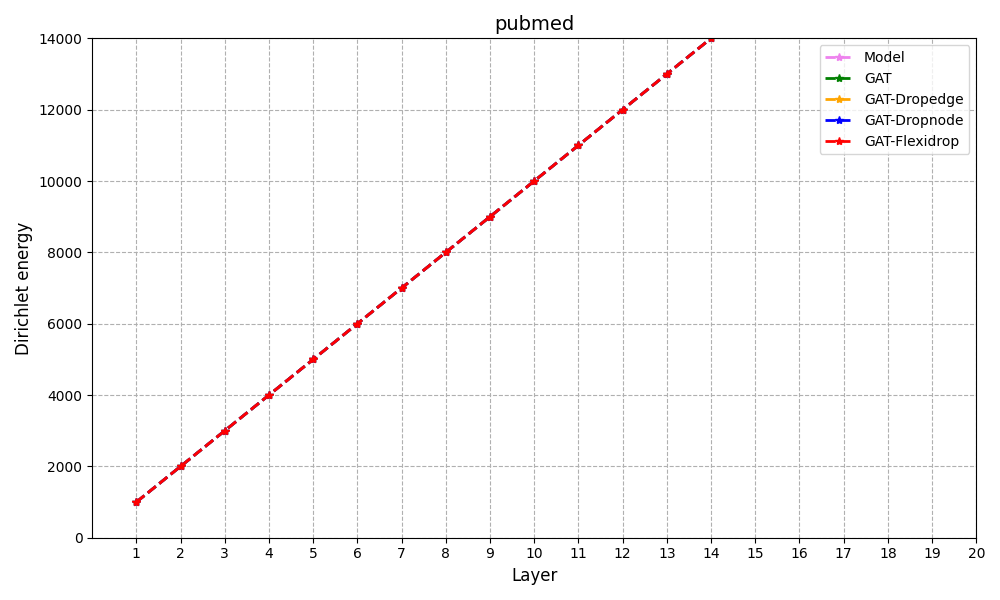} &
    \includegraphics[height=2.1cm,width=\wklength,keepaspectratio]{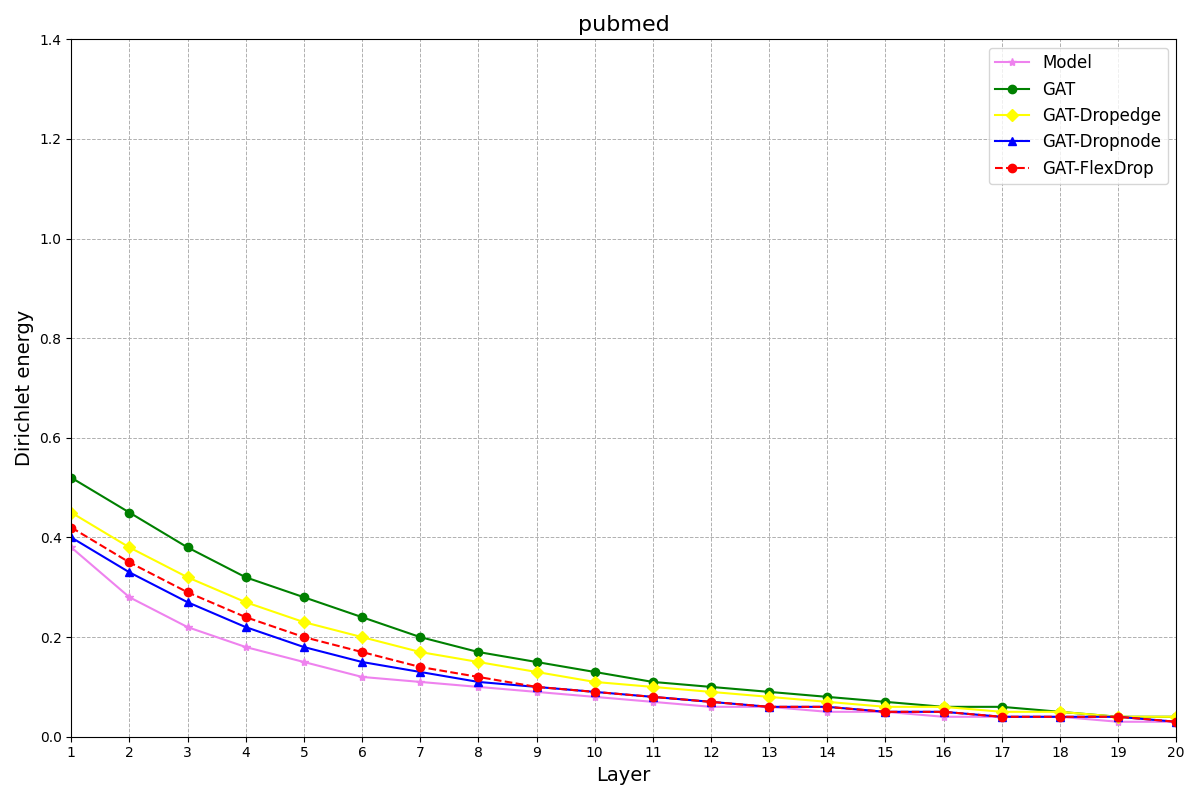} &
    \includegraphics[height=2.1cm,width=\wklength,keepaspectratio]{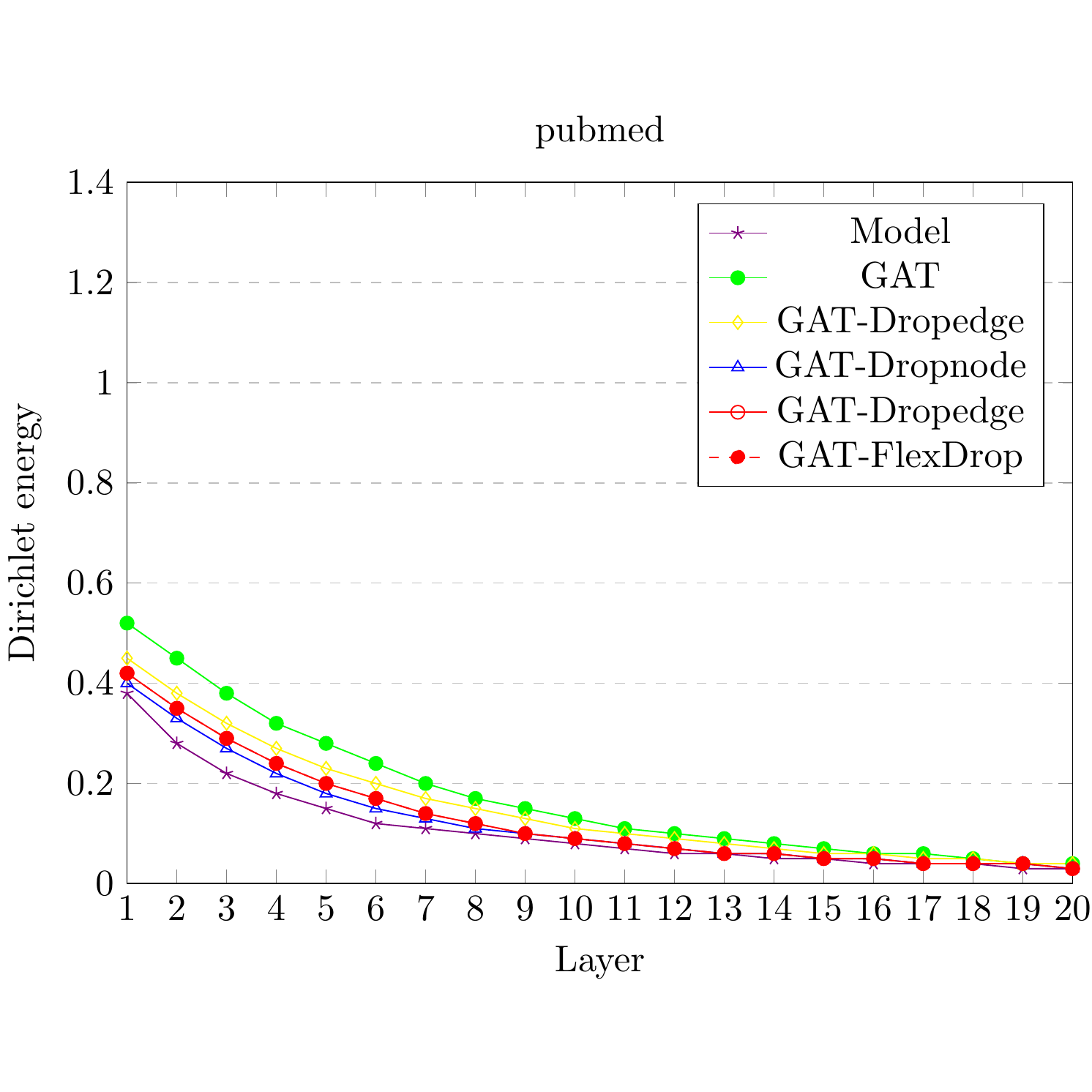} &
    \includegraphics[height=2.1cm,width=\wklength,keepaspectratio]{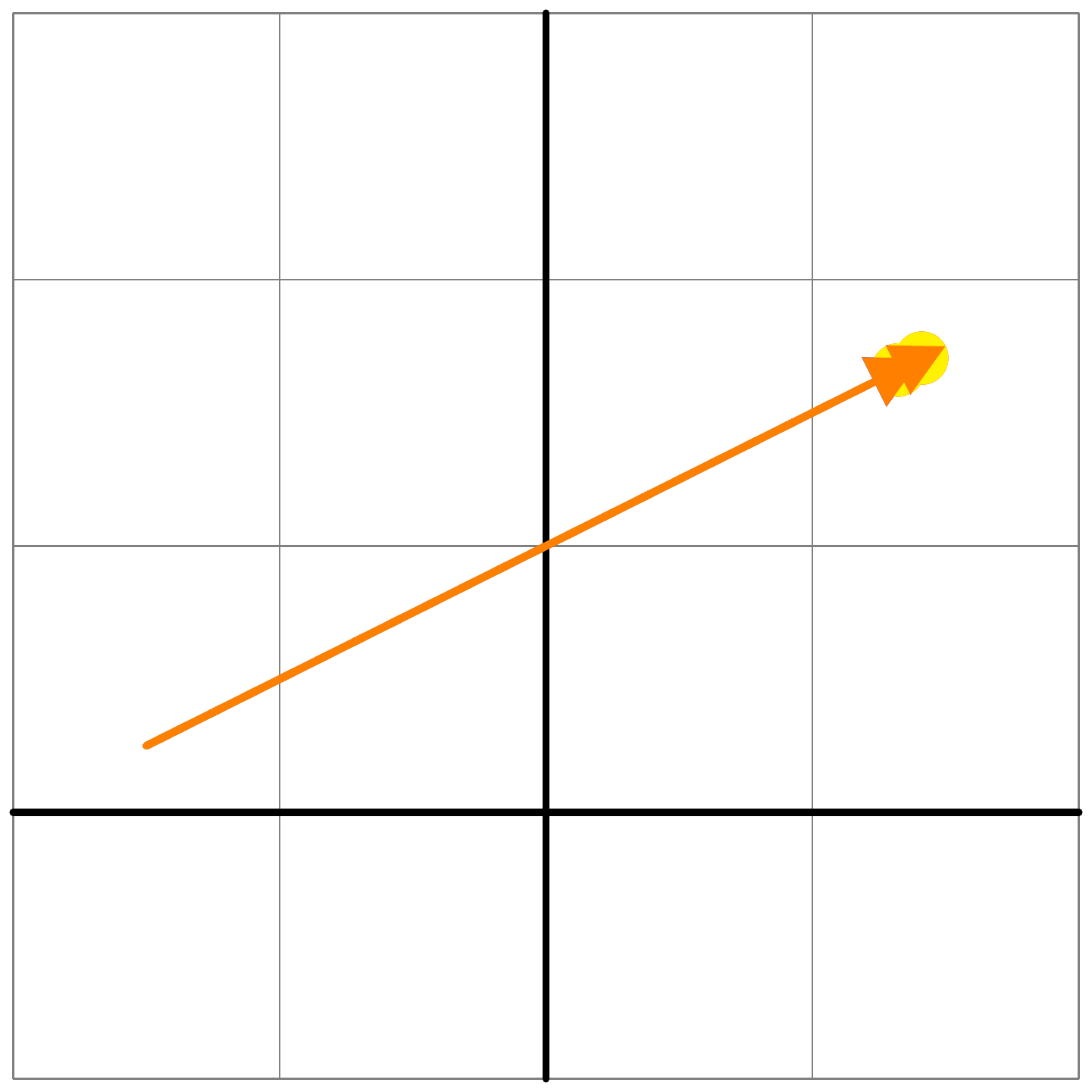} &
    \includegraphics[height=2.1cm,width=\wklength,keepaspectratio]{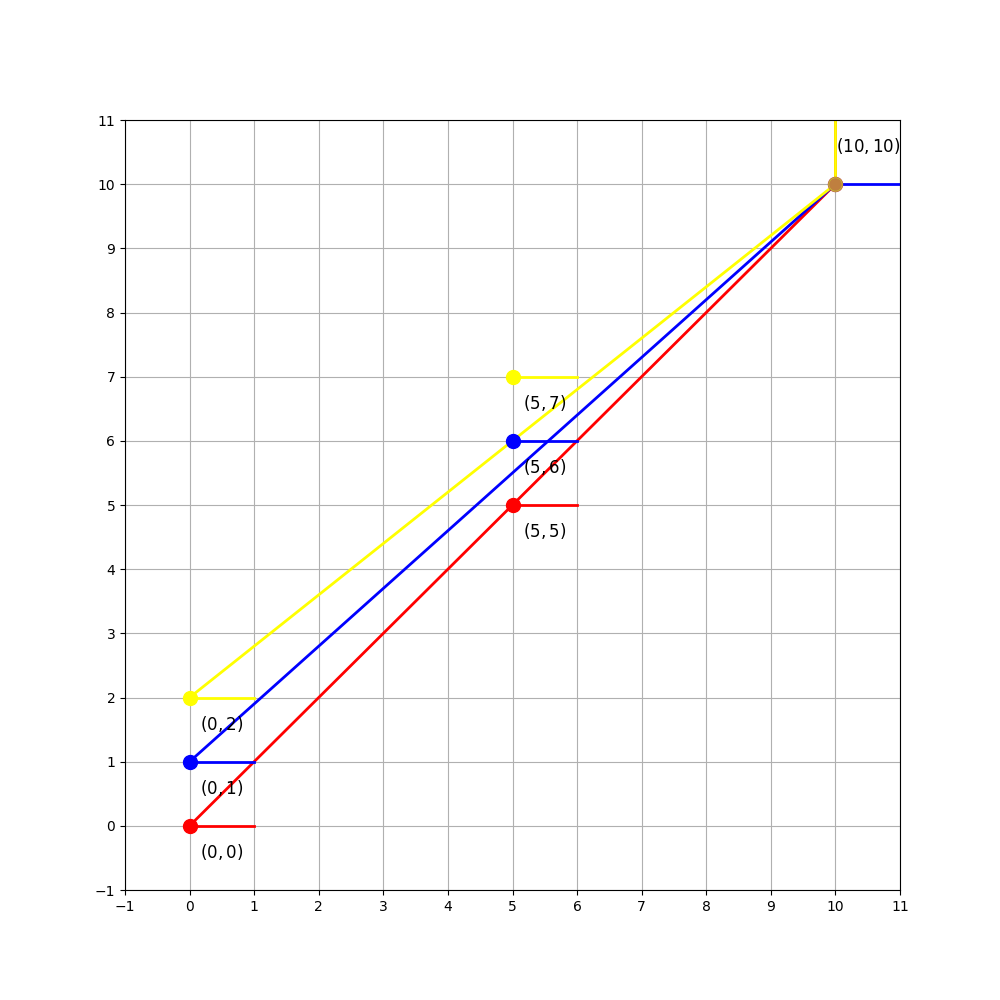} &
    \includegraphics[height=2.1cm,width=\wklength,keepaspectratio]{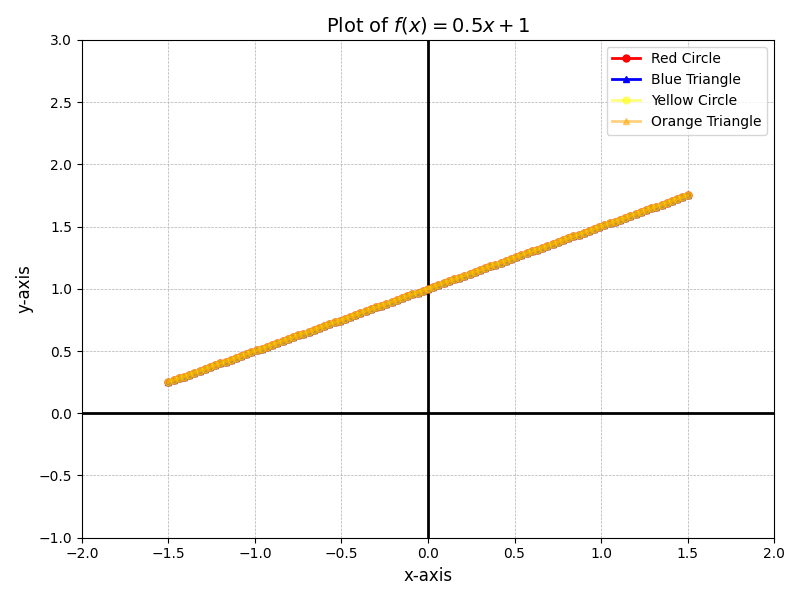} \\
    \midrule
    \includegraphics[height=2.1cm,width=\wklength,keepaspectratio]{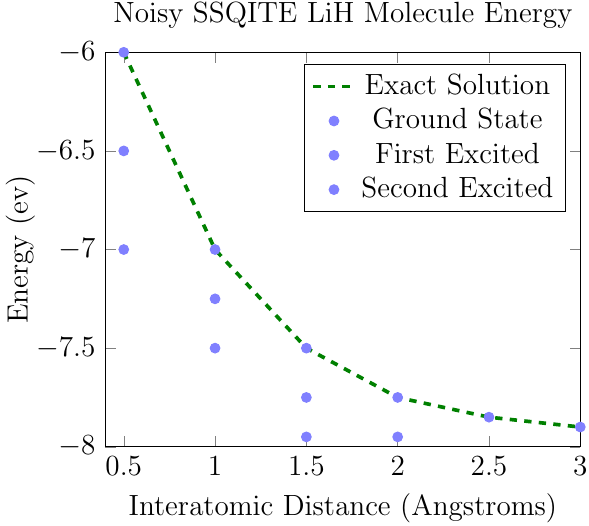} &
    \includegraphics[height=2.1cm,width=\wklength,keepaspectratio]{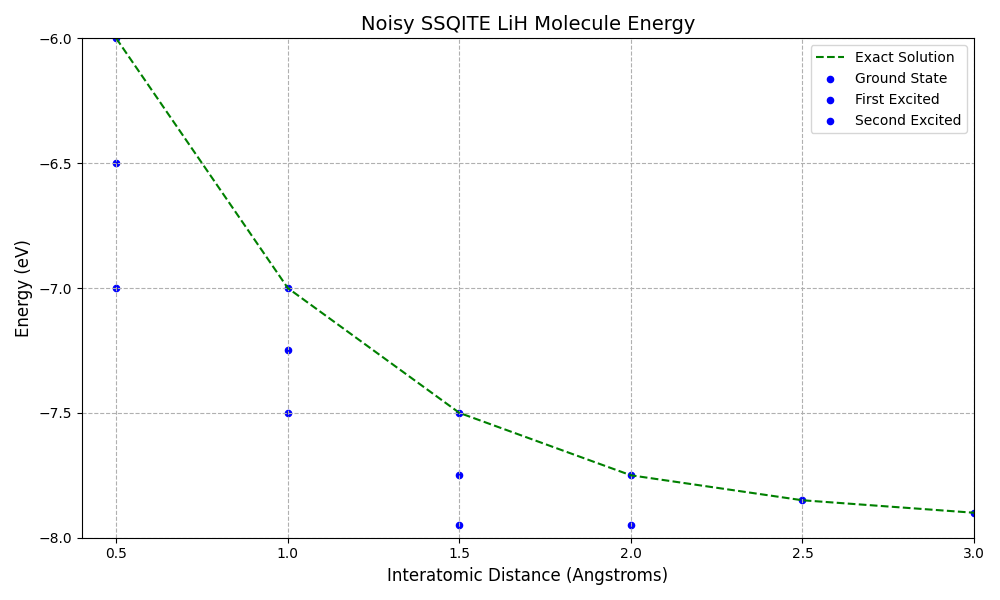} &
    \includegraphics[height=2.1cm,width=\wklength,keepaspectratio]{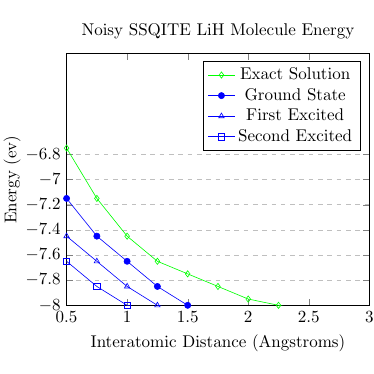} &
    \includegraphics[height=2.1cm,width=\wklength,keepaspectratio]{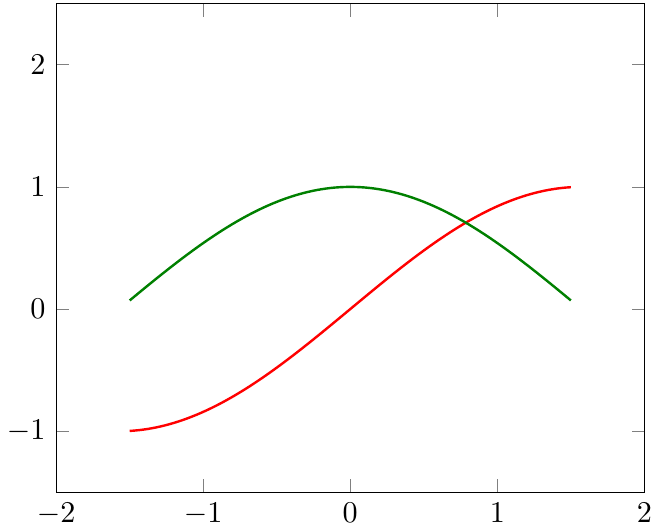} &
    \includegraphics[height=2.1cm,width=\wklength,keepaspectratio]{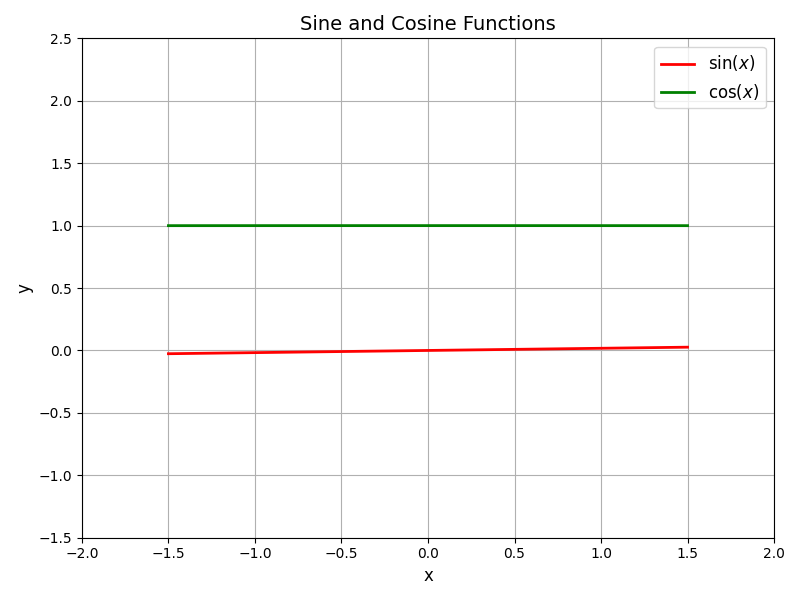} &
    \includegraphics[height=2.1cm,width=\wklength,keepaspectratio]{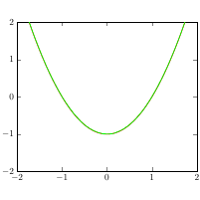} \\
    \midrule
    \includegraphics[height=2.1cm,width=\wklength,keepaspectratio]{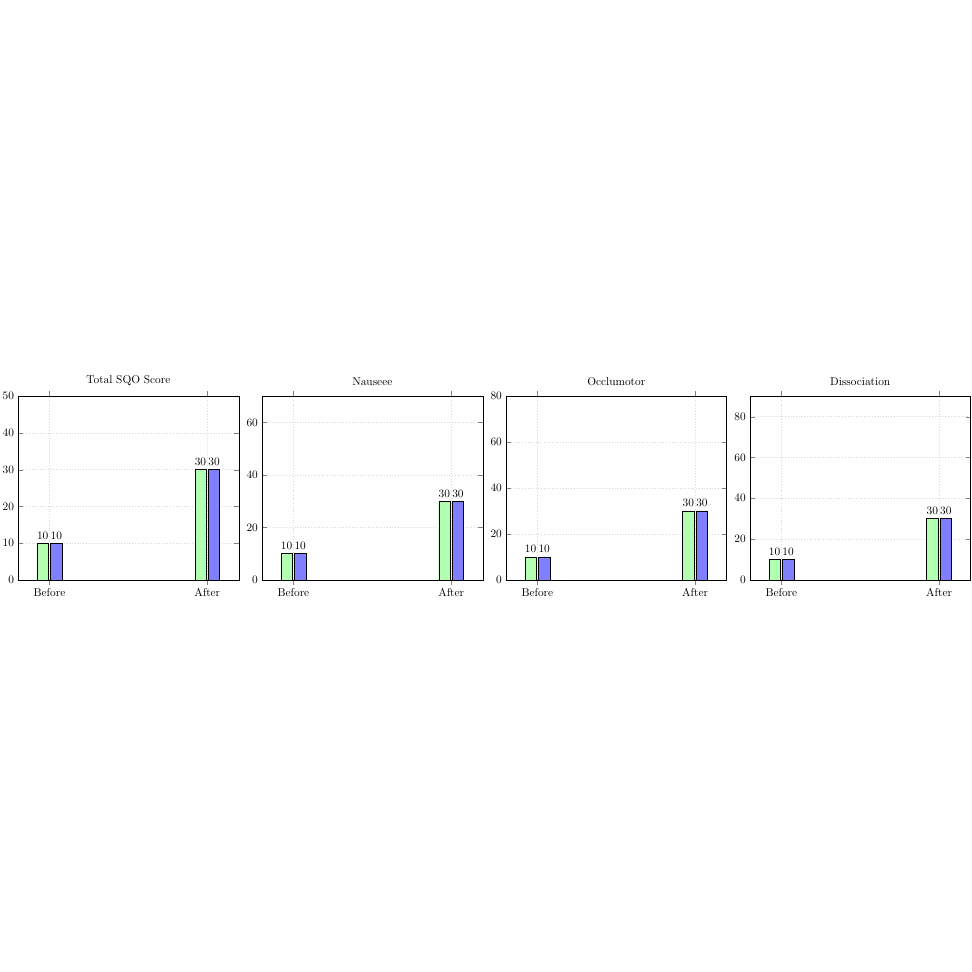} &
    \includegraphics[height=2.1cm,width=\wklength,keepaspectratio]{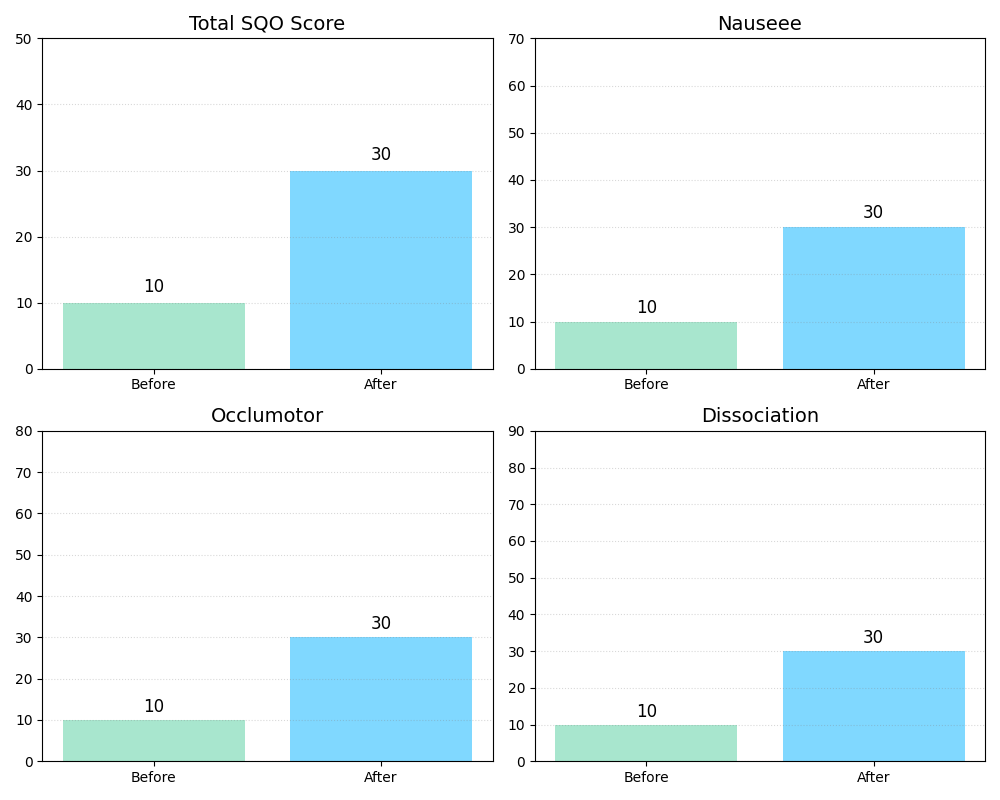} &
    \includegraphics[height=2.1cm,width=\wklength,keepaspectratio]{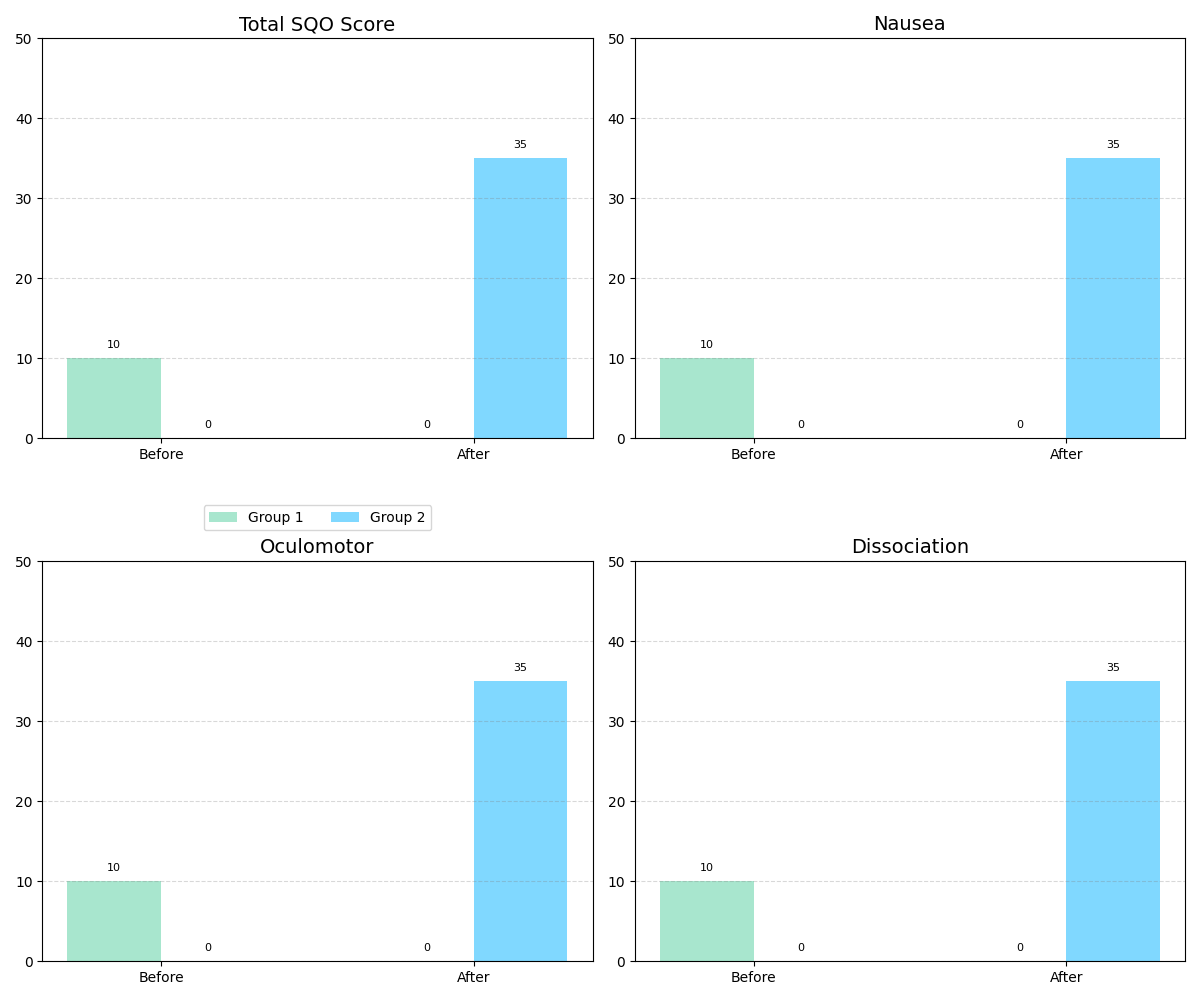} &
    \includegraphics[height=2.1cm,width=\wklength,keepaspectratio]{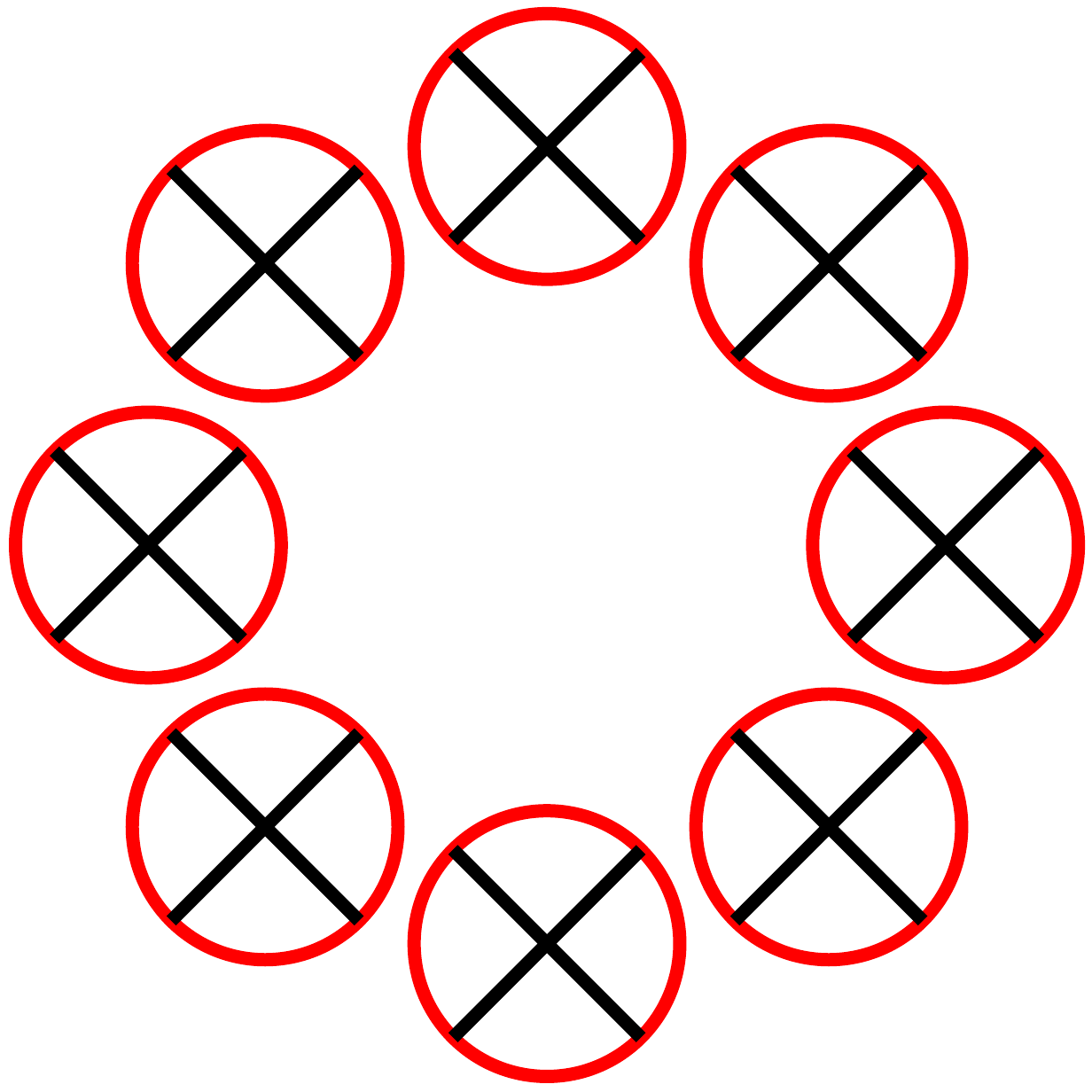} &
    \includegraphics[height=2.1cm,width=\wklength,keepaspectratio]{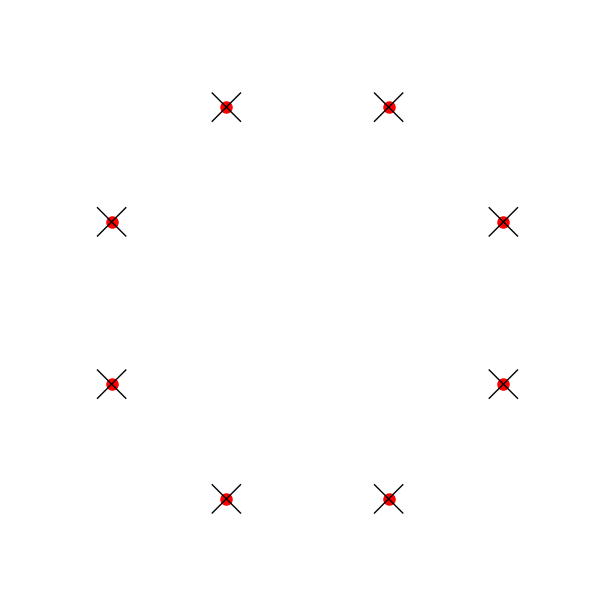} &
    \includegraphics[height=2.1cm,width=\wklength,keepaspectratio]{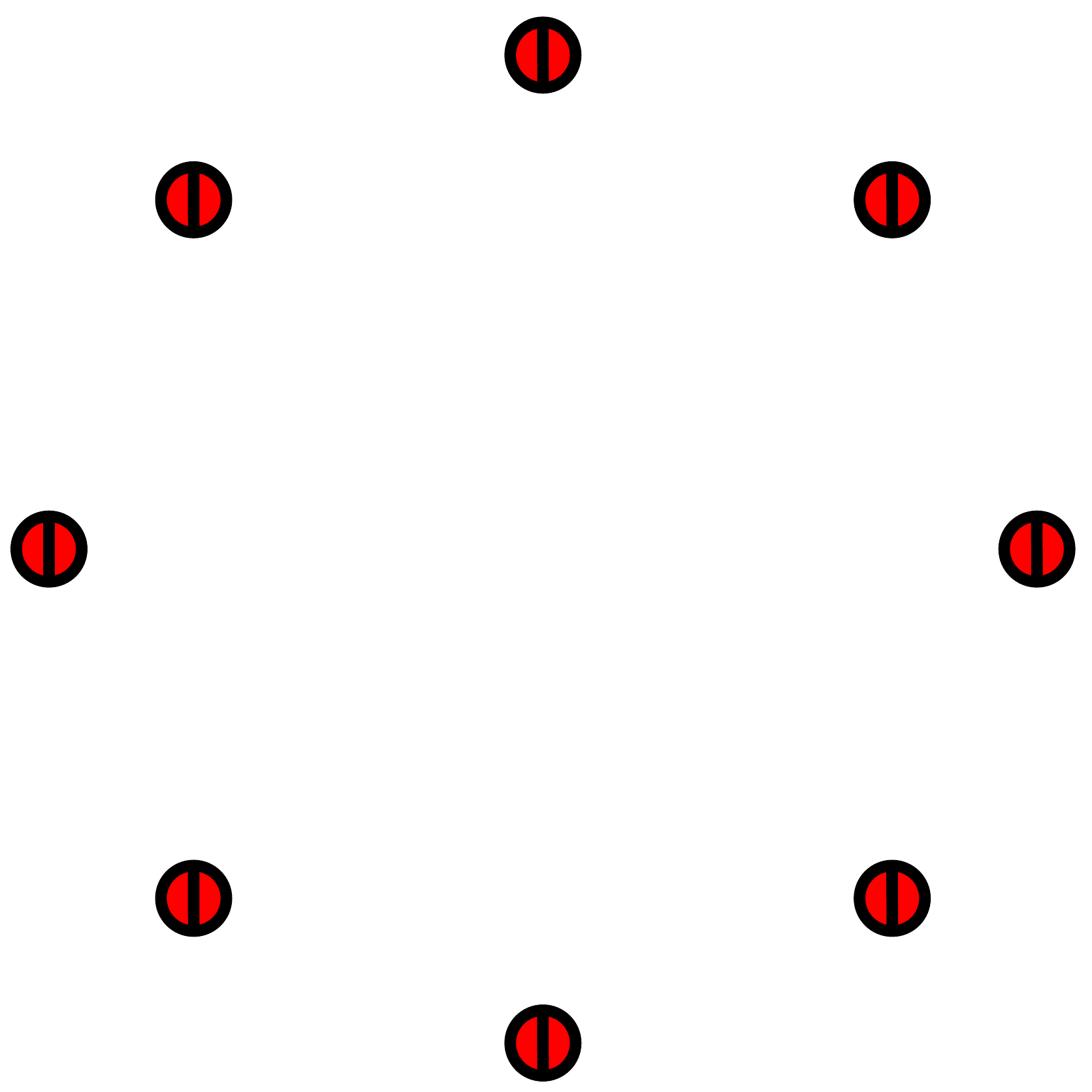} \\
\bottomrule
  \end{tabular}
  \caption{New images synthesized with seed images form arXiv}%
  \label{new_images_from_arxiv_examples_1of1}
\end{table*}

\begin{table*}[htb]
  \scriptsize
  \pgfmathsetlength{\wklength}{\textwidth/6 -2\tabcolsep}
  \begin{tabular}{ccc|ccc}
    \toprule
    \multicolumn{3}{c|}{New Images from One Seed Image} &
    \multicolumn{3}{c}{New Images from One Seed Image} \\
    \midrule
    \includegraphics[height=2.1cm,width=\wklength,keepaspectratio]{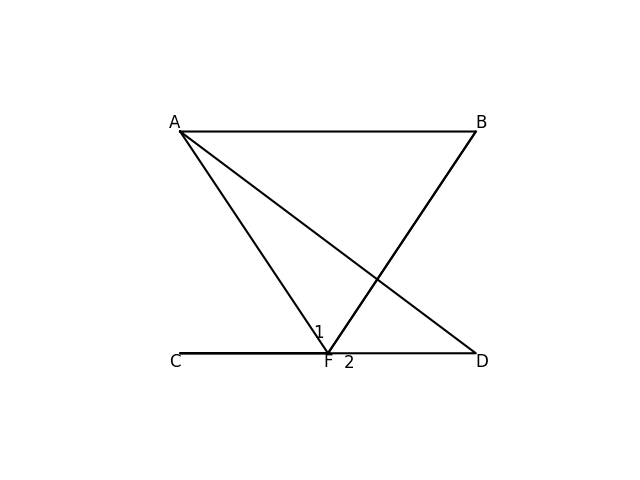} &
    \includegraphics[height=2.1cm,width=\wklength,keepaspectratio]{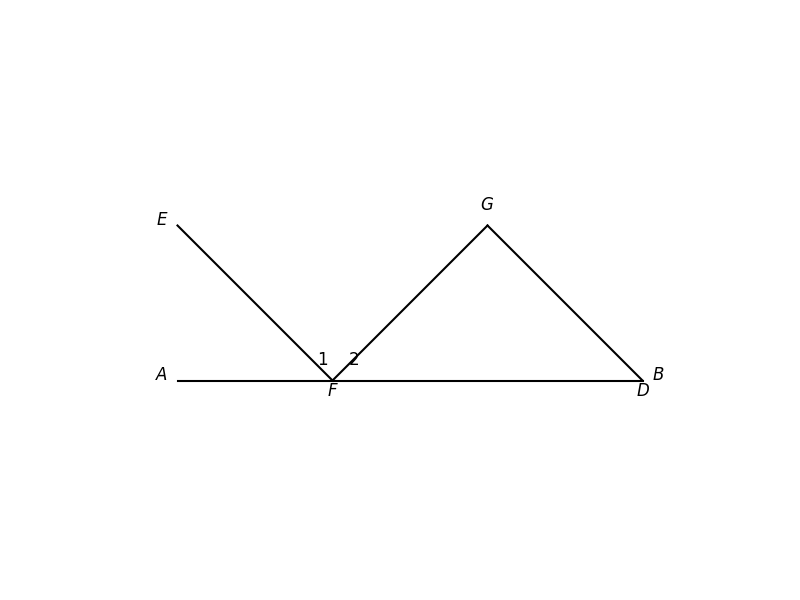} &
    \includegraphics[height=2.1cm,width=\wklength,keepaspectratio]{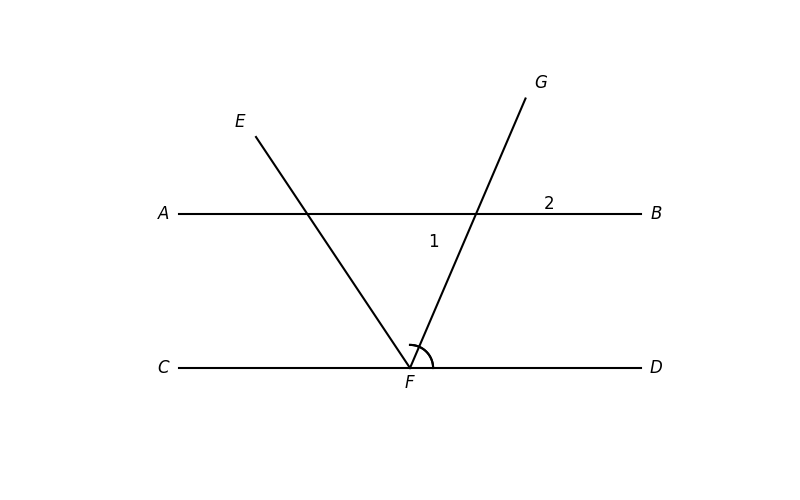} &
    \includegraphics[height=2.1cm,width=\wklength,keepaspectratio]{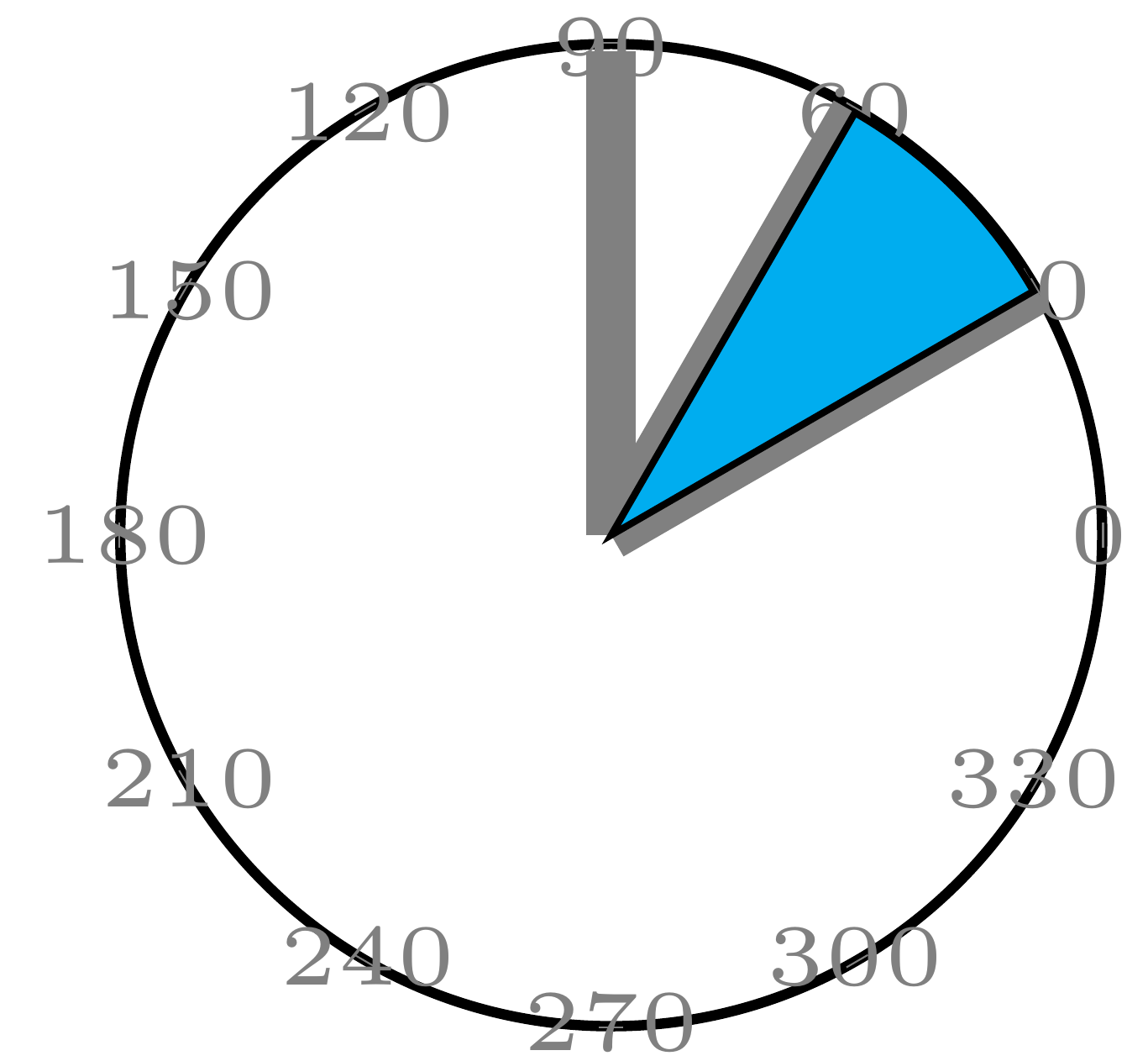} &
    \includegraphics[height=2.1cm,width=\wklength,keepaspectratio]{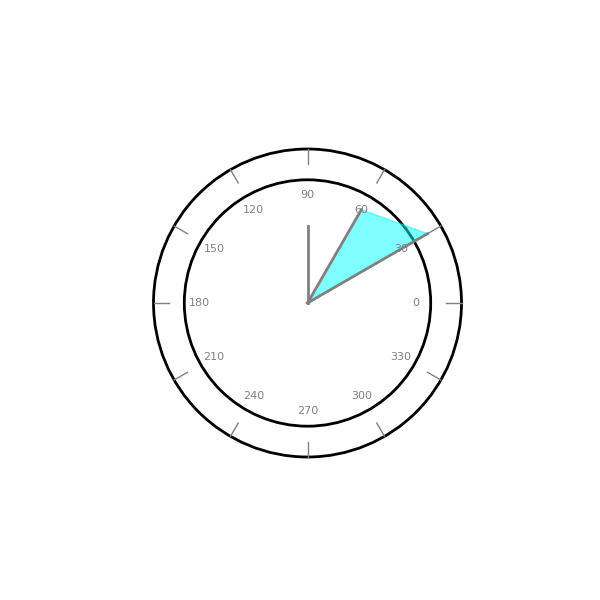} &
    \includegraphics[height=2.1cm,width=\wklength,keepaspectratio]{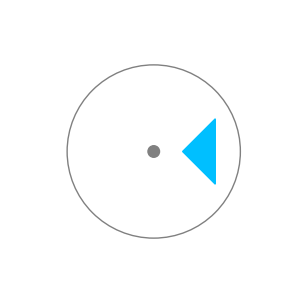} \\
    \midrule
    \includegraphics[height=2.1cm,width=\wklength,keepaspectratio]{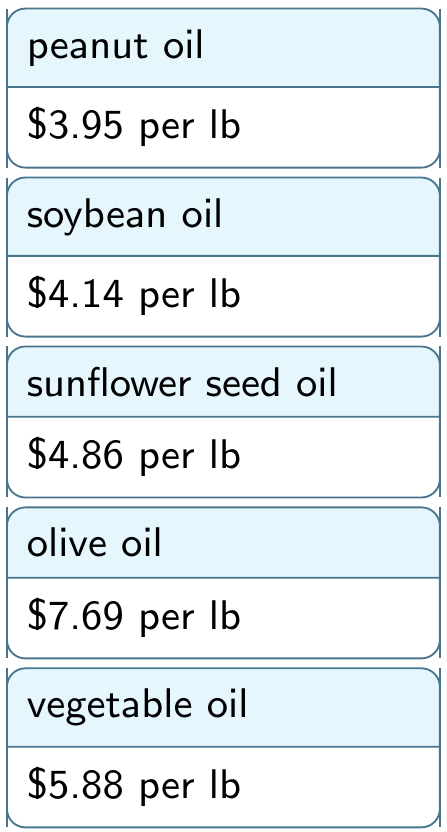} &
    \includegraphics[height=2.1cm,width=\wklength,keepaspectratio]{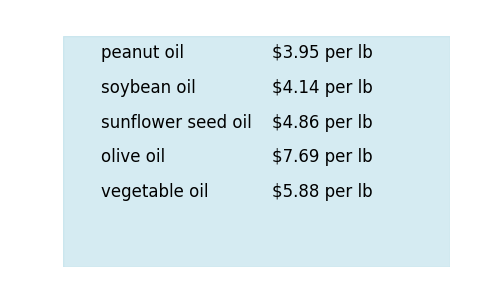} &
    \includegraphics[height=2.1cm,width=\wklength,keepaspectratio]{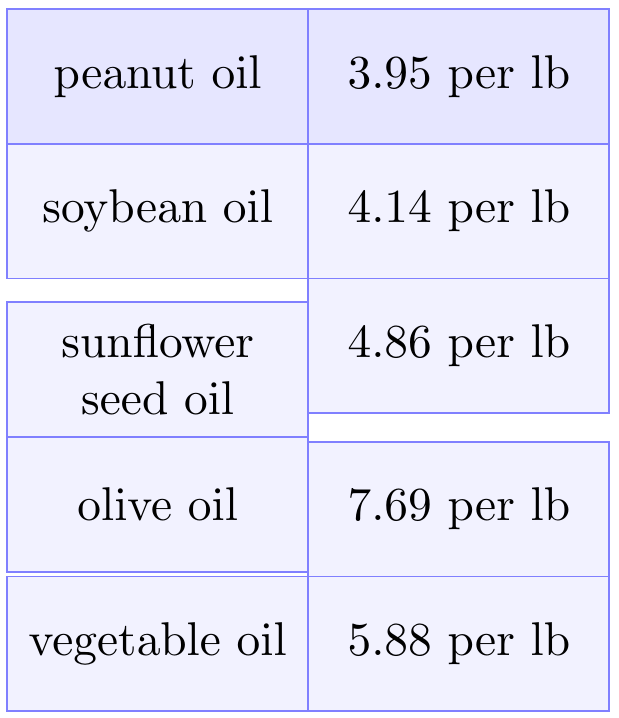} &
    \includegraphics[height=2.1cm,width=\wklength,keepaspectratio]{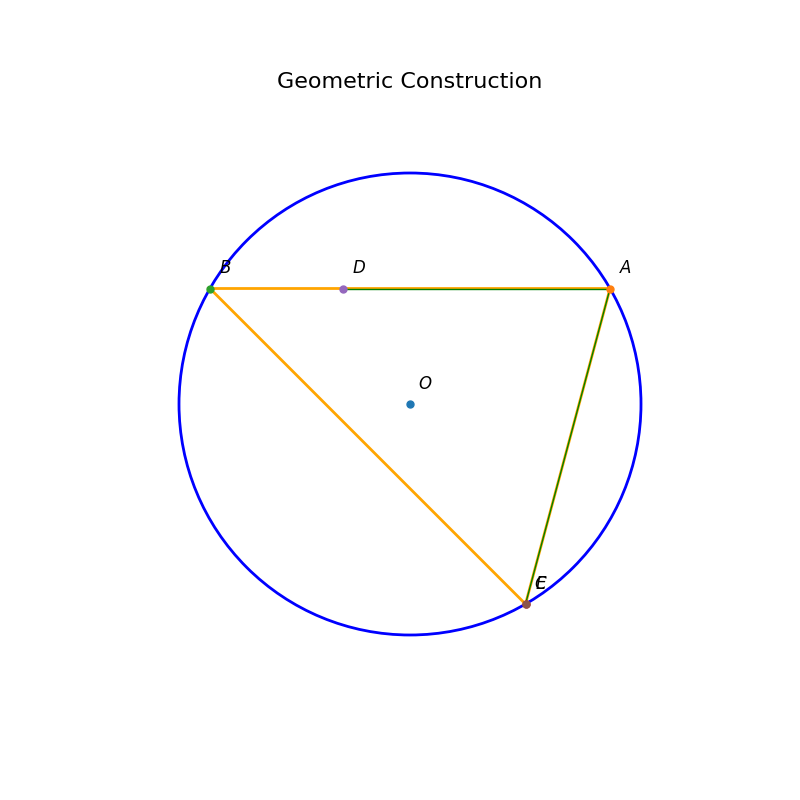} &
    \includegraphics[height=2.1cm,width=\wklength,keepaspectratio]{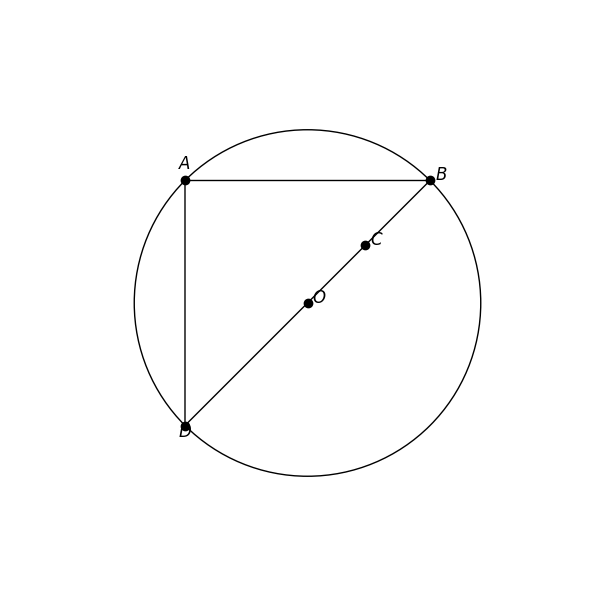} &
    \includegraphics[height=2.1cm,width=\wklength,keepaspectratio]{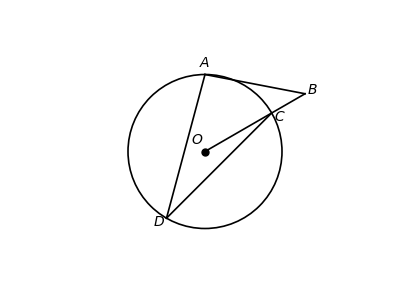} \\
\bottomrule
  \end{tabular}
  \caption{New images synthesized with seed images form MathV360k}%
  \label{new_images_from_mathv360k_examples_1of2}
\end{table*}

\begin{table*}[htb]
  \scriptsize
  \pgfmathsetlength{\wklength}{\textwidth/6 -2\tabcolsep}
  \begin{tabular}{ccc|ccc}
    \toprule
    \multicolumn{3}{c|}{New Images from One Seed Image} &
    \multicolumn{3}{c}{New Images from One Seed Image} \\
    \midrule
    \includegraphics[height=2.1cm,width=\wklength,keepaspectratio]{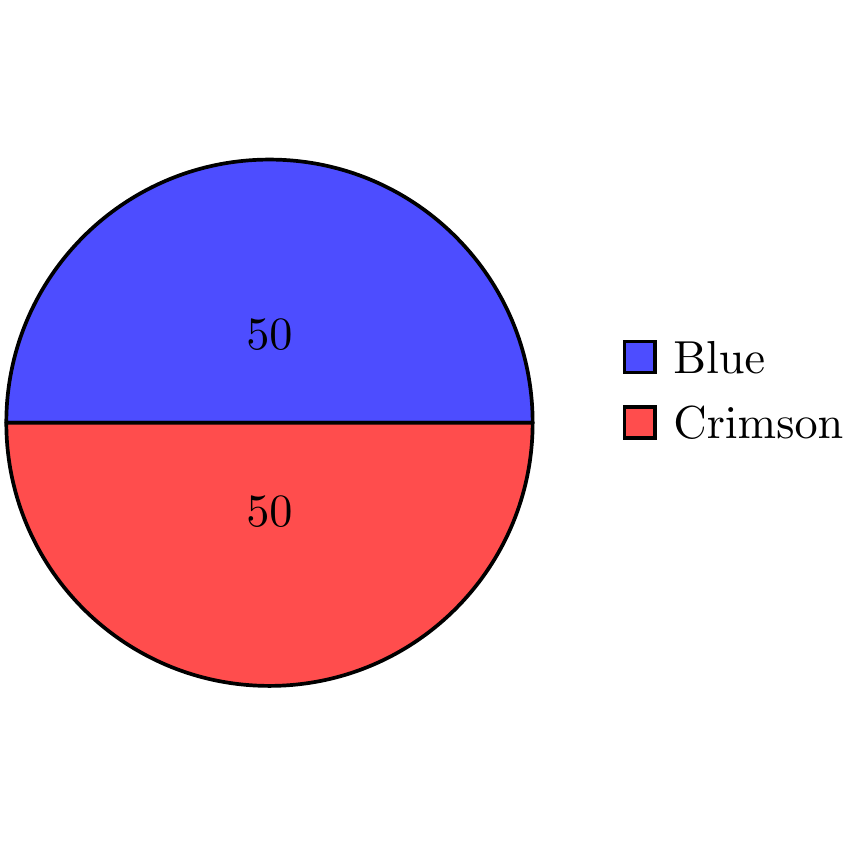} &
    \includegraphics[height=2.1cm,width=\wklength,keepaspectratio]{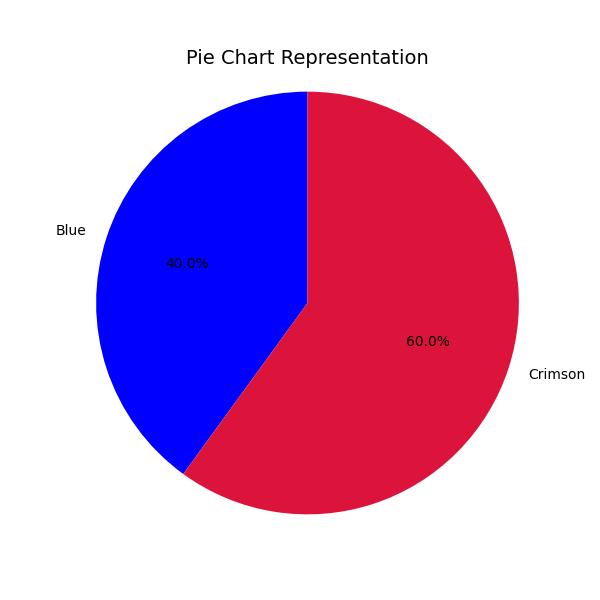} &
    \includegraphics[height=2.1cm,width=\wklength,keepaspectratio]{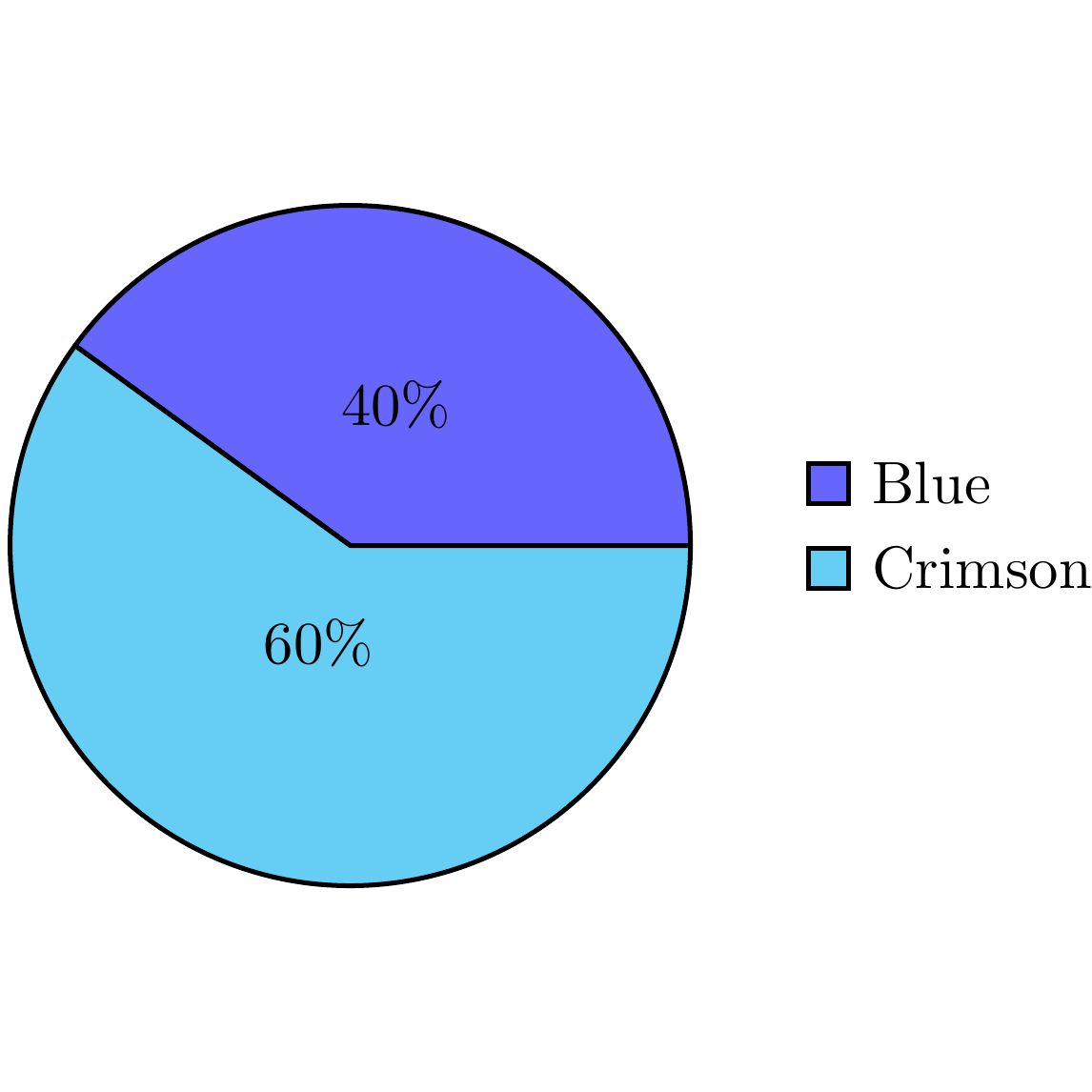} &
    \includegraphics[height=2.1cm,width=\wklength,keepaspectratio]{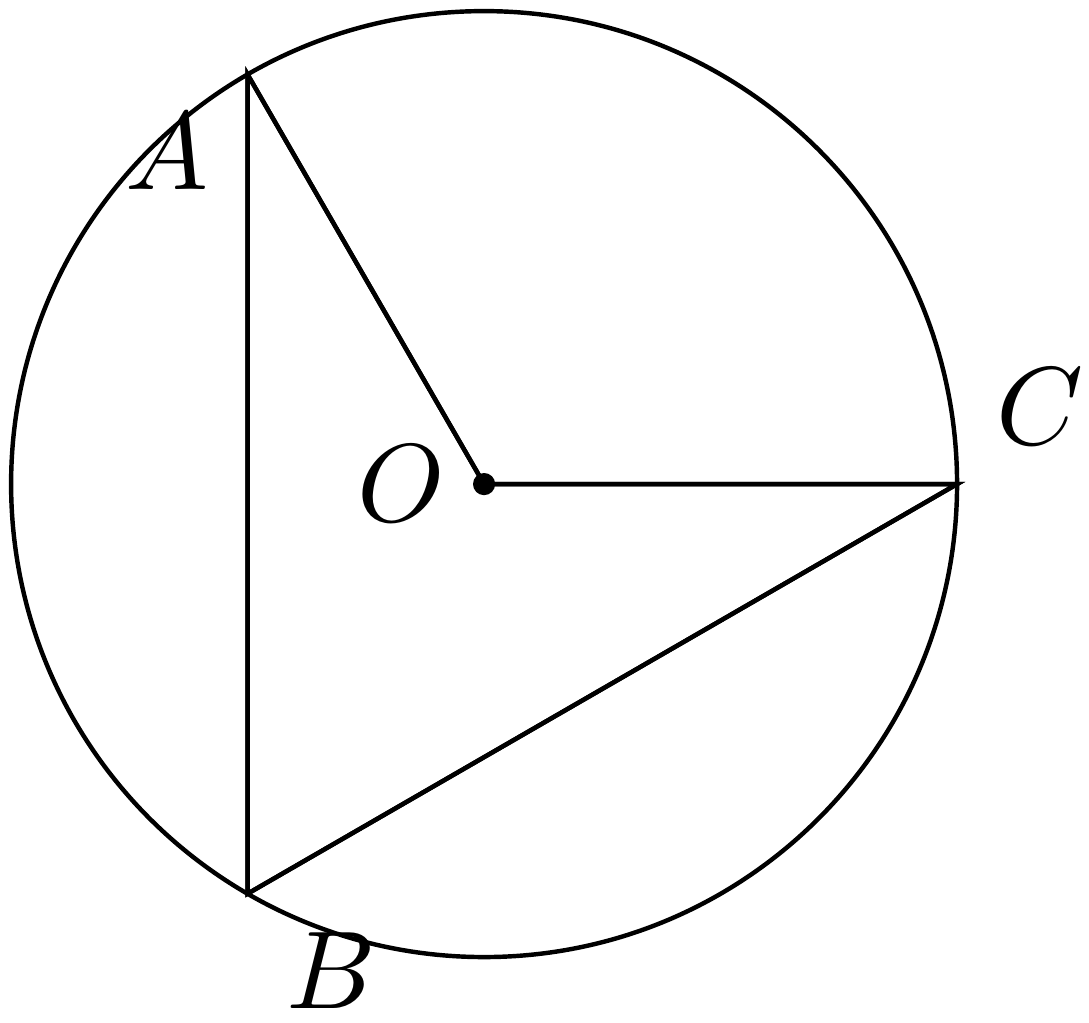} &
    \includegraphics[height=2.1cm,width=\wklength,keepaspectratio]{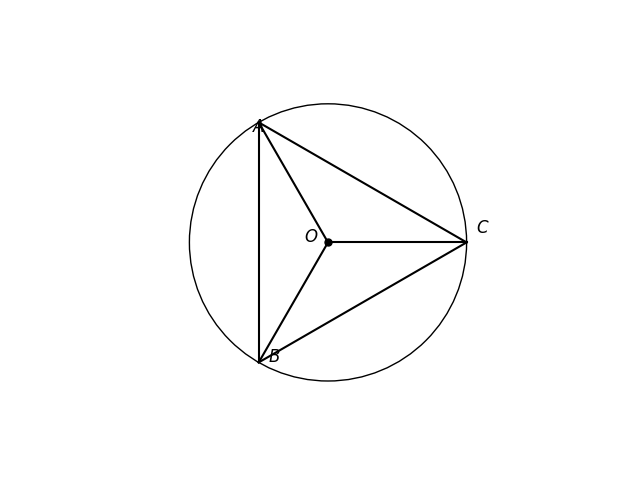} &
    \includegraphics[height=2.1cm,width=\wklength,keepaspectratio]{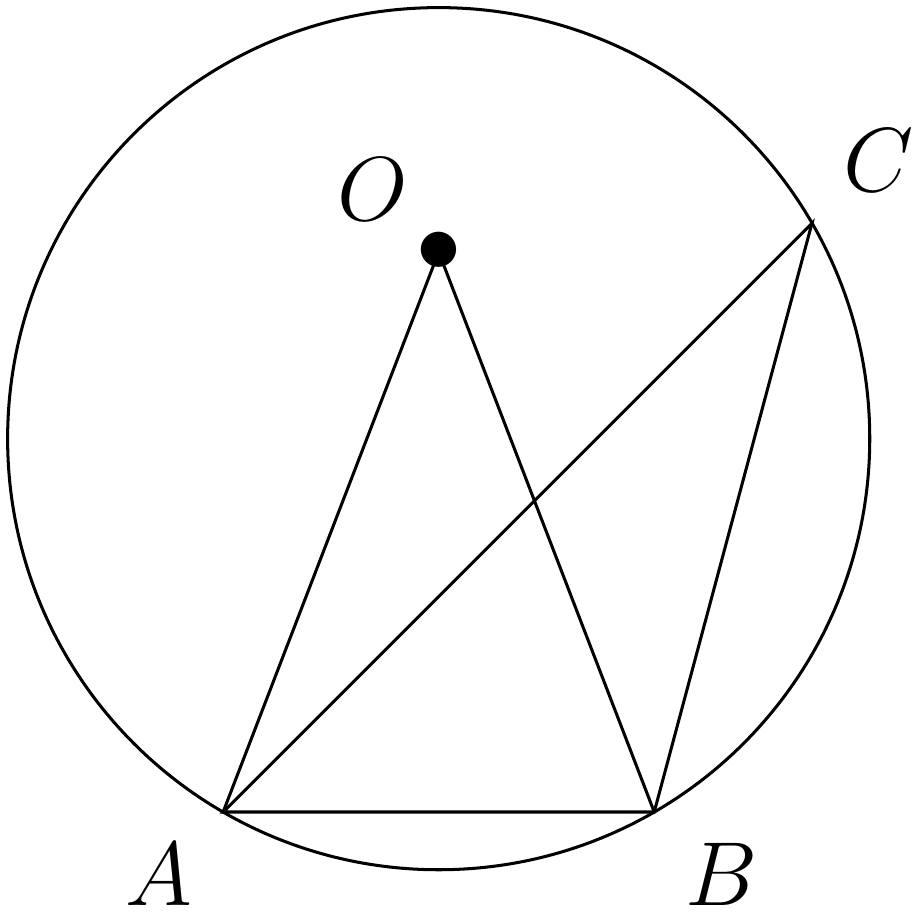} \\
    \midrule
    \includegraphics[height=2.1cm,width=\wklength,keepaspectratio]{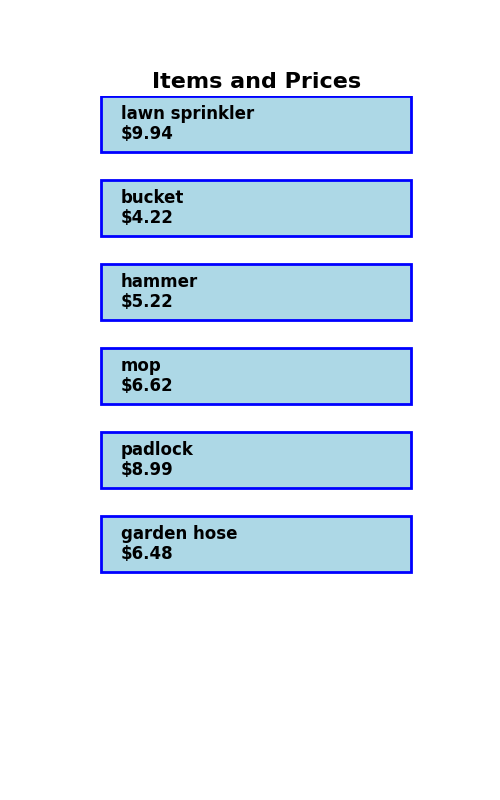} &
    \includegraphics[height=2.1cm,width=\wklength,keepaspectratio]{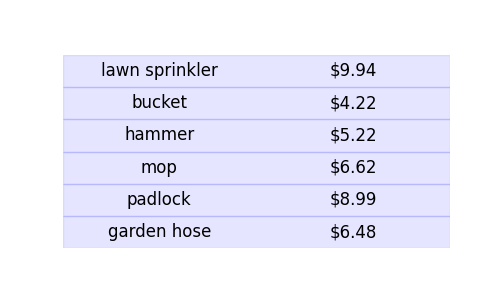} &
    \includegraphics[height=2.1cm,width=\wklength,keepaspectratio]{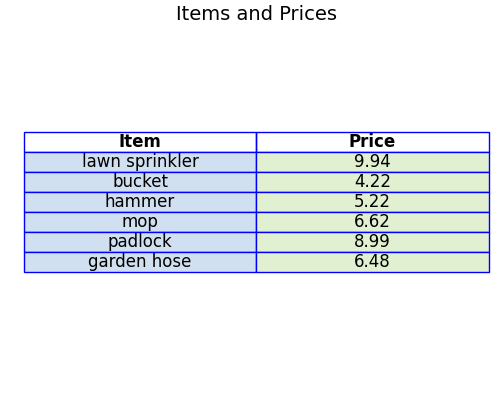} &
    \includegraphics[height=2.1cm,width=\wklength,keepaspectratio]{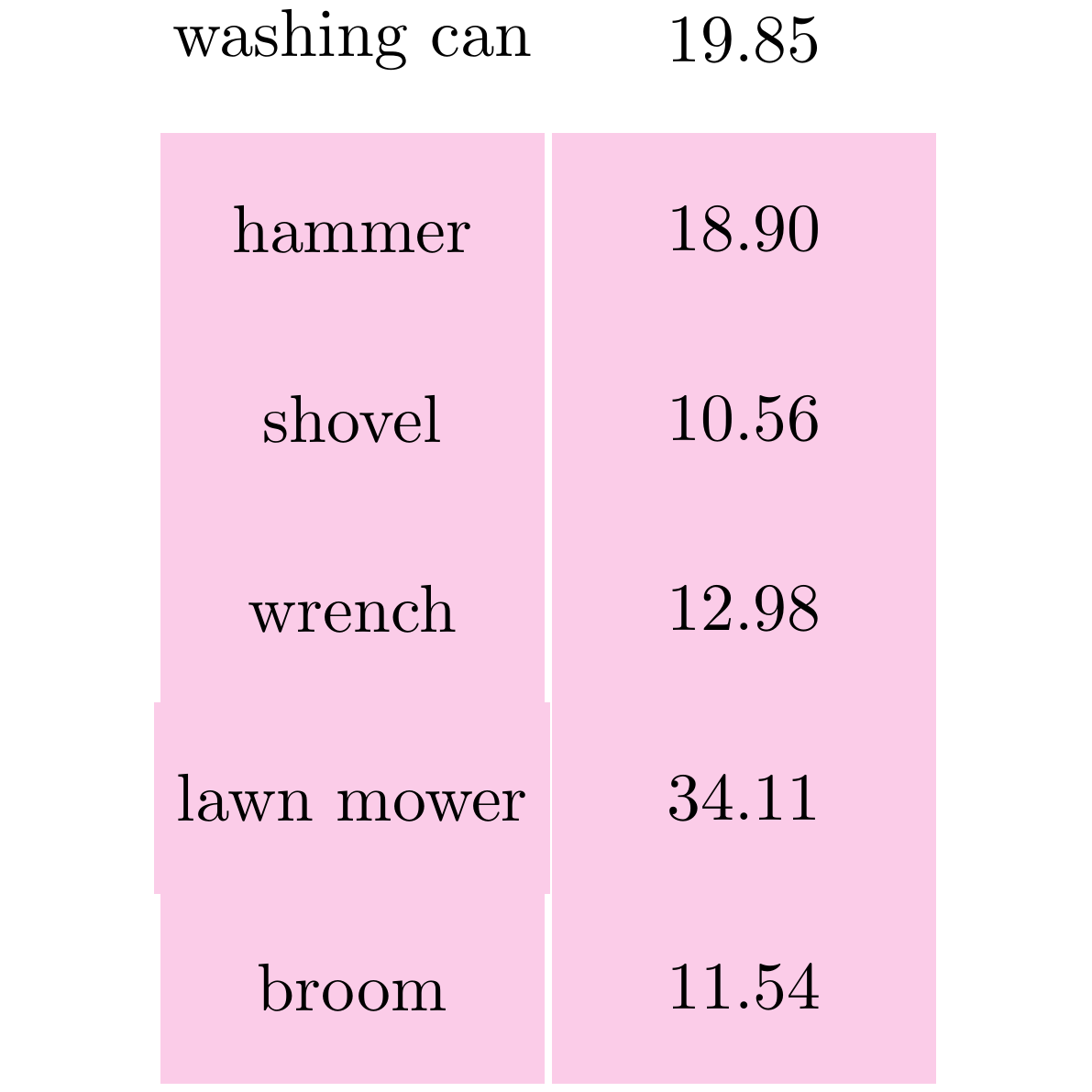} &
    \includegraphics[height=2.1cm,width=\wklength,keepaspectratio]{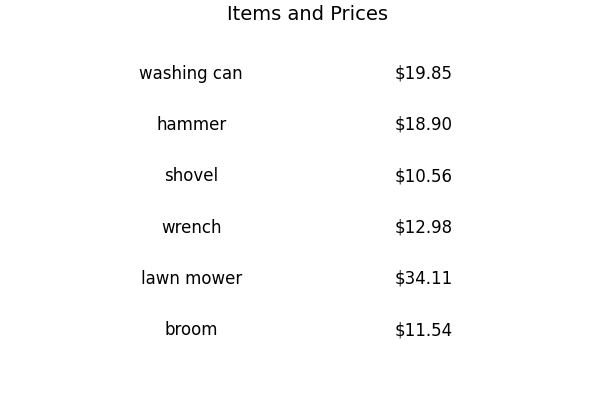} &
    \includegraphics[height=2.1cm,width=\wklength,keepaspectratio]{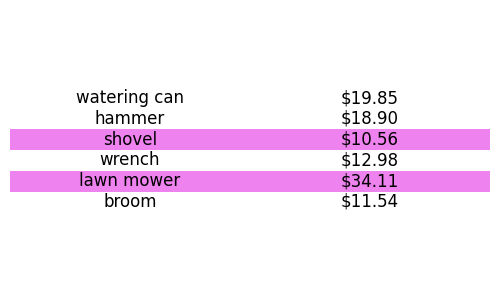} \\
\bottomrule
  \end{tabular}
  \caption{New images synthesized with seed images form MathV360k}%
  \label{new_images_from_mathv360k_examples_2of2}
\end{table*}